\PassOptionsToPackage{table}{xcolor}
\PassOptionsToPackage{hyphens}{url}
\documentclass{article}
\usepackage{iclr2027_conference,times}
\usepackage[table]{xcolor}
\usepackage{hyperref}
\usepackage{xurl}
\hypersetup{hidelinks,colorlinks=false,pdfborder={0 0 0},
  pdftitle={Positions Are Not Facts: The Mismatch Between KV Caches and Memory},
  pdfauthor={Changhai Zhou, Yuhua Zhou, Shiyang Zhang, Jun Gao, Zhen Li, Hua Wu, Hanchao Yu, Haifeng Wang}}
\usepackage{graphicx}
\usepackage{float}
\usepackage{placeins}
\usepackage{caption}
\usepackage{booktabs}
\usepackage{tabularx}
\usepackage{array}
\usepackage{multirow}
\usepackage{amsmath}
\usepackage{amssymb}
\usepackage{microtype}
\usepackage{tcolorbox}
\usepackage{tikz}
\usetikzlibrary{arrows.meta}
\usepackage{etoolbox}
\iclrfinalcopy
\makeatletter
\newcommand{\arxivauthorblock}{%
  {\normalsize
    Changhai Zhou$^{1,2}$, Yuhua Zhou$^{3}$, Shiyang Zhang$^{4}$, Jun Gao$^{3}$\par
    Zhen Li$^{5}$, Hua Wu$^{2}$, Hanchao Yu$^{2}$, Haifeng Wang$^{2}$\par}
  \vskip 3pt
  {\normalsize
    $^{1}$Fudan University\quad $^{2}$Baidu\quad $^{3}$Zhejiang University\par
    $^{4}$Yale University\quad $^{5}$Hong Kong Polytechnic University\par}}
\def\@maketitle{%
  \vbox{\hsize\textwidth
    {\LARGE\sc \@title\par}
    \vskip 0.22in minus 0.08in
    {\centering\arxivauthorblock\par}
    \vskip 0.25in minus 0.1in}}
\makeatother
\definecolor{figureBlue}{HTML}{4477AA}
\definecolor{figureRose}{HTML}{CC6677}
\definecolor{figureGray}{HTML}{777777}
\colorlet{tableInk}{black!85}
\colorlet{tableHeader}{figureGray!14!white}
\colorlet{tableNeutral}{figureGray!8!white}
\colorlet{tableGroup}{figureGray!12!white}
\colorlet{tableWhole}{figureBlue!11!white}
\colorlet{tableNumber}{figureRose!11!white}
\colorlet{tablePass}{tableNeutral}
\colorlet{tableRisk}{tableNeutral}
\colorlet{paletteMint}{figureBlue!22!white}
\colorlet{paletteSky}{figureBlue!35!white}
\colorlet{paletteCream}{figureGray!16!white}
\colorlet{paletteCoral}{figureRose!35!white}
\newcommand{\passcell}[1]{\cellcolor{tableNeutral}#1}
\newcommand{\riskcell}[1]{\cellcolor{tableNeutral}#1}
\newcommand{\neutralcell}[1]{\cellcolor{tableNeutral}#1}

\newcolumntype{Y}{>{\centering\arraybackslash}X}
\newcolumntype{L}[1]{>{\centering\arraybackslash}m{#1}}
\newcolumntype{C}[1]{>{\centering\arraybackslash}m{#1}}
\newcolumntype{R}[1]{>{\centering\arraybackslash}m{#1}}
\newcommand{\TableStyle}{%
  \centering\small
  \arrayrulecolor{tableInk}%
  \setlength{\heavyrulewidth}{0.65pt}%
  \setlength{\lightrulewidth}{0.35pt}%
  \setlength{\cmidrulewidth}{0.25pt}%
  \setlength{\tabcolsep}{4pt}%
  \renewcommand{\arraystretch}{1.08}%
}
\newcommand{\WideTableStyle}{\TableStyle\footnotesize\setlength{\tabcolsep}{3pt}}
\newcommand{\TableHead}[1]{\cellcolor{tableHeader}\textbf{#1}}
\newcommand{\TableLabel}[1]{\textbf{#1}}
\newcommand{\TableStub}[1]{\cellcolor{tableNeutral}#1}
\newcommand{\WholeCell}[1]{\cellcolor{tableWhole}#1}
\newcommand{\NumberCell}[1]{\cellcolor{tableNumber}#1}
\newcommand{\FullCell}[1]{\cellcolor{tableNeutral}#1}
\newcommand{\TableGroupRow}[2]{\multicolumn{#1}{c}{\cellcolor{tableGroup}\textbf{#2}}\\}
\newcommand{\TableSecondary}[1]{\textcolor{figureGray}{#1}}

\newcommand{\TableNoteAlign}{\leftskip=0pt\rightskip=0pt\parfillskip=0pt plus .30\linewidth}
\newcommand{\TableNoteEnd}{\par}

\newcommand{\AnswerPct}[2]{\shortstack[c]{#1\%\\[-.5pt]\textcolor{figureGray}{(#2)}}}
\newcommand{\AnswerPctInline}[2]{#1\%\,\textcolor{figureGray}{(#2)}}
\newcommand{\AnswerPP}[2]{\shortstack[c]{\MainNumber{#1}\,pp\\[-.5pt]\textcolor{figureGray}{(\MainNumber{#2})}}}
\newcommand{\AnswerPPInline}[2]{\MainNumber{#1}\,pp\,\textcolor{figureGray}{(\MainNumber{#2})}}
\newcommand{\WholeHeader}[1]{\textcolor{figureBlue}{\textbf{#1}}}
\newcommand{\NumberHeader}[1]{\textcolor{figureRose}{\textbf{#1}}}
\newcommand{\FullHeader}[1]{\textcolor{figureGray}{\textbf{#1}}}
\newcommand{\TableInterval}[6]{%
  \begin{tikzpicture}[x=38pt,y=1pt,baseline=-0.6ex]
    \pgfmathsetmacro{\Tpoint}{(#1-#4)/(#5-#4)}%
    \pgfmathsetmacro{\Tlow}{(#2-#4)/(#5-#4)}%
    \pgfmathsetmacro{\Thigh}{(#3-#4)/(#5-#4)}%
    \pgfmathsetmacro{\Tzero}{(0-#4)/(#5-#4)}%
    \useasboundingbox (0,-3.4) rectangle (1,3.4);
    \draw[figureGray!45,line width=.3pt] (\Tzero,-3.4)--(\Tzero,3.4);
    \draw[#6,line width=.7pt] (\Tlow,0)--(\Thigh,0);
    \draw[#6,line width=.5pt] (\Tlow,-1.7)--(\Tlow,1.7) (\Thigh,-1.7)--(\Thigh,1.7);
    \fill[#6] (\Tpoint,0) circle[radius=1.25pt];
  \end{tikzpicture}%
}

\newcommand{\TableDivider}{\noalign{\vskip 1pt}\specialrule{.25pt}{0pt}{1pt}}
\newcommand{\MainTableSpacing}{\setlength{\aboverulesep}{1.2pt}\setlength{\belowrulesep}{1.2pt}\renewcommand{\arraystretch}{1.02}}

\newcolumntype{Z}[1]{>{\hsize=#1\hsize\linewidth=\hsize\centering\arraybackslash}X}
\definecolor{mainComparison}{HTML}{ECE4F5}
\newcommand{\MainFocusCell}[1]{\cellcolor{mainComparison}#1}
\newcommand{\MainTableSmall}{\fontsize{8}{9.2}\selectfont}
\newcommand{\MainTableHeaderSize}{\fontsize{8.5}{10}\selectfont}
\newcommand{\MainTableStyle}{%
  \TableStyle\small
  \setlength{\tabcolsep}{3pt}%
  \renewcommand{\arraystretch}{1.04}%
  \setlength{\aboverulesep}{1.5pt}%
  \setlength{\belowrulesep}{1.5pt}%
  \renewcommand{\FullCell}[1]{##1}%
  \renewcommand{\FullHeader}[1]{\textcolor{figureGray}{{\rmfamily\bfseries\MainTableSmall ##1}}}%
  \renewcommand{\WholeHeader}[1]{\textcolor{figureBlue}{{\rmfamily\bfseries\MainTableSmall ##1}}}%
  \renewcommand{\NumberHeader}[1]{\textcolor{figureRose}{{\rmfamily\bfseries\MainTableSmall ##1}}}%
}
\newcommand{\MainTableHead}[1]{{\rmfamily\bfseries\MainTableHeaderSize #1}}
\newcommand{\MainColumnHead}[1]{{\rmfamily\bfseries\MainTableSmall #1}}
\newcommand{\MainNumber}[1]{\MainNumberSign#1\relax}
\def\MainNumberSign#1#2\relax{%
  \ifx-#1\ensuremath{-}#2\else
    \ifx+#1\ensuremath{+}#2\else#1#2\fi
  \fi
}
\newcommand{\MainEstimate}[3]{%
  \shortstack[c]{\MainNumber{#1}\\[-.5pt]{\MainTableSmall\color{figureGray}[\MainNumber{#2},\,\MainNumber{#3}]}}%
}

\makeatletter
\let\iclr@addcontentsline\addcontentsline
\renewcommand{\addcontentsline}[3]{}
\makeatother

\DeclareCaptionJustification{compact}{\leftskip=0pt\rightskip=0pt\parfillskip=0pt plus .30\linewidth}
\title{Positions Are Not Facts:\\The Mismatch Between KV Caches and Memory}
\author{Changhai Zhou, Yuhua Zhou, Shiyang Zhang, Jun Gao, Zhen Li, Hua Wu, Hanchao Yu, Haifeng Wang}

\begin{document}

\maketitle

\setlength{\parfillskip}{0pt plus .50\textwidth}
\begin{abstract}
When a fact changes, how should a language model update the history
stored in its key--value (KV) cache? Hiding the old record is cheap,
but it may still contain needed details or answer questions about the
past. We compare hiding whole records, hiding only replaced values,
and deleting old text and recomputing the cache. In a controlled
quantity task, masking makes all eight models prefer the new value
more strongly, yet six lose complete answers through unit errors or
failure to stop; keeping the unit preserves all current answers.
Later states also retain useful information from earlier records:
on multi-hop updates, rebuilding these states at unchanged positions
lowers historical accuracy by 20--41 percentage points, whereas moving
the existing states has little effect. Keeping object dependencies and
unchanged revision passages prevents many losses. Recognition is a
separate challenge. Learned readouts recover distinctions missed by
fixed cache similarities on synthetic record pairs. On natural text,
text-detector-selected masks show no clear advantage over random masks
at the same rate in 14 same-model detector--generator comparisons.
Query-dependent access can avoid some losses, with additional storage
or access costs. These findings identify what must be preserved beyond
the replaced value when using a KV cache as updatable memory.
\end{abstract}

\section{Introduction}
\label{sec:introduction}

A language model's memory must change when the world or a user's
instructions change. A new deadline should govern current plans, while
the previous deadline may still answer a question about an earlier
agreement. Updating memory therefore requires recognizing what changed
and preserving information that other questions still need. Text-memory
systems address this through retrieval, record revision and temporal
relations \citep{mem02025,amem2025,zep2025}.

Key--value (KV) caches make this problem both practical and subtle.
Serving systems reuse states computed from earlier requests
\citep{promptcache2024,lmcache2025}, and Memory$^3$ retrieves cached
representations of reference texts during inference \citep{memory32024}.
Hiding the positions of an outdated record can avoid rereading the
retained text and rebuilding its cache. Query-aware selection and
learned retention already show that not every cached position must be
read for every question \citep{quest2024,kvp2026,spkv2026}. But deciding
which positions to hide after a factual update requires knowing which
uses of the earlier information remain valid.

Consider \emph{Duration = 12 hours}, followed by \emph{Duration = 18;
use the earlier unit}. The current answer is \emph{18 hours}; the
historical answer is \emph{12 hours}. Hiding the first record removes
access to both the outdated number and the still-needed unit. Moreover,
its influence extends beyond its own positions: later tokens were
processed while it was visible. Recent preprints show that later states
can retain earlier information after the source's own states are edited
or hidden \citep{modelsnotes2026,kveraser2026,eventkv2026}. Recomputing
those states may remove an outdated association, but may also remove a
relation needed to answer a historical question.

\begin{figure}[t]
\centering
\includegraphics[width=\linewidth]{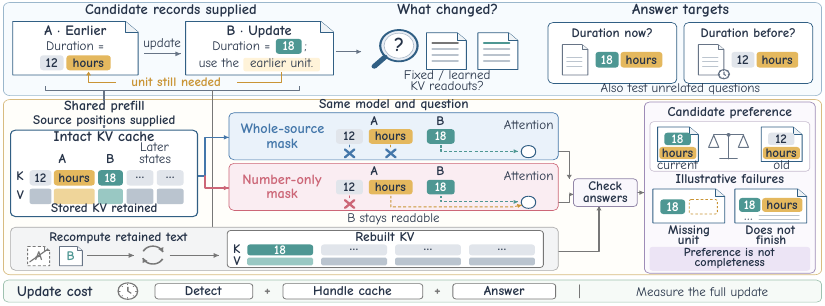}
\caption{Updating a cached history. Whole-source and number-only masks
change access to the same prefilled cache; recomputation rebuilds states
from the retained text. We compare complete current, historical and
unrelated answers, alongside preference between supplied candidates.
Record recognition and the cost of applying a decision are separate
parts of an update.}
\label{fig:overview}
\end{figure}

We study how to recognize a replacement and what happens when it is
applied to a KV cache. Recognition experiments compare fixed cache
similarities, learned readouts and text detectors. Intervention
experiments compare masking whole records, masking replaced values and
recomputing retained text, measuring current, historical and unrelated
answers. We supply record pairs or source positions to study these
decisions independently of retrieval. The comparisons yield four
contributions:
\begin{itemize}
\setlength{\itemsep}{2pt}
\setlength{\parskip}{0pt}
\item \textbf{Preference and completeness can diverge.} In the quantity
task, masking strengthens preference for the new value in all eight
models, but six lose complete answers. Unit-preserving masks repair
these answers; narrative questions show why reference agreement can
also miss lost details (Section~\ref{sec:repair_boundary}).
\item \textbf{Later states preserve useful relations.} On multi-hop
updates, rebuilding states at fixed positions loses 20--41 historical
accuracy points, whereas compacting existing states has little effect.
Object dependencies and unchanged revision passages likewise explain
which older information should remain accessible
(Sections~\ref{sec:later_states}--\ref{sec:main_preservation}).
\item \textbf{Recognition and beneficial intervention are separate.}
Fixed key similarities can match related records while fixed value
scores fail to recognize replacement. Learned readouts recover some
of these distinctions, yet better detection does not establish better
masking decisions (Section~\ref{sec:v45_recognition}).
\item \textbf{Access should be evaluated for the question and reader.}
Query-dependent selection and routing avoid some losses. Repeated-query
timings and direct transfers between related models show why a cheap
cache edit or similar states alone cannot establish efficient,
answer-preserving reuse (Section~\ref{sec:practical}).
\end{itemize}

\section{Tasks, Cache Operations, and Evaluation}
\label{sec:method}
\label{sec:panel}

\paragraph{Records and replacements.}
\label{sec:invalidkv_g_method}
An earlier record A and later record B concern an entity and attribute.
B replaces A's value if it supplies a different current value under the
task's precedence rule. Confirmations, other-entity updates and
other-attribute updates do not replace it. The replacement concerns the
value, leaving the record's other details and historical uses to be
evaluated separately. We supply source positions or derive them from
explicit record relations; detector-selected policies include
recognition errors, but exclude retrieval.

\paragraph{Changing access or recomputing states.}
\label{sec:interventions}
\label{sec:decisive_audit}
Full access, whole-source masking and value masking share a history
prefilled with A and B. Masks set attention logits at selected positions
to negative infinity throughout question processing, candidate scoring
and generation. They leave stored K/V, positions and later states
unchanged. Whole-source masking hides A; value masking hides its old
target string, keeping the unit in the quantity task. Matched controls
hide equally many tokens from a neutral note containing neither answer
nor unit. We also use full access, since control masks can change answers.

Text recomputation deletes selected source text and prefills the retained
history, changing later states, positions and sometimes tokenization.
Section~\ref{sec:later_states} separates state rebuilding at fixed
positions from compaction of existing cache rows. Logical masking frees
no storage. Qwen3.8-27B masks only full-attention layers, retaining
prefilled Gated DeltaNet states; Gemma keeps its native local/global
attention pattern \citep{gateddeltanet2025,qwen38modelcard2026,gemma4modelcard2026}.
Other operations and interfaces are in Appendices~\ref{app:v52_glossary}
and~\ref{app:architecture_audit}.

\paragraph{Questions and complete answers.}
\label{sec:natural_transfer}
The quantity task uses 60 replacement groups and 20 each of
confirmations, other-attribute updates and other-entity updates. Later
records either state the complete quantity or refer to the earlier unit.
Eight models answer current, historical and unrelated questions. A
complete answer has the correct number, unit and format, and ends with
an end-of-sequence (EOS) token. Wrong quantities, omitted units, format
failures and token-limit stops are recorded separately. We use native
chat templates and greedy decoding; generation budgets and sensitivity
analyses are in Appendix~\ref{app:v33_models}.

Public tasks cover multi-hop updates, object locations, article revisions
and accumulated changes. Reference matching checks the supplied answer
or an allowed alias. Source-aware grading instead gives a model grader
the question and original source to check requested details, unsupported
claims and justified abstention. EOS completion and token-limit stops
are measured separately from content. Wikipedia source answerability
is assessed before viewing generated answers; OAKS and FactConsolidation
assess it together with answer quality. Task-specific rules and counts
accompany the results.

\paragraph{Paired measurements.}
For fixed current and old candidates, let
$m(c)=\log p(y_{\mathrm{current}}\mid c)-\log p(y_{\mathrm{old}}\mid c)$.
We measure the change relative to the matched mask:
\begin{equation}
\Delta_m=m(\mathrm{source\ mask})-m(\mathrm{control\ mask}).
\label{eq:margin}
\end{equation}
Scores sum teacher-forced token log probabilities in nats, without length
normalization or EOS. Generated-answer corrections and losses are
measured separately. Paired 95\% bootstrap intervals keep variants,
paraphrases and operations within their base group, case, article or
shared-source component; intervals are pointwise across models.
FactConsolidation's two histories share most facts in the shorter one,
so we report descriptive contrasts without population intervals.
Study sizes are in Appendix~\ref{app:protocol_map}.

\section{Preferring the New Value Does Not Ensure a Complete Answer}
\label{sec:repair_boundary}
\label{sec:score_influence}

The quantity task tests whether a current answer still needs an old
record. When the later record refers to the earlier unit, whole-source
masking blocks that unit along with the outdated number. Candidate
scoring and generation share the history, question and access condition.

\begin{figure}[!htbp]
\centering
\includegraphics[width=\linewidth]{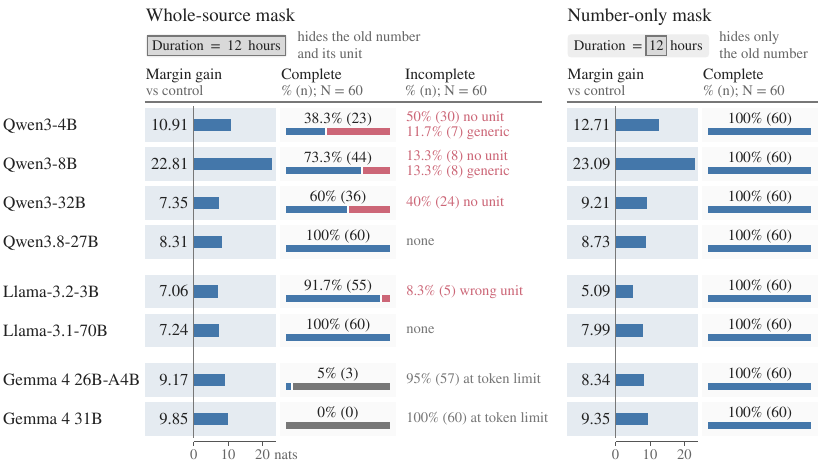}
\caption{Preference and complete answers on the same 60 replacements
whose later record refers to the earlier unit. Whole-source masking
increases the current-minus-old candidate margin in all eight models,
yet six lose complete answers; number-only masking keeps 100\% (60)
in all eight. Rates use 60 questions per model; counts are in parentheses.
Margin changes use each mask's equal-token control, with all paired
95\% group-bootstrap intervals above zero. Full access and source-mask
controls both give 98.3--100\% (59--60) complete answers per model;
the same six lose answers against either. Generic means ``unit(s)''. }
\label{fig:v33_score_generation}
\end{figure}

\paragraph{The margin and the answer can move in different directions.}
Whole-source masking raises the current-versus-old margin in all eight
generators, by 7.06--22.81 nats relative to the control, yet six lose
complete answers relative to full access
(Figure~\ref{fig:v33_score_generation}). The Qwen3 models omit or generalize units, and Llama-3.2-3B
sometimes gives the wrong unit. The Gemmas reach the 256-token limit
in 117 of 120 answers; 103 first give the complete correct quantity
and then continue without a second answer. Qwen3.8-27B keeps 60/60,
and Llama-3.1-70B rises from 59 to 60. These distinguish content errors
from failures to stop.

A larger margin need not mean a more probable correct answer: the
margin also rises when the old answer becomes less probable. All 198
lost answers still favor the current candidate over the old one under
masking. Across the 60 groups, however, the current candidate's mean
log probability falls relative to the matched control in seven of eight
models; the old candidate falls more in all eight.

Can the current answer's probability drop warn of generation failures?
It ranks lost above kept answers with AUC 0.89--0.98 in the four Qwen
and Llama models that lose answers. A threshold chosen on the other
seven models detects 90.5\% (86/95) of losses involving wrong, generic or missing units, but only
1.0\% (1/103) of failures to stop after a complete first answer.
It also flags 41.1\% (115/280) of kept answers. It therefore warns of content losses, but misses stopping failures
and raises false alarms. The held-out-model comparison reuses the same
60 question groups (Appendix~\ref{app:v52_margin_likelihood}).

\paragraph{Preserving the needed detail repairs these current answers.}
Number-only masking leaves the unit visible and preserves 60/60
current answers in every model, as does supplying the unit in the later
record. This comparison uses controls containing no unit information.
Replacing the repeated background note with varied text leaves the
Qwen3-4B and Qwen3-32B losses under whole-source masking, while
number-only masking still gives 60/60
(Appendix~\ref{app:v41_controls}). The result concerns these inputs:
each attribute has one unit, so successful answers under whole-source
masking may also use that cue. In separate quantity constructions,
explicitly requesting a unit repairs many omissions, and Gemma 4 31B
answers every cell correctly. The complete wording and source
comparisons delimit the effect
(Appendices~\ref{app:v17_scope} and~\ref{app:v35_factorial}).

\paragraph{Completeness also matters beyond quantities.}
\label{sec:control_conditions}
Masking improves selection among supplied OAKS narrative answers in
many models (Appendix~\ref{app:v35_oaks}). Removing the
options asks a different question: does the generated answer retain the
details supported by the excerpts? All 1,044 open answers end normally.
On 92 questions judged source-answerable, Qwen3-32B's complete-answer rate
falls from 55.4\% (51) to 45.7\% (42) with masking, $-9.8$ points $[-18.5,-1.1]$.
Reference agreement for the same outputs and grader instead changes by
$+2.2$ points $[-5.4,9.5]$; the paired difference between the two
criteria is $12.0$ points $[2.1,22.1]$. A short reference can omit details that the question requests and the
source supplies, which is why the two criteria can disagree. For
Qwen3-4B and Llama-3.2-3B, the source-based assessment finds no clear
decrease in complete answers (Appendix~\ref{app:v41_oaks_open}). 

\section{Later Cache States Can Preserve Earlier Relations}
\label{sec:later_states}
\label{sec:v45_benefit}
\label{sec:v41_reasoning}

Later cache states are computed while earlier records are visible.
Masking an old record leaves those states intact; deleting it and
recomputing changes them. We test how this distinction affects current
and historical answers.


\newcommand{\MQuakeAnswersTable}{%
\begin{table}[!htbp]
\MainTableStyle
\caption{MQuAKE reference matches as percentages (counts), requiring normalized exact reference or alias agreement and EOS. Current questions have three paraphrases per case.}
\label{tab:main_mquake_answers}
\begin{tabularx}{\linewidth}{C{64pt}Z{.9}Z{.8}Z{1}Z{1.3}Z{.9}Z{.8}Z{1}Z{1.3}}
\toprule
& \multicolumn{4}{c}{\MainTableHead{Current ($N=1,800$)}}
& \multicolumn{4}{c}{\MainTableHead{Historical ($N=600$)}} \\
\cmidrule(lr){2-5}\cmidrule(lr){6-9}
\MainColumnHead{Model} & \FullHeader{Full} & \NumberHeader{Value} & \MainFocusCell{\WholeHeader{Source}} & \MainFocusCell{\FullHeader{Recompute}}
& \FullHeader{Full} & \NumberHeader{Value} & \MainFocusCell{\WholeHeader{Source}} & \MainFocusCell{\FullHeader{Recompute}} \\\midrule
Qwen3-4B & \FullCell{\AnswerPct{81.2}{1,462}} & \AnswerPct{85.7}{1,543} & \MainFocusCell{\AnswerPct{84.2}{1,515}} & \MainFocusCell{\AnswerPct{88.3}{1,590}} & \FullCell{\AnswerPct{88.2}{529}} & \AnswerPct{75}{450} & \MainFocusCell{\AnswerPct{74}{444}} & \MainFocusCell{\AnswerPct{33.5}{201}}\\Llama-3.2-3B & \FullCell{\AnswerPct{44.2}{795}} & \AnswerPct{53.8}{969} & \MainFocusCell{\AnswerPct{63.9}{1,151}} & \MainFocusCell{\AnswerPct{70.6}{1,270}} & \FullCell{\AnswerPct{85.8}{515}} & \AnswerPct{78.8}{473} & \MainFocusCell{\AnswerPct{80.2}{481}} & \MainFocusCell{\AnswerPct{60}{360}}\\Qwen3-32B & \FullCell{\AnswerPct{81.8}{1,473}} & \AnswerPct{85.3}{1,535} & \MainFocusCell{\AnswerPct{83.2}{1,498}} & \MainFocusCell{\AnswerPct{92.8}{1,671}} & \FullCell{\AnswerPct{97}{582}} & \AnswerPct{82.8}{497} & \MainFocusCell{\AnswerPct{82.3}{494}} & \MainFocusCell{\AnswerPct{41.3}{248}}\\
\bottomrule
\end{tabularx}
\par\smallskip{\MainTableSmall\TableNoteAlign
Source hides old records; Value hides their old target strings;
Recompute deletes the records before prefill. Paired comparisons resample
600 cases with paraphrases kept together. All five operations and
intervals: Appendix~\ref{app:v41_stream}.
Shading in Tables~\ref{tab:main_mquake_answers}--\ref{tab:main_preservation}
marks the main comparisons.
\TableNoteEnd}
\end{table}%
}

\newcommand{\MQuakeStatesTable}{%
\begin{table}[!htbp]
\MainTableStyle
\caption{MQuAKE reference matches as percentages (counts) on 600 current
and 600 historical questions. Rebuilding states keeps positions fixed;
compaction preserves their previously computed contents.}
\label{tab:main_mquake_states}
\begin{tabular}{C{62pt}*{6}{C{43pt}}}
\toprule
& \multicolumn{3}{c}{\MainTableHead{Current ($N=600$)}}
& \multicolumn{3}{c}{\MainFocusCell{\MainTableHead{Historical ($N=600$)}}} \\
\cmidrule(lr){2-4}\cmidrule(lr){5-7}
\MainColumnHead{Model}
& \WholeHeader{Source} & \FullHeader{\shortstack{Rebuild\\states}} & \FullHeader{\shortstack{Compact\\rows}}
& \MainFocusCell{\WholeHeader{Source}} & \MainFocusCell{\FullHeader{\shortstack{Rebuild\\states}}} & \MainFocusCell{\FullHeader{\shortstack{Compact\\rows}}} \\
\midrule
Qwen3-4B & \AnswerPct{86.3}{518} & \AnswerPct{93.7}{562} & \AnswerPct{84.8}{509} & \MainFocusCell{\AnswerPct{74}{444}} & \MainFocusCell{\AnswerPct{32.7}{196}} & \MainFocusCell{\AnswerPct{74}{444}} \\
\addlinespace[1.5pt]
Llama-3.2-3B & \AnswerPct{66.3}{398} & \AnswerPct{69.7}{418} & \AnswerPct{66.8}{401} & \MainFocusCell{\AnswerPct{80.2}{481}} & \MainFocusCell{\AnswerPct{60.3}{362}} & \MainFocusCell{\AnswerPct{81.3}{488}} \\
\addlinespace[1.5pt]
Qwen3-32B & \AnswerPct{84.2}{505} & \AnswerPct{95.7}{574} & \AnswerPct{82.5}{495} & \MainFocusCell{\AnswerPct{82.3}{494}} & \MainFocusCell{\AnswerPct{44.8}{269}} & \MainFocusCell{\AnswerPct{82.2}{493}} \\
\bottomrule
\end{tabular}
\par\smallskip{\MainTableSmall\TableNoteAlign
Rebuild recomputes later states at fixed positions. Compact removes masked
rows and repositions later keys while reusing their contents. Current uses
the first paraphrase; paired differences and intervals are in
Appendix~\ref{app:v52_factorial}.
\TableNoteEnd}
\end{table}%
}

\paragraph{One history supports two different questions.}
We adapt 600 MQuAKE cases~\citep{mquake2023} to updates supplied in context,
with 200 each at two, three and four hops. Each history gives earlier
relations, the remaining links needed to answer the questions, and
later updates. Model weights stay fixed. Each case has three paraphrases
of a current question and one historical question; the old records to
mask are supplied (Appendix~\ref{app:v41_stream}).

For example, one hypothetical history says that Abe Sapien was created
by Mike Mignola, who works for Dark Horse Comics. It also says that
Andrew Stanton works for Pixar. An update changes the creator to Andrew
Stanton. ``Who employs the creator of Abe Sapien?'' now requires Pixar;
the same question about the earlier records requires Dark Horse Comics.
Masking the old creator record hides the first link in the historical
chain. The company name remains visible, but the link needed to select
it is no longer directly accessible. Later cache states, however, were
originally computed with that link present.

\paragraph{Masking and recomputation favor different answers.}
Source masking increases current reference matches in all three models
(Table~\ref{tab:main_mquake_answers}). The paired intervals exclude zero
for Qwen3-4B and Llama-3.2-3B, but include zero for Qwen3-32B.
Recomputing the history after deleting the same records improves current
matches by a further 4.2--9.6 percentage points, while losing
20.2--41.0 historical points. Current gains and historical losses need not occur in the same case (Appendix~\ref{app:v52_mquake_retained}).

\MQuakeAnswersTable

Masking only the old value is less consistent. Compared with source
masking, it adds 28 and 37 current matches in the two Qwen models, but
loses 182 in Llama-3.2-3B. Historical matches change little and remain
below full access in all three models. Thus the advantage of number-only
masking in the quantity task does not extend to every kind of update.
Reference/alias matching with EOS cannot measure details omitted
by a multi-part question's reference.

In 520 of the 600 cases, the historical answer string remains somewhere
in the retained text. Almost all of recomputation's net historical loss
occurs in this subset, where it scores 23.5--47.5 points below source
masking. As the creator example shows, retaining the answer name need
not retain the chain connecting it to the question. The masking advantage also persists on questions each model answers
incorrectly without any history, supporting its dependence on the
supplied context (Appendix~\ref{app:v52_weights}).

\paragraph{Does the loss come from rebuilding states or moving tokens?}
Deleting text changes both the later cache states and their positions.
We separate these changes with two operations, each starting from the
source-masked cache. \emph{Rebuild states} recomputes the history from
the first masked record onward, with those records hidden and all
positions fixed. \emph{Compact rows} removes the masked cache rows and
adjusts the rotary positions of later keys, while preserving the content
computed with the old record visible. Both keep the same old records
inaccessible. We test one current question, using the first paraphrase,
and one historical question per case.

\MQuakeStatesTable

Rebuilding states at the original positions lowers historical accuracy
by 19.8--41.3 points. Compacting the existing states changes it by only
$-0.2$ to $+1.2$ points, with all three intervals including zero
(Table~\ref{tab:main_mquake_states}; Appendix~\ref{app:v52_factorial}).
Thus historical loss persists at fixed positions. Rebuilding also
improves current answers in both Qwen models; Llama-3.2-3B's smaller gain has an interval
including zero. In these histories, states computed with the earlier
record present help answer historical questions but can interfere with
current ones.

Rebuilding and then compacting uses different attention distances from
recomputing shortened text, so their effects need not add up
(Appendix~\ref{app:v52_factorial}). Deleting old text can also remove a needed unit: on the 60 unit-dependent Qwen3-32B
cases, it gives 5.0\% (3) complete answers, compared with 60.0\% (36)
under source masking and 100\% (60) when only the old number is masked.
\FloatBarrier

Component edits also differ from removing a fact. Across nine models,
no single key, value or joint-zeroing operation both improves the
targeted margin and meets the accuracy-preservation criterion for
confirmed and unrelated candidate choices in every model (Appendix~\ref{app:v39_core}). In a
separate derived-note task, also hiding the old conclusion raises Llama-3.1-70B's current accuracy
from 92\% (46/50) under source masking to 100\% (50/50);
four other added models are already at ceiling
(Appendix~\ref{app:v39_propagation_extension}).

\section{Retaining Information Needed by Other Questions}
\label{sec:main_preservation}

We test two ways to preserve useful information in earlier records:
keeping movements needed to locate objects and retaining unchanged
passages in revised articles.

\subsection{Keeping Movements Needed for Object Histories}
\label{sec:main_babilong}

BABILong embeds simulated movement and object events in book text
\citep{babilong2024}. If a person drops an object and then moves elsewhere,
the earlier movement still tells us where the object was left. We compare
a mask that hides all but each person's last movement with a rule that
also keeps movements needed to reconstruct object events. The rule uses
only the input events, before seeing a question or reference answer.
A symbolic check confirms that it preserves object trajectories, event
locations and final ownership in all 400 examples across the 8k and 32k
contexts.

At 32k, whole-source masking loses 17--25 correct historical answers per
100 questions, whereas the dependency rule finishes within one to three answers
of full access (Table~\ref{tab:main_preservation}). Equal-token masks of unrelated
book text change these counts by at most three. The rule also removes fewer
records than the source mask. The comparison changes both which
movements are kept and how much text remains available.
For current object locations, the source mask improves all three point estimates
at 32k, although each paired interval includes zero. Retaining information for
historical questions and improving current answers are distinct objectives
(Appendix~\ref{app:v41_babilong}).

\begin{table}[!htbp]
\MainTableStyle
\caption{Preserving answers when earlier records have other uses. Entries give percentages (counts) and
require normal EOS. The two panels use different public tasks and answer criteria.}
\label{tab:main_preservation}
\begin{minipage}[t]{.505\linewidth}
\vspace{0pt}\centering\setlength{\tabcolsep}{2.5pt}
\begin{tabularx}{\linewidth}{C{60pt}YYY}
\toprule
\multicolumn{4}{c}{\MainTableHead{BABILong (32k)}}\\
\multicolumn{4}{c}{\small Historical object locations ($N=100$)}\\
\midrule
\MainColumnHead{Operation}
& \MainColumnHead{\shortstack{Qwen3-\\4B}}
& \MainColumnHead{\shortstack{Llama-\\3.2-3B}}
& \MainColumnHead{\shortstack{Qwen3-\\32B}}\\\midrule
\FullHeader{Full} & \AnswerPctInline{33}{33} & \AnswerPctInline{25}{25} & \AnswerPctInline{40}{40}\\\rowcolor{mainComparison}
\WholeHeader{Whole source} & \AnswerPctInline{16}{16} & \AnswerPctInline{6}{6} & \AnswerPctInline{15}{15}\\\rowcolor{mainComparison}
\FullHeader{\shortstack{Keep\\dependencies}} & \AnswerPctInline{31}{31} & \AnswerPctInline{22}{22} & \AnswerPctInline{39}{39}\\
\bottomrule
\end{tabularx}
\end{minipage}\hfill
\begin{minipage}[t]{.475\linewidth}
\vspace{0pt}\centering\setlength{\tabcolsep}{2.5pt}
\begin{tabularx}{\linewidth}{C{62pt}YY}
\toprule
\multicolumn{3}{c}{\MainTableHead{Wikipedia revisions}}\\
\multicolumn{3}{c}{\small Complete answers ($N=853$)}\\
\midrule
\MainColumnHead{Operation}
& \MainColumnHead{\shortstack{Qwen3.8-\\27B}}
& \MainColumnHead{\shortstack{Gemma 4\\26B-A4B}}\\\midrule
\FullHeader{Full} & \AnswerPctInline{68.3}{583} & \AnswerPctInline{54.9}{468}\\\rowcolor{mainComparison}
\WholeHeader{Whole source} & \AnswerPctInline{14.1}{120} & \AnswerPctInline{10.4}{89}\\\rowcolor{mainComparison}
\NumberHeader{Changed text} & \AnswerPctInline{66.2}{565} & \AnswerPctInline{52.1}{444}\\\FullHeader{Recompute} & \AnswerPctInline{21.3}{182} & \AnswerPctInline{22.2}{189}\\
\bottomrule
\end{tabularx}
\end{minipage}
\par\smallskip{\MainTableSmall\TableNoteAlign
BABILong uses the official location criterion; its dependency rule removes fewer
records. Wikipedia uses source-aware model grades on an independently assessed
answerable subset: 413 historical, 436 change/aggregation and four current
questions. Changed-text masking hides replaced or deleted passages in earlier
revisions; recomputation removes whole non-final bodies. Full protocols and
paired intervals: Appendices~\ref{app:v41_babilong} and
\ref{app:wiki_source_quality}.\TableNoteEnd}
\end{table}

\subsection{Questions about Wikipedia Revision Histories}
\label{sec:v53_wiki}

The Wikipedia-history subset of MINTEval \citep{minteval2026} asks about
earlier versions of articles and the changes between them. Whole-source
masking hides every article body except the final revision. Changed-text
masking hides only passages replaced or deleted between adjacent revisions,
leaving unchanged text accessible. Both masks keep revision headers and
are fixed before the questions arrive. Recomputation deletes the non-final
bodies and processes the remaining history again.

Source review identifies 853 answerable questions among 1,161 from
152 articles. \mbox{GPT-5.6} grades answer completeness against the full
revision histories for all operations
(Appendix~\ref{app:wiki_source_quality}).

Whole-source masking reduces complete answers by 54.3 percentage points for
Qwen and 44.4 for Gemma; changed-text masking reduces them by 2.1 and 2.8
points. The paired 95\% article-bootstrap intervals exclude zero for all four
decreases. Recomputation also remains well below full access
(Table~\ref{tab:main_preservation}). The losses include failures to finish:
whole-source masking raises cap rates from 11.8\% to 40.7\% for Qwen
and from 18.9\% to 64.4\% for Gemma under the same 1,024-token ceiling.
Some outputs therefore lose credit because they continue past the budget,
even when part of the response is correct.

Most answerable questions concern historical states (413) or
changes/aggregation (436); only four concern the current state.
Keeping unchanged passages retains far more of this requested history,
but also keeps more tokens accessible.

\subsection{Why Masking Has Opposite Effects in Longer Fact Histories}
\label{sec:v45_long}

\begin{figure}[!htbp]
\centering
\includegraphics[width=\linewidth]{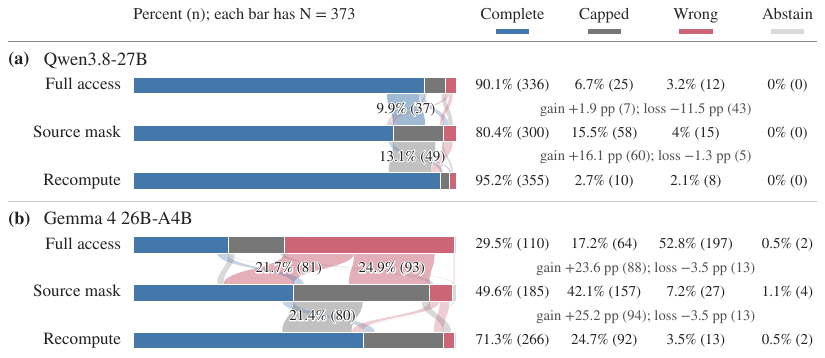}
\caption{Answer outcomes on 373 source-answerable FactConsolidation questions from two overlapping histories. Numbers are percentages (counts), with gain/loss contributions in percentage points. Bands connect the same questions across operations. Capped outputs reach 1,024 tokens and may contain incorrect claims; the other categories require normal completion recorded by an EOS token.}
\label{fig:v45_mab_outcomes}
\end{figure}

FactConsolidation extends MQuAKE to accumulated updates and longer
inputs~\citep{memoryagentbench2025}. We use two overlapping histories
of approximately 39k--80k prompt tokens and 400 questions, of which
GPT-5.6 judges 373 answerable. With the same 1,024-token budget,
source masking lowers Qwen3.8-27B's complete answers with EOS from
90.1\% (336) to 80.4\% (300), but raises Gemma 4 26B-A4B from
29.5\% (110) to 49.6\% (185). Paired transitions explain the opposite signs
(Figure~\ref{fig:v45_mab_outcomes}).

\paragraph{Gemma corrects many outdated answers.}
With full access, 52.8\% (197) of Gemma's answers are wrong despite
ending normally, compared with 3.2\% (12) for Qwen. Of Gemma's wrong
answers, 89.3\% (176/197) give an old-answer alias without a supported
current-answer alias in the final box. This group supplies 87.7\%
(71/81) of Gemma's wrong-to-complete corrections under masking.
For example, a hypothetical update changes Artur Boruc's position
from goalkeeper to midfielder; masking switches Gemma's citation
and answer from the old record to the new one. Its gain largely
corrects outdated answers that Qwen rarely gives at baseline.

\paragraph{Qwen loses multi-hop answers during prolonged search.}
Qwen's multi-hop completeness falls from 80.3\% (139/173) to
61.3\% (106/173), whereas single-hop completeness changes from
98.5\% (197/200) to 97.0\% (194/200). Among initially complete answers,
the loss rate rises from 3.8\% (6/159) at 21--200 generated tokens
to 52.0\% (13/25) above 500 tokens.  Of its 43 lost complete
answers, 86.0\% (37) reach the token limit while repeatedly checking
records. Among these, 73.0\% (27/37) already mention the supported
answer, but the source review labels their responses incorrect or
unsupported. One such output repeatedly attributes Austin to an old fact that
actually says Mexico City. The losses involve unfinished generation and unreliable use of facts (Appendix~\ref{app:v52_fc_truncation}). The opposite
net effects remain when the EOS requirement is removed.

On the 109 questions both models answer completely with full access,
masking preserves 96.3\% (105) for Qwen and 88.1\% (96) for Gemma.
Gemma gains overall while preserving fewer of these same complete answers.

Recomputation raises completeness to 95.2\% (355) for Qwen and
71.3\% (266) for Gemma. Previously capped outputs account for
81.7\% (49/60) and 85.1\% (80/94) of their gains over masking;
Qwen recovers 92.6\% (25/27) of the responses that had mentioned the
right answer without producing a valid final answer. The net effect thus depends on initial errors and subsequent generation (Appendices~\ref{app:v43_mab}
and~\ref{app:v45_mab_analysis}).

\section{Choosing and Applying an Update}
\label{sec:practical}

\subsection{Recognizing a Replacement and Choosing What to Hide}
\label{sec:v45_recognition}

\paragraph{Record similarity does not determine replacement.}
We first ask whether cached representations can identify records about
the same entity and attribute, then whether the later record replaces
the earlier value. On 1,200 role-matched record pairs, fixed key
similarity meets the matching criterion in 20 of 22 model variants,
but fixed value distance gives replacement AUCs of only 0.454--0.493.
These scores use separately encoded records; the two Gemma models are
exceptions to matching. Attention
and lexical relevance to a question also need not identify replacement
(Appendices~\ref{app:paraphrase_robustness} and~\ref{app:signal_details}).

The failed fixed score does not imply that K/V lacks this information.
On a separate test with held-out entities, values and templates,
a classifier using jointly encoded K/V from Qwen3.8-27B reaches
role-discrimination
AUC 0.862 $[0.819,0.913]$, against 0.409 for fixed value distance and
0.514 for the separate-K/V classifier. Llama-3.1-70B instead performs
best with a development-selected head (AUC 0.841); the strongest
readout varies by model. These are predictions from supplied record
pairs, distinct from both retrieval and use during generation
\citep{hewitt2019controls}
(Appendix~\ref{app:v35_role_readouts}).

\paragraph{Better detection need not produce better answers.}
Qwen3.8 and Gemma 4 31B text detectors classify all 300 held-out
synthetic cases correctly, yet even true replacement labels lose
historical answers when used to mask whole records. On 75 natural pairs with model-assessed change labels, adding the
question improves recall of changes in the queried attribute in all
seven detectors but lowers accuracy in six, by labeling more compatible
records as changed
(Appendix~\ref{app:v39_semantic_extension}). On the full 116 natural answer transitions, 14 same-model
detector--generator comparisons show
no clear advantage for detector-selected masks over random selection
at the same masking rate
(Appendix~\ref{app:v39_natural_action}). Calibrating synthetic detection
thresholds on development data gives no positive net current-answer
gain over full access in eight models; optimizing development answer
utility selects no action in
all eight, under conservative ties and high baseline replacement
accuracy (Appendix~\ref{app:v41_calibration}).

\subsection{Other Access Choices Can Avoid Some Losses}
\label{sec:natural_transfer_results}
\label{sec:v8_public}
\label{sec:guarded_results}

We compare query-aware alternatives Quest, FINCH and a reproduction of
RefreshKV \citep{quest2024,finch2024,refreshkv2025}. On 300 controlled
conditions, Qwen3-32B answers 98.3\% (295) with full access, 83.0\%
(249) with source masking, 98.3\% (295) with FINCH and 98.7\% (296)
with RefreshKV, matching initial token budgets to source masking.
Much of the recovery comes from retaining needed records in
confirmations and unrelated updates: on the 150 changed-value
conditions, the respective counts are 150, 149, 150 and 150.

FINCH reads the question before selecting and repositioning rows;
RefreshKV keeps the full cache and can revisit it during decoding.
Matching initial access therefore leaves storage, latency and later
access different. At 512-token capacity on LongMemEval-Oracle, FINCH
raises Qwen3-32B's official-rubric accuracy from 73.2\% (278/380) to
76.1\% (289/380), but falls below full access in the two smaller models
(Appendix~\ref{app:v41_dynamic}).

Question type can also guide cache choice. Offline routing of existing
MQuAKE answers improves accuracy by 4.5--6.6 points over the best single
policy on 3,000 questions per model: questions with the benchmark's
explicit historical prefix use full access, and the others use
recomputation. Deploying this rule requires both caches and temporal
intent. In contrast, recomputing only after a capped FactConsolidation
answer does not beat always recomputing
(Appendix~\ref{app:v52_mquake_retained}).

Access strength matters too. Soft attenuation of older LongMemEval-Oracle
sessions can improve update answers while losing some other answers.
Hard masks on the same spans lower accuracy by 2.9--11.3 points across
three models, with all intervals below zero
(Appendices~\ref{app:guarded_validation} and~\ref{app:v41_public_hard}).

\paragraph{Local edit cost and total request cost differ.}
\label{sec:v45_cost}
With source positions supplied, applying a mask is cheaper than
recomputation with prefix reuse (Appendix~\ref{app:operation_costs}).
But on 30 histories of about 6k tokens, one detected update followed
by ten requests adds 0.47--1.87 seconds over full access across five
models. Per-token decoding stays within 2\%; detection takes
0.61--0.69 seconds, and longer answers drive most extra request time.
Repeated reuse does not remove this overhead when masking leaves
per-request work unchanged. These timings supply source locations and
query types; logical masking frees no storage
(Appendix~\ref{app:v39_cost_extension}).

\paragraph{Cache compatibility must be checked through answers.}
\label{sec:v51_crossversion}
Direct reuse across five related model pairs can help or hurt. On 240
current questions, Qwen3-8B falls from 84.2\% (202) to 76.2\% (183)
on its base variant's cache despite identical first generated tokens.
Qwen3.8-27B instead rises from 72.5\% (174) to 82.9\% (199) on
Qwen3.6-27B's cache, but source masking removes that gain. Similar
states ensure neither stable answers nor stable update effects
(Appendix~\ref{app:v51_crossversion}).

\section{Related Work}
\label{sec:related_work}

AToKe studies historical knowledge after parameter edits, while
RippleEdits tests their consequences for related facts
\citep{atoke2024,rippleedits2024}. Text-memory systems such as Zep retain
temporal relations and superseded records \citep{zep2025}.
We examine these remaining uses of a history encoded in a KV cache.

Context dependence also matters for cache reuse. CacheBlend and EPIC
selectively recompute reused chunks; KVLink trains link tokens between
independently cached segments \citep{cacheblend2025,epic2025,kvlink2025}.
\emph{Models Take Notes at Prefill} reports later states retaining
conclusions after an earlier field's own states are edited in its tested
models \citep{modelsnotes2026}. We examine the consequences for
current and historical answers.

Recent cache-editing preprints address related questions. KVEraser
learns local steering states to approximate behavior after removing a
supplied prompt span \citep{kveraser2026}. Event-KV studies information
remaining downstream after omitting its source and reports that appended
corrections can interfere with historical queries \citep{eventkv2026}.
Our comparisons distinguish masking from recomputation, measuring
answer completeness and termination alongside candidate preference.

FINCH and RefreshKV select context without identifying replacement
relations \citep{finch2024,refreshkv2025}. DroidSpeak and Activated LoRA
support reuse across model variants \citep{droidspeak2026,alora2025};
our direct transfers omit their compatibility mechanisms.
Appendix~\ref{app:extended_related_work} discusses further approaches.

\section{Discussion and Conclusion}
\label{sec:conclusion}

Updating a fact requires preserving other uses of the old information.
Our comparisons separate recognizing replacement, choosing what remains
accessible, and reusing or rebuilding later states. Keeping units,
object dependencies and unchanged passages avoids many losses; long
histories show that correcting outdated answers can also disrupt
previously complete ones.

Fixed keys match related records, and learned readouts and text
detectors recognize replacements in the tested settings, but their
predictions do not reliably select beneficial masks
(Section~\ref{sec:v45_recognition}). Query-dependent selection and
offline routing offer improvements with additional access, storage or
computation (Section~\ref{sec:practical}). Retrieving candidate records
from a larger memory remains outside these supplied-pair experiments.
Model grading, format and generation budget also affect completeness;
FactConsolidation has two overlapping histories and Wikipedia chiefly
tests past states and changes. A useful updater must preserve the
answers still supported by the history while adopting the new fact.

\FloatBarrier
\clearpage
\setlength{\parfillskip}{0pt plus .30\textwidth}
\section*{Reproducibility Statement}

The appendices specify the datasets, model configurations, cache operations,
grading protocols and resampling procedures. An anonymized code supplement
provides the core implementations, selected task inputs, generated answers
and grading labels, with offline scripts for recomputing the main
comparisons. Public datasets and model weights must be obtained separately.

\section*{AI Use Statement}
AI assisted with literature research, language polishing, and code debugging.

\begingroup
\setlength{\parfillskip}{0pt plus 1fil}
\setlength{\emergencystretch}{1em}
\bibliography{References}
\bibliographystyle{iclr2027_conference}
\endgroup

\clearpage
\appendix
\makeatletter
\colorlet{tableWhole}{figureBlue!10!white}
\colorlet{tableNumber}{figureRose!10!white}
\newcommand{\TableNoFill}{\global\let\CT@cell@color\relax}
\newlength{\TableSeam}\setlength{\TableSeam}{1.2pt}
\gdef\TableLastTint{}
\newcommand{\TableTint}[1]{\def\TableThisTint{#1}%
  \ifx\TableThisTint\TableLastTint
    \gdef\CT@cell@color{\CT@color{#1}\global\let\CT@cell@color\relax}%
  \else
    \gdef\CT@cell@color{\CT@color{#1}\advance\@tempdimb-\TableSeam
      \global\let\CT@cell@color\relax}%
  \fi
  \global\let\TableLastTint\TableThisTint}
\renewcommand{\WholeCell}[1]{\TableTint{tableWhole}#1}
\renewcommand{\NumberCell}[1]{\TableTint{tableNumber}#1}
\renewcommand{\TableHead}[1]{\TableNoFill\textbf{#1}}
\renewcommand{\TableLabel}[1]{\textbf{#1}}
\renewcommand{\WholeHeader}[1]{\TableNoFill\textcolor{figureBlue}{\textbf{#1}}}
\renewcommand{\NumberHeader}[1]{\TableNoFill\textcolor{figureRose}{\textbf{#1}}}
\renewcommand{\FullHeader}[1]{\TableNoFill\textcolor{figureGray}{\textbf{#1}}}
\renewcommand{\TableStub}[1]{#1}
\renewcommand{\FullCell}[1]{#1}
\renewcommand{\neutralcell}[1]{#1}
\renewcommand{\passcell}[1]{#1}
\renewcommand{\riskcell}[1]{#1}
\let\TableOrigShortstack\shortstack
\newcommand{\TableLeftStacks}{\def\shortstack{\@ifnextchar[{\TableOrigShortstack}{\TableOrigShortstack[c]}}}
\renewcommand{\TableGroupRow}[2]{\multicolumn{#1}{c}{\TableLeftStacks\textbf{#2}}\\}
\newcommand{\TableSecondarySize}{%
  \fontsize{\strip@pt\dimexpr\ifdim\f@size pt>7.8pt 0.9\dimexpr\f@size pt\relax\else
    \ifdim\f@size pt>7pt 7pt\else\f@size pt\fi\fi\relax}{\f@baselineskip}\selectfont}
\renewcommand{\TableSecondary}[1]{%
  \ifmmode\text{\color{figureGray}\TableSecondarySize$#1$}\else{\color{figureGray}\TableSecondarySize#1}\fi}
\makeatother

\setcounter{secnumdepth}{3}
\makeatletter
\let\addcontentsline\iclr@addcontentsline
\makeatother
\setcounter{tocdepth}{2}
\begingroup
\renewcommand{\contentsname}{Appendix Contents}
\tableofcontents
\endgroup
\clearpage
\renewcommand{\topfraction}{0.95}
\renewcommand{\bottomfraction}{0.95}
\renewcommand{\textfraction}{0.05}
\renewcommand{\floatpagefraction}{0.85}
\setcounter{topnumber}{8}
\setcounter{bottomnumber}{8}
\setcounter{totalnumber}{12}
\makeatletter
\setlength{\@fptop}{0pt}
\setlength{\@fpsep}{9pt}
\setlength{\@fpbot}{0pt plus 1fil}
\makeatother
\setlength{\textfloatsep}{6pt plus 1pt minus 1pt}
\setlength{\intextsep}{6pt plus 1pt minus 1pt}
\setlength{\floatsep}{6pt plus 1pt minus 1pt}
\section{Effects of Masking on Current and Historical Answers}
\label{app:details}

\paragraph{Guide to the supporting studies.}
\label{app:guide}
The appendices collect the protocols, comparisons and sensitivity
analyses behind the main-text studies of answer preservation, later
states and access choices. Recognition, cost and checkpoint reuse
provide additional comparisons. Appendix~\ref{app:protocol_map} summarizes the
inputs and comparison units; Appendix~\ref{app:architecture_audit} specifies
the model interfaces. Appendix~\ref{app:v52_glossary} defines the terms
used throughout.

\paragraph{When does masking help?}
Appendix~\ref{app:details} supports Section~\ref{sec:v45_benefit} with
contextual updates (Appendix~\ref{app:v41_stream}) and accumulated fact
histories (Appendix~\ref{app:v43_mab}); Appendix~\ref{app:v45_mab_analysis}
separates corrections, losses and unfinished answers. Retained text on
contextual MQuAKE (Appendix~\ref{app:v52_mquake_retained}) and the
content of unfinished outputs (Appendix~\ref{app:v52_fc_truncation})
qualify these gains. Controlled histories
and source comparisons (Appendices~\ref{app:panel}--\ref{app:oaks_transfer})
measure candidate preferences. Their full-access baselines
(Appendix~\ref{app:absolute_accuracy}), prompt-format comparison
(Appendix~\ref{app:native_format}) and model sensitivities
(Appendix~\ref{app:v35_scoring}) qualify how those scores should be read.
Object dependencies provide a different test of which old information
remains useful (Appendix~\ref{app:v41_babilong}).

\paragraph{Can the system recognize what changed?}
Appendix~\ref{app:recognition_studies} supports Section~\ref{sec:v45_recognition}.
Its fixed-score comparison appears in
Appendices~\ref{app:paraphrase_robustness} and~\ref{app:v35_fixed_scoring}.
The surrounding studies distinguish
replacement from question relevance (Appendix~\ref{app:signal_details})
and show why a perfect surface baseline limits a comparison
(Appendix~\ref{app:v39_native_pairs}). The definitions, disjoint data split,
and text detectors are in Appendices~\ref{app:exact_definitions},
\ref{app:r570_recognition} and~\ref{app:v8_recognition}.
Learned readouts and explicit pairwise features test alternatives to fixed
distances (Appendices~\ref{app:v35_representation} and~\ref{app:v17_pairs});
calibration tests whether detection scores select useful masks
(Appendix~\ref{app:v41_calibration}). Natural-text detection
(Appendix~\ref{app:v5_natural}) is connected to masking outcomes in
Appendices~\ref{app:natural_join} and~\ref{app:v39_natural_action}.
Query routing and source-label sensitivity test different reasons why a
correct relation label may not predict a helpful action
(Appendices~\ref{app:v39_semantic_extension} and~\ref{app:v17_sources}).
Appendix~\ref{app:v41_complementary} summarizes the distinct effects of
task construction, readout choice and question conditioning.

\paragraph{What must an update preserve?}
Appendix~\ref{app:operation_studies} supports Section~\ref{sec:repair_boundary}.
The principal quantity comparison uses 120 groups with six measurement
units (Appendix~\ref{app:v18_unified}); Appendix~\ref{app:v33_models}
reports all eight models. Appendix~\ref{app:v52_margin_likelihood}
compares the candidate margin with the current answer's own likelihood.
Open OAKS answers test detail preservation
beyond quantities (Appendix~\ref{app:v41_oaks_open}). Source information,
wording and unrelated controls are varied in
Appendices~\ref{app:v17_scope}, \ref{app:v35_generation} and~\ref{app:v41_controls}.
The separate 300-condition record experiment supplies supporting
protocol, unit-error and paired score/generation analyses
(Appendices~\ref{app:revision_protocol}, \ref{app:v8_units} and~\ref{app:v7_bridge}).
Component interventions test whether changed scores preserve answers
(Appendices~\ref{app:component_attribution} and~\ref{app:v39_core_extension}).
Derived notes and state recomputation probe what remains after editing
the original source (Appendices~\ref{app:v5_propagation}
and~\ref{app:v39_propagation_extension}).

\paragraph{Does the complete procedure save computation?}
Appendix~\ref{app:public_studies} supports Section~\ref{sec:v45_cost} with
operation costs
(Appendix~\ref{app:operation_costs}) and repeated queries using a text
detector (Appendices~\ref{app:v5_serial} and~\ref{app:v39_cost_extension}).
Conversational-memory inputs, soft attenuation and answer evaluation are
specified in Appendices~\ref{app:v8_public}, \ref{app:guarded_validation}
and~\ref{app:v17_judge}. Matching the selected spans for hard masks
(Appendix~\ref{app:v41_public_hard}) and comparing query-aware selection
(Appendix~\ref{app:v41_dynamic}) connect access choices to both answer
quality and cost.

\paragraph{Can another checkpoint read the cache?}
Appendix~\ref{app:v51_crossversion} supports Section~\ref{sec:v51_crossversion}.
It specifies the checkpoint pairs and the tokenization checks, compares
answers read from each checkpoint's own cache and its sibling's, and
reports key and value similarity and readout transfer between the paired
checkpoints.

\paragraph{Wikipedia histories and related approaches.}
Appendix~\ref{app:wiki_source_quality} extends the preservation comparison
to questions about changes and past states in Wikipedia revisions. It
separates independently assessed source answerability from generated-answer
quality and reports the article exclusion and paired comparisons.
Appendix~\ref{app:extended_related_work} expands the connections to
temporal knowledge editing, representation readouts, cache reuse and
recent preprints discussed in Section~\ref{sec:related_work}.

\subsection{Terms}
\label{app:v52_glossary}

Each term below keeps one meaning throughout the paper. Unless
recomputation is stated, compared conditions share the same prefilled
history.

\begin{description}
\setlength{\itemsep}{1pt}
\item[Full access.] The unmodified prefilled cache, in which every record
remains visible. Keep-all conditions in the earlier record comparisons are
full access.
\item[Retain.] Keep the cache unchanged when an update arrives; this is the
cost baseline in Table~\ref{tab:v39_operations}, and it answers with full
access.
\item[Whole-source mask (source mask).] Attention logits to every position of the old
record are set to negative infinity during question processing, candidate
scoring and every generated token. History K/V, positions and later
prefilled states are unchanged.
\item[Number-only mask.] Hides only the old number and leaves its unit
visible; tokenizer offsets verify that unit tokens are excluded. On the
stream benchmarks and FactConsolidation, the corresponding value mask
hides old target strings, which need not be numbers.
\item[Matched control.] Hides the same number of tokens from the start of
an inserted neutral note that contains neither the answer nor its unit.
Each mask has its own equal-token control, and controls are also compared
with full access.
\item[Physical removal.] Discards the selected cache rows. Appendix~\ref{app:v52_factorial}
compares answers after compaction, which also repositions later keys;
the cost study separately times physical removal.
\item[Text recomputation (text recompute, text deletion).] Deletes the selected source text
and prefills the retained history again. Full recompute re-encodes all
retained text; stored-prefix recompute reuses the states before the
removed span.
\item[Online mask.] Appends records in order and masks a superseded record
once its replacement has been processed, so later states no longer attend
to it.
\item[Selective refresh.] Reuses the prefix through the old record A and
recomputes every later state with A hidden. All original text, including
derived notes, is kept.
\item[Detector-selected recompute.] Re-encodes the history without the
source selected by a text detector before the questions are answered.
\item[Routing.] Chooses a cache by question type. In the cost study,
supplied historical and unrelated question types read full access and
current questions read the masked cache. The offline MQuAKE analysis
(Appendix~\ref{app:v52_mquake_retained}) sends questions with the
historical prefix to full access and all others to the recomputed cache.
\item[Source-answerable.] A question that the grader judges answerable
from its supplied history or excerpt: 373 of 400 FactConsolidation
questions and 92 of 116 OAKS questions. These source judgments were made
together with the answer grades.
\item[Asserted-value detector.] A text detector with a 128-token limit
that returns the fields \texttt{same\_slot}, \texttt{old\_current} and
\texttt{later\_current}; it predicts a replacement when
\texttt{same\_slot} is true and the two values are nonempty and differ.
\item[Derived note.] A later note C that repeats information from the old
record A.
\item[Templateless prompt; unwrapped answer.] ``Bare text'' has two
meanings, which the paper keeps apart. A templateless prompt omits the
chat template, as in the 22-checkpoint record scoring. An unwrapped answer
is an answer given without its JSON wrapper, as in the cross-checkpoint
comparison.
\item[Complete answer.] In the quantity tasks, the correct number, unit
and format, ending with EOS. Wrong quantities, missing units, format
failures and token-limit termination are counted separately. On
FactConsolidation and OAKS, completeness is judged against the
source-supported details (Appendices~\ref{app:v43_mab}
and~\ref{app:v41_oaks_open}).
\item[Changed answer.] In the cross-checkpoint comparison, a current
answer whose content differs from the reader's own read, ignoring JSON
whitespace and code fences.
\item[Candidate margin.] $m=\log p(y_{\mathrm{current}})-\log
p(y_{\mathrm{old}})$ for fixed candidate strings, in nats summed over
candidate tokens. Equation~\eqref{eq:margin} defines its change $\Delta_m$
relative to the matched control.
\end{description}

\subsection{Experimental Inputs and Interventions}
\label{app:protocol_map}

\begin{table}[htbp]
\TableStyle
\caption{Study sizes and comparison units. $n$ counts the listed units, not replicated model outputs. Text or question variants stay together during resampling; source sharing requires the grouped analyses specified for each study. The public rows use the original remaining partition.}
\label{tab:protocol_map}
{\setlength{\tabcolsep}{\dimexpr\tabcolsep*6/8\relax}
\begin{tabular}{>{\centering\arraybackslash}m{.245\linewidth}>{\centering\arraybackslash}m{.155\linewidth}C{.065\linewidth}>{\centering\arraybackslash}m{.455\linewidth}}
\toprule
\TableHead{Study} & \TableHead{Unit} & \TableHead{$n$} & \TableHead{Compared conditions} \\
\midrule
Candidate scores & Event & 320 & Source mask; matched control \\
Synthetic records & Base group & 50 & Source mask; matched control \\
Quantity factorial & Base group & 10 & Whole/number masks; controls \\
Different units & Base group & 120 & Two later-record versions; full and masked access \\
Detector calibration & Base group & 120 & Fixed threshold; fitted on 18 development groups \\
OAKS narratives & Question & 116 & Option-ID and open answers; source-complete subset: 92 \\
MINTEval state & Source component & 83 & 99 histories, 5,656 questions; five operations \\
FactConsolidation & Length tier & 2 & 400 questions; overlapping facts; 396-question sensitivity \\
MQuAKE updates & Case & 600 & 2--4 hops; 3,000 questions; five operations \\
BABILong qa2/qa3 & Source story & 78, 80 & 100 questions/task/length; 8k and 32k; seven policies \\
Dynamic cache methods & Base group & 50 & 300 controlled questions; fixed and matched capacities \\
Public memory & Question group & 365 & 380 questions; soft/hard and dynamic protocols separately \\
Repeated-query cost & History & 30 & Masking; recomputation; full access \\
\bottomrule
\end{tabular}}
\end{table}

Source masks applied after caching the history block subsequent attention
to selected positions. The history is prefilled with full access and then cloned for
each mask condition. Controlled continuation experiments separately specify
whether masking occurs before or after question and continuation processing
(Appendix~\ref{app:runtime}). Changing B, the neutral note, or the answer
requirements creates a new text condition; only conditions with identical
text share the same prefilled cache. All interventions leave pretrained
model weights unchanged. Reference answers are used only for evaluation
unless explicitly listed below as inputs to a selection signal.

\paragraph{Inputs to source-influence and representation analyses.}
The matched source-influence experiment receives source and unrelated
spans, together with two answer candidates. Query processing and candidate
scoring use masks over the same prefilled history. The measured outcomes
are summed log probabilities and candidate choices, not open-answer
performance or speed. Fixed representation scores encode each record
separately at position zero and pool its states before computing a distance.
Representation capture and score computation have separate costs. Attention
scores require the attention computation; signed leave-one-out scores also
require reference and alternative answers plus additional masked scoring.

\paragraph{Synthetic records and factorial comparisons.}
The intervention study on the synthetic record dataset prefills the history before the question and
uses either predicted or supplied replacement labels. Some policies also
receive the entity--attribute identity or query role. We compare masking
with edited-text recomputation on three query types, measuring exact-match
answers, output format, answer preservation, and intervention time.
The joint scoring-and-generation experiment adds the same neutral note
16 times, prefills the modified history, and compares three mask conditions
using canonical answer scores and EOS-stopped JSON generation. The factorial
experiment varies the information in B, the question, and the note for ten
quantity groups and ten endpoint groups. Each text condition has its own
prefill and five access conditions. The outcomes comprise exact
scores, magnitude and unit correctness, UNKNOWN responses, format validity,
and paired contrasts.

\paragraph{Multi-unit and narrative experiments.}
The multi-unit experiment uses new entity--attribute pairs, supplied source
spans, and explicit units. Reference answers are evaluation inputs. Across
120 test groups, five mask conditions and three question types measure
exact matches, complete quantities under scoring that accepts equivalent unit expressions, answer
preservation, and net changes from selecting masks. The natural-text
experiment instead receives short source excerpts, a question, and two
candidates. We prefill the history using each model's chat template and share that
cache across full access, earlier-source masking, and control masking. The outputs for 116 questions
from five books are scored as A, B, or UNKNOWN. Generated answers are not
available for the constructed negative pairs.

\paragraph{Recomputation and public-workload policies.}
The derived-note experiments receive the source, replacement, and
derived-note positions. They compare masking those positions with
recomputing later states, and report candidate scores, generated answers,
and operation cost (Appendix~\ref{app:v5_propagation}). In the public
workload, the rule for selectively reducing attention uses session order
and question words, without entity--attribute or reference labels. It
retains all KV and processes each question and complete answer
independently, measuring judged correctness and answer latency both
excluding and including initial history preparation. The hard-mask
comparison uses the same selected spans. Query-aware cache methods have
separate schedules for processing the question and selecting the first
generated token, specified in Appendix~\ref{app:v41_dynamic}.

\paragraph{Serial queries.}
The repeated-query experiment is given source spans and query types.
Its timing includes running the detector once when the update arrives. Each sequence contains one update
followed by questions over an unchanged history; previous answers are not
appended. The measured sequences cover 30 histories, up to ten questions,
and up to 6.4k tokens in the 40-token generation comparison. The
256-token comparison reports its history lengths and timings separately
(Appendix~\ref{app:v39_cost_extension}).

\paragraph{Repeated updates and query-aware comparisons.}
The repeated-update histories, contextual multi-hop edits and longer object
reasoning tasks are specified in Appendices~\ref{app:v41_stream} and
\ref{app:v41_babilong}. The online policy changes attention after each update;
text recomputation rebuilds the retained context, unlike a post-prefill mask.
Open OAKS removes the options while keeping the original questions and
references (Appendix~\ref{app:v41_oaks_open}). The matched-span public
comparison shares target positions between hard and soft operations
(Appendix~\ref{app:v41_public_hard}); the dynamic-method comparison uses its
own question/first-token schedules and paired baselines
(Appendix~\ref{app:v41_dynamic}). These are distinct protocols, not additional
rows pooled into a common public score.

\paragraph{Interpretation shared across experiments.}
A logical mask blocks selected attention reads but does not free KV memory
or remove information already incorporated into later states. Zeroing K,
zeroing V, selecting cache rows, and deleting text change different
computations. In particular, K zeroing leaves the position in the attention
normalization. Answer preservation must be compared with both full access
and the equal-token control. Complete-answer correctness, value identity,
and output-format validity are distinct measures. We report pooled output
counts together with the numbers of base groups or questions.

\subsection{Model Architectures and Inference Protocols}

\label{app:architecture_audit}

Architecture measurements use revision-specific configuration files and
runtime cache shapes for 19 of the 22 candidate-scoring models; Qwen3.8-27B
and the two Gemma 4 checkpoints are listed in
Table~\ref{tab:v33_model_inventory}. The 19 contain 17 GQA, one MHA,
and one MLA checkpoint, with 17 dense and two routed-expert models
(Table~\ref{tab:model-architecture-inventory}). Standard-attention entries
follow the checkpoint configurations. In the table, $L$ is the layer count,
$d_{\mathrm{model}}$ the hidden width, Q:KV the query-to-KV head counts, and
$d_K/d_V$ the key/value head dimensions. For DeepSeek-V2-Lite, the
Transformers~5.12 eager runtime exposes 16 key heads of dimension 192 and 16 value heads of
dimension 128 in each of 27 layers, giving 138,240 K/V elements per token.
This is the exposed tensor interface; a specialized MLA inference kernel
can retain a different, compressed latent representation.

The number of runtime-exposed K/V elements per token is
$S_{\mathrm{KV}}=\sum_{\ell}h^{\mathrm{KV}}_{\ell}(d^K_{\ell}+d^V_{\ell})$; ordinary symmetric
attention gives $2Lh^{\mathrm{KV}}d_{\mathrm{head}}$. This counts exposed tensor elements rather than allocated memory.
Architecture, family, size, training, and post-training vary together.
These comparisons describe checkpoint differences and do not isolate the
effect of parameter count or cache width.

Dense checkpoints report independent parameters, counting tied input/output
embedding weights once; sparse checkpoints report total/active parameters.
Qwen3-0.6B and Qwen3-1.7B tie these weights in the evaluated revisions.
Counting the tied weights once gives 596,049,920 and 1,720,574,976
independent parameters, respectively.

\begin{table}[!htbp]
\WideTableStyle
\caption{Architectures of 19 of the 22 candidate-scoring models and the additional Mistral Small 24B checkpoint.}
\label{tab:model-architecture-inventory}

{\setlength{\tabcolsep}{\dimexpr\tabcolsep*14/16\relax}
\begin{tabular}{>{\centering\arraybackslash}m{0.22\linewidth}C{0.115\linewidth}C{0.035\linewidth}C{0.07\linewidth}C{0.09\linewidth}C{0.095\linewidth}C{0.15\linewidth}>{\centering\arraybackslash}m{0.085\linewidth}}
\toprule
\TableHead{Model} & \TableHead{\shortstack{Total/active\\(B)}} & \TableHead{$L$} & \TableHead{$d_{\mathrm{model}}$} & \TableHead{\shortstack{Q:KV\\or MLA}} & \TableHead{$d_K/d_V$} & \TableHead{\shortstack{$S_{\mathrm{KV}}$\\(elem./token)}} & \TableHead{FFN} \\
\midrule
Qwen2.5-0.5B & 0.49 & 24 & 896 & $14:2$ & 64/64 & $6\,144$ & Dense \\
Qwen3-0.6B & 0.60 & 28 & $1\,024$ & $16:8$ & 128/128 & $57\,344$ & Dense \\
Llama-3.2-1B & 1.24 & 16 & $2\,048$ & $32:8$ & 64/64 & $16\,384$ & Dense \\
Qwen2.5-1.5B & 1.54 & 28 & $1\,536$ & $12:2$ & 128/128 & $14\,336$ & Dense \\
Qwen3-1.7B & 1.72 & 28 & $2\,048$ & $16:8$ & 128/128 & $57\,344$ & Dense \\
Qwen2.5-3B & 3.09 & 36 & $2\,048$ & $16:2$ & 128/128 & $18\,432$ & Dense \\
Llama-3.2-3B & 3.21 & 28 & $3\,072$ & $24:8$ & 128/128 & $57\,344$ & Dense \\
\bottomrule
\end{tabular}}
\par\smallskip{\footnotesize\TableNoteAlign\textsuperscript{$\dagger$}Additional checkpoint. \textsuperscript{$\ddagger$}Runtime-exposed MLA tensors.\TableNoteEnd}
\end{table}

Mistral Small 24B is additionally evaluated on the same 1,200
recognition pairs, 320 causal events, and 336 OAKS events. It is excluded
from these 19-model counts, and its results are not reported in this paper. The checkpoint has 81,920 exposed K/V elements
per token, versus 65,536 for Mistral 7B. Parameter count, architecture,
and training change together; their comparison does not isolate scale.

\label{app:runtime}
These 19-model comparisons and the Mistral Small 24B evaluation use bfloat16
parameters, eager attention, and raw prompts without provider chat templates
or system messages. The native-chat and generated-answer experiments specify their prompts and decoding rules separately. Model revisions, licenses, and
software versions are provided in the machine-readable supplement;
users obtain model weights separately from their original providers.

For the controlled-history and component experiments, the raw prefix is the exact
concatenation of \texttt{History:\textbackslash n}, the history,
\texttt{\textbackslash nQuestion:\textbackslash n}, the query, and
\texttt{\textbackslash nReasoning:\textbackslash n}.  The model appends exactly
8, 32, or 128 greedy continuation tokens without EOS stopping, followed by
\texttt{\textbackslash nAnswer:\textbackslash n}; the two options are then
scored by unnormalized teacher-forced log-probability sum.  The full prompt uses
the tokenizer's special tokens, while the answer cue and options do not.

The natural-source and role-matched candidate-scoring panels generate no continuation.
Their prefix concatenates \texttt{Earlier source:\textbackslash n\{old\}},
\texttt{\textbackslash nUnrelated source:\textbackslash n\{neutral\}},
\texttt{\textbackslash nLater source:\textbackslash n\{later\}}, and
\texttt{\textbackslash nQuestion: \{query\}\textbackslash nAnswer:}, followed
by scoring the same two answer options. The comparison using records without a question
uses \texttt{\{old\}\textbackslash n\{later\}}. No special tokens are added.
For tokenizers that preserve segment encodings when concatenated, segments
are encoded independently. Mistral prefixes are instead encoded once;
exact decoding is checked and character offsets map source spans to tokens.
The representation analysis encodes old and later records in separate forward
passes, each starting at position zero. Mistral Small 24B uses the
Transformers 5.12 \texttt{fix\_mistral\_regex} compatibility option throughout
its evaluation. Scores compare the resulting K/V representations, rather
than reading a single cache after both observations.

Visibility masking blocks selected attention edges while keeping the cache
resident and its tensors unchanged.  After the condition-specific transition,
downstream query, continuation, answer-cue, and option-scoring positions receive
negative-infinity attention bias to the selected source-token positions through
a four-dimensional additive mask.  Cached tensors, sequence length, and
absolute positions remain unchanged.  The old-span condition masks the supplied
contiguous old-source interval; the control applies the identical operation to
the same number of token positions from a specified unrelated span.  In the
natural-source panel, the old interval covers the stripped source body and its
trailing newline, not the fixed heading.  Late interventions retain computation
produced before the transition.  All-layer $K$, $V$, and $K{+}V$ zeroing instead changes the stored tensors.

Each intervention uses a contiguous token interval, preserves token positions,
and leaves tensors outside the selected span unchanged. Attention-mask
dimensions match the resulting cache. Each treatment starts from a separate
cache copy, preventing changes from carrying between conditions. These
constraints preserve the intended intervention but do not eliminate numerical
differences between cached and one-shot floating-point computation.

\subsection{Repeated Updates and Contextual Multi-Hop Questions}
\label{app:v41_stream}

The stream experiments evaluate Qwen3-4B, Llama-3.2-3B, and Qwen3-32B
with five policies and unchanged model weights. They ask whether a rule
that identifies an old current-state record also preserves other uses of
that record. These are benchmark histories, not unconstrained agent
trajectories or a full long-horizon memory-system leaderboard.

\paragraph{MINTEval state tracking.}
We use all 99 released state-tracking sessions and all 5,656 questions
\citep{minteval2026}. Histories contain 18--148 ordered events. Current
location, historical location, counting, ordering, and multi-hop questions
retain their supplied references. A source rule masks every explicit
person-movement record except that person's latest movement; acquisition,
drop, and transfer events remain. Old destination strings define the
value-only mask. The rule does not inspect the question or gold answer.
A movement obsolete for a person's current location can still locate an
object at an earlier drop or transfer event.

\begin{figure}[!htbp]
\centering
\includegraphics[width=\linewidth]{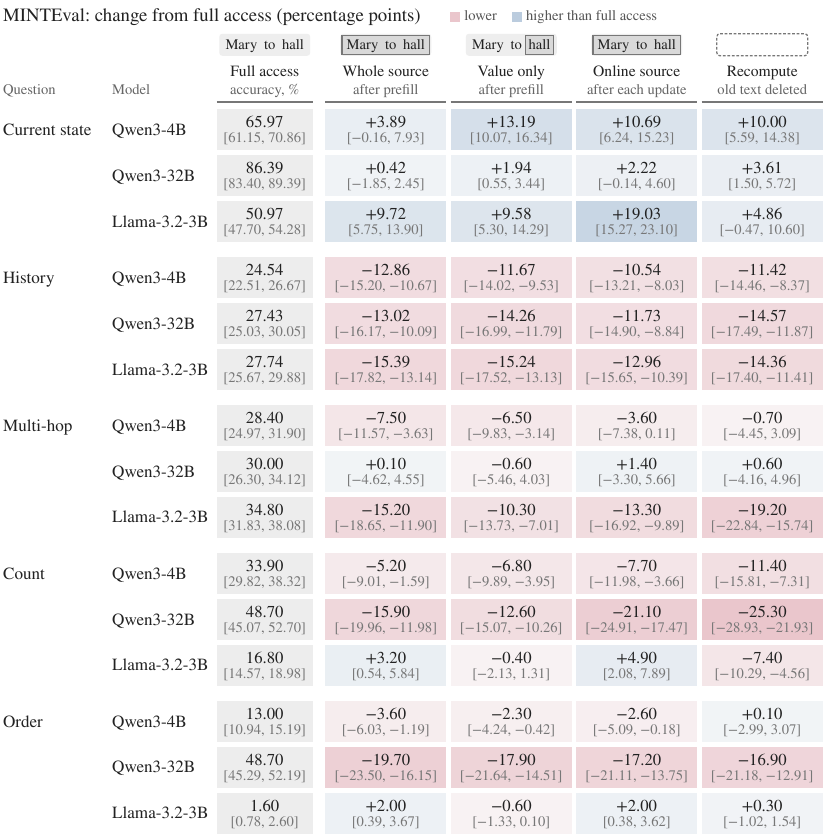}
\caption{MINTEval changes relative to full access across five question types. Cells print the change in normalized reference exact match with EOS (percentage points) and, beneath it, the pointwise 95\% paired interval over 83 shared-source components; rose cells fall below full access, blue cells rise above it. The gray column gives full-access accuracy (\%) with its interval. Each model uses the same 99 histories; value masks hide destination strings, online masks act after each update, and recomputation rebuilds the retained text.}
\label{fig:v41_minteval_operations}
\end{figure}

The 99 histories match original bAbI source stories and form 83 connected
components under shared source identity. Primary uncertainty resamples
these components, keeping every question, wording, and operation together.
The original session-based calculation is retained as a sensitivity.
Under this resampling, whole-source masking lowers all-question accuracy
in all three models: source minus full is $-6.79$ [$-8.35$, $-5.31$]
percentage points for Qwen3-4B, $-5.80$ [$-7.04$, $-4.62$] for
Llama-3.2-3B and $-10.68$ [$-12.89$, $-8.61$] for Qwen3-32B. Only
Llama-3.2-3B's current-location gain, $+9.72$ [$5.75$, $13.90$], has an
interval above zero. History answers fall in all three models (by
12.86, 15.39 and 13.02 points), multi-hop answers fall in Qwen3-4B and Llama-3.2-3B, and count and order
answers fall in both Qwen models but rise in Llama-3.2-3B.
Table~\ref{tab:v41_minteval} gives the counts and
Figure~\ref{fig:v41_minteval_operations} the per-type intervals. The
state subset is short despite its many updates and is not MINTEval's
four-domain, million-token evaluation. BABILong separately tests larger
contexts below.

\paragraph{MQuAKE contextual updates.}
We select 200 cases each at two, three, and four hops from the corrected
MQuAKE-CF-3k-v2 release, using seeds $20260920+h$ for hop count $h$ before
outcome inspection \citep{mquake2023}. Each of the 600 cases retains the three
supplied current multi-hop paraphrases. We add a historical version of
the first, prefixed with ``According to earlier records, before updates:'',
and the first edited relation's supplied single-hop question, giving
3,000 questions. The old and new chains'
unedited links are supplied before later edit records. No edited new value
is prematurely asserted as the old value. The source mask hides earlier
records for edited subject/relation pairs; the value mask hides their
old target strings, which may be multi-token entities rather than numbers.

This adapts MQuAKE to contextual record updates and does not reproduce a
parameter-editing leaderboard. Primary scoring requires normalized exact
match to a supplied answer or alias and normal EOS. Some supplied questions
ask multiple fields while their reference contains only one; for example,
a country-and-language question may have only a language answer. Reference
matching therefore cannot certify every requested detail. Alias presence,
strict closed-box compliance, and success on any of the three current
paraphrases remain separate sensitivities. Paired intervals resample the
600 cases, not 1,800 independent paraphrases; hop count and edit count remain
stratified in the numerical supplement.
Appendix~\ref{app:v52_mquake_retained} separates cases by whether the
historical answer remains in the retained text.

\paragraph{Five access policies.}
Full, source-mask, and value-mask conditions share a full-history prefill
and act during query and decoding. Text recomputation removes the source
spans before rebuilding the history. The online policy appends records in
order and masks a superseded record only after its replacement is
processed, so later states no longer attend to that record. Questions are
asked after the history; this is not an online agent interaction. All
policies use native chat templates, greedy EOS decoding, and a 128-token
answer ceiling. Boxed-answer instructions and the original normalization
are retained, but a correct bare reference answer is not labeled
semantically incomplete merely for lacking a box.

\begin{figure}[!htbp]
\centering
\includegraphics[width=\linewidth]{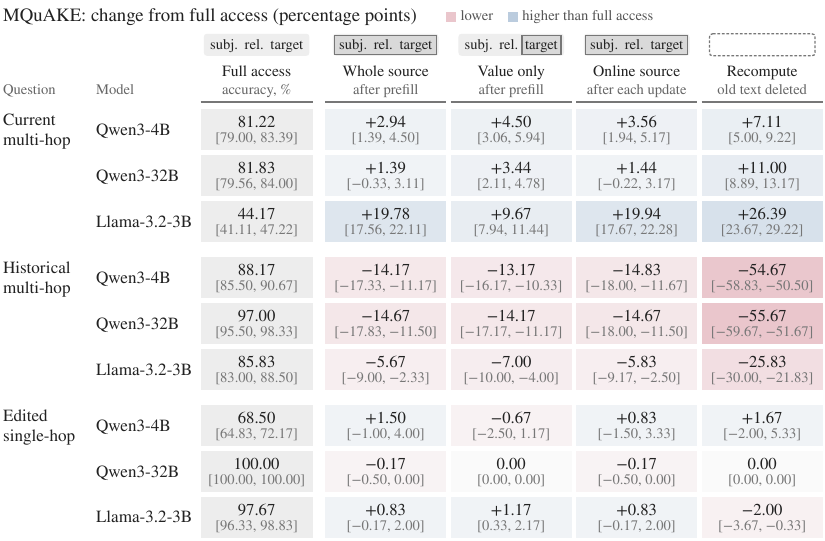}
\caption{Contextual MQuAKE changes relative to full access. Cells print the change in supplied-reference exact match with EOS (percentage points) and, beneath it, the pointwise 95\% interval over 600 cases, retaining all current paraphrases together; rose cells fall below full access, blue cells rise above it. The gray column gives full-access accuracy (\%) with its interval. Current multi-hop, historical multi-hop and edited single-hop questions remain separate; reference matching does not certify every detail in multi-field questions. Value masks hide target strings, which need not be numbers.}
\label{fig:v41_mquake_operations}
\end{figure}

\begin{figure}[!htbp]
\centering
\includegraphics[width=\linewidth]{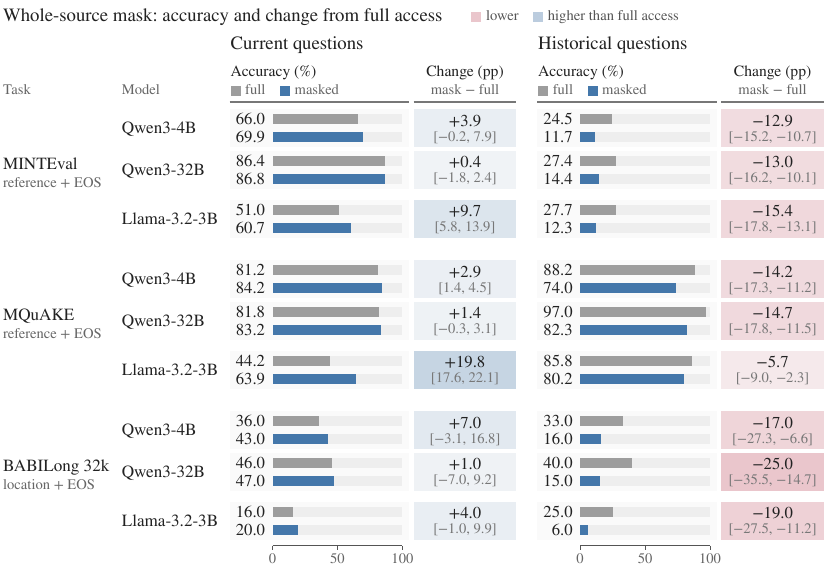}
\caption{Current and historical answers under whole-source masking. Left
block: current questions; right block: historical questions. Paired bars
and numbers give full-access (gray) and masked (blue) accuracy in percent.
Tinted cells give the paired change in points with its pointwise 95\%
interval below, blue for gains and rose for losses, using 83 MINTEval source
components, 600 MQuAKE cases and 78/80 BABILong qa2/qa3 stories; changes
are computed before rounding. Scores use the stated reference or location
metric with EOS. The models and paired questions are fixed.}
\label{fig:v41_reasoning}
\end{figure}

Cache invariance is checked for access-only branches; recomputation and
online history formation intentionally change later states. Each retained
session group includes all five paired policies. The Qwen results combine
complete groups from two H200 execution settings, so aggregate timings are
not interpreted as single-platform performance. Tables~\ref{tab:v41_minteval} and~\ref{tab:v41_mquake} retain all five
policies; Figures~\ref{fig:v41_minteval_operations} and
\ref{fig:v41_mquake_operations} show their paired effects, and
Figure~\ref{fig:v41_reasoning} contrasts current and historical questions
under whole-source masking across the three benchmarks. The numerical
supplement additionally retains termination/format sensitivities,
hop/edit-count strata, and macro-session estimates.

\begin{table}[!htbp]
\WideTableStyle
\caption{MINTEval state tracking over 99 histories. Percentages (counts) use the question-type denominators in the $n$ column. Answers require normalized reference exact match and normal EOS.}
\label{tab:v41_minteval}
{\setlength{\tabcolsep}{\dimexpr\tabcolsep*14/16\relax}
\begin{tabular}{cccccccc}
\toprule
\TableHead{Model} & \TableHead{Query} & \TableHead{$n$} & \FullCell{\FullHeader{Full}} & \WholeCell{\WholeHeader{Source}} & \NumberCell{\NumberHeader{Value}} & \TableHead{Recompute} & \TableHead{Online} \\\midrule
Qwen3-4B & Current state & 720 & \FullCell{\AnswerPct{66}{475}} & \WholeCell{\AnswerPct{69.9}{503}} & \NumberCell{\AnswerPct{79.2}{570}} & \AnswerPct{76}{547} & \AnswerPct{76.7}{552}\\Qwen3-4B & History & 1936 & \FullCell{\AnswerPct{24.5}{475}} & \WholeCell{\AnswerPct{11.7}{226}} & \NumberCell{\AnswerPct{12.9}{249}} & \AnswerPct{13.1}{254} & \AnswerPct{14}{271}\\Qwen3-4B & Count & 1000 & \FullCell{\AnswerPct{33.9}{339}} & \WholeCell{\AnswerPct{28.7}{287}} & \NumberCell{\AnswerPct{27.1}{271}} & \AnswerPct{22.5}{225} & \AnswerPct{26.2}{262}\\Qwen3-4B & Order & 1000 & \FullCell{\AnswerPct{13}{130}} & \WholeCell{\AnswerPct{9.4}{94}} & \NumberCell{\AnswerPct{10.7}{107}} & \AnswerPct{13.1}{131} & \AnswerPct{10.4}{104}\\Qwen3-4B & Multi-hop & 1000 & \FullCell{\AnswerPct{28.4}{284}} & \WholeCell{\AnswerPct{20.9}{209}} & \NumberCell{\AnswerPct{21.9}{219}} & \AnswerPct{27.7}{277} & \AnswerPct{24.8}{248}\\\addlinespace[2pt]
Llama-3.2-3B & Current state & 720 & \FullCell{\AnswerPct{51}{367}} & \WholeCell{\AnswerPct{60.7}{437}} & \NumberCell{\AnswerPct{60.6}{436}} & \AnswerPct{55.8}{402} & \AnswerPct{70}{504}\\Llama-3.2-3B & History & 1936 & \FullCell{\AnswerPct{27.7}{537}} & \WholeCell{\AnswerPct{12.3}{239}} & \NumberCell{\AnswerPct{12.5}{242}} & \AnswerPct{13.4}{259} & \AnswerPct{14.8}{286}\\Llama-3.2-3B & Count & 1000 & \FullCell{\AnswerPct{16.8}{168}} & \WholeCell{\AnswerPct{20}{200}} & \NumberCell{\AnswerPct{16.4}{164}} & \AnswerPct{9.4}{94} & \AnswerPct{21.7}{217}\\Llama-3.2-3B & Order & 1000 & \FullCell{\AnswerPct{1.6}{16}} & \WholeCell{\AnswerPct{3.6}{36}} & \NumberCell{\AnswerPct{1}{10}} & \AnswerPct{1.9}{19} & \AnswerPct{3.6}{36}\\Llama-3.2-3B & Multi-hop & 1000 & \FullCell{\AnswerPct{34.8}{348}} & \WholeCell{\AnswerPct{19.6}{196}} & \NumberCell{\AnswerPct{24.5}{245}} & \AnswerPct{15.6}{156} & \AnswerPct{21.5}{215}\\\addlinespace[2pt]
Qwen3-32B & Current state & 720 & \FullCell{\AnswerPct{86.4}{622}} & \WholeCell{\AnswerPct{86.8}{625}} & \NumberCell{\AnswerPct{88.3}{636}} & \AnswerPct{90}{648} & \AnswerPct{88.6}{638}\\Qwen3-32B & History & 1936 & \FullCell{\AnswerPct{27.4}{531}} & \WholeCell{\AnswerPct{14.4}{279}} & \NumberCell{\AnswerPct{13.2}{255}} & \AnswerPct{12.9}{249} & \AnswerPct{15.7}{304}\\Qwen3-32B & Count & 1000 & \FullCell{\AnswerPct{48.7}{487}} & \WholeCell{\AnswerPct{32.8}{328}} & \NumberCell{\AnswerPct{36.1}{361}} & \AnswerPct{23.4}{234} & \AnswerPct{27.6}{276}\\Qwen3-32B & Order & 1000 & \FullCell{\AnswerPct{48.7}{487}} & \WholeCell{\AnswerPct{29}{290}} & \NumberCell{\AnswerPct{30.8}{308}} & \AnswerPct{31.8}{318} & \AnswerPct{31.5}{315}\\Qwen3-32B & Multi-hop & 1000 & \FullCell{\AnswerPct{30}{300}} & \WholeCell{\AnswerPct{30.1}{301}} & \NumberCell{\AnswerPct{29.4}{294}} & \AnswerPct{30.6}{306} & \AnswerPct{31.4}{314}\\
\addlinespace[2pt]
\bottomrule
\end{tabular}}
\par\smallskip{\footnotesize\TableNoteAlign Source/Value mask old record spans or their old target strings after full history prefill. Recompute forms the retained text states afresh; Online masks old records after each explicit update. Box-format compliance is a separate sensitivity, not semantic completeness. Paired intervals use 83 shared-source components for MINTEval and 600 cases for MQuAKE; the numerical supplement retains hop/edit strata and all five operations. The equal Full counts for Qwen3-4B (current state and history, 475) and Qwen3-32B (count and order, 487) are coincidences in the recorded outputs.\TableNoteEnd}
\end{table}
\begin{table}[!htbp]
\WideTableStyle
\caption{Outcome entries are percentages (counts), using the row-specific $n$. Contextual MQuAKE updates on 600 cases, with 200 each of two, three, and four hops. Counts require normalized supplied-reference exact match and normal EOS; this is not parameter editing.}
\label{tab:v41_mquake}
{\setlength{\tabcolsep}{\dimexpr\tabcolsep*14/16\relax}
\begin{tabular}{cccccccc}
\toprule
\TableHead{Model} & \TableHead{Query} & \TableHead{$n$} & \FullCell{\FullHeader{Full}} & \WholeCell{\WholeHeader{Source}} & \NumberCell{\NumberHeader{Value}} & \TableHead{Recompute} & \TableHead{Online} \\\midrule
Qwen3-4B & Current multi-hop & 1800 & \FullCell{\AnswerPct{81.2}{1462}} & \WholeCell{\AnswerPct{84.2}{1515}} & \NumberCell{\AnswerPct{85.7}{1543}} & \AnswerPct{88.3}{1590} & \AnswerPct{84.8}{1526}\\Qwen3-4B & Historical multi-hop & 600 & \FullCell{\AnswerPct{88.2}{529}} & \WholeCell{\AnswerPct{74}{444}} & \NumberCell{\AnswerPct{75}{450}} & \AnswerPct{33.5}{201} & \AnswerPct{73.3}{440}\\Qwen3-4B & Edited single-hop & 600 & \FullCell{\AnswerPct{68.5}{411}} & \WholeCell{\AnswerPct{70}{420}} & \NumberCell{\AnswerPct{67.8}{407}} & \AnswerPct{70.2}{421} & \AnswerPct{69.3}{416}\\\addlinespace[2pt]
Llama-3.2-3B & Current multi-hop & 1800 & \FullCell{\AnswerPct{44.2}{795}} & \WholeCell{\AnswerPct{63.9}{1151}} & \NumberCell{\AnswerPct{53.8}{969}} & \AnswerPct{70.6}{1270} & \AnswerPct{64.1}{1154}\\Llama-3.2-3B & Historical multi-hop & 600 & \FullCell{\AnswerPct{85.8}{515}} & \WholeCell{\AnswerPct{80.2}{481}} & \NumberCell{\AnswerPct{78.8}{473}} & \AnswerPct{60}{360} & \AnswerPct{80}{480}\\Llama-3.2-3B & Edited single-hop & 600 & \FullCell{\AnswerPct{97.7}{586}} & \WholeCell{\AnswerPct{98.5}{591}} & \NumberCell{\AnswerPct{98.8}{593}} & \AnswerPct{95.7}{574} & \AnswerPct{98.5}{591}\\\addlinespace[2pt]
Qwen3-32B & Current multi-hop & 1800 & \FullCell{\AnswerPct{81.8}{1473}} & \WholeCell{\AnswerPct{83.2}{1498}} & \NumberCell{\AnswerPct{85.3}{1535}} & \AnswerPct{92.8}{1671} & \AnswerPct{83.3}{1499}\\Qwen3-32B & Historical multi-hop & 600 & \FullCell{\AnswerPct{97}{582}} & \WholeCell{\AnswerPct{82.3}{494}} & \NumberCell{\AnswerPct{82.8}{497}} & \AnswerPct{41.3}{248} & \AnswerPct{82.3}{494}\\Qwen3-32B & Edited single-hop & 600 & \FullCell{\AnswerPct{100}{600}} & \WholeCell{\AnswerPct{99.8}{599}} & \NumberCell{\AnswerPct{100}{600}} & \AnswerPct{100}{600} & \AnswerPct{99.8}{599}\\
\addlinespace[2pt]
\bottomrule
\end{tabular}}
\par\smallskip{\footnotesize\TableNoteAlign Source/Value mask old record spans or their old target strings after full history prefill. Recompute forms the retained text states afresh; Online masks old records after each explicit update. Box-format compliance is a separate sensitivity, not semantic completeness. Paired intervals use 83 shared-source components for MINTEval and 600 cases for MQuAKE; the numerical supplement retains hop/edit strata and all five operations.\TableNoteEnd}
\end{table}

\subsection{Long-Context Object Dependencies}
\label{app:v41_babilong}

BABILong embeds simulated movement and object events in book text
\citep{babilong2024}. We use all 100 published examples in each of qa2 and
qa3 at 8k and 32k, retaining the original prompt bodies, questions, and
targets. All 400 extracted fact sequences, questions, and answers match
the original bAbI \emph{test} release exactly. qa2 asks current object
locations; qa3 requires earlier object locations and their sequence.
Repeated queries share 78 original qa2 stories and 80 original qa3 stories,
which define the bootstrap clusters. Reuse of background book text remains
a further dependence beyond these story groups.

The naive source policy hides all but each person's final movement record.
A more conservative rule also retains movements needed to locate an object
when it is dropped or picked up, and to preserve its historical
path. An input-only symbolic trace verifies unchanged object trajectories,
event locations and final ownership under that rule for all 400 cases.
It is not question/gold-conditioned. The naive and conservative policies
remove different amounts of text, so their contrast evaluates whole
policies rather than isolating the effect of a dependency predictor.
The equal-token neutral mask selects non-fact distractor positions,
excluding recognized movements, object events, and prompt framing.

\begin{table}[!htbp]
\WideTableStyle
\caption{Outcome entries are percentages (counts). BABILong object reasoning at 8k and 32k. Counts out of 100 require the official location-answer criterion and normal EOS. Earlier movement records can remain necessary for object history, particularly on qa3.}
\label{tab:v41_babilong}
{\setlength{\tabcolsep}{\dimexpr\tabcolsep*16/18\relax}
\begin{tabular}{ccccccccc}
\toprule
\multirow{2}{*}[-0.6ex]{\TableLabel{Model}} & \multirow{2}{*}[-0.6ex]{\TableLabel{Task/length}} & \multicolumn{4}{c}{\TableHead{Full history or supplied-span masks}} & \multicolumn{3}{c}{\TableHead{Dynamic capacity 512}} \\
\cmidrule(lr){3-6}\cmidrule(l){7-9}
 &  & \FullCell{\FullHeader{Full}} & \WholeCell{\WholeHeader{Source}} & \TableHead{Neutral} & \TableHead{Rule} & \TableHead{Quest} & \TableHead{FINCH} & \TableHead{RefreshKV} \\\midrule
Qwen3-4B & qa2/8k & \FullCell{\AnswerPct{46}{46}} & \WholeCell{\AnswerPct{57}{57}} & \AnswerPct{46}{46} & \AnswerPct{50}{50} & \AnswerPct{32}{32} & \AnswerPct{37}{37} & \AnswerPct{38}{38}\\Qwen3-4B & qa2/32k & \FullCell{\AnswerPct{36}{36}} & \WholeCell{\AnswerPct{43}{43}} & \AnswerPct{36}{36} & \AnswerPct{39}{39} & \AnswerPct{10}{10} & \AnswerPct{30}{30} & \AnswerPct{45}{45}\\Qwen3-4B & qa3/8k & \FullCell{\AnswerPct{28}{28}} & \WholeCell{\AnswerPct{13}{13}} & \AnswerPct{29}{29} & \AnswerPct{31}{31} & \AnswerPct{18}{18} & \AnswerPct{30}{30} & \AnswerPct{28}{28}\\Qwen3-4B & qa3/32k & \FullCell{\AnswerPct{33}{33}} & \WholeCell{\AnswerPct{16}{16}} & \AnswerPct{32}{32} & \AnswerPct{31}{31} & \AnswerPct{9}{9} & \AnswerPct{29}{29} & \AnswerPct{30}{30}\\\addlinespace[2pt]
Llama-3.2-3B & qa2/8k & \FullCell{\AnswerPct{28}{28}} & \WholeCell{\AnswerPct{29}{29}} & \AnswerPct{29}{29} & \AnswerPct{28}{28} & \AnswerPct{20}{20} & \AnswerPct{26}{26} & \AnswerPct{30}{30}\\Llama-3.2-3B & qa2/32k & \FullCell{\AnswerPct{16}{16}} & \WholeCell{\AnswerPct{20}{20}} & \AnswerPct{18}{18} & \AnswerPct{20}{20} & \AnswerPct{0}{0} & \AnswerPct{21}{21} & \AnswerPct{18}{18}\\Llama-3.2-3B & qa3/8k & \FullCell{\AnswerPct{25}{25}} & \WholeCell{\AnswerPct{13}{13}} & \AnswerPct{25}{25} & \AnswerPct{22}{22} & \AnswerPct{13}{13} & \AnswerPct{23}{23} & \AnswerPct{23}{23}\\Llama-3.2-3B & qa3/32k & \FullCell{\AnswerPct{25}{25}} & \WholeCell{\AnswerPct{6}{6}} & \AnswerPct{26}{26} & \AnswerPct{22}{22} & \AnswerPct{3}{3} & \AnswerPct{22}{22} & \AnswerPct{13}{13}\\\addlinespace[2pt]
Qwen3-32B & qa2/8k & \FullCell{\AnswerPct{52}{52}} & \WholeCell{\AnswerPct{61}{61}} & \AnswerPct{52}{52} & \AnswerPct{56}{56} & \AnswerPct{39}{39} & \AnswerPct{52}{52} & \AnswerPct{47}{47}\\Qwen3-32B & qa2/32k & \FullCell{\AnswerPct{46}{46}} & \WholeCell{\AnswerPct{47}{47}} & \AnswerPct{46}{46} & \AnswerPct{45}{45} & \AnswerPct{22}{22} & \AnswerPct{43}{43} & \AnswerPct{39}{39}\\Qwen3-32B & qa3/8k & \FullCell{\AnswerPct{32}{32}} & \WholeCell{\AnswerPct{14}{14}} & \AnswerPct{31}{31} & \AnswerPct{33}{33} & \AnswerPct{21}{21} & \AnswerPct{33}{33} & \AnswerPct{25}{25}\\Qwen3-32B & qa3/32k & \FullCell{\AnswerPct{40}{40}} & \WholeCell{\AnswerPct{15}{15}} & \AnswerPct{37}{37} & \AnswerPct{39}{39} & \AnswerPct{12}{12} & \AnswerPct{33}{33} & \AnswerPct{17}{17}\\
\addlinespace[2pt]
\bottomrule
\end{tabular}}
\par\smallskip{\footnotesize\TableNoteAlign Source hides all but the last movement per person; Rule additionally retains movements needed for object-event locations. Rule removes fewer records, so its contrast is a whole-policy comparison. Neutral hides an equal count of non-fact tokens. Dynamic budgets are much smaller than source-visible context and do not match physical storage. Uncertainty uses 78 qa2/80 qa3 original story groups; shared distractor-text dependence remains.\TableNoteEnd}
\end{table}

\begin{figure}[!htbp]
\centering
\includegraphics[width=\linewidth]{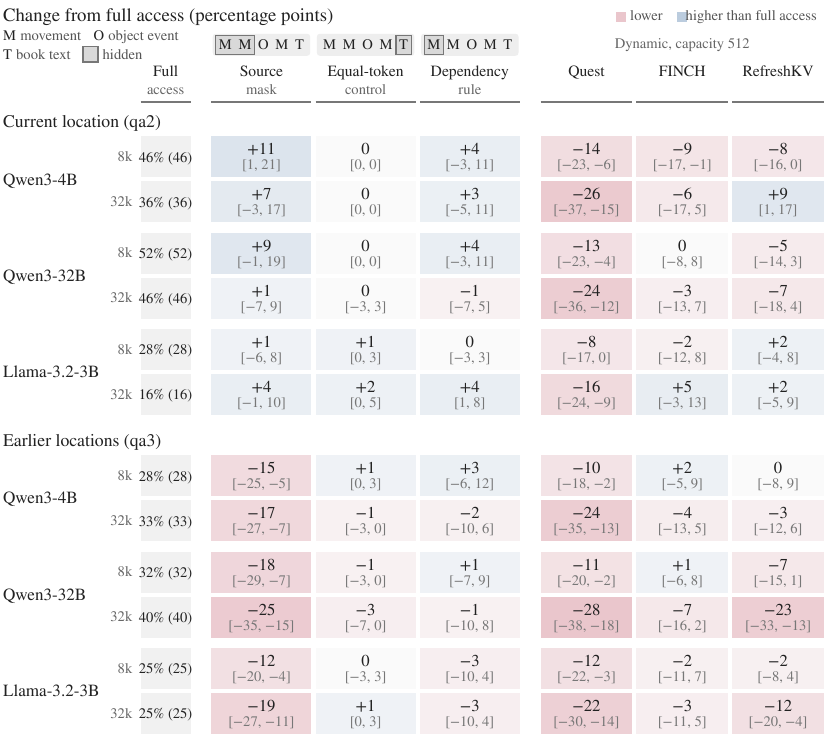}
\caption{BABILong answer changes at two context lengths, relative to full access at the same length. Each model has an 8k row above a 32k row; blocks separate current locations (qa2) from earlier locations (qa3). Cells give the change in percentage points with its pointwise paired 95\% interval beneath, over 78 qa2 or 80 qa3 source stories, requiring official location match and EOS; rose marks losses and blue gains, on one shade scale. The gray column gives full-access percentages (counts), with 100 questions per cell. Header chips shade the records each mask hides among movements (M), an object event (O) and book text (T). The dependency rule removes fewer records; the dynamic methods use capacity 512. These policies do not match visible context or physical storage.}
\label{fig:v53_babilong_lengths}
\label{fig:v41_babilong_8k}
\label{fig:v41_babilong_32k}
\end{figure}

Figure~\ref{fig:v53_babilong_lengths} compares both context lengths on a common scale, with paired differences and source-story intervals.
Each of the three models produces 2,800 responses across seven policies.
Quest, FINCH and RefreshKV use a fixed 512-token capacity at both lengths,
with the method-specific schedules in Appendix~\ref{app:v41_dynamic}.
Source masks leave nearly all distractor context visible, whereas the
dynamic methods select a much smaller subset; these settings do not compare
compression efficiency at equal resources. Native model limits suffice:
no truncation or rotary-position extension is used. Greedy generation
allows 128 tokens rather than the benchmark's original 20-token ceiling,
with EOS and ceiling hits recorded separately.

The primary metric follows the official location parser, excluding location
labels appearing in the question, and additionally requires normal EOS.
Strict bare-location exact match is a formatting sensitivity. For qa3 at
32k, full/source/conservative-rule counts are 33/16/31 for Qwen3-4B,
25/6/22 for Llama-3.2-3B, and 40/15/39 for Qwen3-32B. On qa2 at the same
length, naive masking improves the point counts from 36 to 43, 16 to 20,
and 46 to 47. These differences delimit the harm: information that is old
for the person's current state can still be required for the object's past.
Keeping its symbolic trajectory intact does not guarantee unchanged
language-model accuracy.

\subsection{Current Answers after Accumulated Fact Updates}
\label{app:v43_mab}

MemoryAgentBench FactConsolidation tests current answers after many facts
have been updated \citep{memoryagentbench2025}. Derived from MQuAKE, it
extends the number of accumulated updates and the input length, rather
than adding an independent source domain. We evaluate fixed questions
after ingesting the full history; this is a cache-intervention adaptation,
not execution of the upstream interactive memory-agent harness.

\paragraph{Complete inputs and four operations.}
We retain both released nominal 32k/64k histories, each with 100 single-hop
and 100 multi-hop questions. Qwen3.8-27B and Gemma 4 26B-A4B each have a
262,144-token native window. Their maximum full prompt lengths are
39,593/79,685 and 38,734/78,098 tokens, respectively; no history is
truncated. The histories contain 2,310/4,580 numbered facts, and 2,244
shorter-history fact sentences recur in the longer history. Each mapped
subject--relation is replaced at most once. The task therefore tests many
updates across different facts, not repeated replacement of one attribute.
Four questions exposed during runtime qualification are retained in the
400-question analysis and excluded in a separate 396-question sensitivity.

The original MQuAKE triples and relation templates map numbered sentences
to their source relations independently of questions and references.
The latest numbered conflicting fact prevails. Source masking selects
834/1,682 old fact sentences; value masking selects their exact old target
strings. Facts associated with ambiguous same-name entity keys remain
visible, so this mapping does not claim complete conflict coverage.
Hidden entity identifiers are not supplied to generators or evaluators.
Full, source-mask and value-mask answers branch from the same prefilled
history. Masks act on the question and every generated token while
preserving stored states. Recompute deletes the selected old sentences
and prefills the retained text again. These whole-policy comparisons lack
an equal-token unrelated mask: selected content, age and token count
vary together.

Both generators use native templates with thinking disabled, greedy decoding and a fixed
1,024-token maximum, stopping earlier at EOS. The instruction requests
only a boxed final answer. The budget was selected before test generation
after development continuations reached 256 tokens. Prefill uses
128-token chunks; Qwen retains its recurrent states while masks act on
full-attention layers, and Gemma uses its original local/global attention
implementation. Cache digests verify unchanged shared prefixes across
conditions. Each model generates 1,600 answers across questions and
operations, giving 3,200 answers with recorded termination and cache checks.

\paragraph{Source support, reference agreement and termination.}
GPT-5.6 reviews all 400 questions and 1,411 distinct displayed
answers in forty batches. Each batch contains a complete numbered history,
ten original questions and references, and anonymous answers from all model
and operation conditions. Exact repeats share a judgment. Generator,
operation, previous scores and termination are hidden. The parser extracts
the final boxed answer when a box marker occurs; otherwise it displays
the complete original response. Full responses and displayed text are
retained separately.

Source assessments and answer labels are produced together, rather than
by an independent source-only assessment. The review identifies 373 answerable,
24 unanswerable and three ambiguous questions; 355 original references
are supported, 38 unsupported and seven mismatched to the question.
The rubric applies the latest-fact rule at each reasoning step, requires
quotations from the stated fact number, and separates reference agreement,
complete or partial source-supported answers, wrong claims and abstention.
Refusal justification and explanation accuracy are separate: a reasonable
refusal does not excuse a false substantive assertion. After an initial
run exposed a conflict in that precedence rule, all 400 questions were
reviewed uniformly under the corrected rule; earlier labels are preserved
but excluded from the final analysis. Under the pre-revision rule (370 validator-accepted questions), Complete+EOS changes by at most 6 per cell and all six operation-minus-full contrasts keep their sign; on EOS answers to those questions the two rule versions agree on 97.4\% of distinct source grades (601/617). Authored controls check adherence to the rubric, response format, and
source quotations; they do not measure grading accuracy on these benchmark
answers.

\begin{table}[!htbp]
\WideTableStyle
\caption{Outcome entries give percentages, with the original count/denominator beneath. Current FactConsolidation answers over the full nominal 32k and 64k histories.}
\label{tab:v43_mab_long}
{\setlength{\tabcolsep}{\dimexpr\tabcolsep*10/12\relax}
\begin{tabular}{cccccc}
\toprule
\TableHead{Model} & \TableHead{Access} & \TableHead{\shortstack{Official match\\/400}} & \TableHead{\shortstack{Final match\\/400}} & \TableHead{\shortstack{Reference agrees\\/400}} & \TableHead{\shortstack{Source complete\\/answerable}} \\\midrule
Qwen3.8-27B & \FullCell{Full} & \AnswerPct{90}{360/400} & \AnswerPct{81.5}{326/400} & \AnswerPct{81.5}{326/400} & \AnswerPct{90.1}{336/373}\\ & \WholeCell{Source} & \AnswerPct{85.5}{342/400} & \AnswerPct{73.8}{295/400} & \AnswerPct{73.5}{294/400} & \AnswerPct{80.4}{300/373}\\ & \NumberCell{Old value} & \AnswerPct{87.5}{350/400} & \AnswerPct{76}{304/400} & \AnswerPct{75}{300/400} & \AnswerPct{82.8}{309/373}\\ & \FullCell{Recompute} & \AnswerPct{88.8}{355/400} & \AnswerPct{85.5}{342/400} & \AnswerPct{85.5}{342/400} & \AnswerPct{95.2}{355/373}\\\midrule
Gemma 4 26B-A4B & \FullCell{Full} & \AnswerPct{31.5}{126/400} & \AnswerPct{28}{112/400} & \AnswerPct{28}{112/400} & \AnswerPct{29.5}{110/373}\\ & \WholeCell{Source} & \AnswerPct{54}{216/400} & \AnswerPct{45.3}{181/400} & \AnswerPct{45.3}{181/400} & \AnswerPct{49.6}{185/373}\\ & \NumberCell{Old value} & \AnswerPct{47.8}{191/400} & \AnswerPct{42.5}{170/400} & \AnswerPct{42.5}{170/400} & \AnswerPct{45.6}{170/373}\\ & \FullCell{Recompute} & \AnswerPct{71.5}{286/400} & \AnswerPct{65}{260/400} & \AnswerPct{65}{260/400} & \AnswerPct{71.3}{266/373}\\
\bottomrule
\end{tabular}}
\par\smallskip{\footnotesize\TableNoteAlign Official match uses the pinned full-response substring score without requiring EOS; Final match applies it to the extracted final answer with EOS. The two GPT-5.6 semantic columns separately assess original-reference agreement and source completeness with EOS. Source targets and answer grades are assigned together, with generator and operation identities hidden. Tiers share most shorter-history facts; all contrasts are descriptive, with no independent-question confidence interval.\TableNoteEnd}
\end{table}

Source completeness with normal EOS uses the 373 source-answerable
questions, while original-reference metrics retain all 400. Under
full/source/value/recompute, Qwen's complete+EOS counts are
336/300/309/355; Gemma's are 110/185/170/266. Source masking changes the
counts by $-36$ and $+75$, respectively, and on this measure
recomputation exceeds both full baselines. Under the official match in
Table~\ref{tab:v43_mab_long}, Qwen's recompute count of 355/400 stays
below its full-access 360/400. These are descriptive comparisons on two highly
overlapping histories; we do not report independent-question confidence
intervals or population intervals based on two histories.

\paragraph{Separating completeness and termination.}
Appendix~\ref{app:v45_mab_analysis} decomposes the gains and losses,
including the additional capped outputs in both models.
Appendix~\ref{app:v52_fc_truncation} describes what the capped outputs
contain. No answer in
this review receives the partial-answer label, so these losses do not
replicate the complete-to-partial endpoint measured on OAKS.

This decomposition assigns capped outputs to one category before
classifying normally terminated answers. It preserves their semantic
labels separately; a capped response may also contain a wrong assertion.
The original grade-based transition counts and capped counts can overlap
and must not be summed. Complete+EOS gains from capped baselines need not
represent repaired factual errors.

\begin{table}[!htbp]
\WideTableStyle
\caption{FactConsolidation termination and output form at the fixed 1,024-token ceiling. Entries are percentages (counts) of all 400 questions per model and operation.}
\label{tab:v43_mab_termination}
{\setlength{\tabcolsep}{\dimexpr\tabcolsep*10/12\relax}
\begin{tabular}{ccccc}
\toprule
\TableHead{Model} & \TableHead{Access} & \TableHead{\shortstack{EOS\\(all valid boxes)}} & \TableHead{Capped} & \TableHead{\shortstack{Only box\\+EOS}} \\\midrule
Qwen3.8-27B & \FullCell{Full} & \AnswerPctInline{87.3}{349} & \AnswerPctInline{12.8}{51} & \AnswerPctInline{11.3}{45}\\ & \WholeCell{Source} & \AnswerPctInline{78.8}{315} & \AnswerPctInline{21.3}{85} & \AnswerPctInline{11}{44}\\ & \NumberCell{Old value} & \AnswerPctInline{80.5}{322} & \AnswerPctInline{19.5}{78} & \AnswerPctInline{10.5}{42}\\ & \FullCell{Recompute} & \AnswerPctInline{91.5}{366} & \AnswerPctInline{8.5}{34} & \AnswerPctInline{52.3}{209}\\\midrule
Gemma 4 26B-A4B & \FullCell{Full} & \AnswerPctInline{79}{316} & \AnswerPctInline{21}{84} & \AnswerPctInline{0}{0}\\ & \WholeCell{Source} & \AnswerPctInline{54.8}{219} & \AnswerPctInline{45.3}{181} & \AnswerPctInline{0}{0}\\ & \NumberCell{Old value} & \AnswerPctInline{70.5}{282} & \AnswerPctInline{29.5}{118} & \AnswerPctInline{0}{0}\\ & \FullCell{Recompute} & \AnswerPctInline{71.3}{285} & \AnswerPctInline{28.8}{115} & \AnswerPctInline{0}{0}\\
\bottomrule
\end{tabular}}
\par\smallskip{\footnotesize\TableNoteAlign Counts are out of 400 questions for each condition. Every normally terminated response has a valid closed answer box, so EOS and Valid box+EOS counts coincide. A valid box permits surrounding text; Only box requires the entire response to be one box. These checks are separate from factual accuracy. The two released length tiers share a fact pool; counts are descriptive.\TableNoteEnd}
\end{table}

The official full-response substring metric credits 236 outputs that do
not pass final-answer substring matching with EOS: 198 are capped without
a box, and 38 terminate normally but mention the reference only outside
the extracted final answer. Substring matches can also accept conflicting
values: six Qwen masked answers list English alongside the current Danish,
Italian, French or German answer. Source-aware grading marks those six
incorrect. These distinctions explain why the table retains separate
official, final-reference and source-completeness measurements.

\paragraph{Question type and exposure sensitivity.}
Qwen's source-mask losses are concentrated in multi-hop questions:
complete+EOS counts change from 71 to 57 of 84 source-answerable questions
in the shorter tier, and from 68 to 49 of 89 in the longer tier.
Single-hop counts change from 99 to 97 and from 98 to 97, each out of 100.
Gemma improves in all four tier/question-type groups, from lower baselines.
Length and question groups differ, so these are observational strata,
not an isolated causal effect of context length.

\begin{table}[!htbp]
\WideTableStyle
\caption{Outcome entries give percentages, with the original count/denominator beneath. FactConsolidation sensitivity excluding the four questions seen during runtime qualification.}
\label{tab:v43_mab_exposure}
{\setlength{\tabcolsep}{\dimexpr\tabcolsep*10/12\relax}
\begin{tabular}{cccccc}
\toprule
\TableHead{Model} & \TableHead{Access} & \TableHead{\shortstack{Official match\\/396}} & \TableHead{\shortstack{Reference agrees\\/396}} & \TableHead{\shortstack{Source complete\\excluding four}} & \TableHead{\shortstack{Source complete\\all questions}} \\\midrule
Qwen3.8-27B & \FullCell{Full} & \AnswerPct{89.9}{356/396} & \AnswerPct{81.3}{322/396} & \AnswerPct{90}{332/369} & \AnswerPct{90.1}{336/373}\\ & \WholeCell{Source} & \AnswerPct{85.4}{338/396} & \AnswerPct{73.7}{292/396} & \AnswerPct{80.8}{298/369} & \AnswerPct{80.4}{300/373}\\ & \NumberCell{Old value} & \AnswerPct{87.4}{346/396} & \AnswerPct{75}{297/396} & \AnswerPct{82.9}{306/369} & \AnswerPct{82.8}{309/373}\\ & \FullCell{Recompute} & \AnswerPct{88.9}{352/396} & \AnswerPct{85.6}{339/396} & \AnswerPct{95.4}{352/369} & \AnswerPct{95.2}{355/373}\\\midrule
Gemma 4 26B-A4B & \FullCell{Full} & \AnswerPct{31.3}{124/396} & \AnswerPct{27.8}{110/396} & \AnswerPct{29.3}{108/369} & \AnswerPct{29.5}{110/373}\\ & \WholeCell{Source} & \AnswerPct{54}{214/396} & \AnswerPct{45.5}{180/396} & \AnswerPct{49.9}{184/369} & \AnswerPct{49.6}{185/373}\\ & \NumberCell{Old value} & \AnswerPct{47.5}{188/396} & \AnswerPct{42.2}{167/396} & \AnswerPct{45.3}{167/369} & \AnswerPct{45.6}{170/373}\\ & \FullCell{Recompute} & \AnswerPct{71.5}{283/396} & \AnswerPct{64.9}{257/396} & \AnswerPct{71.3}{263/369} & \AnswerPct{71.3}{266/373}\\
\bottomrule
\end{tabular}}
\par\smallskip{\footnotesize\TableNoteAlign Each condition excludes the same four questions without using their correctness. The source-complete denominator further requires source answerability and is printed explicitly; both semantic endpoints require EOS. This exposure sensitivity preserves the dependency between the two released length tiers.\TableNoteEnd}
\end{table}
\begin{table}[!htbp]
\WideTableStyle
\caption{FactConsolidation answers by question type and history length, shown as percentages with the original counts and denominators. The panels distinguish the official score from source-based completeness.}
\label{tab:v43_mab_singlehop}
\label{tab:v43_mab_multihop}
{\setlength{\tabcolsep}{\dimexpr\tabcolsep*10/12\relax}
\begin{tabular}{cccccc}
\toprule
\multirow{2}{*}{\TableLabel{Model}} & \multirow{2}{*}{\TableLabel{Access}} & \multicolumn{2}{c}{\TableHead{32k history}} & \multicolumn{2}{c}{\TableHead{64k history}} \\
\cmidrule(lr){3-4}\cmidrule(l){5-6}
 & & \TableHead{\shortstack{Official\\/100}} & \TableHead{\shortstack{Complete\\/answerable}} & \TableHead{\shortstack{Official\\/100}} & \TableHead{\shortstack{Complete\\/answerable}} \\\midrule
\TableGroupRow{6}{A. Single-hop questions}
Qwen3.8-27B & \FullCell{Full} & \AnswerPct{98}{98/100} & \AnswerPct{99}{99/100} & \AnswerPct{98}{98/100} & \AnswerPct{98}{98/100}\\ & \WholeCell{Source} & \AnswerPct{97}{97/100} & \AnswerPct{97}{97/100} & \AnswerPct{99}{99/100} & \AnswerPct{97}{97/100}\\ & \NumberCell{Old value} & \AnswerPct{98}{98/100} & \AnswerPct{98}{98/100} & \AnswerPct{98}{98/100} & \AnswerPct{98}{98/100}\\ & \FullCell{Recompute} & \AnswerPct{98}{98/100} & \AnswerPct{100}{100/100} & \AnswerPct{98}{98/100} & \AnswerPct{100}{100/100}\\\midrule
Gemma 4 26B-A4B & \FullCell{Full} & \AnswerPct{64}{64/100} & \AnswerPct{59}{59/100} & \AnswerPct{48}{48/100} & \AnswerPct{41}{41/100}\\ & \WholeCell{Source} & \AnswerPct{84}{84/100} & \AnswerPct{79}{79/100} & \AnswerPct{71}{71/100} & \AnswerPct{60}{60/100}\\ & \NumberCell{Old value} & \AnswerPct{73}{73/100} & \AnswerPct{72}{72/100} & \AnswerPct{71}{71/100} & \AnswerPct{62}{62/100}\\ & \FullCell{Recompute} & \AnswerPct{91}{91/100} & \AnswerPct{92}{92/100} & \AnswerPct{83}{83/100} & \AnswerPct{78}{78/100}\\\midrule
\TableGroupRow{6}{B. Multi-hop questions}
Qwen3.8-27B & \FullCell{Full} & \AnswerPct{83}{83/100} & \AnswerPct{84.5}{71/84} & \AnswerPct{81}{81/100} & \AnswerPct{76.4}{68/89}\\ & \WholeCell{Source} & \AnswerPct{72}{72/100} & \AnswerPct{67.9}{57/84} & \AnswerPct{74}{74/100} & \AnswerPct{55.1}{49/89}\\ & \NumberCell{Old value} & \AnswerPct{75}{75/100} & \AnswerPct{66.7}{56/84} & \AnswerPct{79}{79/100} & \AnswerPct{64}{57/89}\\ & \FullCell{Recompute} & \AnswerPct{79}{79/100} & \AnswerPct{95.2}{80/84} & \AnswerPct{80}{80/100} & \AnswerPct{84.3}{75/89}\\\midrule
Gemma 4 26B-A4B & \FullCell{Full} & \AnswerPct{8}{8/100} & \AnswerPct{6}{5/84} & \AnswerPct{6}{6/100} & \AnswerPct{5.6}{5/89}\\ & \WholeCell{Source} & \AnswerPct{39}{39/100} & \AnswerPct{36.9}{31/84} & \AnswerPct{22}{22/100} & \AnswerPct{16.9}{15/89}\\ & \NumberCell{Old value} & \AnswerPct{24}{24/100} & \AnswerPct{22.6}{19/84} & \AnswerPct{23}{23/100} & \AnswerPct{19.1}{17/89}\\ & \FullCell{Recompute} & \AnswerPct{71}{71/100} & \AnswerPct{76.2}{64/84} & \AnswerPct{41}{41/100} & \AnswerPct{36}{32/89}\\
\bottomrule
\end{tabular}}
\par\smallskip{\footnotesize\TableNoteAlign Each original length/task cell contains 100 questions. Official uses the original full-response substring score; source-complete final answers require EOS and are reported over explicitly printed answerable denominators. Length labels are nominal; actual input lengths and source overlap appear in the protocol. These descriptive cells are not independent replications.\TableNoteEnd}
\end{table}

The exclusion sensitivity retains 369 source-answerable questions and
the same direction of source masking for each model. It does not remove
the shared-fact dependence between tiers. The supplementary results retain
all generated answers, displayed text, labels and mappings. Original
benchmark histories, source quotations and API reasoning are not
redistributed with these results.

\subsection{Why Net Accuracy Changes Differ between the Two Models}
\label{app:v45_mab_analysis}

We analyze the same 3,200 generated answers and frozen source-aware
judgments as Appendix~\ref{app:v43_mab}, without new generation or
regrading. Comparisons retain the 373 source-answerable questions.
A response that reaches 1,024 tokens is assigned to the capped category
before its content is considered. Among normally terminated answers,
complete, wrong and abstaining answers remain distinct. The separate
semantic label is retained for every capped response, which can also
contain an incorrect or unsupported assertion.

\begin{table}[!htbp]
\WideTableStyle
\caption{FactConsolidation answer categories as percentages (counts) of the same 373 source-answerable questions. Each row separates content-complete answers, wrong answers, token-limit stops and abstentions.}
\label{tab:v45_mab_outcomes}
{\setlength{\tabcolsep}{\dimexpr\tabcolsep*10/12\relax}
\begin{tabular}{cccccc}
\toprule
\TableHead{Model} & \TableHead{Access} & \TableHead{Complete + EOS} & \TableHead{Wrong + EOS} & \TableHead{Capped} & \TableHead{Abstain + EOS} \\\midrule
Qwen3.8-27B & \FullCell{Full} & \AnswerPctInline{90.1}{336} & \AnswerPctInline{3.2}{12} & \AnswerPctInline{6.7}{25} & \AnswerPctInline{0}{0}\\ & \WholeCell{Source mask} & \AnswerPctInline{80.4}{300} & \AnswerPctInline{4}{15} & \AnswerPctInline{15.5}{58} & \AnswerPctInline{0}{0}\\ & \NumberCell{Old value} & \AnswerPctInline{82.8}{309} & \AnswerPctInline{3.5}{13} & \AnswerPctInline{13.7}{51} & \AnswerPctInline{0}{0}\\ & \FullCell{Recompute} & \AnswerPctInline{95.2}{355} & \AnswerPctInline{2.1}{8} & \AnswerPctInline{2.7}{10} & \AnswerPctInline{0}{0}\\\midrule
Gemma 4 26B-A4B & \FullCell{Full} & \AnswerPctInline{29.5}{110} & \AnswerPctInline{52.8}{197} & \AnswerPctInline{17.2}{64} & \AnswerPctInline{0.5}{2}\\ & \WholeCell{Source mask} & \AnswerPctInline{49.6}{185} & \AnswerPctInline{7.2}{27} & \AnswerPctInline{42.1}{157} & \AnswerPctInline{1.1}{4}\\ & \NumberCell{Old value} & \AnswerPctInline{45.6}{170} & \AnswerPctInline{27.9}{104} & \AnswerPctInline{26}{97} & \AnswerPctInline{0.5}{2}\\ & \FullCell{Recompute} & \AnswerPctInline{71.3}{266} & \AnswerPctInline{3.5}{13} & \AnswerPctInline{24.7}{92} & \AnswerPctInline{0.5}{2}\\
\bottomrule
\end{tabular}}
\par\smallskip{\footnotesize\TableNoteAlign The categories partition each row. Token caps take precedence even when the displayed response also contains a wrong assertion. Wrong includes contradictions and unsupported claims. No partially complete or unverifiable answer terminates normally in this subset.\TableNoteEnd}
\end{table}

\paragraph{Separate new corrections from lost complete answers.}
Qwen's complete-answer rate falls from 90.1\% (336) to 80.4\% (300)
under source masking. Gains of 1.9\% (7) and losses of 11.5\% (43)
give the net change of $-9.7$ percentage points. Gemma rises from
29.5\% (110) to 49.6\% (185): gains of 23.6\% (88) exceed losses of
3.5\% (13), giving $+20.1$ points. These rates all use 373 questions.
Qwen's gains comprise four formerly wrong answers with EOS and three
capped answers; Gemma's comprise 81 wrong answers with EOS and seven
capped answers. Of Qwen's 43 losses, 86.0\% (37) become capped and
14.0\% (6) become wrong answers with EOS. For Gemma, the corresponding
fractions of 13 losses are 92.3\% (12) and 7.7\% (1).

The starting populations explain much of the difference in net change.
Full access gives only 3.2\% (12/373) wrong answers with EOS in Qwen,
but 52.8\% (197/373) in Gemma. Masking corrects 33.3\% (4/12) and
41.1\% (81/197) of those wrong answers, respectively. Among initially
complete answers, loss rates are 12.8\% (43/336) and 11.8\% (13/110).
These conditional rates use different question subsets. Restricting the
comparison to the 109 questions complete in both models under full
access, masking preserves 96.3\% (105) for Qwen and 88.1\% (96)
for Gemma. The model with the larger overall gain thus preserves fewer
of their commonly correct answers.

\paragraph{Gemma's corrections largely replace outdated answers.}
Among Gemma's 197 wrong answers with EOS under full access, 89.3\%
(176) have a final box containing an old-answer alias and no supported
current-answer alias. Old aliases come from the mapped benchmark
relations; names overlapping a current alias are excluded. We also check the frozen source review's supported answer
and accepted aliases: none of these 176 final answers contains one.
After masking, 40.3\% (71/176) become complete, 47.7\% (84/176)
reach the token cap, 11.4\% (20/176) remain wrong with EOS, and
0.6\% (1/176) abstain. This group supplies 87.7\% (71/81) of all
wrong-to-complete corrections. Masking frequently removes use of an
outdated answer, but many affected questions still fail to finish.

For example, a hypothetical update in the shorter benchmark history
changes Artur Boruc's playing position. Fact 1173 gives goalkeeper;
the later Fact 2052 gives midfielder. Full access makes Gemma cite
Fact 1173 and return a boxed goalkeeper answer. Source masking hides
that old record, and Gemma cites Fact 2052 and returns midfielder.
Recomputation also returns the supported current answer. The comparison
checks the original numbered facts and the actual selected mask spans.

\paragraph{Qwen's losses concentrate in longer multi-hop answers.}
On 200 single-hop questions, Qwen changes from 98.5\% (197) complete
to 97.0\% (194); on 173 source-answerable multi-hop questions, it
changes from 80.3\% (139) to 61.3\% (106). Gemma instead improves
from 50.0\% (100) to 69.5\% (139) on single-hop questions and from
5.8\% (10) to 26.6\% (46) on multi-hop questions. The multi-hop
direction is the same at both released lengths for each model.
The histories share facts and have different question sets, so these
strata describe where the effects occur.

Within Qwen's initially complete answers, masking loses 3.0\% (6/197)
of single-hop answers and 26.6\% (37/139) of multi-hop answers.
Loss also increases with the length of the original full-access answer:
0.0\% (0/42) for 0--20 tokens, 3.8\% (6/159) for 21--200 tokens,
21.8\% (24/110) for 201--500 tokens, and 52.0\% (13/25) above
500 tokens. The repeated fact checks, token caps and errors within these
outputs are examined in Appendix~\ref{app:v52_fc_truncation}.

\paragraph{The opposite directions survive changes to the scoring subset.}
Removing the EOS requirement leaves the same directions: Qwen's
source-graded completeness falls from 91.7\% (342/373) to 80.4\%
(300/373), whereas Gemma rises from 29.5\% (110/373) to 49.6\%
(185/373). Thus the disagreement is not created solely by requiring
normal termination. On the 355 questions with source-supported original
references, Qwen changes from 91.5\% (325) to 82.8\% (294) and
Gemma from 30.7\% (109) to 50.7\% (180). Excluding the four
questions seen during development leaves 369 source-answerable questions:
Qwen changes from 90.0\% (332) to 80.8\% (298), and Gemma from
29.3\% (108) to 49.9\% (184). Both checks retain the original signs.

\paragraph{Recomputation restores many previously capped answers.}
Recomputation gives Qwen 16.1\% (60/373) gains and 1.3\% (5/373)
losses relative to masking, raising completeness by 14.7 points.
Of its 60 gains, 81.7\% (49) were capped and 18.3\% (11) were
wrong with EOS. Gemma gains 25.2\% (94/373) and loses 3.5\%
(13/373), a net increase of 21.7 points. Of its 94 gains, 85.1\%
(80) were capped, 12.8\% (12) were wrong with EOS, and 2.1\% (2)
were abstentions. Among all source-masked capped outputs, 84.5\%
(49/58) become complete with EOS in Qwen and 51.0\% (80/157) in Gemma.

Instructions and answer budgets stay fixed. Recomputation removes old
sentences, retokenizes the retained history and rebuilds its states at
shorter positions, including Qwen's recurrent states and Gemma's local
attention computations. The parser grades the final box when present and otherwise the complete
response. Recovery therefore measures a supported final answer within
budget, with the content and termination criteria both satisfied.

The recovered questions differ between models. Of Qwen's 49
cap-to-complete recoveries, 71.4\% (35) had been complete under full
access. Of Gemma's 80 recoveries, 55.0\% (44) had been wrong with EOS
under full access, 32.5\% (26) had already been capped, and 12.5\%
(10) had been complete. Recomputing Qwen's history thus mainly restores
answers that masking had disrupted; many of Gemma's recoveries instead
concern questions it had already answered incorrectly with full access.
\FloatBarrier

\subsection{Controlled Histories and Paired Analysis}

\label{app:panel}

We construct 400 history clusters, equally split across location, API endpoint,
preference, deadline, and incident-cause updates. A paired \textsc{Current}
continuation repeats the old value, while \textsc{Superseded} replaces it.
Each continuation has one target and one unrelated query, giving 1,600 cases
per model before interventions and timing conditions. Statements share entity
and attribute phrases within each family; unrelated queries concern another
entity with the same attribute. Answer order, old-span position, and value
direction are counterbalanced. Relation labels, answer keys, and span offsets
are stored separately from the text presented to the model.

Every model uses the same constructed histories, intervention definitions, and continuation budgets.  Model-specific adapters are limited to prompt assembly, tokenizer
offset mapping, layer/head indexing, and cache/attention tensor layout.  They
do not change signal definitions or decision criteria.

Confidence intervals use 10,000 paired bootstrap replicates. Observation
comparisons resample connected pair groups; role-matched recognition
resamples matched groups within update family; causal transfer resamples
groups within wording pattern; OAKS resamples questions within book; and
controlled or component interventions resample history clusters. KV-based and surface scores use the same resamples. Bonferroni corrections apply
within each model to the $K$, $V$ and $K{+}V$ component intervals
(Appendix~\ref{app:component_attribution}); all other intervals are
pointwise.
\subsection{Role-Matched Source Influence}
\label{app:paraphrase_causal}

The role-matched comparison contains 400 groups and 1,200 observation pairs
across five update families and eight wording patterns. Both candidate
values occur in every record about the same entity and attribute. \textsc{Duplicate} and
\textsc{Supersedes} exchange which value is current, preserving token
inventory and text length within each group. The corresponding recognition
results are reported in Appendix~\ref{app:decisive_audit}.

The causal experiment compares access conditions on the same 400
role-matched groups, sharing the prefilled cache within each group. One balanced replicate across five update
families, eight wording patterns, and two value directions supplies 80
development events. Whole-source masking changes only 6/80 selected answers, and four
of eight wording patterns have 100\% matched-control accuracy. We therefore
select the current-minus-old log-probability margin before evaluating the
remaining four replicates, comprising 320 events. The estimator averages the
eight wording patterns and uses 10,000 matched-group bootstrap draws within
each pattern. All conditions use the same prefilled source cache, with identical token counts
for the old-source and unrelated-source masks.

\begin{figure}[!ht]
  \centering
  \includegraphics[width=\linewidth]{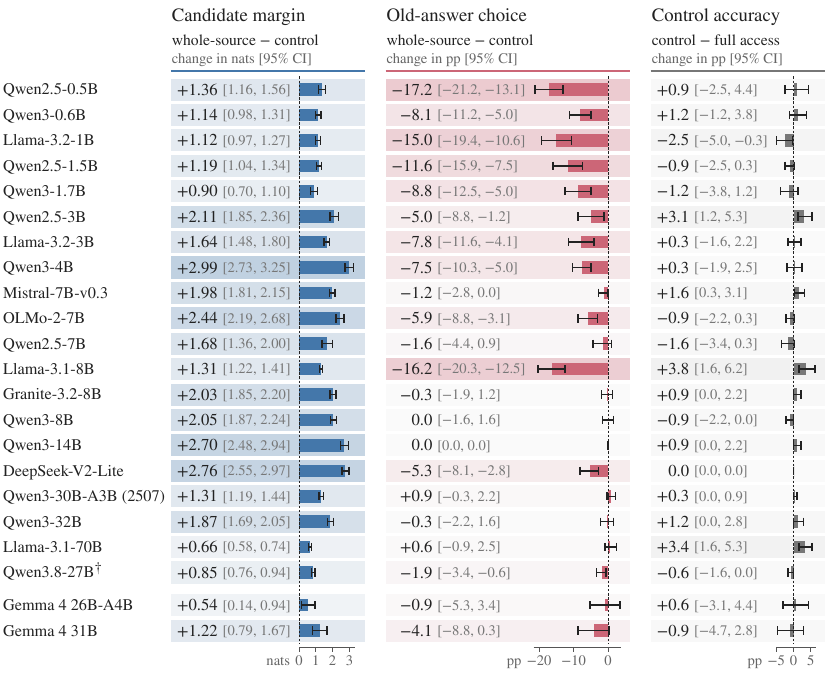}
  \caption{Controlled source effects across 22 checkpoints on 320 events. Each cell prints the estimate and its original pointwise paired 95\% interval, with groups resampled within wording pattern; the bar and whisker draw the same values to scale, and dashed lines mark no change. Left and middle: whole-source minus matched-control changes in candidate margin (nats) and old-answer selection (pp). Right: control-minus-full candidate accuracy (pp), separating source influence from control damage; the two pp blocks share one scale. The separate Mistral Small 24B study is excluded. $\dagger$ Qwen3.8 retains recurrent states.}
  \label{fig:model_causal_matrix}
\end{figure}

The decomposition relative to full access for the same 22 checkpoints appears in
Table~\ref{tab:v23_absolute_margins}; Appendix~\ref{app:v35_scoring} reports model exceptions and prompt-format comparisons.

\subsection{Natural Sources and Book Transfer}
\label{app:oaks_transfer}

We use the OAKS-Novel dataset release at commit
\texttt{0bbc4b3877ea}.  Its 19 public-domain
novels contain 435 questions, 1,354 ordered chunks, and 1,722 answer changes.
We retain 1,316 changes for which the old answer has a nonempty earlier source
sentence and the new answer has a nonempty source sentence at the change
chunk.  A hash-ordered book split assigns 9/5/5 books to development,
calibration, and held-out evaluation; the held-out books contribute 336 events from 116
questions.  Each event identifies the book, chunk order, answers at every chunk, and
source sentences. The dataset is used under its noncommercial research license.

For each event, unrelated text is selected deterministically from another
book and excludes nontrivial tokens from either answer. It supports equal-token controls for each evaluated tokenizer. In the original 19-model evaluation, across 6,384
model--events, prefixes have median length 289 tokens and maximum 1,951;
old spans have median length 41 and maximum 432 tokens. Each treatment
uses a separately allocated copy of the same cache.

The reduction in old-answer selection is
\begin{equation}
 R_{\mathrm{obs}}=\mathbb E_e\!\left[
 \mathbf1\{\hat y_e(\mathrm{control\ mask})=\mathrm{old}_e\}
 -\mathbf1\{\hat y_e(\mathrm{old\ mask})=\mathrm{old}_e\}\right],
 \label{eq:causal}
\end{equation}
where $\hat y$ is the higher sequence-scored candidate.

\begin{figure}[!ht]
  \centering
  \includegraphics[width=\linewidth]{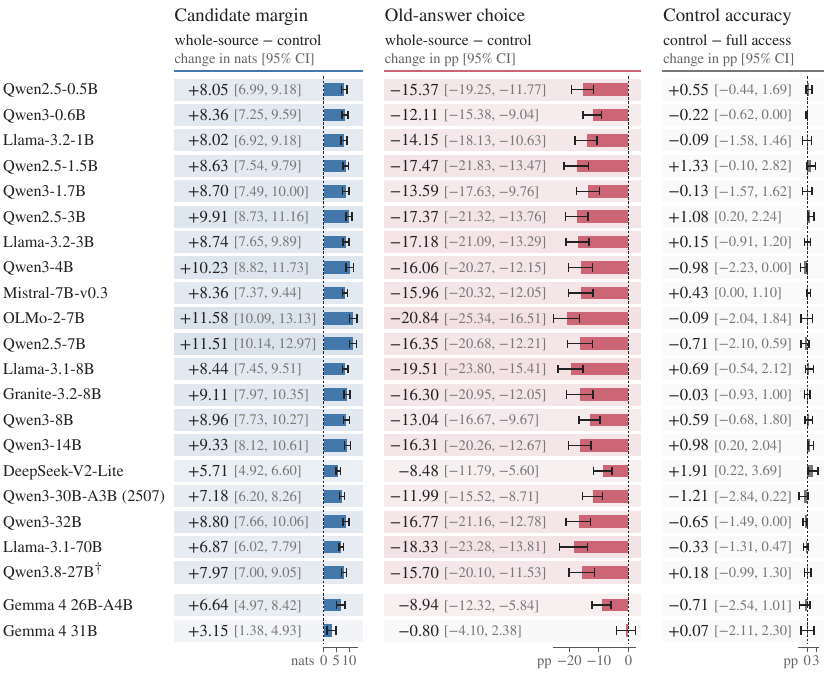}
  \caption{Candidate decisions on 336 OAKS events from 116 questions in five books. Blocks show whole-source minus matched-control margin and old-answer choice changes, and control-minus-full accuracy. Bars and whiskers draw the printed estimates and pointwise 95\% intervals (question resampling within book; equal book weights); the pp blocks share one scale. Gemma 4 31B's old-answer choice interval includes zero. $\dagger$ Qwen3.8 retains recurrent states.}
  \label{fig:model_oaks_matrix}
\end{figure}

\subsection{Absolute Forced-Choice Accuracies}
\label{app:absolute_accuracy}
The model chooses between the same old and current answer strings using their sequence log probabilities. These accuracies evaluate that choice, rather than the semantic correctness of freely generated answers. Controlled entries use 320 events. For OAKS, event contributions are averaged within each book, then the five books are averaged equally; uncertainty resamples whole questions within each book.
Four controlled comparisons fail the specified $-3$-point lower-bound criterion:
Llama-3.2-1B ($-4.38$), Qwen3-1.7B ($-3.44$), Qwen2.5-7B ($-3.13$),
and Gemma 4 31B ($-4.06$). All 22 checkpoints meet the OAKS control criterion,
whose lower bounds range from $-2.55$ to $+0.49$ points.
The separate Mistral 24B evaluation is outside this inventory.

\begin{table}[!htbp]
\WideTableStyle
\caption{Current-answer candidate accuracy (\%) for 22 checkpoints before and after source masking. Full recomputation of edited text is not part of these comparisons.}

{\setlength{\tabcolsep}{\dimexpr\tabcolsep*12/14\relax}
\begin{tabular}{ccccccc}
\toprule
\multirow{2}{*}[-0.6ex]{\TableLabel{Model}} & \multicolumn{3}{c}{\TableHead{Controlled}} & \multicolumn{3}{c}{\TableHead{OAKS excerpts}} \\
\cmidrule(lr){2-4}\cmidrule(lr){5-7}
 & \FullCell{\FullHeader{Full}} & \WholeCell{\WholeHeader{Whole mask}} & \FullCell{\FullHeader{Control}} & \FullCell{\FullHeader{Full}} & \WholeCell{\WholeHeader{Whole mask}} & \FullCell{\FullHeader{Control}} \\
\midrule
Qwen2.5-0.5B & \FullCell{36.2} & \WholeCell{54.4} & \FullCell{37.2} & \FullCell{44.4} & \WholeCell{60.3} & \FullCell{45.0} \\
Qwen2.5-1.5B & \FullCell{75.6} & \WholeCell{86.2} & \FullCell{74.7} & \FullCell{45.0} & \WholeCell{63.8} & \FullCell{46.3} \\
Qwen2.5-3B & \FullCell{76.9} & \WholeCell{85.0} & \FullCell{80.0} & \FullCell{44.7} & \WholeCell{63.2} & \FullCell{45.8} \\
Qwen2.5-7B & \FullCell{92.8} & \WholeCell{92.8} & \FullCell{91.2} & \FullCell{50.3} & \WholeCell{66.0} & \FullCell{49.6} \\
Qwen3-0.6B & \FullCell{69.1} & \WholeCell{78.4} & \FullCell{70.3} & \FullCell{50.4} & \WholeCell{62.3} & \FullCell{50.2} \\
Qwen3-1.7B & \FullCell{60.9} & \WholeCell{68.4} & \FullCell{59.7} & \FullCell{46.6} & \WholeCell{60.1} & \FullCell{46.5} \\
Qwen3-14B & \FullCell{97.8} & \WholeCell{98.8} & \FullCell{98.8} & \FullCell{48.8} & \WholeCell{66.1} & \FullCell{49.7} \\
Qwen3-30B-A3B-2507 & \FullCell{97.8} & \WholeCell{97.2} & \FullCell{98.1} & \FullCell{54.1} & \WholeCell{64.8} & \FullCell{52.9} \\
Qwen3-32B & \FullCell{96.6} & \WholeCell{98.1} & \FullCell{97.8} & \FullCell{52.1} & \WholeCell{68.3} & \FullCell{51.5} \\
Qwen3-4B & \FullCell{87.2} & \WholeCell{95.0} & \FullCell{87.5} & \FullCell{49.0} & \WholeCell{64.1} & \FullCell{48.0} \\
Qwen3-8B & \FullCell{94.7} & \WholeCell{93.8} & \FullCell{93.8} & \FullCell{50.2} & \WholeCell{63.8} & \FullCell{50.8} \\
Qwen3.8-27B$^{\dagger}$ & \FullCell{93.4} & \WholeCell{94.7} & \FullCell{92.8} & \FullCell{48.4} & \WholeCell{64.3} & \FullCell{48.6} \\
OLMo-2-7B & \FullCell{92.2} & \WholeCell{97.2} & \FullCell{91.2} & \FullCell{45.5} & \WholeCell{66.2} & \FullCell{45.4} \\
DeepSeek-V2-Lite & \FullCell{81.2} & \WholeCell{86.6} & \FullCell{81.2} & \FullCell{50.5} & \WholeCell{60.9} & \FullCell{52.4} \\
\addlinespace[2pt]
Gemma 4 26B-A4B & \FullCell{71.9} & \WholeCell{73.4} & \FullCell{72.5} & \FullCell{57.3} & \WholeCell{65.5} & \FullCell{56.6} \\
Gemma 4 31B & \FullCell{70.3} & \WholeCell{73.4} & \FullCell{69.4} & \FullCell{58.5} & \WholeCell{59.4} & \FullCell{58.6} \\
\TableLabel{granite-3.2-8b} & \FullCell{90.3} & \WholeCell{91.6} & \FullCell{91.2} & \FullCell{47.3} & \WholeCell{63.5} & \FullCell{47.2} \\
Llama-3.1-70B & \FullCell{92.5} & \WholeCell{95.3} & \FullCell{95.9} & \FullCell{50.7} & \WholeCell{68.7} & \FullCell{50.4} \\
Llama-3.2-1B & \FullCell{59.1} & \WholeCell{71.6} & \FullCell{56.6} & \FullCell{48.5} & \WholeCell{62.5} & \FullCell{48.4} \\
\addlinespace[2pt]
Llama-3.2-3B & \FullCell{82.5} & \WholeCell{90.6} & \FullCell{82.8} & \FullCell{47.6} & \WholeCell{64.9} & \FullCell{47.7} \\
Llama-3.1-8B & \FullCell{64.4} & \WholeCell{84.4} & \FullCell{68.1} & \FullCell{45.8} & \WholeCell{66.0} & \FullCell{46.5} \\
Mistral-7B-v0.3 & \FullCell{95.6} & \WholeCell{98.4} & \FullCell{97.2} & \FullCell{50.1} & \WholeCell{66.5} & \FullCell{50.5} \\
\bottomrule
\end{tabular}}
\end{table}
\subsection{Paired Prompt-Format Sensitivity}
\label{app:native_format}

We compare prompt formats on the same 320 controlled test events for Qwen3-4B, Qwen3-8B, and Llama-3.2-3B, holding
old/unrelated/later source strings, questions, answer options and the scoring
rule fixed. Raw input is their literal concatenation; native input wraps
that concatenation as a single user message, without a system message or
added instruction. Qwen thinking is disabled. Both conditions tokenize the complete text at once, which can differ from segmented tokenization at
boundaries; we therefore report the raw-text condition separately from the
segmented-tokenization measurements.

Each format prefills its history once, then evaluates full access,
old masking and equal-token unrelated masking, keeping history K/V
unchanged and summing each answer option's log probabilities.
All 320 events and eight wording patterns are included. This is a
retrospective robustness analysis on existing data: the protocol was
specified before evaluating the format comparison, with prior knowledge
of the results on the synthetic record dataset.

\begin{table}[!htbp]
\TableStyle
\caption{Format sensitivity on the same 320 controlled test events. The same scorer and literal text are used for both formats. Accuracy is percent; margin contrast is old-mask minus control-mask, with a paired 95\% interval.}
\label{tab:native_format}

{\setlength{\tabcolsep}{\dimexpr\tabcolsep*10/12\relax}
\begin{tabular}{cccccc}
\toprule
\TableHead{Model} & \TableHead{Format} & \FullCell{\FullHeader{Full}} & \WholeCell{\WholeHeader{Whole mask}} & \FullCell{\FullHeader{Control}} & \TableHead{Margin contrast} \\
\midrule
Qwen3-4B & raw & \FullCell{87.2} & \WholeCell{95.0} & \FullCell{87.5} & $2.985\;\TableSecondary{[2.722,\,3.253]}$ \\
Qwen3-4B & native & \FullCell{59.1} & \WholeCell{68.4} & \FullCell{63.1} & $1.683\;\TableSecondary{[1.173,\,2.233]}$ \\
Qwen3-8B & raw & \FullCell{94.7} & \WholeCell{93.8} & \FullCell{93.8} & $2.051\;\TableSecondary{[1.867,\,2.233]}$ \\
Qwen3-8B & native & \FullCell{58.1} & \WholeCell{70.0} & \FullCell{57.2} & $1.040\;\TableSecondary{[0.926,\,1.156]}$ \\
Llama-3.2-3B & raw & \FullCell{82.5} & \WholeCell{90.6} & \FullCell{82.8} & $1.636\;\TableSecondary{[1.474,\,1.803]}$ \\
Llama-3.2-3B & native & \FullCell{81.6} & \WholeCell{89.1} & \FullCell{82.2} & $1.539\;\TableSecondary{[1.389,\,1.689]}$ \\
\bottomrule
\end{tabular}}
\end{table}

Native-format margin effects remain positive in all three models.
Native-minus-raw contrasts are $-1.302$, $-1.011$, and $-0.098$;
their paired 95\% intervals are [$-1.804,-0.780$], [$-1.182,-0.833$],
and [$-0.264,0.068$]. With native chat formatting, old-source masking improves accuracy relative to control masking by
5.31/12.81/6.88 points, with intervals [2.19,8.44], [9.06,16.56],
and [3.75,10.00]. Under native chat, the lower confidence bounds for the accuracy difference between control masking and full access are
1.88/$-2.81$/$-0.94$ points, above the $-3$-point tolerance.
All intervals use 10,000 paired resamples of event groups within wording
pattern (seed 20260910), shared across formats. This supports the direction of the masking effect under native chat, but does not establish a general improvement in generated answers.

Appendix~\ref{app:v35_native_scoring} applies this paired protocol to five additional generators and compares all eight.

\subsection{Candidate Decisions and Input Format}
\label{app:v35_scoring}
The candidate-scoring studies evaluate 22 distinct checkpoints, including
Qwen3-32B, Llama-3.1-70B, Qwen3.8-27B, the two Gemma 4 checkpoints, and
17 others. Mistral Small 24B is evaluated separately and is outside this
count. Each checkpoint contributes one evaluation to the aggregate figures.
The format comparison uses three smaller generators and five larger ones.

These measurements ask how masking changes preference between supplied
answers and how that effect depends on prompt format. None directly measures free-generation
completeness. Appendices~\ref{app:v35_generation} and
\ref{app:v33_models} provide those separate comparisons.
Uncertainty estimates use the sampling unit and estimator specified for
each study. Intervals across 22 checkpoints are pointwise and do not
provide simultaneous coverage.

\subsubsection{Candidate-Margin Changes Across Models}
\label{app:v35_controlled_scoring}
The controlled experiment retains 320 events held out after the 80-event
set used to develop the metrics (Appendix~\ref{app:paraphrase_causal}). Each of
eight wording patterns supplies 40 matched groups. The estimator averages
patterns equally and resamples groups within each pattern, retaining the
paired full-access, source-mask, and control-mask scores for every group.

All 22 source-minus-control candidate-margin intervals are above zero;
point estimates span 0.539--2.985 nats. That is narrower than uniform
improvement in selected answers. Eighteen models meet the specified
control-preservation requirement and 21 show a positive margin change in
at least six of eight wording patterns. Seventeen meet all three specified
criteria together. These requirements describe the effect of this specific
intervention and control; they are not measures of overall model quality.

Several checkpoints qualify this pattern. Gemma 4 26B-A4B has a positive
aggregate margin change but only five positive wording-pattern effects,
short of the required six. Gemma 4 31B has six, but its control-minus-full
accuracy lower bound is below the allowed three-point loss. Its source-mask
candidate accuracy is higher than control accuracy, yet that does not
make the control behavior irrelevant. Llama-3.1-70B's candidate margin
rises while its correct-choice rate is slightly lower under source masking
than under the matched control. The selected answer and the margin can
therefore diverge before considering free generation.

\begin{table}[!htbp]
\WideTableStyle

\caption{Outcome entries give percentages, with the original count/denominator beneath. Candidate accuracy and margin changes for five checkpoints on 320 controlled events. Accuracy is supplied-candidate choice accuracy (\%). Margin change compares the source mask with its matched control. The final column is the one-sided 95\% lower bound for control-minus-full accuracy (pp); the preservation criterion requires this lower bound to exceed $-3$ percentage points.}
\label{tab:v35_controlled_added}
{\setlength{\tabcolsep}{\dimexpr\tabcolsep*12/14\relax}
\begin{tabular}{ccccccc}
\toprule
\multirow{2}{*}[-0.6ex]{\TableLabel{Model}} & \multicolumn{3}{c}{\TableHead{Candidate accuracy}} & \multicolumn{3}{c}{\TableHead{Source-access comparison}} \\
\cmidrule(lr){2-4}\cmidrule(l){5-7}
 & \FullCell{\FullHeader{Full}} & \FullCell{\FullHeader{Control}} & \WholeCell{\WholeHeader{Source}} & \TableHead{$\Delta m$ [95\% CI]} & \TableHead{Patterns} & \TableHead{$L_{.95}$} \\\midrule
Qwen3-32B & \FullCell{96.56} & \FullCell{97.81} & \WholeCell{98.12} & +1.87 \TableSecondary{[1.69,\,2.05]} & \AnswerPct{100}{8/8} & +0.00\\Qwen3.8-27B$^{\dagger}$ & \FullCell{93.44} & \FullCell{92.81} & \WholeCell{94.69} & +0.85 \TableSecondary{[0.76,\,0.94]} & \AnswerPct{100}{8/8} & $-$1.25\\Llama-3.1-70B & \FullCell{92.50} & \FullCell{95.94} & \WholeCell{95.31} & +0.66 \TableSecondary{[0.58,\,0.74]} & \AnswerPct{75}{6/8} & +1.88\\Gemma 4 26B-A4B & \FullCell{71.88} & \FullCell{72.50} & \WholeCell{73.44} & +0.54 \TableSecondary{[0.14,\,0.94]} & \AnswerPct{62.5}{5/8} & $-$2.50\\Gemma 4 31B & \FullCell{70.31} & \FullCell{69.38} & \WholeCell{73.44} & +1.22 \TableSecondary{[0.79,\,1.67]} & \AnswerPct{75}{6/8} & $-$4.06\\
\bottomrule
\end{tabular}}
\end{table}

\subsubsection{Separating Each Mask from Full Access}
\label{app:v23_margin}
With $m=\log p(y_{\mathrm{current}})-\log p(y_{\mathrm{old}})$, define
$\Delta_A=m_{\mathrm{source\ mask}}-m_{\mathrm{full}}$ and
$\Delta_N=m_{\mathrm{control\ mask}}-m_{\mathrm{full}}$.
Their difference is the matched contrast. Table~\ref{tab:v23_absolute_margins}
uses exactly the 22-checkpoint inventory of
Figure~\ref{fig:model_causal_matrix}. All rows use the complete 320-event
outputs and within-pattern bootstrap estimates. The five checkpoints
listed separately above and the other 17 use the same contrast definitions.
This decomposition shows whether an apparent benefit reflects the source
mask, the control, or both, without treating a larger relative margin as
an increase in the current candidate's own probability.

\begin{table}[!htbp]
\TableStyle

\caption{Absolute-margin decomposition for the same 22-checkpoint inventory as Figure~\ref{fig:model_causal_matrix}. $m_F$ is the full-access margin; $\Delta_A$ and $\Delta_N$ are source-mask and control-mask changes relative to it. Brackets give paired pointwise 95\% intervals in nats. The final column is their difference.}
\label{tab:v23_absolute_margins}
{\setlength{\tabcolsep}{\dimexpr\tabcolsep*8/10\relax}
\begin{tabular}{ccccc}
\toprule
\TableHead{Model} & \FullCell{\FullHeader{$m_F$}} & \WholeCell{\WholeHeader{$\Delta_A$ [95\% CI]}} & \FullCell{\FullHeader{$\Delta_N$ [95\% CI]}} & \TableHead{$\Delta_A-\Delta_N$} \\
\midrule
Qwen2.5-0.5B & \FullCell{$-$1.05} & \WholeCell{+1.51 \TableSecondary{[1.32,\,1.70]}} & \FullCell{+0.15 \TableSecondary{[0.09,\,0.20]}} & +1.36 \\
Qwen3-0.6B & \FullCell{1.21} & \WholeCell{+1.19 \TableSecondary{[1.03,\,1.35]}} & \FullCell{+0.05 \TableSecondary{[0.01,\,0.08]}} & +1.14 \\
Llama-3.2-1B & \FullCell{0.37} & \WholeCell{+1.10 \TableSecondary{[0.96,\,1.24]}} & \FullCell{$-$0.02 \TableSecondary{[$-$0.06,\,0.02]}} & +1.12 \\
Qwen2.5-1.5B & \FullCell{1.89} & \WholeCell{+1.16 \TableSecondary{[1.03,\,1.31]}} & \FullCell{$-$0.03 \TableSecondary{[$-$0.06,\,0.01]}} & +1.19 \\
Qwen3-1.7B & \FullCell{1.29} & \WholeCell{+0.94 \TableSecondary{[0.74,\,1.16]}} & \FullCell{+0.05 \TableSecondary{[$-$0.01,\,0.11]}} & +0.90 \\
Qwen2.5-3B & \FullCell{3.73} & \WholeCell{+2.13 \TableSecondary{[1.89,\,2.36]}} & \FullCell{+0.02 \TableSecondary{[$-$0.07,\,0.11]}} & +2.11 \\
\addlinespace[2pt]
Llama-3.2-3B & \FullCell{1.53} & \WholeCell{+1.83 \TableSecondary{[1.67,\,2.00]}} & \FullCell{+0.20 \TableSecondary{[0.14,\,0.25]}} & +1.64 \\
Qwen3-4B & \FullCell{4.88} & \WholeCell{+3.20 \TableSecondary{[2.94,\,3.46]}} & \FullCell{+0.21 \TableSecondary{[0.13,\,0.29]}} & +2.99 \\
Mistral-7B-v0.3 & \FullCell{4.22} & \WholeCell{+2.19 \TableSecondary{[2.02,\,2.36]}} & \FullCell{+0.21 \TableSecondary{[0.16,\,0.26]}} & +1.98 \\
OLMo-2-7B & \FullCell{6.38} & \WholeCell{+2.80 \TableSecondary{[2.59,\,3.02]}} & \FullCell{+0.37 \TableSecondary{[0.28,\,0.45]}} & +2.44 \\
Qwen2.5-7B & \FullCell{6.85} & \WholeCell{+1.59 \TableSecondary{[1.27,\,1.91]}} & \FullCell{$-$0.09 \TableSecondary{[$-$0.20,\,0.03]}} & +1.68 \\
Llama-3.1-8B & \FullCell{0.44} & \WholeCell{+1.45 \TableSecondary{[1.36,\,1.54]}} & \FullCell{+0.13 \TableSecondary{[0.11,\,0.16]}} & +1.31 \\
Granite-3.2-8B & \FullCell{3.73} & \WholeCell{+2.10 \TableSecondary{[1.93,\,2.27]}} & \FullCell{+0.08 \TableSecondary{[0.04,\,0.11]}} & +2.03 \\
Qwen3-8B & \FullCell{5.79} & \WholeCell{+2.41 \TableSecondary{[2.22,\,2.60]}} & \FullCell{+0.36 \TableSecondary{[0.30,\,0.41]}} & +2.05 \\
Qwen3-14B & \FullCell{7.24} & \WholeCell{+2.80 \TableSecondary{[2.57,\,3.03]}} & \FullCell{+0.10 \TableSecondary{[0.05,\,0.14]}} & +2.70 \\
DeepSeek-V2-Lite & \FullCell{3.98} & \WholeCell{+2.79 \TableSecondary{[2.58,\,2.99]}} & \FullCell{+0.03 \TableSecondary{[$-$0.04,\,0.09]}} & +2.76 \\
Qwen3-30B-A3B-2507 & \FullCell{7.33} & \WholeCell{+1.45 \TableSecondary{[1.33,\,1.57]}} & \FullCell{+0.14 \TableSecondary{[0.08,\,0.19]}} & +1.31 \\
Qwen3-32B & \FullCell{3.73} & \WholeCell{+1.63 \TableSecondary{[1.46,\,1.80]}} & \FullCell{$-$0.24 \TableSecondary{[$-$0.30,\,$-$0.18]}} & +1.87 \\
Llama-3.1-70B & \FullCell{2.51} & \WholeCell{+0.79 \TableSecondary{[0.72,\,0.86]}} & \FullCell{+0.13 \TableSecondary{[0.10,\,0.16]}} & +0.66 \\
Qwen3.8-27B$^{\dagger}$ & \FullCell{2.61} & \WholeCell{+0.86 \TableSecondary{[0.77,\,0.95]}} & \FullCell{+0.01 \TableSecondary{[$-$0.02,\,0.04]}} & +0.85 \\
\addlinespace[2pt]
Gemma 4 26B-A4B & \FullCell{2.56} & \WholeCell{+0.51 \TableSecondary{[0.17,\,0.86]}} & \FullCell{$-$0.03 \TableSecondary{[$-$0.37,\,0.31]}} & +0.54 \\
Gemma 4 31B & \FullCell{4.61} & \WholeCell{+1.08 \TableSecondary{[0.68,\,1.49]}} & \FullCell{$-$0.14 \TableSecondary{[$-$0.45,\,0.17]}} & +1.22 \\
\bottomrule
\end{tabular}}
\end{table}

\subsubsection{Native Formatting Does Not Move Every Model in the Same Direction}
\label{app:v35_native_scoring}
The paired format comparison follows Appendix~\ref{app:native_format}:
identical literal records, question, options, and scorer are either
presented as raw text or wrapped in one native user message, with no
added system instruction. Thinking is disabled where applicable. Both
formats tokenize the entire text once. This differs from the segmented
assembly in the broad raw-prompt study, so raw cells from the two
protocols are reported separately. For Qwen3.8, full-access candidate
choices differ on two events between those raw assembly procedures.

Gemma 4 31B's full-access accuracy rises from 70.31\% to 99.69\%, while
Gemma 4 26B-A4B rises from 71.88\% to 88.75\%. The reverse occurs for
Qwen3-32B (96.56\% to 85.94\%) and Llama-3.1-70B (92.50\% to 76.25\%).
Native formatting is therefore an experimental factor, not a uniformly
beneficial correction to raw results. Model family, size, and training
also vary together; these comparisons do not isolate any one of them.

\begin{table}[!htbp]
\WideTableStyle

\caption{Paired raw/native format comparison on the same 320 events. Both formats tokenize their complete text at once; their raw results are separate from segmented-tokenization results. Accuracy columns are percentages; $L_{.95}$ is the control-minus-full accuracy lower bound in pp.}
\label{tab:v35_native_added}
{\setlength{\tabcolsep}{\dimexpr\tabcolsep*12/14\relax}
\begin{tabular}{ccccccc}
\toprule
\multirow{2}{*}[-0.6ex]{\TableLabel{Model}} & \multirow{2}{*}[-0.6ex]{\TableLabel{Format}} & \multicolumn{3}{c}{\TableHead{Candidate accuracy}} & \multicolumn{2}{c}{\TableHead{Source-access comparison}} \\
\cmidrule(lr){3-5}\cmidrule(l){6-7}
 &  & \FullCell{\FullHeader{Full}} & \FullCell{\FullHeader{Control}} & \WholeCell{\WholeHeader{Source}} & \TableHead{$\Delta m$ [95\% CI]} & \TableHead{$L_{.95}$} \\
\midrule
Qwen3-32B & Raw & \FullCell{96.56} & \FullCell{97.81} & \WholeCell{98.12} & +1.87 \TableSecondary{[1.69,\,2.05]} & +0.00 \\
Qwen3-32B & Native & \FullCell{85.94} & \FullCell{85.00} & \WholeCell{90.31} & +2.65 \TableSecondary{[2.33,\,2.97]} & $-$1.88 \\
Qwen3.8-27B$^{\dagger}$ & Raw & \FullCell{94.06} & \FullCell{92.81} & \WholeCell{94.69} & +0.84 \TableSecondary{[0.75,\,0.94]} & $-$2.19 \\
Qwen3.8-27B$^{\dagger}$ & Native & \FullCell{95.00} & \FullCell{93.12} & \WholeCell{95.00} & +1.92 \TableSecondary{[1.70,\,2.14]} & $-$3.75 \\
Llama-3.1-70B & Raw & \FullCell{92.50} & \FullCell{95.94} & \WholeCell{95.31} & +0.66 \TableSecondary{[0.58,\,0.74]} & +1.88 \\
Llama-3.1-70B & Native & \FullCell{76.25} & \FullCell{76.88} & \WholeCell{80.00} & +1.62 \TableSecondary{[1.47,\,1.79]} & +0.00 \\
Gemma 4 26B-A4B & Raw & \FullCell{71.88} & \FullCell{72.50} & \WholeCell{73.44} & +0.54 \TableSecondary{[0.14,\,0.94]} & $-$2.50 \\
Gemma 4 26B-A4B & Native & \FullCell{88.75} & \FullCell{89.38} & \WholeCell{89.06} & +1.73 \TableSecondary{[1.31,\,2.17]} & $-$0.63 \\
Gemma 4 31B & Raw & \FullCell{70.31} & \FullCell{69.38} & \WholeCell{73.44} & +1.22 \TableSecondary{[0.79,\,1.67]} & $-$4.06 \\
Gemma 4 31B & Native & \FullCell{99.69} & \FullCell{100.00} & \WholeCell{99.06} & +0.78 \TableSecondary{[0.60,\,0.97]} & +0.00 \\
\bottomrule
\end{tabular}}
\end{table}
\begin{figure}[!htbp]
\centering
\includegraphics[width=\linewidth]{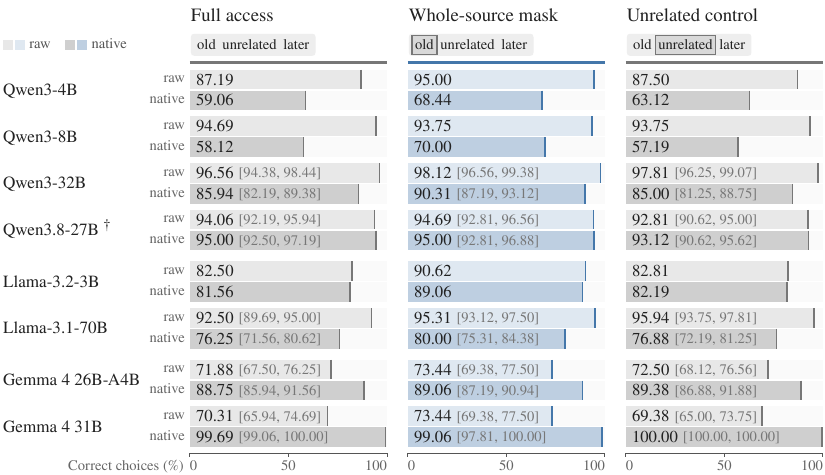}
\caption{Supplied-candidate accuracy (\%) on the same 320 events; each model's raw-text row (light) sits above its native chat-template row (dark). Blue is whole-source masking; full access and the equal-token unrelated-mask control are gray, as in the tables. Chips box the hidden span. Bars start at 0; chance is 50\%. Paired 95\% intervals follow the values for five checkpoints; Qwen3-4B, Qwen3-8B, and Llama-3.2-3B provide point estimates only. The comparison changes input format, not the answer candidates. $\dagger$ Qwen3.8 retains recurrent states.}
\label{fig:v35_native_scoring}
\end{figure}

Every added model retains a positive source-minus-control margin change
under native formatting, but this is again distinct from better choices.
Gemma 4 31B's native control answers every event correctly, while source
masking introduces three obsolete choices. Gemma 4 26B-A4B has one net
additional obsolete choice. Qwen3.8's native control-minus-full lower
bound is $-3.75$ points, so it fails the original $-3$-point preservation
criterion even though its raw condition passed. A failed criterion means
that preservation is not established at the declared tolerance, not that
an arbitrary harm threshold has been proved. The paired format contrast
and obsolete-choice estimates are shown together in
Table~\ref{tab:v35_native_differences}.

\begin{table}[!htbp]
\WideTableStyle
\begin{minipage}{0.84\linewidth}\centering

\caption{Format-dependent changes. The first column subtracts the raw mask-minus-control margin contrast from its native-format counterpart, retaining each event pair. The second reports source-minus-control obsolete-choice change under native formatting; negative means fewer obsolete choices.}
\label{tab:v35_native_differences}
{\setlength{\tabcolsep}{\dimexpr\tabcolsep*4/6\relax}
\begin{tabular}{ccc}
\toprule
\TableHead{Model} & \TableHead{Native minus raw $\Delta m$ [95\% CI]} & \TableHead{Native obsolete change (pp)} \\
\midrule
Qwen3-32B & +0.78 \TableSecondary{[0.51,\,1.04]} & $-$5.31 \TableSecondary{[$-$7.81,\,$-$3.12]} \\
Qwen3.8-27B$^{\dagger}$ & +1.08 \TableSecondary{[0.88,\,1.27]} & $-$1.87 \TableSecondary{[$-$4.38,\,0.63]} \\
Llama-3.1-70B & +0.97 \TableSecondary{[0.85,\,1.09]} & $-$3.12 \TableSecondary{[$-$5.00,\,$-$1.56]} \\
Gemma 4 26B-A4B & +1.19 \TableSecondary{[0.60,\,1.79]} & +0.31 \TableSecondary{[$-$1.88,\,2.50]} \\
Gemma 4 31B & $-$0.44 \TableSecondary{[$-$0.93,\,0.04]} & +0.94 \TableSecondary{[0.00,\,2.19]} \\
\bottomrule
\end{tabular}}
\end{minipage}
\end{table}
\FloatBarrier

\subsubsection{Natural-Source Candidate Decisions and the Gemma 4 31B Exception}
\label{app:v35_oaks_scoring}
This panel retains 336 OAKS transitions from 116 questions in five held-out
books (Appendix~\ref{app:oaks_transfer}). Several events can come from one
question. Within each book, event contributions are averaged and whole
questions are resampled with all their events; the final estimate averages
books equally. Consequently, its percentages are not simply counts divided
by 336 and cannot be substituted for the 116-question generated-answer
percentages in Appendix~\ref{app:v35_oaks}.

Twenty-one of 22 checkpoints meet the specified requirements of fewer
obsolete choices, preserved control accuracy, and improvement across the
required number of books. The direction is favorable in 107 of 110
model--book cells. All 22 preserve control accuracy at the specified
criterion. Gemma 4 31B is the exception to reliable obsolete-choice
reduction: its change is $-0.80$ points with interval $[-4.10,2.38]$.
It improves in only two books, ties in two, and reverses direction in one,
despite positive margin changes in all five. Positive candidate-margin changes therefore do not establish improved
answer choices across books.

\begin{table}[!htbp]
\TableStyle

\caption{OAKS candidate decisions on 336 events from 116 questions. Percentages are equal-weight averages of within-book event rates, not raw totals divided by 336. The obsolete-choice change is source minus control (pp); intervals resample whole questions within each of five books.}
\label{tab:v35_oaks_candidates_added}
{\setlength{\tabcolsep}{\dimexpr\tabcolsep*10/12\relax}
\begin{tabular}{cccccc}
\toprule
\TableHead{Model} & \FullCell{\FullHeader{Full}} & \FullCell{\FullHeader{Control}} & \WholeCell{\WholeHeader{Source}} & \TableHead{Obsolete change [95\% CI]} & \TableHead{Books} \\
\midrule
Qwen3-32B & \FullCell{52.14} & \FullCell{51.49} & \WholeCell{68.26} & $-$16.77 \TableSecondary{[$-$21.16,\,$-$12.78]} & 5/5 \\
Qwen3.8-27B$^{\dagger}$ & \FullCell{48.41} & \FullCell{48.59} & \WholeCell{64.29} & $-$15.70 \TableSecondary{[$-$20.10,\,$-$11.53]} & 5/5 \\
Llama-3.1-70B & \FullCell{50.72} & \FullCell{50.39} & \WholeCell{68.72} & $-$18.33 \TableSecondary{[$-$23.28,\,$-$13.81]} & 5/5 \\
Gemma 4 26B-A4B & \FullCell{57.27} & \FullCell{56.55} & \WholeCell{65.49} & $-$8.94 \TableSecondary{[$-$12.32,\,$-$5.84]} & 5/5 \\
Gemma 4 31B & \FullCell{58.49} & \FullCell{58.56} & \WholeCell{59.36} & $-$0.80 \TableSecondary{[$-$4.10,\,2.38]} & 2/5 \\
\bottomrule
\end{tabular}}
\end{table}
\begin{table}[!htbp]
\WideTableStyle
\begin{minipage}{0.78\linewidth}\centering

\caption{Gemma 4 31B book-specific point estimates. Source masking increases margins in every book, but obsolete choices decrease in two, remain unchanged in two, and increase in one. These descriptive book points have no separate uncertainty intervals.}
\label{tab:v35_gemma31_books}
{\setlength{\tabcolsep}{\dimexpr\tabcolsep*6/8\relax}
\begin{tabular}{cccc}
\toprule
\TableHead{Book} & \TableHead{Obsolete change (pp)} & \TableHead{$\Delta m$ (nats)} & \TableHead{Control minus full (pp)} \\
\midrule
P12 & +1.08 & +4.30 & +3.23 \\
P22 & $-$1.23 & +0.18 & $-$4.94 \\
P25 & +0.00 & +5.21 & +0.00 \\
P27 & $-$3.85 & +1.05 & $-$1.92 \\
P37 & +0.00 & +5.01 & +4.00 \\
\bottomrule
\end{tabular}}
\end{minipage}
\end{table}

The contrast concerns which of two supplied answers has higher sequence
probability after a narrative update. It does not require generating the
answer in the requested format. Gemma 4 26B-A4B's clear candidate-choice
improvement here therefore does not contradict its frequent format and
completion failures in the separate generation experiment. The measurements expose different behaviors under different response
protocols. Appendix~\ref{app:absolute_accuracy} reports the absolute
full-access and control baselines behind these contrasts.
\FloatBarrier

These comparisons distinguish candidate preference, selected-answer
preservation and prompt sensitivity. Appendix~\ref{app:v35_fixed_scoring}
examines fixed-score recognition alongside the other recognition studies.

\FloatBarrier
\section{Recognizing Which Memories Have Changed}
\label{app:recognition_studies}

These comparisons ask whether record representations identify that a later
statement replaces an earlier one, and whether the resulting scores help
choose an intervention. Supplied record pairs isolate relation recognition;
they do not include the search needed to locate those records in a larger
memory.

\subsection{Recognizing Replacement from Record Representations}
\label{app:decisive_audit}

The dataset of record pairs without questions contains 5,328 distinct text pairs and 9,216 labeled comparisons.  Its held-out record-pair evaluation contributes 216 connected dependency
components across groups with the same or different aliases; the controlled-history evaluation contributes
400 history components across the five update families.  Each distinct pair is scored once; repeated occurrences share its score.
Resampling preserves these dependencies, so repeated labeled comparisons do
not count as independent AUC observations.

On the record-pair dataset, the KV-based AUC minus the surface-baseline AUC is $-0.054$, $0.040$,
$0.063$, and $0.097$ in model order; controlled-history differences are
$-0.002$, $0.000$, $-0.007$, and $-0.003$. We require an advantage on both
sets, using the smaller difference and a lower confidence bound strictly
above zero. No model satisfies this requirement, including Qwen3-8B with
its zero controlled-history difference. However, the token and character
baselines each reach AUC 1.0 on every controlled-history stratum, so a
strict advantage on that panel is mathematically impossible. Failure of
this conjunction does not establish absence of relational information.
In contrast, the AUC lower bounds for entity--attribute matching and KV-based replacement recognition exceed $0.60$ for the original four models
(Table~\ref{tab:v39_native_pairs}).

\subsubsection{Fixed Scores on Role-Matched Record Pairs}
\label{app:paraphrase_robustness}
The role-matched inputs in Appendix~\ref{app:paraphrase_causal} hold
both candidate values fixed while exchanging their current and rejected roles.
The strongest surface-similarity baseline, allowing either score direction, has macro-AUC 0.579,
with no family above 0.589.

Across 22 checkpoints, K-based entity--attribute matching AUC ranges from $0.558$ to $0.879$;
20 meet its specified lower-bound criterion, with both Gemma models as exceptions.
Replacement recognition using the fixed V distance ranges from $0.454$ to $0.493$.
None meets the $0.60$ criterion or exceeds surface similarity. These results concern the fixed
pooling and comparison rule; Appendix~\ref{app:v35_representation} evaluates alternative learned representation readouts.
\begin{figure}[!ht]
  \centering
  \includegraphics[width=\linewidth]{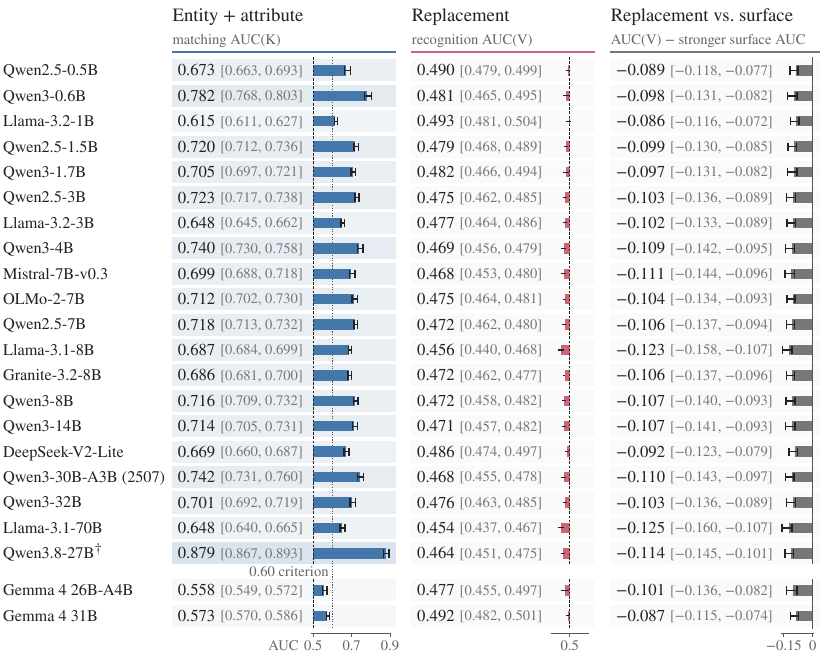}
  \caption{Fixed readouts on 1,200 role-matched record pairs across 22 checkpoints. Columns compare entity--attribute matching, replacement recognition, and the replacement score's advantage over the stronger surface baseline. Each cell prints the estimate and paired 95\% interval; bars and whiskers share one scale and start at chance AUC (dashed) or zero. Matching succeeds under the specified criterion (dotted line) in 20 models, but both Gemmas are exceptions. These results concern the fixed readouts. $\dagger$ Qwen3.8 uses full-attention K/V coordinates.}
  \label{fig:model_coordinate_matrix}
\end{figure}

\subsubsection{Fixed-Score Details and Direction Sensitivity}
\label{app:v35_fixed_scoring}
The representation experiment uses 1,200 supplied pairs from 400 matched
groups across five update families. Records are encoded separately. A
fixed comparison of K vectors is evaluated for matching the same entity
and attribute; a fixed pooled V distance is evaluated for whether the
value was replaced, restricting that task to matching entity--attribute
pairs. The record variants preserve the lexical inventory while exchanging
which candidate is current. AUC measures how often a randomly selected
positive pair receives a larger score than a negative pair, counting ties
as one half. The two tasks have different labels and eligible pairs.

Results average family AUCs equally and resample complete matched groups
within family. Both recognition criteria require the one-sided 95\% AUC
lower bound to exceed 0.60. A stronger interpretation additionally requires
the V score to outperform the strongest token/character surface baseline
on the same bootstrap draws. The surface baseline may use either score direction and reaches
AUC 0.579 when the five families receive equal weight.

\begin{table}[!htbp]
\WideTableStyle
\caption{Fixed readouts and retrospective reversal of the V-distance direction on the same 1,200 pairs from 400 groups. AUC cells include the one-sided 95\% lower bound; the V-minus-surface cell has a paired two-sided 95\% interval. Recognition requires the lower bound to exceed 0.60. The surface baseline is 0.579.}
\label{tab:v35_fixed_added}
\begin{tabular}{ccccc}
\toprule
\multirow{2}{*}{\TableLabel{Model}} & \multicolumn{3}{c}{\TableHead{Specified readouts}} & \TableHead{Retrospective} \\
\cmidrule(lr){2-4}\cmidrule(l){5-5}
 & \TableHead{K matching} & \TableHead{V replacement} & \TableHead{V minus surface} & \TableHead{Reversed V} \\
\midrule
Qwen3-32B & 0.701 \TableSecondary{[0.694]} & 0.476 \TableSecondary{[0.464]} & $-$0.103 \TableSecondary{[$-$0.136,\,$-$0.089]} & 0.524 \TableSecondary{[0.517]} \\
Qwen3.8-27B$^{\dagger}$ & 0.879 \TableSecondary{[0.869]} & 0.464 \TableSecondary{[0.453]} & $-$0.114 \TableSecondary{[$-$0.145,\,$-$0.101]} & 0.536 \TableSecondary{[0.527]} \\
Llama-3.1-70B & 0.648 \TableSecondary{[0.641]} & 0.454 \TableSecondary{[0.440]} & $-$0.125 \TableSecondary{[$-$0.160,\,$-$0.107]} & 0.546 \TableSecondary{[0.535]} \\
Gemma 4 26B-A4B & 0.558 \TableSecondary{[0.551]} & 0.477 \TableSecondary{[0.458]} & $-$0.101 \TableSecondary{[$-$0.136,\,$-$0.082]} & 0.523 \TableSecondary{[0.506]} \\
Gemma 4 31B & 0.573 \TableSecondary{[0.571]} & 0.492 \TableSecondary{[0.483]} & $-$0.087 \TableSecondary{[$-$0.115,\,$-$0.074]} & 0.508 \TableSecondary{[0.500]} \\
\bottomrule
\end{tabular}
\end{table}

K-based matching meets its criterion in 20 of 22 checkpoints. The two Gemma
checkpoints have point estimates of 0.558 and 0.573, both below 0.60.
Qwen3.8 reaches
0.879 on K matching while its fixed V replacement score is 0.464. These
are different task/readout combinations; a strong result on one does not
supply a replacement detector for the other.

The fixed V replacement AUCs span 0.454--0.493, with no model meeting the
criterion or exceeding the surface baseline. Reversing the score direction
is reported as a retrospective diagnostic, separately from the specified
primary direction. For the five larger checkpoints it yields 0.508--0.546, still below
both the 0.60 criterion and the surface baseline. This tests one fixed
pooling and distance rule. It neither proves that K/V lacks relation
information nor rules out a different decoder trained on independent data.
The learned-probe experiment uses a different representation and
selection protocol.

\subsection{Attention, Lexical Relevance, and Leave-One-Out Signals}
\label{app:signal_details}

We measure question relevance as the fraction of clusters in which a score is higher for a question requiring the target information than for an unrelated question. The attention, lexical, and leave-one-out
ranges cover four models at 8, 32, and 128 continuation tokens; the Qwen3-8B
V-distance score is reported separately. This score compares the V representations of the earlier and later records after encoding them together without the question.

\begin{table}[!htbp]
\TableStyle
\begin{minipage}{0.82\linewidth}\centering
  \caption{Replacement-recognition AUC and the fraction of pairs receiving a higher score for the relevant question. Ranges span four
  models and three budgets; the Qwen3-8B row gives a 95\% cluster-bootstrap interval.}
  \label{tab:signal_boundary}

  {\setlength{\tabcolsep}{\dimexpr\tabcolsep*4/6\relax}
\begin{tabular}{c c c}
    \toprule
\TableHead{Signal} & \TableHead{Replacement AUC} & \TableHead{Question relevance} \\
\midrule
    Lexical overlap & 0.500 & 1.000 \\
    Attention to the old source & 0.356--0.525 & 0.316--0.953 \\
    Signed leave-one-out & 0.466--0.724 & 0.350--0.636 \\
    \midrule
    Qwen3-8B V distance & $0.999\;\TableSecondary{[0.998,\,1.000]}$ & -- \\
    \bottomrule
  \end{tabular}}
\end{minipage}
\end{table}

Lexical overlap identifies which question requires the target information in this construction, but recognizes replacement at chance. Attention and signed leave-one-out results
vary across models and continuation budgets.
The analysis of replacement recognition within each update family evaluates 39 model--budget
combinations: 12 lexical scores, 12 attention scores, 12 signed
leave-one-out scores, and three repeated-budget Qwen3-8B V-distance scores.
Twenty-six satisfy the criterion for poor replacement recognition, eight support the opposite
conclusion, and five are inconclusive. The eight contrary results comprise
five leave-one-out comparisons and the three Qwen3-8B V-distance scores.
Of 36 question-relevance comparisons, lexical overlap satisfies the criterion in
12/12, attention in 4/12, and leave-one-out sensitivity in 1/12.

The Qwen3-8B V-distance score has unconditional AUC 0.8542 and conditional AUC
0.99909 after stratifying by update family. Its AUC is 1.000 for location,
endpoint, deadline, and incident cause, and 0.99547 for preference.
Token-Jaccard and character-sequence distance each reach 1.000 in every
family. This V-distance score therefore does not improve on surface similarity
under this construction.

\subsection{Text Similarity Perfectly Separates the Controlled-History Pairs}
\label{app:v39_native_pairs}

The 5,328-pair study is distinct from the 1,200-pair role-matched experiment.
It evaluates record pairs without a question, using native forward paths
and the connected-component estimator. All nine models use the same
pairs and labels, with the comparison criteria in
Appendix~\ref{app:decisive_audit}.

\begin{table}[!htbp]
\WideTableStyle
\caption{Native record-pair recognition across nine models, including the original four-model comparison. Matching and relation cells show the weaker-panel AUC and its one-sided 95\% lower bound. The surface comparisons retain both panels; the last column retains the lower bound for the smaller difference where it was reported.}
\label{tab:v39_native_pairs}
\begin{tabular}{cccccc}
\toprule
\multirow{2}{*}{\TableLabel{Model}} & \multicolumn{2}{c}{\TableHead{AUC [one-sided lower bound]}} & \multicolumn{3}{c}{\TableHead{Relation AUC minus surface baseline}} \\
\cmidrule(lr){2-3}\cmidrule(l){4-6}
 & \TableHead{Matching} & \TableHead{Relation} & \TableHead{Intrinsic} & \TableHead{History} & \TableHead{$L_{.95}(\min\Delta)$} \\
\midrule
Qwen3-4B & 0.974 \TableSecondary{[0.968]} & 0.696 \TableSecondary{[0.682]} & $-$0.054 & $-$0.002 & $-$0.076 \\
Qwen3-8B & 0.968 \TableSecondary{[0.963]} & 0.790 \TableSecondary{[0.766]} & +0.040 & 0.000 & 0.000 \\
\addlinespace[3pt]
Llama-3.2-3B & 0.959 \TableSecondary{[0.951]} & 0.813 \TableSecondary{[0.792]} & +0.063 & $-$0.007 & $-$0.010 \\
Llama-3.1-8B & 0.950 \TableSecondary{[0.941]} & 0.847 \TableSecondary{[0.830]} & +0.097 & $-$0.003 & $-$0.006 \\
\addlinespace[3pt]
Qwen3-32B & 0.931 \TableSecondary{[0.919]} & 0.785 \TableSecondary{[0.766]} & +0.035 & 0.000 & \textemdash \\
Qwen3.8-27B$^{\dagger}$ & 0.997 \TableSecondary{[0.995]} & 0.855 \TableSecondary{[0.833]} & +0.105 & $-$0.036 & \textemdash \\
\addlinespace[3pt]
Gemma 4 26B-A4B & 0.726 \TableSecondary{[0.709]} & 0.709 \TableSecondary{[0.683]} & $-$0.041 & $-$0.239 & \textemdash \\
Gemma 4 31B & 0.755 \TableSecondary{[0.734]} & 0.726 \TableSecondary{[0.705]} & $-$0.024 & $-$0.027 & \textemdash \\
\addlinespace[3pt]
Llama-3.1-70B & 0.770 \TableSecondary{[0.752]} & 0.706 \TableSecondary{[0.687]} & $-$0.044 & $-$0.006 & \textemdash \\
\bottomrule
\end{tabular}
\par\smallskip{\footnotesize\TableNoteAlign Each model uses 5,328 distinct pairs and 9,216 labeled comparisons. Resampling preserves the 216 intrinsic and 400 history dependency components. The smaller-difference lower bound was reported for the original four models only; a dash does not imply zero. Differences are rounded to three decimals, with ties away from zero. $\dagger$ Qwen3.8 pools full-attention K/V and retains recurrent states.\TableNoteEnd}
\end{table}

The matching lower bound exceeds 0.60 in all nine models and the
relation lower bound exceeds 0.60 in all nine models. These readouts distinguish
the constructed labels above the specified threshold.
The stricter comparison requires
native relation scores to exceed the surface baseline on both panels.
Every model fails that comparison, but the controlled-history token and
character baselines both have AUC 1.0 by construction. A strictly positive
native-minus-surface lower bound is consequently unattainable on that panel.
The failure of this conjunction cannot discriminate between models that
encode more or less useful relational information.

The intrinsic panel remains informative because its surface baseline is
0.75, the macro-average of the different-alias stratum (surface AUC 0.5)
and the same-alias stratum (1.0). For every model the intrinsic panel is
the weaker one, so the relation column reports its AUC. Five of nine
models exceed the baseline: Qwen3.8 reaches 0.855, Llama-3.1-8B 0.847,
Llama-3.2-3B 0.813, Qwen3-8B 0.790 and Qwen3-32B 0.785. Because the
baseline is fixed by construction, the printed one-sided lower bounds
minus 0.75 bound their excess at $+0.083$, $+0.080$, $+0.042$, $+0.016$
and $+0.016$. Qwen3-4B (0.696), Gemma 4 26B-A4B (0.709), Gemma 4 31B
(0.726) and Llama-3.1-70B (0.706) stay below it. In every model the
native score exceeds the surface baseline only among different-alias
pairs (0.619--0.766 versus 0.5) and falls below it among same-alias pairs
(0.758--0.961 versus 1.0). Llama-3.1-70B reaches only 0.706 on intrinsic
pairs despite its controlled-history relation AUC of 0.994, and Gemma 4
31B reaches 0.726 and 0.973. These differences show
why the panels must be reported separately: success on easily distinguished
history pairs does not establish generalization to the intrinsic pairs,
and a pooled failure does not erase a positive result on one panel.

This is still a fixed V-based score derived from a specified representation and
comparison rule. Its results need not agree with learned joint-KV or
selected-head readouts on the separate role-matched study. Neither a
negative surface comparison here nor a low pooled V-distance AUC elsewhere
establishes that the model cannot represent the relation under another
readout. The results apply to the tested scores and pair constructions, with
uncertainty calculated from their shared dependency groups.

\subsection{Exact Scores and Detector Prompts}
\label{app:exact_definitions}

\paragraph{Key similarity for entity--attribute matching.}
A and B are encoded separately, each at positions $0,\dots,n_X-1$. Let $k_{X,i}^{\ell,h}$ denote token $i$'s key in
layer $\ell$ and KV head $h$ as stored in the cache, that is, after the
rotary position embedding and after q/k normalization where the model
applies it. Both records therefore share the same position offsets, and
key comparisons are position-matched.
Here $L$ is the number of evaluated attention layers, $H_\ell$ the number of
KV heads in layer $\ell$, $n_X$ the token count of record $X$, and $d$ the
dimension of each head's value vector.
Normalize each key by $\max(\lVert k_{X,i}^{\ell,h}\rVert_2,10^{-8})$
and let $c_{ij}^{\ell,h}$ be the dot product of the normalized A and B keys.
We compute cosine similarity by finding each token's closest match in the other record, then averaging in both directions:
\begin{equation}
 s_K(A,B)=\frac{1}{L}\sum_{\ell=1}^{L}\frac{1}{2H_\ell}
 \sum_{h=1}^{H_\ell}
 \left[
 \frac{1}{n_A}\sum_i\max_j c_{ij}^{\ell,h}
 +\frac{1}{n_B}\sum_j\max_i c_{ij}^{\ell,h}
 \right].
\end{equation}
The implementation clips each layer's score to $[-1,1]$ before averaging
layers. Larger scores predict that the records concern the same entity and attribute. Best matches are computed
within a head and need not be one-to-one. The V-distance below instead averages the token vectors before computing the distance between records.

\paragraph{V-distance between the earlier and later records.}
Let $V_X^{\ell,h}\in\mathbb R^{n_X\times d}$ be the V rows of
record $X\in\{A,B\}$ in layer $\ell$ and head $h$, encoded independently
with positions starting at zero and without the query. With
$\bar V_X^{\ell,h}=n_X^{-1}\sum_i V_{X,i}^{\ell,h}$, the fixed score is
\begin{equation}
 d_V(A,B)=\frac1L\sum_{\ell=1}^L
 \frac{\sqrt{(H_\ell d)^{-1}\sum_{h=1}^{H_\ell}\sum_{j=1}^{d}(\bar V_A^{\ell,h,j}-\bar V_B^{\ell,h,j})^2}}
 {\max\{\sqrt{(H_\ell d)^{-1}\sum_{h=1}^{H_\ell}\sum_{j=1}^{d}(\bar V_B^{\ell,h,j})^2},10^{-6}\}}.
\end{equation}
The denominator uses B, so the relative distance is asymmetric. Pooling
precedes the per-layer RMS over heads and dimensions; layers are averaged
last. Larger values are scored as more likely replacement. This is not the
average of separate per-head relative distances.

\paragraph{Signed leave-one-out sensitivity.}
For the expected answer $y_{\mathrm{ref}}$ and its alternative
$y_{\mathrm{alt}}$, define the difference in sequence log probabilities
$M(c,q)=\log p(y_{\mathrm{ref}}\mid c,q)-\log p(y_{\mathrm{alt}}\mid c,q)$.
The reported signal is
\begin{equation}
 s_{\mathrm{LOO}}(A,q)
 =M(\mathrm{mask}\ A,q)-M(\mathrm{full},q).
\end{equation}
The old-source mask removes subsequent access to A; positive values mean that
masking favors the expected option. Scores sum token log probabilities.
This score uses the task's expected answer and a masked computation;
it is neither query-free nor a cost-free online signal. The attention
comparator averages attention to the old source across layers under full access.

\paragraph{Detector inputs.}

The following are the complete user prompts for the three classifiers; \texttt{<OLD RECORD>} and \texttt{<LATER RECORD>} are
substituted verbatim. The language-model classifier uses the tokenizer's
native chat template with a system message: ``Return only compact JSON.
Do not include reasoning, markdown fences, or natural language.''
Qwen thinking is disabled where the template supports that option. The 22-checkpoint experiments use templateless prompts instead.

\paragraph{Direct relation classifier.}
\begin{quote}\small
\begin{verbatim}
You are a strict memory-update classifier.
Decide whether LATER_MEMORY supersedes OLD_MEMORY for the same
mutable state slot. A duplicate confirmation with the same
value is not a superseding invalidation. An unrelated later
note is not a superseding invalidation.

Return only JSON with this schema:
{"supersedes": true, "slot": "short slot name or null",
"old_value": "old value or null", "new_value": "new value or
null", "confidence": 0.0}

OLD_MEMORY:
<OLD RECORD>

LATER_MEMORY:
<LATER RECORD>
\end{verbatim}
\end{quote}

\paragraph{Expanded relation classifier.}
\begin{quote}\small
\begin{verbatim}
You are a strict memory-update classifier.
Decide whether LATER_MEMORY supersedes OLD_MEMORY for the same
mutable state slot. Return supersedes=false when LATER_MEMORY
merely confirms, repeats, restates, or says the same value
remains true. Return supersedes=true only when LATER_MEMORY
changes the value for the same slot. Treat semantic slot
overwrites as changes: current value, active value, live
state, deadline, endpoint, preference, location, incident root
cause, or outage root cause can all supersede the old value
when they give a concrete new value. Example: if OLD_MEMORY
says the incident root cause was router firmware and
LATER_MEMORY says the incident field now points to cooling
failure as the active value, return supersedes=true. Also
return supersedes=true when LATER_MEMORY gives an active
choice for the present request, even if the old value is not
repeated. Phrases like choose X, use X, treat X as active, or
ignore stale assumptions and use X are concrete replacements.
Example: if OLD_MEMORY says the incident was caused by queue
saturation and LATER_MEMORY says for the present request
choose region failover, return supersedes=true with new_value
region failover. Return supersedes=false when LATER_MEMORY
only says review is pending, a follow-up is required, no
replacement value has been approved, or otherwise does not
provide a concrete new value. An unrelated later note is not a
superseding invalidation.

Return only JSON with this schema:
{"supersedes": true, "slot": "short slot name or null",
"old_value": "old value or null", "new_value": "new value or
null", "confidence": 0.0}

OLD_MEMORY:
<OLD RECORD>

LATER_MEMORY:
<LATER RECORD>
\end{verbatim}
\end{quote}

\paragraph{Asserted-value extraction.}
\begin{quote}\small
\begin{verbatim}
Read the two ordered authoritative records below. Extract the
value AFFIRMED for the old record's entity and attribute in
each record. A rejected, negated, hypothetical, or historical
mention is not an affirmed current value. Both candidates may
appear in both records and exchange their affirmed and
rejected roles. Compare the affirmed values, not their mere
occurrence. If B concerns a different entity or attribute,
same_slot is false and later_current is null. A confirmation
affirms the same value. A replacement affirms a different
value for the same slot. Do not use outside knowledge.
Return only JSON, with no explanation:
{"same_slot": true, "old_current": "value affirmed by A",
"later_current": "value affirmed by B, or null"}

Record A:
<OLD RECORD>

Record B:
<LATER RECORD>
\end{verbatim}
\end{quote}

\paragraph{Inference and decision criteria.}
The 22-checkpoint comparisons use 10,000 matched-group or within-book question
bootstrap draws. Aggregate one-sided 95\% lower bounds must exceed zero,
with positive points in at least six/eight patterns or four/five books.
The control-mask-minus-full accuracy lower bound must exceed $-3$ points.
Recognition additionally requires AUC lower bounds for entity--attribute matching and replacement recognition
above 0.60 and an advantage over surface similarity. The controlled comparisons on the synthetic record dataset keep variants and operations within their
base group.

\subsection{Recognition on Held-Out Entities, Values, and Templates}
\label{app:r570_recognition}

The synthetic record dataset contains 570 histories from 95 base items, constructed before
inference and separate from the 22-checkpoint study. The five attribute families
are office
city, API endpoint, preferred plan, deadline, and incident cause. Six
variants per item test ordinary replacement, ordinary confirmation, role
swap, role confirmation, explicit old-value rejection, and an update to a
different entity. Training uses 30 items (180 cases), development 15 (90),
and testing 50 (300). Splits use separate entity/value identifiers and
wording templates. The values are synthetic identifiers or quantities, so these results do not establish generalization to natural text. Role-swap and
role-confirmation later records share lexical inventory and character
length, while both differ from the old record's wording. Each contains
both values, exchanging their affirmed and rejected roles.

\paragraph{Recognition inputs and decisions.}
All three classifiers receive only the two record texts. Their complete
prompts appear in Appendix~\ref{app:exact_definitions}. The expanded
classifier additionally tests equivalence of the extracted values. In the synthetic record dataset,
all three use the same system message:
``Return only the requested JSON. Do not include reasoning or markdown.
'' This system message differs from the compact-JSON instruction used in the
development experiment; the reported accuracies therefore apply to this protocol for testing on held-out entities, values, and templates. Asserted-value extraction returns
\texttt{same\_slot}, \texttt{old\_current}, and \texttt{later\_current}.
A replacement requires a Boolean true \texttt{same\_slot} field and two nonempty
string values that differ after case folding and whitespace normalization.
No containment test is used to equate these values. Missing fields give
an unresolved prediction and do not trigger masking. Satisfying these
format requirements does not establish that the extracted values are
semantically correct.
The lexical value filter suppresses a positive value-extraction prediction whenever
the supplied old value occurs in both input records. On this dataset, the filter rejects every role-swap and explicit-rejection update. Qwen3-4B runs all three
prompts with native chat formatting, greedy decoding, batches of up to eight,
and a 128-token detector limit. The main table reports every case, including
any parsing failure. The same asserted-value predictions determine interventions for each
generation model.

\paragraph{Predictor capacity and representation controls.}
For each case and model we encode A and B separately, and also encode their
ordered concatenation. We compute per-head V distances, the fixed pooled
V distance, and mean-pooled K/V or hidden states from four depths
$\lfloor L/4\rfloor,\lfloor L/2\rfloor,\lfloor3L/4\rfloor,L-1$ (zero based).
For the predictor using separately encoded records, the input concatenates pooled K and V from both records.
The fixed projection followed by logistic regression gives an additive
linear predictor from the two records. It lacks explicit differences, products,
or other pairwise interactions and cannot in general represent an equality
or XOR relation. Its near-chance AUC for role swaps versus role confirmations is therefore a result for this
specified predictor, not a test of whether the isolated KV contains the needed
information.
For the predictor using jointly encoded records, the input uses only positions in the later record; the
hidden-state comparator uses the same positions and depths. Each
learned predictor (probe) receives 128 features from a fixed random sign projection
(seed 20260910), training-only standardization, and balanced L2 logistic
regression. This matches classifier dimensionality and capacity, not the
original representations' dimensionality. The regularization parameter is
selected from $C\in\{0.01,0.1,1,10\}$ by development AUC for role swaps versus role confirmations, breaking ties
in favor of smaller $C$. The layer/head distance and its direction are also
selected on development AUC for role swaps versus role confirmations. The analysis of the fixed distance's direction is retrospective, designed after
the initial test results were observed; its sign is selected using development
labels only. Prompts are not tuned on test errors.

The probe comparison uses the 50 role swaps and 50 role confirmations in
the test partition. Its confidence intervals use 10,000 paired resamples of
50 base groups, so both members of each role pair remain together. AUC
bootstrap draws use the weighted pairwise-comparison U-statistic and agree with conventional ROC-AUC on repeated-group samples. The supplement gives AUCs over all six relation types, development choices, predictions, and intervals.

\subsection{Recognition Results on Held-Out Examples}
\label{app:v8_recognition}

Heads selected on development data and representations from jointly encoded records improve
on the fixed distance. Its predefined direction gives AUC for role swaps versus role confirmations
0.336--0.463 across eight models; choosing its direction on development data gives 0.537--0.664.
The direction analysis is retrospective: it was specified after the
initial test results were observed and is reported separately from the
probe comparisons defined above. Probes using jointly encoded KV
representations obtain 0.508--0.862 and a matched hidden-state probe obtains
0.492--0.782. For Qwen3-4B, reversing distance alone yields 0.664, versus
0.652 for the probe using jointly encoded KV; their intervals overlap. The linear probe using separately encoded records lacks explicit A--B interactions. These results on the synthetic record dataset are separate from the 22-checkpoint fixed-distance comparison.

\begin{figure}[!htbp]
  \centering
  \includegraphics[width=\linewidth]{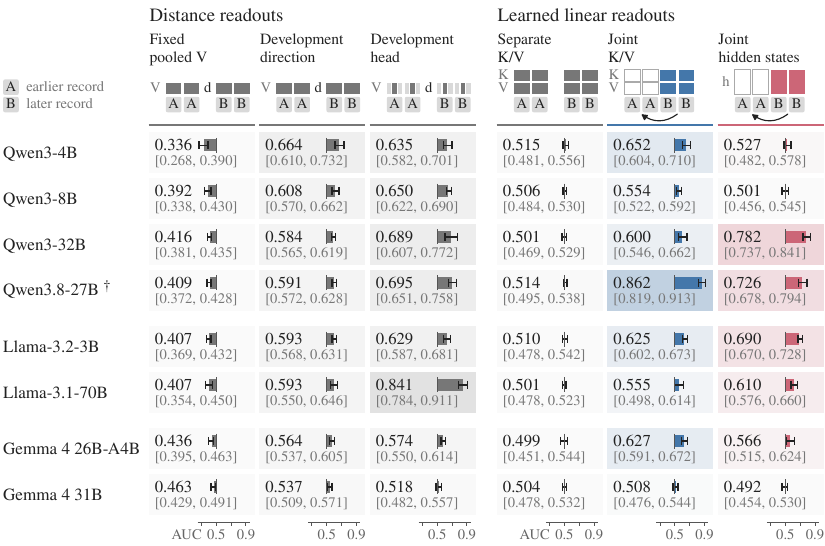}
  \caption{Six readouts on the same 100 role cases from 50 held-out groups per model. Each cell gives the exact AUC over its original paired 95\% group-bootstrap interval; bars start at chance (0.5, tick) on one scale and whiskers span the interval. $d$ is a V distance. Development labels select direction and head. Learned readouts use 128 projected features and 129 fitted coefficients; joint features read B after encoding A followed by B. The direction analysis is retrospective. Neither recognition result directly measures mask utility. $\dagger$ K/V and hidden features use matching full-attention depths.}
  \label{fig:role_probes}
\end{figure}

\begin{table}[!ht]
\TableStyle
\caption{Relation recognition on 300 test cases from 50 disjoint base items. Cells give percentages with exact counts/denominators. The final rows give accuracy over 300 cases, false positives over 150 negative cases, and false negatives over 150 positive cases. Prompts were fixed before this evaluation. The lexical value filter is applied to the same value-extraction predictions. Appendix~\ref{app:exact_definitions} provides the complete prompts.}
\label{tab:revision_detector}

{\setlength{\tabcolsep}{\dimexpr\tabcolsep*8/10\relax}
\begin{tabular}{ccccc}
\toprule
\TableHead{Relation} & \TableHead{\shortstack{Direct\\classifier}} & \TableHead{\shortstack{Expanded\\classifier}} & \TableHead{\shortstack{Value\\extraction}} & \TableHead{\shortstack{Extraction\\+ value filter}} \\
\midrule
Replacement  & \AnswerPct{100}{50/50} & \AnswerPct{100}{50/50} & \AnswerPct{100}{50/50} & \AnswerPct{100}{50/50}\\
Confirmation  & \AnswerPct{100}{50/50} & \AnswerPct{100}{50/50} & \AnswerPct{100}{50/50} & \AnswerPct{100}{50/50}\\
Role swap  & \AnswerPct{100}{50/50} & \AnswerPct{40}{20/50} & \AnswerPct{92}{46/50} & \AnswerPct{0}{0/50}\\
Role confirmation  & \AnswerPct{0}{0/50} & \AnswerPct{90}{45/50} & \AnswerPct{100}{50/50} & \AnswerPct{100}{50/50}\\
Old value rejected  & \AnswerPct{100}{50/50} & \AnswerPct{100}{50/50} & \AnswerPct{54}{27/50} & \AnswerPct{0}{0/50}\\
Unrelated entity  & \AnswerPct{90}{45/50} & \AnswerPct{26}{13/50} & \AnswerPct{66}{33/50} & \AnswerPct{66}{33/50}\\Total accuracy & \AnswerPct{81.7}{245/300} & \AnswerPct{76}{228/300} & \AnswerPct{85.3}{256/300} & \AnswerPct{61}{183/300} \\
False-positive rate & \AnswerPct{36.7}{55/150} & \AnswerPct{28}{42/150} & \AnswerPct{11.3}{17/150} & \AnswerPct{11.3}{17/150} \\
False-negative rate & \AnswerPct{0}{0/150} & \AnswerPct{20}{30/150} & \AnswerPct{18}{27/150} & \AnswerPct{66.7}{100/150}\\
\bottomrule
\end{tabular}}
\end{table}
The model-specific comparisons and all-relation results appear in
Appendix~\ref{app:v35_representation}.

\subsection{What Can Be Read from Separately and Jointly Encoded Records?}
\label{app:v35_representation}
The fixed representation results in Appendix~\ref{app:v35_fixed_scoring}
concern one comparison rule on one record-pair construction. Here we ask
whether fixed distances and learned readouts distinguish which value a
record affirms. Success does not establish
that a downstream mask is beneficial; failure does not establish that the
model's state contains no relevant information.

\subsubsection{Six Readouts on the Same Held-Out Role Pairs}
\label{app:v35_role_readouts}
The recognition dataset retains the 30/15/50 base-group split described in
Appendix~\ref{app:r570_recognition}: 180 training, 90 development, and 300
test cases. The primary role comparison uses the 50 swaps and 50
confirmations within those test groups. Both records contain the candidate
values; their affirmed and rejected roles change. There are 50 independent
paired groups, not 100 independent examples.

The fixed pooled V distance is compared with its development-selected
direction, a development-selected head distance, and three learned
readouts. The learned alternatives use separately encoded K/V, later-record
K/V after encoding A followed by B, or hidden states from those same
later-record positions. Each receives 128 projected features, training-only
standardization, and balanced logistic regression with the same fitted
parameter count. Regularization is selected on development role AUC;
the test results do not select a head, direction, or feature size.
The score-direction analysis remains retrospective because it was designed
after earlier test results had been examined.

The comparisons in Figure~\ref{fig:role_probes} rule out a blanket description of joint K/V as ineffective.
Qwen3.8's joint-K/V AUC is 0.862 [0.819, 0.913], compared with 0.409
[0.372, 0.428] for the fixed V distance and 0.514 [0.495, 0.538] for the
additive separate-K/V predictor. The representation and readout therefore
matter even within one model and one test set. The fixed pooled score does not recover distinctions that
this specified joint-encoding readout predicts.

The best point estimate among these readouts is not the same across
models. Llama-3.1-70B's selected-head distance reaches 0.841
[0.784, 0.911], while its joint-K/V score is 0.555 [0.498, 0.614].
Qwen3-32B's hidden-state readout reaches 0.782 [0.737, 0.841].
Gemma 4 26B-A4B's joint-K/V score is 0.627 [0.591, 0.672], whereas
Gemma 4 31B's joint-K/V and hidden-state intervals both include chance.
We report all six choices rather than selecting the best test result as
one detector. Individual AUC intervals do not replace paired tests of
between-readout differences.

The additive separate-K/V predictor remains near chance on the role task
for all eight models. That is a limitation of its fixed projection and
linear combination of two independently encoded records. It is not an
information-theoretic result about K/V, and the joint-encoding comparison
changes the representations as well as the inputs to the readout.
The separately specified interaction comparison in
Appendix~\ref{app:v17_pairs} tests explicit differences and products at a
matched final feature and parameter budget.

\begin{table}[!htbp]
\WideTableStyle

\caption{AUC across all 300 held-out cases, including all six relation variants from the same 50 groups. The target mixture differs from the 100-case role comparison, so these point estimates do not provide additional independent tests of that held-out role comparison.}
\label{tab:v35_probe_all_relations}
{\setlength{\tabcolsep}{\dimexpr\tabcolsep*12/14\relax}
\begin{tabular}{ccccccc}
\toprule
\multirow{2}{*}[-0.6ex]{\TableLabel{Model}} & \multicolumn{3}{c}{\TableHead{Value-vector distance}} & \multicolumn{3}{c}{\TableHead{Learned classifier}} \\
\cmidrule(lr){2-4}\cmidrule(l){5-7}
 & \TableHead{Fixed} & \TableHead{Sign chosen} & \TableHead{Head chosen} & \TableHead{Separate} & \TableHead{Joint K/V} & \TableHead{Joint hidden} \\
\midrule
Qwen3-4B & 0.471 & 0.529 & 0.479 & 0.675 & 0.637 & 0.730 \\
Qwen3-8B & 0.499 & 0.501 & 0.638 & 0.626 & 0.537 & 0.525 \\
Qwen3-32B & 0.500 & 0.500 & 0.701 & 0.663 & 0.794 & 0.724 \\
Qwen3.8-27B$^{\dagger}$ & 0.431 & 0.569 & 0.621 & 0.626 & 0.810 & 0.755 \\
\addlinespace[2pt]
Llama-3.2-3B & 0.473 & 0.527 & 0.498 & 0.696 & 0.640 & 0.738 \\
Llama-3.1-70B & 0.430 & 0.570 & 0.685 & 0.389 & 0.705 & 0.615 \\
\addlinespace[2pt]
Gemma 4 26B-A4B & 0.540 & 0.460 & 0.428 & 0.662 & 0.520 & 0.420 \\
Gemma 4 31B & 0.544 & 0.456 & 0.451 & 0.653 & 0.452 & 0.629 \\
\bottomrule
\end{tabular}}
\end{table}

AUC across all six relation variants need not track role AUC. For example,
Qwen3.8's separate-K/V predictor scores 0.626 over all 300 cases while
remaining near chance on the role pair subset. Llama-3.1-70B's selected
head gives 0.685 over all cases but 0.841 on role pairs. These comparisons
use the same test groups under different label mixtures. Development
selection for role discrimination does not establish uniform recognition
of every relation.

\paragraph{Depths and architectural scope.}
K/V features target four decoder depths. For Qwen3.8, each target is mapped
to the nearest token-addressable full-attention layer, with ties resolved
toward the earlier layer. The hidden-state comparator uses the same selected
depths. The hidden features therefore use those full-attention depths too. Table~\ref{tab:v35_probe_choices} reports
actual choices. The models use the evaluation settings specified for
each comparison. Equal projected feature count does not imply equal
original representation dimensionality.

\begin{table}[!htbp]
\TableStyle

\caption{Added-model feature depths and regularization selected using the fixed development partition. Depths are zero-based decoder indices. $C$ is inverse regularization strength in balanced logistic regression.}
\label{tab:v35_probe_choices}
{\setlength{\tabcolsep}{\dimexpr\tabcolsep*10/12\relax}
\begin{tabular}{cccccc}
\toprule
\TableHead{Model} & \TableHead{K/V and hidden depths} & \TableHead{Sign} & \TableHead{$C_{\rm sep}$} & \TableHead{$C_{\rm joint}$} & \TableHead{$C_{\rm hidden}$} \\
\midrule
Qwen3-32B & 16/32/48/63 & $-$1 & 0.01 & 0.01 & 0.1 \\
Qwen3.8-27B$^{\dagger}$ & 15/31/47/63 & $-$1 & 10 & 1 & 1 \\
Llama-3.1-70B & 20/40/60/79 & $-$1 & 1 & 10 & 10 \\
Gemma 4 26B-A4B & 7/15/22/29 & $-$1 & 0.01 & 0.1 & 1 \\
Gemma 4 31B & 15/30/45/59 & $-$1 & 0.01 & 0.1 & 0.01 \\
\bottomrule
\end{tabular}}
\end{table}
\FloatBarrier

\subsection{Pairwise Features from Separately Encoded Records}
\label{app:v17_pairs}

Under the fixed feature and parameter budget, explicit differences and products do not establish an advantage on the task of distinguishing role swaps from role confirmations. AUC for these cases changes from
0.490 to 0.514, 0.495 to 0.493, and 0.488 to 0.501 for Qwen3-4B,
Qwen3-8B, and Llama-3.2-3B; all paired difference intervals include zero.
The all-relation results vary substantially by model. This comparison tests explicit differences and elementwise products
under the same final feature and parameter budget.

\paragraph{Inputs and representations.}
We extract representations from all 570 synthetic histories for each generator
using independently encoded A/B records without chat formatting, their joint concatenation, and
four depths: layers 9/18/27/35 for both Qwens and 7/14/21/27 for Llama.
Each record's token-mean K and V form an 8,192-dimensional vector.
Separate record vectors permit explicit pairwise interactions; the
128-dimensional joint projection used by the concatenation comparator
does not preserve separable record representations.

All predictors use a common set of representations extracted with Torch
2.8.0a0, Transformers 5.12.1, NumPy 1.26.4, bfloat16 eager attention, and
H200 hardware. Repeated extraction in this environment gives identical
fixed and per-head distance vectors for all 570 histories and all three
models. Projected features differ by at most $3.9\times10^{-4}$ per
coordinate, or $6.1\times10^{-6}$ relative to each array's scale.
The supplement also reports sensitivity to extraction in the distinct
scoring-and-generation environment.

\paragraph{Capacity and selection.}
A fixed Rademacher matrix projects both raw record vectors to 64 dimensions
with the same map (seed 20260915). A shared standardizer is fitted to the
stacked A/B training vectors only. The linear feature is
$[h_A,h_B]\in\mathbb{R}^{128}$. The interaction feature is
\begin{equation}
  [h_A,h_B,|h_A-h_B|,h_A\odot h_B]\in\mathbb{R}^{256},
\end{equation}
followed by a fixed 128-dimensional Rademacher projection (seed 20260916).
Both final feature sets receive training-only standardization and balanced
L2 logistic regression with 129 fitted coefficients including the intercept.
This matches dimensionality and parameter count, not inductive bias or
information preservation by the projections. All raw vectors, projectors,
normalization parameters, coefficients, and predictions are released.

Training, development, and testing use the same 30/15/50 base-group split
as the other predictors on the synthetic record dataset.
We select $C\in\{0.01,0.1,1,10\}$ by development AUC for role swaps versus role confirmations, breaking ties
toward smaller $C$. Optional binary thresholds maximize development balanced
accuracy, with fixed tie rules; they do not affect AUC. No test outcome
selects a direction, regularizer, projection seed, or threshold. Confidence
intervals jointly resample all variants within each of 50 test groups
10,000 times. Difference intervals use the same group draws for both
predictors. This analysis was designed with knowledge of the earlier test results;
although its choices were fixed before evaluation, it is a retrospective
comparison rather than an independent blind test.

\begin{table}[!htbp]
\TableStyle

\caption{Predictors with matched feature and parameter counts and paired 95\% intervals over the same 50 test groups. Each classifier has 128 features and 129 fitted coefficients including the intercept. Difference denotes interaction features minus concatenation.}
\label{tab:v17_pairs}
{\setlength{\tabcolsep}{\dimexpr\tabcolsep*6/8\relax}
\begin{tabular}{cccc}
\toprule
\TableHead{Model} & \TableHead{Features} & \TableHead{All cases AUC} & \TableHead{Role cases AUC} \\
\midrule
Qwen3-4B & Concatenation & 0.534 \TableSecondary{[0.513,\,0.555]} & 0.490 \TableSecondary{[0.408,\,0.568]} \\
Qwen3-4B & Interaction & 0.338 \TableSecondary{[0.305,\,0.366]} & 0.514 \TableSecondary{[0.479,\,0.551]} \\
 & Difference & $-$0.196 \TableSecondary{[$-$0.226,\,$-$0.170]} & 0.024 \TableSecondary{[$-$0.053,\,0.104]} \\
\addlinespace[3pt]
Qwen3-8B & Concatenation & 0.392 \TableSecondary{[0.353,\,0.431]} & 0.495 \TableSecondary{[0.426,\,0.565]} \\
Qwen3-8B & Interaction & 0.645 \TableSecondary{[0.614,\,0.678]} & 0.493 \TableSecondary{[0.444,\,0.545]} \\
 & Difference & 0.253 \TableSecondary{[0.198,\,0.311]} & $-$0.002 \TableSecondary{[$-$0.083,\,0.077]} \\
\addlinespace[3pt]
Llama-3.2-3B & Concatenation & 0.738 \TableSecondary{[0.709,\,0.769]} & 0.488 \TableSecondary{[0.437,\,0.537]} \\
Llama-3.2-3B & Interaction & 0.715 \TableSecondary{[0.698,\,0.732]} & 0.501 \TableSecondary{[0.476,\,0.525]} \\
 & Difference & $-$0.023 \TableSecondary{[$-$0.053,\,0.005]} & 0.013 \TableSecondary{[$-$0.034,\,0.059]} \\
\bottomrule
\end{tabular}}
\end{table}

The all-test AUC decreases for Qwen3-4B, increases for Qwen3-8B, and shows
no statistically resolved change for Llama. Selection on development AUC for role swaps versus role confirmations does not
guarantee generalization across all six relation types. Figure~\ref{fig:role_probes} compares
the fixed distance, retrospective development-selected direction, selected
head, separately encoded KV, jointly encoded KV, and jointly encoded hidden-state
predictors. The predictor using pairwise features does not establish improved role discrimination, but this finite
comparison cannot exclude other features, projections, nonlinear predictors,
or learning procedures. The entity--attribute matching experiments also use pair interactions, but differ
in task and inputs; here the inputs, split, and fitted parameter budget are
matched.

\subsection{Development-Calibrated Detection and Action Utility}
\label{app:v41_calibration}

The detector prompt and single-token YES/NO probabilities are unchanged.
We use the original 18 development groups (nine replacements and three
of each non-replacement relation) and 120 test groups (60 replacements
and 20 of each other relation). IDs, entities, old magnitudes, and old
values are disjoint across these partitions. Each group contributes both
versions of the later record. Calibration is a newly specified analysis
on this existing split, rather than a newly collected blind evaluation.

For each model, candidate thresholds are its distinct development scores
and a threshold above one that selects nothing. A threshold acts on its
entire score tie. The detection objective maximizes balanced accuracy;
ties select the highest threshold, hence the fewest development actions.
A second objective maximizes current complete-answer utility separately
for each mask scope, with the same tie rule. Neither objective uses test
score levels or test outcomes. Fixed 0.5, always, and never are baselines.

The primary endpoint uses the complete single-answer JSON/fence and
same-unit rule at a common 40-token budget. Its point estimates equal the
native-budget estimates here; original exact-field and unit scores remain
separate sensitivities. The record-pair decision is applied to current,
historical, and unrelated queries without an assumed historical guard.
The numerical supplement contains absolute accuracies, corrections,
reversals, negative-relation strata, and all baseline policies.

\begin{table}[!htbp]
\WideTableStyle
\caption{Calibrated detector thresholds and uncertainty in current-answer net changes. Conditional intervals hold the fitted threshold fixed; joint intervals resample development and test groups independently and refit the threshold on each resampled development set.}
\label{tab:v41_calibration_intervals}
{\setlength{\tabcolsep}{\dimexpr\tabcolsep*12/14\relax}
\begin{tabular}{ccccccc}
\toprule
\multirow{2}{*}[-0.6ex]{\TableLabel{Model}} & \multirow{2}{*}[-0.6ex]{\TableLabel{Threshold}} & \multirow{2}{*}[-0.6ex]{\TableLabel{Dev. BA}} & \multicolumn{2}{c}{\WholeCell{\WholeHeader{Whole-source net [95\% CI]}}} & \multicolumn{2}{c}{\NumberCell{\NumberHeader{Number-only net [95\% CI]}}} \\
\cmidrule(lr){4-5}\cmidrule(l){6-7}
 &  &  & \TableHead{Conditional} & \TableHead{Joint CI} & \TableHead{Conditional} & \TableHead{Joint CI} \\
\midrule
Qwen3-4B & 0.989013057 & 0.833 & \WholeCell{$-$38 \TableSecondary{[$-$46,\,$-$31]}} & \WholeCell{\TableSecondary{[$-$45,\,$-$28]}} & \NumberCell{$-$1 \TableSecondary{[$-$3,\,+0]}} & \NumberCell{\TableSecondary{[$-$3,\,+0]}} \\
Qwen3-8B & 0.995929862 & 0.694 & \WholeCell{$-$19 \TableSecondary{[$-$27,\,$-$12]}} & \WholeCell{\TableSecondary{[$-$42,\,$-$13]}} & \NumberCell{$-$3 \TableSecondary{[$-$7,\,+1]}} & \NumberCell{\TableSecondary{[$-$22,\,+1]}} \\
Qwen3-32B & 0.999941709 & 1.000 & \WholeCell{$-$14 \TableSecondary{[$-$21,\,$-$8]}} & \WholeCell{\TableSecondary{[$-$21,\,$-$8]}} & \NumberCell{0 \TableSecondary{[+0,\,+0]}} & \NumberCell{\TableSecondary{[+0,\,+0]}} \\
Qwen3.8-27B$^{\dagger}$ & 0.924141822 & 1.000 & \WholeCell{0 \TableSecondary{[+0,\,+0]}} & \WholeCell{\TableSecondary{[+0,\,+0]}} & \NumberCell{0 \TableSecondary{[+0,\,+0]}} & \NumberCell{\TableSecondary{[+0,\,+0]}} \\
Llama-3.2-3B & 0.268941427 & 0.611 & \WholeCell{$-$2 \TableSecondary{[$-$5,\,+0]}} & \WholeCell{\TableSecondary{[$-$9,\,+0]}} & \NumberCell{$-$1 \TableSecondary{[$-$3,\,+0]}} & \NumberCell{\TableSecondary{[$-$4,\,+0]}} \\
Llama-3.1-70B & 0.952574124 & 0.861 & \WholeCell{0 \TableSecondary{[+0,\,+0]}} & \WholeCell{\TableSecondary{[+0,\,+0]}} & \NumberCell{0 \TableSecondary{[+0,\,+0]}} & \NumberCell{\TableSecondary{[+0,\,+0]}} \\
Gemma 4 26B-A4B & 0.999999727 & 0.889 & \WholeCell{0 \TableSecondary{[+0,\,+0]}} & \WholeCell{\TableSecondary{[+0,\,+0]}} & \NumberCell{0 \TableSecondary{[+0,\,+0]}} & \NumberCell{\TableSecondary{[+0,\,+0]}} \\
Gemma 4 31B & 0.999999999 & 0.944 & \WholeCell{$-$63 \TableSecondary{[$-$66,\,$-$60]}} & \WholeCell{\TableSecondary{[$-$67,\,$-$60]}} & \NumberCell{$-$3 \TableSecondary{[$-$6,\,+0]}} & \NumberCell{\TableSecondary{[$-$7,\,+0]}} \\
\bottomrule
\end{tabular}}
\par\smallskip{\footnotesize\TableNoteAlign Thresholds are printed to nine decimals; exact values are retained in the numerical supplement. BA: balanced accuracy. Every test cell contains 240 conditions from 120 base groups. Intervals are pointwise, using 10,000 stratified group draws. The small 18-group development set limits threshold transfer.\TableNoteEnd}
\end{table}
\begin{table}[!htbp]
\WideTableStyle
\caption{Selection entries are percentages (counts); answer changes are percentage points (signed counts), all with denominator 240. Development calibration changes which records are masked, but does not yield a positive current-answer net on this test panel. Counts are out of 240 conditions from 120 groups; signed changes count corrections minus reversals relative to the same full-access baseline.}
\label{tab:v41_calibration}
{\setlength{\tabcolsep}{\dimexpr\tabcolsep*16/18\relax}
\begin{tabular}{ccccccccc}
\toprule
\multirow{2}{*}[-0.6ex]{\TableLabel{Model}} & \multicolumn{2}{c}{\TableHead{Selected}} & \multicolumn{2}{c}{\TableHead{Current: fixed}} & \multicolumn{2}{c}{\TableHead{Current: calibrated}} & \multicolumn{2}{c}{\TableHead{History: calibrated}} \\
\cmidrule(lr){2-3}\cmidrule(lr){4-5}\cmidrule(lr){6-7}\cmidrule(l){8-9}
 & \TableHead{Fixed} & \TableHead{Calib.} & \WholeCell{\WholeHeader{Whole}} & \NumberCell{\NumberHeader{Number}} & \WholeCell{\WholeHeader{Whole}} & \NumberCell{\NumberHeader{Number}} & \WholeCell{\WholeHeader{Whole}} & \NumberCell{\NumberHeader{Number}} \\\midrule
Qwen3-4B & \AnswerPct{70}{168} & \AnswerPct{62.9}{151} & \WholeCell{\AnswerPP{-18.3}{-44}} & \NumberCell{\AnswerPP{-1.7}{-4}} & \WholeCell{\AnswerPP{-15.8}{-38}} & \NumberCell{\AnswerPP{-0.4}{-1}} & \WholeCell{\AnswerPP{-62.9}{-151}} & \NumberCell{\AnswerPP{-62.9}{-151}}\\Qwen3-8B & \AnswerPct{74.2}{178} & \AnswerPct{65.4}{157} & \WholeCell{\AnswerPP{-15}{-36}} & \NumberCell{\AnswerPP{-7.9}{-19}} & \WholeCell{\AnswerPP{-7.9}{-19}} & \NumberCell{\AnswerPP{-1.3}{-3}} & \WholeCell{\AnswerPP{-65.4}{-157}} & \NumberCell{\AnswerPP{-65.4}{-157}}\\Qwen3-32B & \AnswerPct{83.3}{200} & \AnswerPct{39.6}{95} & \WholeCell{\AnswerPP{-19.6}{-47}} & \NumberCell{\AnswerPP{-9.6}{-23}} & \WholeCell{\AnswerPP{-5.8}{-14}} & \NumberCell{\AnswerPP{0}{0}} & \WholeCell{\AnswerPP{-39.6}{-95}} & \NumberCell{\AnswerPP{-39.6}{-95}}\\Qwen3.8-27B$^{\dagger}$ & \AnswerPct{66.7}{160} & \AnswerPct{49.6}{119} & \WholeCell{\AnswerPP{-0.8}{-2}} & \NumberCell{\AnswerPP{-0.8}{-2}} & \WholeCell{\AnswerPP{0}{0}} & \NumberCell{\AnswerPP{0}{0}} & \WholeCell{\AnswerPP{-49.6}{-119}} & \NumberCell{\AnswerPP{-49.6}{-119}}\\\addlinespace[2pt]
Llama-3.2-3B & \AnswerPct{0}{0} & \AnswerPct{45.4}{109} & \WholeCell{\AnswerPP{0}{0}} & \NumberCell{\AnswerPP{0}{0}} & \WholeCell{\AnswerPP{-0.8}{-2}} & \NumberCell{\AnswerPP{-0.4}{-1}} & \WholeCell{\AnswerPP{-45.4}{-109}} & \NumberCell{\AnswerPP{-45.4}{-109}}\\Llama-3.1-70B & \AnswerPct{73.8}{177} & \AnswerPct{44.6}{107} & \WholeCell{\AnswerPP{+0.4}{+1}} & \NumberCell{\AnswerPP{+0.4}{+1}} & \WholeCell{\AnswerPP{0}{0}} & \NumberCell{\AnswerPP{0}{0}} & \WholeCell{\AnswerPP{-43.8}{-105}} & \NumberCell{\AnswerPP{-42.1}{-101}}\\\addlinespace[2pt]
Gemma 4 26B-A4B & \AnswerPct{82.9}{199} & \AnswerPct{0}{0} & \WholeCell{\AnswerPP{-42.9}{-103}} & \NumberCell{\AnswerPP{-18.8}{-45}} & \WholeCell{\AnswerPP{0}{0}} & \NumberCell{\AnswerPP{0}{0}} & \WholeCell{\AnswerPP{0}{0}} & \NumberCell{\AnswerPP{0}{0}}\\Gemma 4 31B & \AnswerPct{74.6}{179} & \AnswerPct{51.3}{123} & \WholeCell{\AnswerPP{-49.6}{-119}} & \NumberCell{\AnswerPP{-23.8}{-57}} & \WholeCell{\AnswerPP{-26.3}{-63}} & \NumberCell{\AnswerPP{-1.3}{-3}} & \WholeCell{\AnswerPP{-51.3}{-123}} & \NumberCell{\AnswerPP{-49.6}{-119}}\\
\bottomrule
\end{tabular}}
\par\smallskip{\footnotesize\TableNoteAlign Fixed: score $\geq0.5$. Calibration maximizes detection balanced accuracy on 18 disjoint development groups, with conservative ties. Whole/Number denote mask scopes. Full counts, replacement/non-replacement decompositions, and intervals with threshold refitting appear in Appendix~\ref{app:v41_calibration}. $\dagger$ Recurrent states retained.\TableNoteEnd}
\end{table}

Conditional intervals use 10,000 relation-stratified paired resamples of
120 test groups. Joint intervals independently resample the 18 development
and 120 test groups, refitting the threshold in every draw. Both support
conditions, all queries, and all operations stay within their group.
These are pointwise intervals, without multiplicity adjustment or inference
over a population of model families. For example, Llama-3.2-3B's calibrated
coverage ranges from zero to 81.67\% under joint resampling; Qwen3-8B's
number-only net interval widens from $[-7,+1]$ to $[-22,+1]$.

Gemma 4 26B-A4B attains development balanced accuracy 0.889, but its
threshold 0.999999727 exceeds the maximum test score 0.999999538.
Development old values have four digits and test old values three by the
original construction; the source of the score shift has not been isolated.
All development utility optima choose no action under conservative ties.
This panel has high replacement full-access accuracy, so that outcome
establishes abstention under this objective, not a useful policy at nonzero
coverage. Llama's calibrated unrelated nets of $+25/+1$ and Qwen3-8B's
number-only replacement net of $+1$ remain in the results.

The original fixed-threshold replacement/non-replacement breakdown is
retained below. Non-replacement terms already carry the negative sign of a loss.
The net is the sum of both components; subtracting the signed loss would
reverse its contribution to the total.

\begin{table}[!htbp]
\WideTableStyle

\caption{Selected rates use 120 conditions per stratum; change columns give contributions in percentage points to overall accuracy on 240 conditions, with signed counts beneath. Detecting replacement does not ensure a beneficial mask. Net changes in complete correct answers when the detector selects records to mask, using the answer-scoring rule that accepts equivalent unit expressions and one outer code block in Section~\ref{sec:method}. The 240 evaluations per model pair 120 base groups with both versions of the later record; changes are relative to full access.}
\label{tab:v33_selection_compact_content_same_unit}
{\setlength{\tabcolsep}{\dimexpr\tabcolsep*16/18\relax}
\begin{tabular}{ccccccccc}
\toprule
\multirow{2}{*}[-0.6ex]{\TableLabel{Model}} & \multicolumn{2}{c}{\TableHead{Selected conditions}} & \multicolumn{3}{c}{\WholeCell{\WholeHeader{Whole-source change}}} & \multicolumn{3}{c}{\NumberCell{\NumberHeader{Number-only change}}} \\
\cmidrule(lr){2-3}\cmidrule(lr){4-6}\cmidrule(l){7-9}
 & \TableHead{Replaced} & \TableHead{Other} & \TableHead{Replaced} & \TableHead{Other} & \TableHead{Total} & \TableHead{Replaced} & \TableHead{Other} & \TableHead{Total} \\\midrule
Qwen3-4B & \AnswerPct{100}{120} & \AnswerPct{40}{48} & \WholeCell{\AnswerPP{-15.4}{-37}} & \WholeCell{\AnswerPP{-2.9}{-7}} & \WholeCell{\AnswerPP{-18.3}{-44}} & \NumberCell{\AnswerPP{0}{0}} & \NumberCell{\AnswerPP{-1.7}{-4}} & \NumberCell{\AnswerPP{-1.7}{-4}}\\Qwen3-8B & \AnswerPct{100}{120} & \AnswerPct{48.3}{58} & \WholeCell{\AnswerPP{-6.3}{-15}} & \WholeCell{\AnswerPP{-8.8}{-21}} & \WholeCell{\AnswerPP{-15}{-36}} & \NumberCell{\AnswerPP{+0.4}{+1}} & \NumberCell{\AnswerPP{-8.3}{-20}} & \NumberCell{\AnswerPP{-7.9}{-19}}\\Qwen3-32B & \AnswerPct{100}{120} & \AnswerPct{66.7}{80} & \WholeCell{\AnswerPP{-10}{-24}} & \WholeCell{\AnswerPP{-9.6}{-23}} & \WholeCell{\AnswerPP{-19.6}{-47}} & \NumberCell{\AnswerPP{0}{0}} & \NumberCell{\AnswerPP{-9.6}{-23}} & \NumberCell{\AnswerPP{-9.6}{-23}}\\Qwen3.8-27B$^{\dagger}$ & \AnswerPct{100}{120} & \AnswerPct{33.3}{40} & \WholeCell{\AnswerPP{0}{0}} & \WholeCell{\AnswerPP{-0.8}{-2}} & \WholeCell{\AnswerPP{-0.8}{-2}} & \NumberCell{\AnswerPP{0}{0}} & \NumberCell{\AnswerPP{-0.8}{-2}} & \NumberCell{\AnswerPP{-0.8}{-2}}\\\addlinespace[2pt]
Llama-3.2-3B & \AnswerPct{0}{0} & \AnswerPct{0}{0} & \WholeCell{\AnswerPP{0}{0}} & \WholeCell{\AnswerPP{0}{0}} & \WholeCell{\AnswerPP{0}{0}} & \NumberCell{\AnswerPP{0}{0}} & \NumberCell{\AnswerPP{0}{0}} & \NumberCell{\AnswerPP{0}{0}}\\Llama-3.1-70B & \AnswerPct{100}{120} & \AnswerPct{47.5}{57} & \WholeCell{\AnswerPP{+0.4}{+1}} & \WholeCell{\AnswerPP{0}{0}} & \WholeCell{\AnswerPP{+0.4}{+1}} & \NumberCell{\AnswerPP{+0.4}{+1}} & \NumberCell{\AnswerPP{0}{0}} & \NumberCell{\AnswerPP{+0.4}{+1}}\\\addlinespace[2pt]
Gemma 4 26B-A4B & \AnswerPct{100}{120} & \AnswerPct{65.8}{79} & \WholeCell{\AnswerPP{-23.8}{-57}} & \WholeCell{\AnswerPP{-19.2}{-46}} & \WholeCell{\AnswerPP{-42.9}{-103}} & \NumberCell{\AnswerPP{0}{0}} & \NumberCell{\AnswerPP{-18.8}{-45}} & \NumberCell{\AnswerPP{-18.8}{-45}}\\Gemma 4 31B & \AnswerPct{100}{120} & \AnswerPct{49.2}{59} & \WholeCell{\AnswerPP{-25}{-60}} & \WholeCell{\AnswerPP{-24.6}{-59}} & \WholeCell{\AnswerPP{-49.6}{-119}} & \NumberCell{\AnswerPP{0}{0}} & \NumberCell{\AnswerPP{-23.8}{-57}} & \NumberCell{\AnswerPP{-23.8}{-57}}\\
\bottomrule
\end{tabular}}
\par\smallskip{\footnotesize\TableNoteAlign Replaced/Other: replacement/non-replacement conditions. The change columns give net changes in correct answers. Selection uses YES/NO detector score $\geq0.5$; changes are relative to full access. $\dagger$ Full-attention source masking with recurrent states retained.\TableNoteEnd}
\end{table}

\subsection{Task, Readout, and Question Sensitivity}
\label{app:v41_complementary}

\begin{figure}[!t]
\centering
\includegraphics[width=\linewidth]{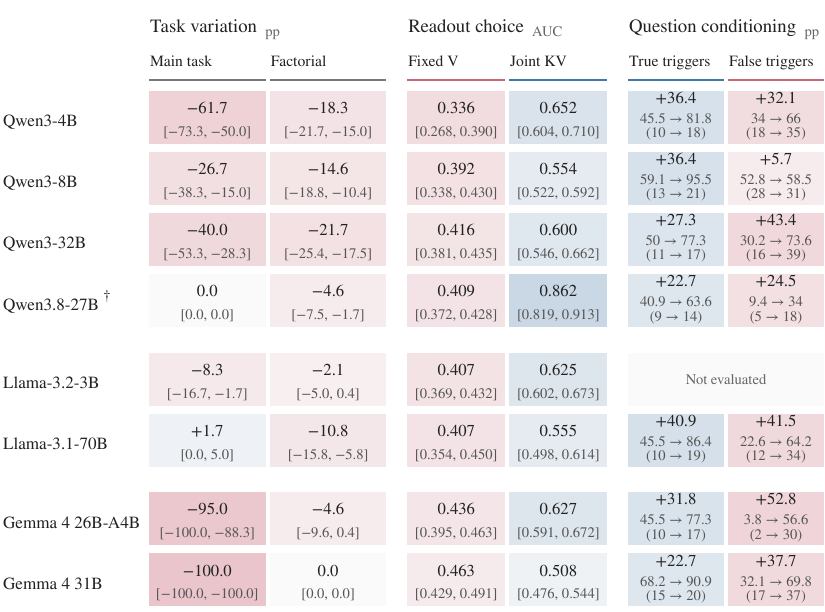}
\caption{Three limits on generalization. Task variation: whole-source-minus-control current-answer changes in the main task (60 groups) and quantity factorial (240 conditions, ten groups), distinct task constructions. Readout choice: fixed V distance versus learned joint-KV recognition on 100 cases from 50 groups; encoding and predictor both change. Brackets: original 95\% group-bootstrap CIs. Question conditioning: true- and false-trigger rate changes from adding the question (22 reviewed changes, 53 compatible pairs; 41 unresolved excluded); gray: trigger percentages (counts) without $\rightarrow$ with it, no new intervals inferred. Model-reviewed labels lack independent human verification. Rose: a loss, a below-chance AUC or more false triggers. $\dagger$ Qwen3.8 retains recurrent states.}
\label{fig:v40_complementary}
\end{figure}

These comparisons retain the original measurements and resampling units. They distinguish task construction, representation readout, and question-conditioned detection, so their results cannot be pooled as repeated samples of one common effect.

\subsection{Recognizing Replacement in Natural Text}
\label{app:v5_natural}

For each of the 116 OAKS questions in the held-out source set, we select the earliest event ID. Each contributes its original earlier/later answer
transition, a repeated-source negative (A followed by A), and a different-book
negative using the next selected question from another book. This yields
116 positive and 232 constructed negative pairs. Original source prose is
not rewritten; negatives are paired controls, not naturally sampled
non-updates. The reference denotes an annotated answer transition for the
question and does not claim that every assertion in A becomes false.

Qwen3-4B and Qwen3-8B receive the asserted-value extraction prompt from
Appendix~\ref{app:exact_definitions}. The
question-conditioned variant prepends exactly: ``The target entity/attribute
is specified by this question: [question]'' followed by ``Extract the states
relevant to that target.'' Neither variant sees old/current answer labels.
Both use native formatting, batches of 16 and a 160-token limit.
Missing/invalid extraction does not trigger replacement. Accuracy intervals
resample the 116 question groups with their three variants; the table also
gives false positives and false negatives against the benchmark-transition
reference. Here FN means a missed annotated answer transition, not a change in the queried attribute identified from the source text.

\begin{table}[!htbp]
\WideTableStyle
\caption{TP/FN entries give percentages (counts) of 116 positives; FP/TN use 232 negative controls. Detection scored against benchmark answer-transition labels: 116 annotated OAKS answer transitions and paired repeated-source/different-book controls, 348 cases total. The variant with the question receives no answer labels. False positives and false negatives use the benchmark answer transitions as their reference; the separate source annotations assess whether the queried attribute changes.}
\label{tab:v5_natural}

{\setlength{\tabcolsep}{\dimexpr\tabcolsep*12/14\relax}
\begin{tabular}{ccccccc}
\toprule
\TableHead{Model} & \TableHead{Inputs} & \TableHead{TP} & \TableHead{FP} & \TableHead{FN} & \TableHead{TN} & \TableHead{Acc. (\%)} \\\midrule
Qwen3-4B & Records & \AnswerPctInline{30.2}{35} & \AnswerPctInline{4.7}{11} & \AnswerPctInline{69.8}{81} & \AnswerPctInline{95.3}{221} & 73.6\\Qwen3-4B & Records + question & \AnswerPctInline{61.2}{71} & \AnswerPctInline{7.3}{17} & \AnswerPctInline{38.8}{45} & \AnswerPctInline{92.7}{215} & 82.2\\Qwen3-8B & Records & \AnswerPctInline{50.9}{59} & \AnswerPctInline{12.1}{28} & \AnswerPctInline{49.1}{57} & \AnswerPctInline{87.9}{204} & 75.6\\Qwen3-8B & Records + question & \AnswerPctInline{67.2}{78} & \AnswerPctInline{3.4}{8} & \AnswerPctInline{32.8}{38} & \AnswerPctInline{96.6}{224} & 86.8\\
\bottomrule
\end{tabular}}
\end{table}

Providing the question during value extraction increases recall to 61.2\% and 67.2\%, but the
remaining 45 and 38 cases are missed benchmark answer transitions. Because
this reference includes compatible descriptions and cases in which the relevant attribute cannot be determined,
these counts mix differences in what the labels represent with detection errors. They are not
semantic false-negative counts. When detection is evaluated against annotations of whether the queried attribute changes,
Table~\ref{tab:v39_natural_reviewed} instead reports recall together with false
triggers on compatible pairs. This does not validate automatic localization
over complete novels.
The supplement distributes event IDs, labels, source hashes, regeneration
instructions, predictions, and analysis. Original source passages require
separately obtained OAKS data and are not redistributed.

\subsection{Natural Relation Detection and Masking Outcomes}
\label{app:natural_join}

The detector inputs and constructed negatives are specified in
Appendix~\ref{app:v5_natural}; the analysis below connects their decisions
to generated-answer outcomes.

We retrospectively analyze detector predictions and generated answers on
the same 116 OAKS questions, source events, and A/B excerpts. Each
detector, given either records alone or records with the question, partitions the questions into
detected and missed transitions. For each of the three generators, we
compare full-access and old-mask answers under the protocol in
Appendix~\ref{app:v7_bridge}, using the benchmark reference labels.

\begin{table}[!htbp]
\TableStyle
\caption{All original detector--generator combinations on the same 116 OAKS transitions. Cells give the selected/missed share of all 116 questions, then correction (C) and reversal (R) percentages within that subset; exact counts are retained for full access $\rightarrow$ old-source masking. Each detector row group shares its generated outputs across both prompt conditions; these are paired views, not independent replications.}
\label{tab:natural_detection_action_all}
\begin{tabular}{ccccc}
\toprule
\multirow{2}{*}{\TableLabel{Generator}} & \multicolumn{2}{c}{\TableHead{Records only}} & \multicolumn{2}{c}{\TableHead{Records and question}} \\
\cmidrule(lr){2-3}\cmidrule(l){4-5}
 & \TableHead{Detected: share/C/R} & \TableHead{Missed: share/C/R} & \TableHead{Detected: share/C/R} & \TableHead{Missed: share/C/R} \\\midrule
\TableGroupRow{5}{Detector: Qwen3-4B}
Qwen3-4B & \shortstack[c]{\AnswerPctInline{30.2}{35}\\ C: \AnswerPctInline{25.7}{9}\\ R: \AnswerPctInline{0}{0}} & \shortstack[c]{\AnswerPctInline{69.8}{81}\\ C: \AnswerPctInline{22.2}{18}\\ R: \AnswerPctInline{1.2}{1}} & \shortstack[c]{\AnswerPctInline{61.2}{71}\\ C: \AnswerPctInline{15.5}{11}\\ R: \AnswerPctInline{1.4}{1}} & \shortstack[c]{\AnswerPctInline{38.8}{45}\\ C: \AnswerPctInline{35.6}{16}\\ R: \AnswerPctInline{0}{0}}\\Qwen3-8B & \shortstack[c]{\AnswerPctInline{30.2}{35}\\ C: \AnswerPctInline{5.7}{2}\\ R: \AnswerPctInline{0}{0}} & \shortstack[c]{\AnswerPctInline{69.8}{81}\\ C: \AnswerPctInline{6.2}{5}\\ R: \AnswerPctInline{2.5}{2}} & \shortstack[c]{\AnswerPctInline{61.2}{71}\\ C: \AnswerPctInline{1.4}{1}\\ R: \AnswerPctInline{0}{0}} & \shortstack[c]{\AnswerPctInline{38.8}{45}\\ C: \AnswerPctInline{13.3}{6}\\ R: \AnswerPctInline{4.4}{2}}\\Llama-3.2-3B & \shortstack[c]{\AnswerPctInline{30.2}{35}\\ C: \AnswerPctInline{14.3}{5}\\ R: \AnswerPctInline{2.9}{1}} & \shortstack[c]{\AnswerPctInline{69.8}{81}\\ C: \AnswerPctInline{16}{13}\\ R: \AnswerPctInline{3.7}{3}} & \shortstack[c]{\AnswerPctInline{61.2}{71}\\ C: \AnswerPctInline{11.3}{8}\\ R: \AnswerPctInline{4.2}{3}} & \shortstack[c]{\AnswerPctInline{38.8}{45}\\ C: \AnswerPctInline{22.2}{10}\\ R: \AnswerPctInline{2.2}{1}}\\
\bottomrule
\end{tabular}
\par\smallskip{\footnotesize\TableNoteAlign Constructed negative pairs lack generated outcomes here, so the counts cannot estimate the population cost when a detector incorrectly masks a non-replaced record.\TableNoteEnd}
\end{table}

Recognized transitions include masking reversals, and missed transitions
include corrections. The counts describe the relationship between these
particular decisions and outputs. They do not estimate deployment benefit:
the sample contains annotated positive transitions, while generated outcomes
on the constructed negative pairs are absent. Reusing one question across
detectors and generators does not add independent questions.

\paragraph{Whether an attribute changes and how the change is described.}
The historical annotation makes two separate judgments: whether the queried
attribute changes and how the text describes the change. For P12\_q24, it
labels an attribute change and temporal progression, whereas the rule with
four mutually exclusive classes labels progression. This example is retained
only as a diagnostic of the original benchmark reference labels: the question
asks about a plan, but the selected reference states an achievement.
The source-aware review in Appendix~\ref{app:v41_oaks_open} identifies this
plan-versus-achievement mismatch. The historical labels and counts remain
unchanged; this case does not add source-aware model support for a semantic reversal.
Appendix~\ref{app:v17_sources} examines sensitivity to the annotation rule.
Source identifiers and hashes permit reconstruction of the original inputs
under the dataset's terms.

\paragraph{Annotating changes in the queried attribute.}
A second model annotator applied the prespecified annotation rules to all
116 source pairs without access to the first annotator's labels,
intervention outcomes, or prior analysis discussions. The annotator
identified the entity, attribute, A/B reference times, and A/B asserted
values, then judged whether the queried attribute changes and, separately, how the text describes the change. Aging, role changes, and goal completion can change an attribute's current value even when the old assertion remains historically true. The annotation identifies changes in 22 pairs, compatible statements in 53, and insufficient context in 41. These labels describe the selected examples from five books; they estimate neither the frequency of relations in natural text nor human-validated ground truth.

Alternative annotations quantify ambiguity in each judgment: 61 pairs have
an alternative label for whether the attribute changes and 60 for how the change is described, with 50 in both
sets. Eleven are ambiguous only about attribute change and ten only about its description.
A judgment is ambiguous when the annotator identifies a plausible label
different from its primary choice. The union
contains 71 pairs. Excluding that union leaves 12 attribute changes,
23 compatible pairs and 10 insufficient pairs, as in the tables below.
Excluding ambiguity about attribute change alone instead leaves 14, 31 and 10,
respectively. Thus, ``unambiguous'' means that the historical annotator supplied
no alternative label for either judgment; it does not certify semantic clarity.
The supplement reports
ambiguity in each judgment and detector results under both exclusion rules;
primary annotations remain fixed.
\begin{table*}[!htbp]
\TableStyle

\caption{Generated outcomes grouped by whether the queried attribute changes. Cells give percentages (counts) of the source-relation group $N$; separate rows report full/masked correctness, corrections and reversals. ``Unambiguous'' excludes ambiguity in either judgment. The Qwen rows for unambiguous attribute changes contain one error under both conditions, so zero change does not imply perfect accuracy.}
\label{tab:v18_slot_outcomes}
\begin{tabular}{cccccc}
\toprule
\TableHead{Source relation} & \TableHead{$N$} & \TableHead{Outcome} & \TableHead{Qwen3-4B} & \TableHead{Qwen3-8B} & \TableHead{Llama-3.2-3B} \\
\midrule
Attribute changes & 22 & Full correct & \AnswerPctInline{81.8}{18} & \AnswerPctInline{86.4}{19} & \AnswerPctInline{86.4}{19} \\
 &  & Masked correct & \AnswerPctInline{90.9}{20} & \AnswerPctInline{90.9}{20} & \AnswerPctInline{81.8}{18} \\
 &  & Corrections & \AnswerPctInline{9.1}{2} & \AnswerPctInline{4.5}{1} & \AnswerPctInline{4.5}{1} \\
 &  & Reversals & \AnswerPctInline{0}{0} & \AnswerPctInline{0}{0} & \AnswerPctInline{9.1}{2} \\
\addlinespace[3pt]
\shortstack[c]{Attribute changes,\\unambiguous} & 12 & Full correct & \AnswerPctInline{91.7}{11} & \AnswerPctInline{91.7}{11} & \AnswerPctInline{83.3}{10} \\
 &  & Masked correct & \AnswerPctInline{91.7}{11} & \AnswerPctInline{91.7}{11} & \AnswerPctInline{66.7}{8} \\
 &  & Corrections & \AnswerPctInline{0}{0} & \AnswerPctInline{0}{0} & \AnswerPctInline{0}{0} \\
 &  & Reversals & \AnswerPctInline{0}{0} & \AnswerPctInline{0}{0} & \AnswerPctInline{16.7}{2} \\
\addlinespace[3pt]
Compatible & 53 & Full correct & \AnswerPctInline{60.4}{32} & \AnswerPctInline{83}{44} & \AnswerPctInline{64.2}{34} \\
 &  & Masked correct & \AnswerPctInline{84.9}{45} & \AnswerPctInline{86.8}{46} & \AnswerPctInline{77.4}{41} \\
 &  & Corrections & \AnswerPctInline{24.5}{13} & \AnswerPctInline{5.7}{3} & \AnswerPctInline{17}{9} \\
 &  & Reversals & \AnswerPctInline{0}{0} & \AnswerPctInline{1.9}{1} & \AnswerPctInline{3.8}{2} \\
\addlinespace[3pt]
Insufficient context & 41 & Full correct & \AnswerPctInline{63.4}{26} & \AnswerPctInline{82.9}{34} & \AnswerPctInline{53.7}{22} \\
 &  & Masked correct & \AnswerPctInline{90.2}{37} & \AnswerPctInline{87.8}{36} & \AnswerPctInline{73.2}{30} \\
 &  & Corrections & \AnswerPctInline{29.3}{12} & \AnswerPctInline{7.3}{3} & \AnswerPctInline{19.5}{8} \\
 &  & Reversals & \AnswerPctInline{2.4}{1} & \AnswerPctInline{2.4}{1} & \AnswerPctInline{0}{0} \\
\addlinespace[3pt]
\bottomrule
\end{tabular}\end{table*}
The two Llama reversals against the benchmark labels within these 12 pairs
are P12\_q24 and P37\_q18. The historical annotator describes both as
progression, but P12\_q24 has a plan-versus-achievement mismatch
(Appendix~\ref{app:v41_oaks_open}); it is a benchmark-label reversal, not a
confirmed semantic reversal or additional source-semantic support.
We retain the original labels and counts. The two Qwen models have no
benchmark-label reversals within the 22 annotated attribute changes.
These results describe masking outcomes for the tested models and protocols;
they do not establish the prevalence of errors on natural supersession.
\paragraph{Detector errors against source annotations.}
Among the 75 resolvable pairs, 22 are annotated attribute changes and 53 are compatible. Detection is compared with the attribute-change annotation, while the 41 insufficient pairs remain unscored. This separates missed semantic changes from missed benchmark answer transitions. All 116 cases were originally selected as benchmark answer transitions, so the 53 compatible pairs are not a representative sample of unchanged-answer histories.
Table~\ref{tab:v39_natural_reviewed} combines the full seven-model comparison with the original two-model confusion counts and ambiguity-exclusion sensitivity.
On all 75 resolvable pairs, adding the query raises Qwen3-4B recall from
$10/22=45.5\%$ to $18/22=81.8\%$, while its compatible-pair false-trigger
rate rises from $18/53=34.0\%$ to $35/53=66.0\%$.
For Qwen3-8B, recall rises from $13/22=59.1\%$ to $21/22=95.5\%$, while
the false-trigger rate rises from $28/53=52.8\%$ to $31/53=58.5\%$.
These are comparisons against the model annotator's attribute-change labels on the
fixed candidate pool, not verified human semantic-error rates.
The four-class rule and the rule that separately judges attribute change and its description define different targets. Of
the 29 strict replacements under the four-class rule, 15 are attribute
changes under the separate judgment, five are compatible, and nine have insufficient context. Of the 32 progressions, seven are attribute changes, 14
are compatible, and 11 have insufficient context. These differences measure
sensitivity to annotation rules, not agreement between raters using the
same rule.
\paragraph{Sensitivity to mutually exclusive relation classes.}
The 116 answer transitions associated with specific questions come from five books. We
retrospectively stratify outcomes under the mutually exclusive four-class
rubric, which assigns aging and goal fulfillment to progression rather than
strict replacement. The separate judgments permit temporal
progression to coincide with a change in the queried attribute: the annotator compares
$(\text{entity},\text{attribute},\text{reference time},\text{asserted value})$
and independently labels how the text describes the change.
Table~\ref{tab:v17_source} brings these detector fractions together with the complete generated outcomes and correction/reversal counts for each class.
\paragraph{Dependence introduced by constructed negatives.}
Connecting question groups that use an identical A or B excerpt anywhere in the three-condition detection experiment produces one connected component containing all 116 questions. Cross-book negatives reuse donor B excerpts, so a component bootstrap cannot estimate uncertainty from independent components. The within-book question bootstrap is conditional on the fixed books and negative construction. We additionally report point-estimate changes after omitting each question book and after omitting each negative donor book. These are descriptive sensitivity analyses. This dependence arises in the detection experiment with constructed
negatives. The generation experiment contains only positive transitions
and does not reuse negative donor excerpts.
\paragraph{Coverage and net corrections.}
For each fixed operation define $U_e=\mathbf{1}\{\text{masked answer correct}\}-\mathbf{1}\{\text{full answer correct}\}$. We rank natural transitions by a binary detector or a measured score change and show expected net corrections at 10\% coverage increments. A binary detector produces two tied score groups. Within each group, we average over uniform selection. Bootstrap draws resample questions and recompute the ranking and boundary tie size.
At 20\% coverage, ranking by the reference-dependent margin score yields expected net corrections of 18.2, 7.0, and 10.0 for Qwen3-4B, Qwen3-8B, and Llama-3.2-3B, compared with 5.2, 1.0, and 2.8 under uniform selection. These scores read current/old reference candidates and require extra masked computation; they support retrospective analysis and are unavailable to a controller
without reference answers and additional computation. The natural candidate pool contains annotated positive answer transitions only. Generated outcomes on constructed non-replacements are absent, so these curves omit answer errors caused by masking those cases and do not establish an overall benefit in deployment.
\begin{table}[!htbp]
\WideTableStyle

\caption{Selected entries are percentages (counts); Net and Random are percentage points (observed or expected counts), using denominator 116. Binary relation selection on all 116 natural transitions, using the same detector and generator model. Random is expected net corrections at the same coverage; the final column is the gain over that baseline in percentage points of all 116 questions.}
\label{tab:v18_binary_selection}
\begin{tabular}{cccccc}
\toprule
\TableHead{Model} & \TableHead{Input} & \TableHead{Selected} & \TableHead{Net} & \TableHead{Random (pp)} & \TableHead{Gain pp, 95\% CI} \\\midrule
Qwen 4B & A/B & \AnswerPctInline{30.2}{35} & \AnswerPPInline{+7.8}{+9} & \AnswerPPInline{+6.8}{+7.84} & +1.00 \TableSecondary{[$-$2.62,\,+4.64]}\\Qwen 4B & A/B + query & \AnswerPctInline{61.2}{71} & \AnswerPPInline{+8.6}{+10} & \AnswerPPInline{+13.7}{+15.91} & $-$5.10 \TableSecondary{[$-$9.01,\,$-$1.13]}\\Qwen 8B & A/B & \AnswerPctInline{50.9}{59} & \AnswerPPInline{+1.7}{+2} & \AnswerPPInline{+2.2}{+2.54} & $-$0.47 \TableSecondary{[$-$2.99,\,+1.92]}\\Qwen 8B & A/B + query & \AnswerPctInline{67.2}{78} & \AnswerPPInline{+2.6}{+3} & \AnswerPPInline{+2.9}{+3.36} & $-$0.31 \TableSecondary{[$-$2.90,\,+2.20]}\\
\bottomrule\end{tabular}\end{table}

\begin{figure}[!htbp]
\centering
\includegraphics[width=\linewidth]{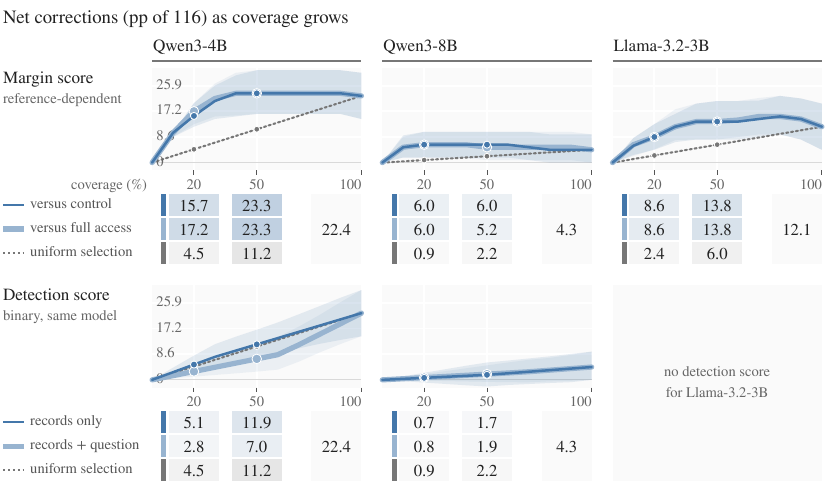}
\caption{Net corrections as the selected fraction increases on 116 annotated natural answer transitions. Top: reference-dependent score changes using additional masked computation. Bottom: binary scores from the same detector/generator model, randomized within ties. Shading is pointwise 95\% question-bootstrap uncertainty conditional on the observed books. The candidate pool excludes constructed negatives; improvement from reference-dependent ranking does not establish a benefit for an online controller.}
\label{fig:v18_natural_utility}
\end{figure}
\FloatBarrier

\subsection{Recognition and Query Routing Answer Different Questions}
\label{app:v39_semantic_extension}

We separate relation recognition from the effect of using its predictions.
Each detector evaluates three prompts on the same 300 test
conditions from 50 groups, and seven policies answer current, historical,
and unrelated questions. The detector and generator have different roles:
Table~\ref{tab:v39_semantic_detection} varies the detector, whereas
Table~\ref{tab:v39_semantic_routing} keeps the Qwen3-4B detector fixed
while varying the generator. A stronger generator's own recognition score
is therefore not the routing accuracy of the latter experiment.

\begin{table}[!htbp]
\TableStyle

\caption{Outcome entries give percentages, with the original count/denominator beneath. Replacement detection on 300 test conditions from 50 independent synthetic groups. The original prompt extracts old and new values; the expanded prompt adds update examples and a duplicate-value guard; the third compares values affirmed by each record. The final columns use the third prompt.}
\label{tab:v39_semantic_detection}
{\setlength{\tabcolsep}{\dimexpr\tabcolsep*10/12\relax}
\begin{tabular}{cccccc}
\toprule
\multirow{2}{*}[-0.6ex]{\TableLabel{Model}} & \multicolumn{3}{c}{\TableHead{Synthetic detector accuracy}} & \multicolumn{2}{c}{\TableHead{Affirmed-value triggers}} \\
\cmidrule(lr){2-4}\cmidrule(l){5-6}
 & \TableHead{Original} & \TableHead{Expanded} & \TableHead{Affirmed values} & \TableHead{True triggers} & \TableHead{False triggers} \\\midrule
Qwen3-32B & \AnswerPct{95.7}{287/300} & \AnswerPct{91.3}{274/300} & \AnswerPct{98}{294/300} & \AnswerPct{100}{150/150} & \AnswerPct{4}{6/150}\\Qwen3.8-27B$^{\dagger}$ & \AnswerPct{100}{300/300} & \AnswerPct{100}{300/300} & \AnswerPct{100}{300/300} & \AnswerPct{100}{150/150} & \AnswerPct{0}{0/150}\\Gemma 4 26B-A4B & \AnswerPct{92}{276/300} & \AnswerPct{89.3}{268/300} & \AnswerPct{93.3}{280/300} & \AnswerPct{99.3}{149/150} & \AnswerPct{12.7}{19/150}\\Gemma 4 31B & \AnswerPct{100}{300/300} & \AnswerPct{98.3}{295/300} & \AnswerPct{100}{300/300} & \AnswerPct{100}{150/150} & \AnswerPct{0}{0/150}\\Llama-3.1-70B & \AnswerPct{96.7}{290/300} & \AnswerPct{93.7}{281/300} & \AnswerPct{98}{294/300} & \AnswerPct{100}{150/150} & \AnswerPct{4}{6/150}\\
\bottomrule
\end{tabular}}
\end{table}

Comparing affirmed values is perfect on these cases for Qwen3.8 and Gemma 4 31B,
but neither perfect synthetic recognition nor an oracle replacement label
guarantees preservation under masking. With oracle masks, historical
correct answers fall from 300 to 280 for Qwen3.8 and from 300 to 247 for
Gemma 4 31B. Conversely, routing that uses the supplied question type restores
every model's full-access historical and unrelated counts while leaving
current-query losses. This is an intervention on which questions receive
the mask; it is not a learned solution to recognizing question type.

\begin{table}[!htbp]
\TableStyle

\caption{Complete answers as percentages (counts), with 300 cases per question type and a 256-token budget. Detected policies use Qwen3-4B; the oracle uses known replacement labels.}
\label{tab:v39_semantic_routing}
{\setlength{\tabcolsep}{\dimexpr\tabcolsep*8/10\relax}
\begin{tabular}{cccccc}
\toprule
\TableHead{Model} & \TableHead{Question} & \FullCell{\FullHeader{Full}} & \WholeCell{\WholeHeader{Detected}} & \TableHead{Routing} & \WholeCell{\WholeHeader{Oracle}} \\
\midrule
Qwen3-32B & Current & \AnswerPctInline{99}{297} & \WholeCell{\AnswerPctInline{93.7}{281}} & \AnswerPctInline{93.7}{281} & \WholeCell{\AnswerPctInline{98.3}{295}} \\
 & Historical & \AnswerPctInline{100}{300} & \WholeCell{\AnswerPctInline{77}{231}} & \AnswerPctInline{100}{300} & \WholeCell{\AnswerPctInline{82.3}{247}} \\
 & Unrelated & \AnswerPctInline{99.3}{298} & \WholeCell{\AnswerPctInline{94.3}{283}} & \AnswerPctInline{99.3}{298} & \WholeCell{\AnswerPctInline{93.7}{281}} \\
\addlinespace[2pt]
Qwen3.8-27B$^{\dagger}$ & Current & \AnswerPctInline{100}{300} & \WholeCell{\AnswerPctInline{94.3}{283}} & \AnswerPctInline{94.3}{283} & \WholeCell{\AnswerPctInline{100}{300}} \\
 & Historical & \AnswerPctInline{100}{300} & \WholeCell{\AnswerPctInline{88}{264}} & \AnswerPctInline{100}{300} & \WholeCell{\AnswerPctInline{93.3}{280}} \\
 & Unrelated & \AnswerPctInline{100}{300} & \WholeCell{\AnswerPctInline{100}{300}} & \AnswerPctInline{100}{300} & \WholeCell{\AnswerPctInline{100}{300}} \\
\addlinespace[2pt]
Gemma 4 26B-A4B & Current & \AnswerPctInline{100}{300} & \WholeCell{\AnswerPctInline{94}{282}} & \AnswerPctInline{94}{282} & \WholeCell{\AnswerPctInline{99.7}{299}} \\
 & Historical & \AnswerPctInline{100}{300} & \WholeCell{\AnswerPctInline{78.7}{236}} & \AnswerPctInline{100}{300} & \WholeCell{\AnswerPctInline{84.3}{253}} \\
 & Unrelated & \AnswerPctInline{98.3}{295} & \WholeCell{\AnswerPctInline{98.3}{295}} & \AnswerPctInline{98.3}{295} & \WholeCell{\AnswerPctInline{98.3}{295}} \\
\addlinespace[2pt]
Gemma 4 31B & Current & \AnswerPctInline{100}{300} & \WholeCell{\AnswerPctInline{94.3}{283}} & \AnswerPctInline{94.3}{283} & \WholeCell{\AnswerPctInline{99.7}{299}} \\
 & Historical & \AnswerPctInline{100}{300} & \WholeCell{\AnswerPctInline{77.3}{232}} & \AnswerPctInline{100}{300} & \WholeCell{\AnswerPctInline{82.3}{247}} \\
 & Unrelated & \AnswerPctInline{100}{300} & \WholeCell{\AnswerPctInline{100}{300}} & \AnswerPctInline{100}{300} & \WholeCell{\AnswerPctInline{100}{300}} \\
\addlinespace[2pt]
Llama-3.1-70B & Current & \AnswerPctInline{83.7}{251} & \WholeCell{\AnswerPctInline{80.3}{241}} & \AnswerPctInline{80.3}{241} & \WholeCell{\AnswerPctInline{82}{246}} \\
 & Historical & \AnswerPctInline{93}{279} & \WholeCell{\AnswerPctInline{70.3}{211}} & \AnswerPctInline{93}{279} & \WholeCell{\AnswerPctInline{74}{222}} \\
 & Unrelated & \AnswerPctInline{78.3}{235} & \WholeCell{\AnswerPctInline{78.7}{236}} & \AnswerPctInline{78.3}{235} & \WholeCell{\AnswerPctInline{78.7}{236}} \\
\addlinespace[2pt]
\bottomrule
\end{tabular}}
\par\smallskip{\footnotesize\TableNoteAlign Query routing permits masking only for a supplied current-query label. It does not include a learned classifier of query type. Shorter 40- and 128-token views are truncations of the same native outputs.\TableNoteEnd}
\end{table}

A legacy guard that rejects any pair repeating the old value reduces the
third detector's true triggers to 50/150 in all five models. Repetition
alone therefore discards valid changes when the old value occurs in a
negation or comparison. The result identifies a limitation of that lexical
guard on this construction, rather than a general failure of the model's
semantic detector.

\paragraph{Question conditioning can increase both sensitivity and false actions.}
\begin{table}[!htbp]
\WideTableStyle
\caption{Confusion entries give percentages (counts): TP/FN use $N_+$, FP/TN use $N_-$, and accuracy uses $N_++N_-$. Question conditioning against annotations of whether the queried attribute changes. Each prompt reports change cases as TP/FN, compatible cases as FP/TN, and the correct-label count. The 22/53 groups combine the seven-model comparison with the original confusion counts; the 12/23 groups give the two-model sensitivity analysis after excluding ambiguous annotations.}
\label{tab:v39_natural_reviewed}
\begin{tabular}{cccccc}
\toprule
\TableHead{Detector} & \TableHead{TP} & \TableHead{FN} & \TableHead{FP} & \TableHead{TN} & \TableHead{Accuracy} \\
\midrule
\TableGroupRow{6}{Records only: $N_+=22$, $N_-=53$}
Qwen3-4B & \AnswerPctInline{45.5}{10} & \AnswerPctInline{54.5}{12} & \AnswerPctInline{34}{18} & \AnswerPctInline{66}{35} & \AnswerPctInline{60}{45} \\
Qwen3-8B & \AnswerPctInline{59.1}{13} & \AnswerPctInline{40.9}{9} & \AnswerPctInline{52.8}{28} & \AnswerPctInline{47.2}{25} & \AnswerPctInline{50.7}{38} \\
Qwen3-32B & \AnswerPctInline{50}{11} & \AnswerPctInline{50}{11} & \AnswerPctInline{30.2}{16} & \AnswerPctInline{69.8}{37} & \AnswerPctInline{64}{48} \\
Qwen3.8-27B$^{\dagger}$ & \AnswerPctInline{40.9}{9} & \AnswerPctInline{59.1}{13} & \AnswerPctInline{9.4}{5} & \AnswerPctInline{90.6}{48} & \AnswerPctInline{76}{57} \\
Gemma 4 26B-A4B & \AnswerPctInline{45.5}{10} & \AnswerPctInline{54.5}{12} & \AnswerPctInline{3.8}{2} & \AnswerPctInline{96.2}{51} & \AnswerPctInline{81.3}{61} \\
Gemma 4 31B & \AnswerPctInline{68.2}{15} & \AnswerPctInline{31.8}{7} & \AnswerPctInline{32.1}{17} & \AnswerPctInline{67.9}{36} & \AnswerPctInline{68}{51} \\
Llama-3.1-70B & \AnswerPctInline{45.5}{10} & \AnswerPctInline{54.5}{12} & \AnswerPctInline{22.6}{12} & \AnswerPctInline{77.4}{41} & \AnswerPctInline{68}{51} \\
\addlinespace[3pt]
\TableGroupRow{6}{Records and question: $N_+=22$, $N_-=53$}
Qwen3-4B & \AnswerPctInline{81.8}{18} & \AnswerPctInline{18.2}{4} & \AnswerPctInline{66}{35} & \AnswerPctInline{34}{18} & \AnswerPctInline{48}{36} \\
Qwen3-8B & \AnswerPctInline{95.5}{21} & \AnswerPctInline{4.5}{1} & \AnswerPctInline{58.5}{31} & \AnswerPctInline{41.5}{22} & \AnswerPctInline{57.3}{43} \\
Qwen3-32B & \AnswerPctInline{77.3}{17} & \AnswerPctInline{22.7}{5} & \AnswerPctInline{73.6}{39} & \AnswerPctInline{26.4}{14} & \AnswerPctInline{41.3}{31} \\
Qwen3.8-27B$^{\dagger}$ & \AnswerPctInline{63.6}{14} & \AnswerPctInline{36.4}{8} & \AnswerPctInline{34}{18} & \AnswerPctInline{66}{35} & \AnswerPctInline{65.3}{49} \\
Gemma 4 26B-A4B & \AnswerPctInline{77.3}{17} & \AnswerPctInline{22.7}{5} & \AnswerPctInline{56.6}{30} & \AnswerPctInline{43.4}{23} & \AnswerPctInline{53.3}{40} \\
Gemma 4 31B & \AnswerPctInline{90.9}{20} & \AnswerPctInline{9.1}{2} & \AnswerPctInline{69.8}{37} & \AnswerPctInline{30.2}{16} & \AnswerPctInline{48}{36} \\
Llama-3.1-70B & \AnswerPctInline{86.4}{19} & \AnswerPctInline{13.6}{3} & \AnswerPctInline{64.2}{34} & \AnswerPctInline{35.8}{19} & \AnswerPctInline{50.7}{38} \\
\addlinespace[3pt]
\TableGroupRow{6}{Records only: $N_+=12$, $N_-=23$}
Qwen3-4B & \AnswerPctInline{58.3}{7} & \AnswerPctInline{41.7}{5} & \AnswerPctInline{39.1}{9} & \AnswerPctInline{60.9}{14} & \AnswerPctInline{60}{21} \\
Qwen3-8B & \AnswerPctInline{66.7}{8} & \AnswerPctInline{33.3}{4} & \AnswerPctInline{60.9}{14} & \AnswerPctInline{39.1}{9} & \AnswerPctInline{48.6}{17} \\
\addlinespace[3pt]
\TableGroupRow{6}{Records and question: $N_+=12$, $N_-=23$}
Qwen3-4B & \AnswerPctInline{83.3}{10} & \AnswerPctInline{16.7}{2} & \AnswerPctInline{73.9}{17} & \AnswerPctInline{26.1}{6} & \AnswerPctInline{45.7}{16} \\
Qwen3-8B & \AnswerPctInline{91.7}{11} & \AnswerPctInline{8.3}{1} & \AnswerPctInline{60.9}{14} & \AnswerPctInline{39.1}{9} & \AnswerPctInline{57.1}{20} \\
\addlinespace[3pt]
\bottomrule
\end{tabular}
\par\smallskip{\footnotesize\TableNoteAlign TP/FN are detected/missed changes; FP/TN are triggered/untriggered compatible pairs. The 22/53 groups exclude 41 unresolved pairs. The 12/23 groups exclude ambiguity in either attribute change or its description. These annotations were produced by models in isolated contexts, without independent human verification; they differ from the 348 constructed detection targets used elsewhere.\TableNoteEnd}
\end{table}

The natural-text comparison uses the same reviewed 22 changes, 53 compatible
pairs, and 41 unresolved pairs as Appendix~\ref{app:natural_join}. These labels answer whether
the queried attribute changes, rather than treating every later narrative
answer as an invalidation of the earlier statement. The complete seven-model
table keeps increased true triggers and increased false triggers visible
instead of reporting only agreement with the original positive construction.
Adding the question increases detected changes in all seven models, but
decreases the number of correct resolvable labels in six. The five larger
models all show that tradeoff: Qwen3.8 falls from 57/75 to 49/75 correct labels,
and Gemma 4 26B-A4B falls from 61/75 to 40/75. Qwen3-8B is the exception,
improving from 38/75 to 43/75. On these annotated cases, a question can identify the relevant attribute
while also encouraging a detector to call ordinary narrative progression
a replacement. The table measures this tradeoff; it does not establish
the detector's reasoning mechanism or human-validated natural accuracy.

Appendix~\ref{app:v41_complementary} compares sensitivities to questions, tasks and readouts.

\subsection{Natural Detection Does Not Reliably Rank Mask Benefits}
\label{app:v39_natural_action}

We compare seven detectors with eight generators on the same natural-text
questions, using the alignment defined in Appendix~\ref{app:natural_join}.
The comparison keeps detector identity separate from
generator identity and retains both prompt variants, the native and
truncated-budget views, and both source-label taxonomies. The 116 positive
answer transitions have 111 connected source components: 106 singletons
and five pairs. Question-resampled intervals are conditional on the fixed
books; the supplement also retains component-resampled outcome intervals.

\begin{table}[!htbp]
\WideTableStyle

\caption{Selected/net cells show selection percentage (count) above net accuracy change in pp (signed count), both over 116 questions. Whether a detector selects helpful masks on 116 narrative answer transitions. For matched detector and generator checkpoints, each prompt shows selected cases / net corrections and the excess net accuracy over random selection at the same coverage, in percentage points with a paired 95\% interval.}
\label{tab:v39_natural_action}
{\setlength{\tabcolsep}{\dimexpr\tabcolsep*8/10\relax}
\begin{tabular}{ccccc}
\toprule
\multirow{2}{*}[-0.6ex]{\TableLabel{Model}} & \multicolumn{2}{c}{\TableHead{Records only}} & \multicolumn{2}{c}{\TableHead{Records and question}} \\
\cmidrule(lr){2-3}\cmidrule(l){4-5}
 & \TableHead{Selected / net} & \TableHead{Excess over random} & \TableHead{Selected / net} & \TableHead{Excess over random} \\\midrule
Qwen3-4B & \shortstack[c]{\AnswerPctInline{30.2}{35}\\\AnswerPPInline{+7.8}{+9}} & +1.00 \TableSecondary{[$-$2.50,\,4.74]} & \shortstack[c]{\AnswerPctInline{61.2}{71}\\\AnswerPPInline{+8.6}{+10}} & $-$5.10 \TableSecondary{[$-$8.86,\,$-$1.43]}\\Qwen3-8B & \shortstack[c]{\AnswerPctInline{50.9}{59}\\\AnswerPPInline{+1.7}{+2}} & $-$0.47 \TableSecondary{[$-$2.99,\,2.00]} & \shortstack[c]{\AnswerPctInline{67.2}{78}\\\AnswerPPInline{+2.6}{+3}} & $-$0.31 \TableSecondary{[$-$2.90,\,2.05]}\\Qwen3-32B & \shortstack[c]{\AnswerPctInline{30.2}{35}\\\AnswerPPInline{+1.7}{+2}} & +0.42 \TableSecondary{[$-$2.07,\,3.05]} & \shortstack[c]{\AnswerPctInline{63.8}{74}\\\AnswerPPInline{+1.7}{+2}} & $-$1.03 \TableSecondary{[$-$3.72,\,1.52]}\\Qwen3.8-27B$^{\dagger}$ & \shortstack[c]{\AnswerPctInline{14.7}{17}\\\AnswerPPInline{0}{+0}} & $-$1.14 \TableSecondary{[$-$2.03,\,$-$0.45]} & \shortstack[c]{\AnswerPctInline{31.9}{37}\\\AnswerPPInline{+0.9}{+1}} & $-$1.61 \TableSecondary{[$-$3.40,\,0.24]}\\Gemma 4 26B-A4B & \shortstack[c]{\AnswerPctInline{13.8}{16}\\\AnswerPPInline{+0.9}{+1}} & $-$0.21 \TableSecondary{[$-$1.66,\,1.63]} & \shortstack[c]{\AnswerPctInline{47.4}{55}\\\AnswerPPInline{+1.7}{+2}} & $-$1.95 \TableSecondary{[$-$5.08,\,1.16]}\\Gemma 4 31B & \shortstack[c]{\AnswerPctInline{34.5}{40}\\\AnswerPPInline{+1.7}{+2}} & +1.13 \TableSecondary{[$-$0.62,\,3.18]} & \shortstack[c]{\AnswerPctInline{64.7}{75}\\\AnswerPPInline{+2.6}{+3}} & +1.47 \TableSecondary{[$-$0.56,\,3.72]}\\Llama-3.1-70B & \shortstack[c]{\AnswerPctInline{25}{29}\\\AnswerPPInline{+0.9}{+1}} & $-$0.43 \TableSecondary{[$-$1.96,\,1.28]} & \shortstack[c]{\AnswerPctInline{61.2}{71}\\\AnswerPPInline{+0.9}{+1}} & $-$2.30 \TableSecondary{[$-$4.75,\,0.01]}\\
\bottomrule
\end{tabular}}
\par\smallskip{\footnotesize\TableNoteAlign Each checkpoint uses its native generation budget: 40 tokens for the two original Qwen models and 256 for the added models. Intervals resample questions within the five fixed books. They are descriptive and do not provide simultaneous coverage across checkpoints.\TableNoteEnd}
\end{table}

Increasing the number of triggered masks does not ensure more corrections.
For Llama-3.1-70B, adding the question increases selected cases from 29 to
71, yet the net correction count stays at one. For Qwen3.8, the corresponding
counts are 17 to 37 and zero to one. At matched coverage, these changes
do not exceed the expected benefit from random selection. None of the
14 matched-checkpoint intervals has a positive lower bound; Qwen3.8's
records-only interval is below zero. This comparison tests whether the
detector locates helpful interventions, rather than only whether it
recognizes the annotated relation.

The full cross-model comparison is less uniform and is retained instead
of extrapolating the matched-checkpoint result. Among the 112 native-budget
detector--generator--prompt cells, question-conditioned Qwen3-8B selecting
Gemma 4 26B-A4B has a positive descriptive interval: $+2.54$ points
$[0.09,5.07]$ above equal-coverage random selection. This isolated pointwise
interval is not a multiplicity-controlled finding or a policy selected on
independent development data. The other cells, exact tie handling, coverage
curves, and rankings based on reference margins are retained in the numerical supplement for every detector--generator
pairing.

The generation pool contains the original positive answer transitions;
constructed repeated-source and different-book negatives have detection
outputs but no corresponding mask-generation outcomes here. The join
therefore cannot estimate the population cost of false actions. Likewise,
ranking by a margin computed from the known answer candidates requires
reference answers and extra masked computation. It is a retrospective
comparison, not a deployable replacement detector.

\subsection{Sensitivity to How Source Relations Are Classified}
\label{app:v17_sources}

The annotation rule with four mutually exclusive classes identifies 29 strict replacements, 32 state
progressions, 40 additive or descriptive changes, and 15 cases with
insufficient context. These labels provide an alternative semantic classification without
changing the benchmark's answer targets. Because progression and strict
replacement are mutually exclusive, this rubric measures a different
target from the annotation of whether the queried attribute changes. An
earlier statement can remain historically true while the current value of that attribute changes.

\paragraph{Labeling rubric and review procedure.}
The four categories were specified before examining case-level
intervention outcomes.
Strict replacement requires incompatible asserted values of the same
question-relevant attribute, including an explicit revision of belief or
status. State progression includes aging, successive activities, process
completion, and fulfillment of an intention. Additive description introduces
compatible details or characterizations. Insufficient context applies when
the supplied extracts cannot securely establish the entity, attribute,
temporal relation, or relevant change. Candidate-answer differences alone
do not establish replacement.

A model annotator classified every A/B pair and question in random
order, without answer candidates, generator identity, detector predictions,
intervention condition, or generated answers. The annotator had previously
seen aggregate results and the Avery example, so annotation was not fully
blind to the study. The four-class annotations have no independent human
or second-rater validation. The separate annotation of attribute changes and their descriptions is reported in
Table~\ref{tab:v18_slot_outcomes}. Each annotation includes a source-based
rationale and ambiguity flag; source identifiers and text hashes permit
reconstruction. Ambiguity flags describe plausible alternative interpretations
and are not inter-rater agreement measurements.

Fifty-seven cases admit a plausible alternative category, including all
15 insufficient-context cases. Excluding these leaves 17 strict replacements,
17 progressions, and 25 additive descriptions. For example, a new teaching
appointment can be viewed as either a change in the person's current job or a career progression;
an overheard conversation may add information without establishing a new
relationship. The labels therefore support a bounded distinction, not a
precise population estimate of relation prevalence.

\paragraph{Connection to detection and generation.}
Source labels were assigned before examining case-level outcomes. We
stratify the 12 detector--prompt--generator combinations by source class:
two Qwen detectors, prompts containing records alone or records with the question, and three
generators. The analysis includes all 1,392 question--combination outcomes,
reporting correct, old, and UNKNOWN answers under full access and masking,
as well as corrections and reversals, within detected and missed subsets.
The supplement also reports results excluding ambiguous cases. The repeated
conditions share the same 116 questions and do not increase the sample size.

\begin{table}[!htbp]
\WideTableStyle
\caption{Detection and generated answers under the same mutually exclusive four-class rubric. All outcome entries give percentages (counts) of the source class. F/M denotes full access/source masking; corrections and reversals use the same class denominator. Detector rows use records-only inputs; the dash denotes an unmeasured detector. Every result uses its source-class denominator; all 116 questions, including ambiguous cases, retain the original benchmark answer targets.}
\label{tab:v17_source}
\begin{tabular}{ccccc}
\toprule
\TableHead{Source class ($N$)} & \TableHead{Outcome} & \TableHead{Qwen3-4B} & \TableHead{Qwen3-8B} & \TableHead{Llama-3.2-3B} \\
\midrule
\shortstack[c]{Strict replacement\\(29)} & Detector trigger & \AnswerPctInline{27.6}{8} & \AnswerPctInline{51.7}{15} & -- \\
 & Correct: F $\to$ M & \shortstack[c]{\AnswerPctInline{79.3}{23} $\to$\\\AnswerPctInline{93.1}{27}} & \shortstack[c]{\AnswerPctInline{89.7}{26} $\to$\\\AnswerPctInline{93.1}{27}} & \shortstack[c]{\AnswerPctInline{75.9}{22} $\to$\\\AnswerPctInline{79.3}{23}} \\
 & Old: F $\to$ M & \shortstack[c]{\AnswerPctInline{20.7}{6} $\to$\\\AnswerPctInline{6.9}{2}} & \shortstack[c]{\AnswerPctInline{10.3}{3} $\to$\\\AnswerPctInline{0}{0}} & \shortstack[c]{\AnswerPctInline{24.1}{7} $\to$\\\AnswerPctInline{20.7}{6}} \\
 & UNKNOWN: F $\to$ M & \shortstack[c]{\AnswerPctInline{0}{0} $\to$\\\AnswerPctInline{0}{0}} & \shortstack[c]{\AnswerPctInline{0}{0} $\to$\\\AnswerPctInline{6.9}{2}} & \shortstack[c]{\AnswerPctInline{0}{0} $\to$\\\AnswerPctInline{0}{0}} \\
 & Corrections & \AnswerPctInline{13.8}{4} & \AnswerPctInline{3.4}{1} & \AnswerPctInline{6.9}{2} \\
 & Reversals & \AnswerPctInline{0}{0} & \AnswerPctInline{0}{0} & \AnswerPctInline{3.4}{1} \\
\addlinespace[3pt]
\shortstack[c]{Progression\\(32)} & Detector trigger & \AnswerPctInline{43.8}{14} & \AnswerPctInline{59.4}{19} & -- \\
\bottomrule
\end{tabular}
\par\smallskip{\footnotesize\TableNoteAlign Detector fractions outside the strict-replacement row are trigger frequencies, not replacement recall. The source classes describe one model annotator's interpretation, without independent human validation. They differ from the attribute-change annotations in Table~\ref{tab:v18_slot_outcomes}.\TableNoteEnd}
\end{table}

Among the 29 strict replacements under this rubric, corrections/reversals from full access to masking
are 4/0, 1/0, and 2/1 for Qwen3-4B, Qwen3-8B, and Llama-3.2-3B.
The additive class accounts for the largest number of Qwen3-4B corrections
(12 of 27). Qwen3-8B's two reversals fall in additive and insufficient-context
classes, one each. These labels clarify what the natural answer changes represent; they do not establish that the relation class predicts whether masking improves an answer.
Generated answers are unavailable for the constructed natural negatives, so the cost of masking incorrectly and net deployment benefit remain unmeasured.

\paragraph{Source reconstruction.}
The public supplement includes question/event IDs, SHA-256 hashes, labels,
and model-written summaries. Original novel excerpts are not redistributed;
they can be reconstructed from separately obtained OAKS data using the
released extraction procedure and verified against the supplied hashes.
Intentions and state progressions are not treated as benchmark errors.

\FloatBarrier
\section{Updating KV States and Preserving Useful Information}
\label{app:operation_studies}

These experiments supply the locations of old records to isolate the
effects of the cache operation. We first examine the information needed
for complete current and historical answers, then vary source details,
question wording and controls. Component interventions and recomputation
finally test alternatives to masking. The quantity studies distinguish
the 120-group six-unit task from the earlier 300-condition record task.

\subsection{Answering Quantity Questions and Selecting Masks}
\label{app:v18_unified}

The protocol and initial generation results below concern Qwen3-4B,
Qwen3-8B, and Llama-3.2-3B. The selection curves and exclusion analysis
include all eight generators on the same groups;
Appendix~\ref{app:v33_models} reports their complete generation results.

This experiment tests whether answer preservation depends on the later
record providing a complete quantity. It comprises 120 independently constructed test groups: 60 strict
replacements and 20 each of confirmation, other-attribute, and other-entity
updates. Each group records an entity, attribute, reference time, and
asserted value for A and B. The test uses minutes, hours, meters, liters,
grams, and volts. The 60 strict groups contain respectively
9, 6, 14, 10, 11, and 10 groups in those units. Eighteen development
groups use separate templates, entities, values, and units. This synthetic test controls the information supplied by each record.
Its design used the findings from the preceding quantity comparisons;
it is not an external sample of natural text.

\paragraph{Source information, queries, and operations.}
In one condition, B explicitly states the complete quantity. In the paired
condition, B refers to A for its unit. Neither the unrelated
inspection-label record nor the repeated control note contains the target
unit. The control note describes a square seal, blank signature box, gray
cover, and closed folder. All queries explicitly request complete units
for quantities, without naming the unit, and forbid conversion.
Current, T1-historical, and unrelated-label questions are evaluated independently
from the same history-only prefix; generated answers are never appended
to subsequent histories. Each condition is full access, a whole-source mask,
its equal-token control, a number-only mask, or its equal-token control.
Tokenizer offsets verify that number-only masks exclude unit tokens. Control
masks select the beginning of the note. The positions to be masked are supplied in this experiment.

\paragraph{Pre-specified protocol and inference settings.}
The specification and all test texts, canonical values, splits, score
formulas, selection rules, and resampling units were fixed before model
outputs were inspected. All three models ran native chat, greedy
EOS-stopped generation, 40 answer tokens, bfloat16 weights, and eager
attention on H200. The runtime uses Torch 2.11.0+cu130, Transformers 5.12.1,
and NumPy 2.4.6; the template date is fixed to 16 September 2026.
The experiment measuring both candidate scores and generated answers in Appendix~\ref{app:v7_bridge}
uses a different prompt construction and Transformers 5.8.0.

The evaluation comprises 10,800 test and 1,620 development outputs.
Across 828 model--history combinations covering both splits and information
conditions, the original K/V tensors remain identical after evaluating
the mask conditions. Nine cached-versus-full-prefill comparisons on
development examples agree in first-token choices and complete answers.
Their maximum bfloat16
first-step logit difference is 0.89, so they establish output
agreement rather than bitwise numerical parity. Every test answer satisfies
the strict single-string-field JSON schema. The released evaluation data
include generated answers, prompt and span hashes, model metadata,
and checksums identifying the evaluated software and data.

\paragraph{Exact strings and complete quantities.}
The pre-specified primary metric is exact equality of the answer field after normalizing case and whitespace. That score counts \emph{261 liter}, \emph{628 min}, and
\emph{181 V} differently from their canonical plural unit strings.
A separate post-hoc scoring rule accepts a fixed list of standard unit
symbols and singular/plural names \citep{nist811}; it performs no numeric
unit conversion and is applied uniformly to all 12,420 outputs.
This secondary scoring rule was introduced after inspecting model outputs
and is reported separately from the pre-specified exact-match score.
It changes 58/10,800 test labels. Bare numbers and generic ``unit(s)''
remain incomplete; unchanged numbers with a different physical unit
remain incorrect. Tables and coverage curves for both scoring rules are included in the supplement.

Table~\ref{tab:v33_current_dependent_exact} compares the two scoring rules for current-value answers, including these three models.
When B refers to A for its unit, whole-source masking retains each target magnitude
but gives 23, 44, and 55 complete answers out of 60 under the scoring rule that accepts equivalent unit expressions for
Qwen3-4B, Qwen3-8B, and Llama. Relative to full access, there are 37,
15, and five reversals, respectively, with zero corrections. Qwen3-8B has
one full-access old-value error corrected by number-only masking.
Its number-only control retains that error, while whole-source controls
score 60/60 in all three models. Appendix~\ref{app:v23_units_signals} separates missing and wrong units
and defines the detector scores.

\input{context/v18_unified_preservation}

\paragraph{Preservation and uncertainty.}
Historical full-access and both control conditions score 240/240 in every model.
Blocking A or its old-number positions degrades historical access. On questions about the unrelated record, both Qwen models score 240/240
under every operation. Llama scores 8/240 under the whole-source control,
compared with 146/240 under full access; all its errors are UNKNOWN. The note includes a color description
but no target label or unit; its masking cannot be assumed semantically
inert for every query. This loss makes the large mask-minus-control
contrast unreliable as a measure of benefit; full access provides a
necessary additional reference. Appendix~\ref{app:v23_controls}
examines this control behavior and the ambiguous auxiliary attribute.

Intervals use 10,000 seeded base-group bootstrap draws. Information conditions,
query roles, and models stay paired within each group; pooled curves
stratify by the four assigned relation categories. Each source-information
condition has 60 replacement groups. A zero observed reversal rate still has a 4.87\%
one-sided 95\% binomial upper bound. The analogous bound in a 20-group
negative class is 13.91\%; zero-width bootstrap intervals at a floor or
ceiling are not guarantees of zero population error.

\paragraph{Selection signals and their input requirements.}
The method for extracting affirmed values in Appendix~\ref{app:exact_definitions}
extracts A/B's affirmed values; the second signal
uses forced YES/NO sequence scores for whether the value of the same entity and attribute changed.
Both see A/B, and neither sees a reference answer. Their threshold is
predeclared as 0.5; no test-set sign, threshold, or prompt tuning is used.
The additional analyses of candidate scores require reference current/old
answer strings and masked forward passes. The table and curves separate
these access requirements. The policy given true replacement labels knows which values were replaced,
but not which masks improve answers. Signal generation time is recorded separately; the curves
are quality analyses and do not claim a latency saving.

\input{context/v18_unified_selection}
\paragraph{Replacement and non-replacement contributions.}

In the three-model analysis, the detector in each Qwen model selects all 120 strict-replacement current-query
conditions, so their strict-replacement contribution equals that of the
policy given true replacement labels. Table~\ref{tab:v28_selection_decomposition}
separates this contribution from actions on non-replacements, using the
unchanged threshold and scoring rule that accepts equivalent unit expressions.

Whole-source losses remain within correctly identified replacements.
When the detector selects number-only masks, the losses arise from masking
non-replacements: strict replacements contribute zero for Qwen3-4B and one
correction with no reversals for Qwen3-8B. Llama selects no conditions
at this threshold and contributes zero under either operation. These
results concern the tested current queries; they do not establish
preservation for historical or unrelated queries.

\input{context/v25_auxiliary_sensitivity}

Llama-3.2-3B selects no operations at either fixed threshold. Its zero net change
therefore results from leaving every history unmodified. Below, we distinguish parsed negative decisions, invalid outputs, and outputs
whose fields violate the prompt instructions; a valid status does not establish a correct semantic judgment.
All other-attribute negatives remain in the pooled analysis despite their
near-zero full baseline. Relation-specific outcomes below separate this
floor from losses on initially correct answers.

\begin{figure}[!htbp]
\centering
\includegraphics[width=\linewidth]{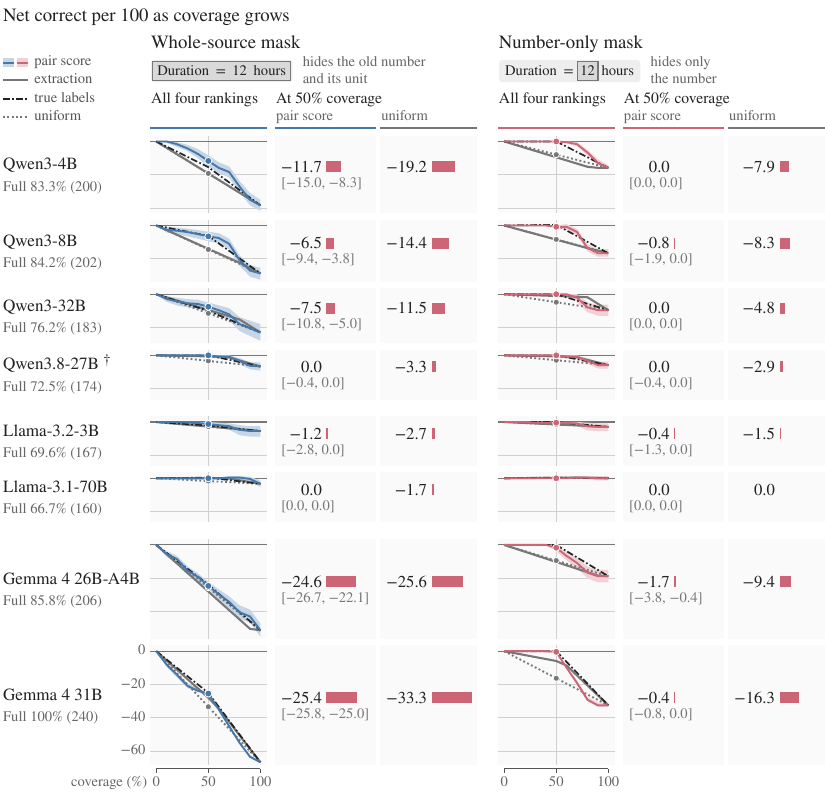}
\caption{Current-query utility over all relations: 240 conditions from 120 groups per model. Rows list full-access percentages (counts), with 240 conditions per model; all plots share one vertical scale. The mask-hue line ranks record-pair scores, with its 95\% group-bootstrap band; solid gray ranks affirmed-value extraction, black dash-dot uses supplied replacement labels, and gray dotted is uniform selection. Exact ties are averaged uniformly. Dots mark 50\% coverage, printed with rose loss bars. Every model uses observed 40-token prefixes and complete-response scoring; these views are not separate timing runs. $\dagger$ Qwen3.8 retains recurrent states.}
\label{fig:v18_unified_utility}
\end{figure}

\subsubsection{Results by Measurement Unit and Detector Scores}
\label{app:v23_units_signals}

For each model, this analysis uses 60 histories in which B replaces the
value but refers to A for its unit. We evaluate current-value questions.
The six units have unequal counts because relation assignment was
randomized before inference: 9 minutes, 14 meters,
10 liters, 10 volts, 11 grams, and 6 hours. Table~\ref{tab:v23_unit_breakdown}
separates exact-string correctness from the
scoring rule that accepts equivalent unit expressions. The latter changes 4 Qwen3-4B and 41 Qwen3-8B labels
in this subset, accepting the previously specified unit symbols and spellings.
It changes no Llama label. Llama returns \texttt{days} in five of the six cases requiring
\texttt{hours}. These are incorrect units, not unit omissions. Per-case raw answers and both labels are
included in the released per-case results.

\begin{table}[!htbp]
\WideTableStyle
\caption{Current answers after whole-source masking when the update refers to the earlier unit. Percentages (counts) use the per-unit $n$ in each row and distinguish complete answers from omitted, generic or wrong units.}
\label{tab:v23_unit_breakdown}

{\setlength{\tabcolsep}{\dimexpr\tabcolsep*14/16\relax}
\begin{tabular}{cccccccc}
\toprule
\TableHead{Model} & \TableHead{Unit} & \TableHead{$n$} & \TableHead{Exact} & \TableHead{Complete} & \TableHead{Missing} & \TableHead{Generic} & \TableHead{Wrong} \\\midrule
Qwen3-4B & minutes & 9 & \AnswerPct{11.1}{1} & \AnswerPct{11.1}{1} & \AnswerPct{88.9}{8} & \AnswerPct{0}{0} & \AnswerPct{0}{0}\\ & meters & 14 & \AnswerPct{42.9}{6} & \AnswerPct{42.9}{6} & \AnswerPct{35.7}{5} & \AnswerPct{21.4}{3} & \AnswerPct{0}{0}\\ & liters & 10 & \AnswerPct{0}{0} & \AnswerPct{0}{0} & \AnswerPct{70}{7} & \AnswerPct{30}{3} & \AnswerPct{0}{0}\\ & volts & 10 & \AnswerPct{70}{7} & \AnswerPct{100}{10} & \AnswerPct{0}{0} & \AnswerPct{0}{0} & \AnswerPct{0}{0}\\ & grams & 11 & \AnswerPct{45.5}{5} & \AnswerPct{54.5}{6} & \AnswerPct{45.5}{5} & \AnswerPct{0}{0} & \AnswerPct{0}{0}\\ & hours & 6 & \AnswerPct{0}{0} & \AnswerPct{0}{0} & \AnswerPct{83.3}{5} & \AnswerPct{16.7}{1} & \AnswerPct{0}{0}\\\midrule
Qwen3-8B & minutes & 9 & \AnswerPct{11.1}{1} & \AnswerPct{66.7}{6} & \AnswerPct{11.1}{1} & \AnswerPct{22.2}{2} & \AnswerPct{0}{0}\\ & meters & 14 & \AnswerPct{0}{0} & \AnswerPct{85.7}{12} & \AnswerPct{14.3}{2} & \AnswerPct{0}{0} & \AnswerPct{0}{0}\\ & liters & 10 & \AnswerPct{0}{0} & \AnswerPct{50}{5} & \AnswerPct{50}{5} & \AnswerPct{0}{0} & \AnswerPct{0}{0}\\ & volts & 10 & \AnswerPct{20}{2} & \AnswerPct{100}{10} & \AnswerPct{0}{0} & \AnswerPct{0}{0} & \AnswerPct{0}{0}\\ & grams & 11 & \AnswerPct{0}{0} & \AnswerPct{100}{11} & \AnswerPct{0}{0} & \AnswerPct{0}{0} & \AnswerPct{0}{0}\\ & hours & 6 & \AnswerPct{0}{0} & \AnswerPct{0}{0} & \AnswerPct{0}{0} & \AnswerPct{100}{6} & \AnswerPct{0}{0}\\\midrule
Llama-3.2-3B & minutes & 9 & \AnswerPct{100}{9} & \AnswerPct{100}{9} & \AnswerPct{0}{0} & \AnswerPct{0}{0} & \AnswerPct{0}{0}\\ & meters & 14 & \AnswerPct{100}{14} & \AnswerPct{100}{14} & \AnswerPct{0}{0} & \AnswerPct{0}{0} & \AnswerPct{0}{0}\\ & liters & 10 & \AnswerPct{100}{10} & \AnswerPct{100}{10} & \AnswerPct{0}{0} & \AnswerPct{0}{0} & \AnswerPct{0}{0}\\ & volts & 10 & \AnswerPct{100}{10} & \AnswerPct{100}{10} & \AnswerPct{0}{0} & \AnswerPct{0}{0} & \AnswerPct{0}{0}\\ & grams & 11 & \AnswerPct{100}{11} & \AnswerPct{100}{11} & \AnswerPct{0}{0} & \AnswerPct{0}{0} & \AnswerPct{0}{0}\\ & hours & 6 & \AnswerPct{16.7}{1} & \AnswerPct{16.7}{1} & \AnswerPct{0}{0} & \AnswerPct{0}{0} & \AnswerPct{83.3}{5}\\
\bottomrule
\end{tabular}}
\par\smallskip{\footnotesize\TableNoteAlign Missing, generic, and wrong refer to units;
complete requires both the correct magnitude and the specified unit.\TableNoteEnd}
\end{table}

The replacement detector uses the native chat template with the system message
\emph{Return only the requested answer. Do not include reasoning or markdown.}
Its instruction asks whether A and B affirm different values of the same
entity and single-valued attribute, explicitly excludes confirmations and
changes to a different entity or attribute, and asks the model to resolve references to A's unit.
The formatted prompt and all candidate token IDs are preserved. Reconstructing
all 828 prompts with tokenizer files matching the inference metadata
reproduces the hashes of the prompts used for inference.

The exact user template is the following (placeholders contain the
unmodified source records):
\begin{quote}\small\ttfamily\raggedright
Do records A and B affirm DIFFERENT values for the SAME entity and the SAME
single-valued attribute at their respective reference times? A later
confirmation of the same quantity is not a replacement; updates to a
different entity or attribute are not replacements of A. Resolve any
explicit reference to A's unit. Answer only YES or NO.

Record A:\\
\{old\}

Record B:\\
\{later\}
\end{quote}
The literal template has two newlines between the instruction and
\texttt{Record A:}, one newline before each record value, two newlines
between A and \texttt{Record B:}, and two trailing newlines. Native chat
formatting supplies the assistant-generation prefix; the date argument,
where supported, is \texttt{16 Sep 2026}. The value extraction prompt is the
template in Appendix~\ref{app:exact_definitions}.

The scorer tokenizes the literal strings \texttt{YES} and \texttt{NO}
separately without special tokens, then appends their token IDs to the
prompt IDs. It adds no whitespace or end-of-sequence token and does not
retokenize the concatenated text. Each candidate is one token: 14004/8996
for Qwen and 14331/9173 for Llama. It sums float32 next-token log
probabilities and forms
\begin{equation}
s=\frac{1}{1+\exp\!\left(\operatorname{clip}
 (\log p(\texttt{NO})-\log p(\texttt{YES}),-80,80)\right)}.
\label{eq:detector_pairwise_score}
\end{equation}
This is normalization over the two candidate strings, rather than the
marginal probability of \texttt{YES}. No length normalization is applied;
with these one-token candidates, the sum equals the per-token mean.
The original action threshold for Equation~\ref{eq:detector_pairwise_score}
is $s\geq0.5$. Llama's development
scores range from 0.148 to 0.407 and test scores from 0.119 to 0.321,
so its zero-action policy follows directly from these scores, without
an invalid-output fallback or a fitted threshold. Appendix~\ref{app:v41_calibration}
separately evaluates thresholds calibrated on the existing development groups.

The parser for extracted values (Appendix~\ref{app:exact_definitions})
distinguishes parsed negative decisions, malformed extraction outputs,
and unresolved values. Its \texttt{unrelated}
status means that the extracted \texttt{same\_slot} field is false;
\texttt{confirmation} means that the normalized value strings are equal.
Neither status establishes that the semantic decision is correct.
All 240 Llama test outputs have valid JSON and
\texttt{same\_slot=false}; none receives the parser status
\texttt{invalid\_schema} or \texttt{unresolved}. However, all 240 retain a
non-null \texttt{later\_current}, contrary to the prompt instruction for
pairs about different entities or attributes. The parser returns \texttt{unrelated} before
checking that field. Four of Qwen3-8B's seven negative
\texttt{same\_slot=false} outputs have the same cross-field violation.
The parser therefore records negative decisions even though these outputs violate the extraction instructions.
Qwen3-8B's 80 test false positives comprise seven confirmations whose two
extracted values differ only by a missing unit, and 73 cases about a different entity or attribute.
We count normalization and entity--attribute errors separately; this error
analysis does not modify the predictions used to evaluate mask selection.

\begin{figure}[!htbp]
\centering
\includegraphics[width=\linewidth]{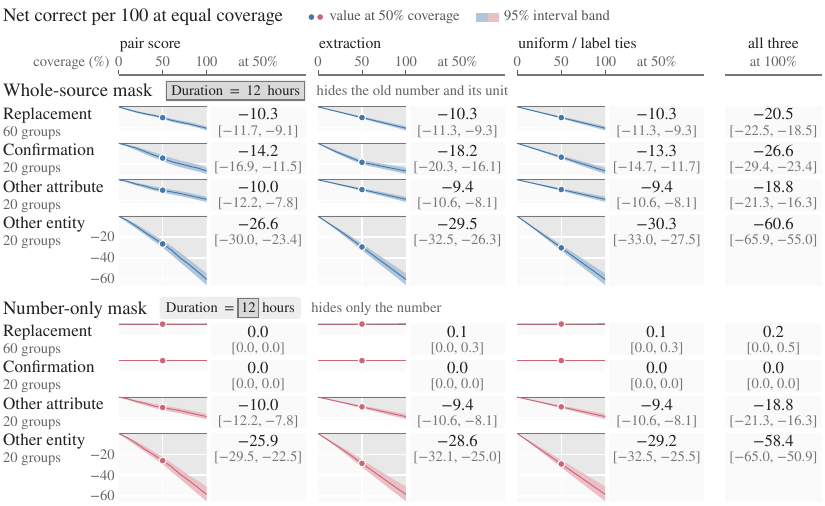}
\caption{Utility at equal coverage within each model, averaged equally over eight models. Rows fix relation and mask scope; columns fix the signal; tiles share one scale. The original 95\% group-bootstrap bands retain all models and supports in each draw. Label ties equal uniform selection within each relation. Full counts are 958/960, 320/320, 60/320 and 194/320; independent groups are 60, 20, 20 and 20. Models use complete-response scoring on observed 40-token prefixes; scores are never ranked across models.}
\label{fig:v23_stratified_utility}
\end{figure}

Relation-stratified results in Figure~\ref{fig:v23_stratified_utility}
retain the other-attribute cases, on which full-access accuracy is low. Group bootstrap draws keep
both information conditions and all three models together. Fixed-threshold
counts report actions separately from outcome changes, including actions
whose full-access answers were already incorrect.

\paragraph{Within-model ranking and uncertainty.}
At each coverage level, we rank candidates separately within each of the
eight generators, pool the two information conditions within each model,
and average the eight net-correct rates equally. Scores are never ranked
across models; an increasing transformation within one model leaves the
estimate unchanged.
Each of 10,000 bootstrap draws resamples base groups once and uses the
same multiplicities for every model and both information conditions.
Within each draw, coverage cutoffs and exact-tie fractions are recomputed
before averaging. The shaded intervals use quantiles of these averages.
A ground-truth replacement score is constant within each relation stratum,
so its ranked curve equals uniform selection. This differs from its
fixed-threshold policy, which acts only on strict replacements. Complete correctness uses the response rule in Section~\ref{sec:method}
under a common 40-token limit. The numerical supplement retains all scoring
rules and each generator's recorded budget.

\begin{table}[!htbp]
\WideTableStyle
\caption{Entries are percentages (counts) of the relation population; Net is pp. C/R is corrections above reversals, each with the same denominator. Pair-score selection at the fixed threshold $s\geq0.5$, pooled across all eight models at the common 40-token response limit. Complete-response correctness retains every candidate.}
\label{tab:v23_relation_selection}

{\setlength{\tabcolsep}{\dimexpr\tabcolsep*12/14\relax}
\begin{tabular}{ccccccc}
\toprule
\TableHead{Relation} & \TableHead{Scope} & \FullCell{\FullHeader{Full/$N$}} & \TableHead{Selected} & \TableHead{C/R} & \TableHead{Net} & \TableHead{False} \\\midrule
\TableLabel{Replacement} & \WholeCell{Whole} & \FullCell{\AnswerPct{99.8}{958/960}} & \AnswerPct{87.5}{840} & \shortstack[c]{\AnswerPctInline{0.1}{1}\\\AnswerPctInline{20.1}{193}} & \AnswerPP{-20}{-192} & \AnswerPct{0}{0}\\ & \NumberCell{Number-only} & \FullCell{\AnswerPct{99.8}{958/960}} & \AnswerPct{87.5}{840} & \shortstack[c]{\AnswerPctInline{0.2}{2}\\\AnswerPctInline{0}{0}} & \AnswerPP{+0.2}{+2} & \AnswerPct{0}{0}\\\TableLabel{Confirmation} & \WholeCell{Whole} & \FullCell{\AnswerPct{100}{320/320}} & \AnswerPct{1.3}{4} & \shortstack[c]{\AnswerPctInline{0}{0}\\\AnswerPctInline{0.9}{3}} & \AnswerPP{-0.9}{-3} & \AnswerPct{1.3}{4}\\ & \NumberCell{Number-only} & \FullCell{\AnswerPct{100}{320/320}} & \AnswerPct{1.3}{4} & \shortstack[c]{\AnswerPctInline{0}{0}\\\AnswerPctInline{0}{0}} & \AnswerPP{0}{+0} & \AnswerPct{1.3}{4}\\\TableLabel{Other attribute} & \WholeCell{Whole} & \FullCell{\AnswerPct{18.8}{60/320}} & \AnswerPct{80}{256} & \shortstack[c]{\AnswerPctInline{0}{0}\\\AnswerPctInline{11.9}{38}} & \AnswerPP{-11.9}{-38} & \AnswerPct{80}{256}\\ & \NumberCell{Number-only} & \FullCell{\AnswerPct{18.8}{60/320}} & \AnswerPct{80}{256} & \shortstack[c]{\AnswerPctInline{0}{0}\\\AnswerPctInline{11.9}{38}} & \AnswerPP{-11.9}{-38} & \AnswerPct{80}{256}\\\TableLabel{Other entity} & \WholeCell{Whole} & \FullCell{\AnswerPct{60.6}{194/320}} & \AnswerPct{50.3}{161} & \shortstack[c]{\AnswerPctInline{0}{0}\\\AnswerPctInline{36.6}{117}} & \AnswerPP{-36.6}{-117} & \AnswerPct{50.3}{161}\\ & \NumberCell{Number-only} & \FullCell{\AnswerPct{60.6}{194/320}} & \AnswerPct{50.3}{161} & \shortstack[c]{\AnswerPctInline{0}{0}\\\AnswerPctInline{35.3}{113}} & \AnswerPP{-35.3}{-113} & \AnswerPct{50.3}{161}\\
\bottomrule
\end{tabular}}
\par\smallskip{\footnotesize\TableNoteAlign C/R: corrections/reversals versus full visibility. False: actions on annotated non-replacements. Counts repeat eight models and two support conditions: 960 outputs from 60 replacement groups, and 320 outputs from 20 groups in each other class. They are not independent sample counts. No prediction or threshold was changed.\TableNoteEnd}
\end{table}

\subsubsection{Attribute ambiguity and effects of the note control}
\label{app:v23_controls}

Analysis of all 12,420 generated answers, including 10,800 test outputs,
identifies two limitations of the experiment with six measurement units: ambiguity in the
different-attribute condition and a loss of unrelated-query accuracy under
the note control. This analysis uses the same prompts, references, and
correctness criteria as the aggregate results. The released per-case data
include A, B, the inspection record, neutral note, question, reference,
all five generated answers, and masked text. Aggregating these data recovers
the reported counts under exact matching and equivalent unit expressions for all 360 combinations of model,
relation, source information, query, and mask.

\paragraph{The different-attribute baseline follows the auxiliary value.}
For the 20 different-attribute groups, full-access current accuracy is
0/20 for every model when B states the complete quantity. When B refers to A for its unit, it is
0/20, 3/20, and 1/20 for Qwen3-4B, Qwen3-8B, and Llama-3.2-3B. All 116
incorrect answers among these 120 model-by-information-condition outputs reproduce B's
auxiliary quantity exactly; the four remaining answers match the original
reference. None is UNKNOWN or malformed JSON. These are repeated outputs
from 20 base groups, not 120 independent questions.
For the same different-attribute groups, every model answers all 40
full-access historical queries correctly, including both source-information
conditions.

The intended negative changes attribute $X$ to a different field named
``auxiliary $X$'' in B, while the question still asks about $X$. The system
instruction says that a later entry changes only its own entity and
attribute, and the original reference therefore retains A's value. The
construction supplies only this modifier to distinguish the fields; it
does not explicitly define them as independent. Moreover, persistence from
T1 to T2 for an untouched field follows the system instruction rather than
an explicit persistence sentence. The outputs establish substitution of
the auxiliary quantity under this wording. They do not isolate lexical
attribute conflation from an alternative reading of the specification.
Consequently, this negative class is a difficult, narrow condition with a
poor full baseline; its results cannot establish a general operation cost
on otherwise well-answered different-attribute questions.

\paragraph{The neutral-note control increases errors on unrelated questions.}
Llama's whole-source control changes 138 correct answers to unrelated questions into
UNKNOWN, with no corrections (Table~\ref{tab:v23_control_counts}). Every
incorrect answer in all five Llama unrelated-query conditions is UNKNOWN. All
10,800 test outputs satisfy the strict single-field JSON schema; none
reaches the 40-token generation cap. Output lengths are 6--26, 6--13, and
6--15 tokens for Qwen3-4B, Qwen3-8B, and Llama-3.2-3B, respectively.
Thus malformed output and cap truncation do not explain this contrast.

\begin{table}[!htbp]
\TableStyle
\begin{minipage}{0.72\linewidth}\centering

\caption{Outcome entries give percentages, with the original count/denominator beneath. Correct Llama answers to unrelated-label questions. Each information condition has
120 questions; the two conditions share 120 base groups. All errors are
UNKNOWN.}
\label{tab:v23_control_counts}
{\setlength{\tabcolsep}{\dimexpr\tabcolsep*6/8\relax}
\begin{tabular}{cccc}
\toprule
\TableHead{Mask condition} & \TableHead{Unit in B} & \TableHead{Unit from A} & \TableHead{Both} \\\midrule
\FullCell{Full} & \AnswerPct{68.3}{82/120} & \AnswerPct{53.3}{64/120} & \AnswerPct{60.8}{146/240}\\\WholeCell{Whole-source mask} & \AnswerPct{98.3}{118/120} & \AnswerPct{67.5}{81/120} & \AnswerPct{82.9}{199/240}\\\FullCell{Whole-source control} & \AnswerPct{6.7}{8/120} & \AnswerPct{0}{0/120} & \AnswerPct{3.3}{8/240}\\\NumberCell{Number-only mask} & \AnswerPct{70}{84/120} & \AnswerPct{55.8}{67/120} & \AnswerPct{62.9}{151/240}\\\FullCell{Number-only control} & \AnswerPct{67.5}{81/120} & \AnswerPct{54.2}{65/120} & \AnswerPct{60.8}{146/240}\\
\bottomrule
\end{tabular}}
\end{minipage}
\end{table}

Token positions and character offsets were validated against all 2,484
native prompt texts, whose SHA256 hashes agree with those used for inference.
The positions align with the texts in all 4,140 history-to-condition mappings.
Whole and value controls
have the same token counts as their respective masks and touch only the
neutral-note body; neither control overlaps the inspection record, B, or
the question. This alignment analysis uses the tokenization from the generation
experiment; it does not independently test attention computations.
The control matches the number of masked tokens but changes model answers. Its large unrelated-query loss limits source-versus-control
interpretation; full access remains an essential reference.

\paragraph{Consistency across visibility implementations.}
All 828 model--history combinations retain unchanged prefix tensors,
equal treatment/control lengths, and a shared query prefix.
Cached-versus-uncached numerical comparisons cover nine development examples:
the first current/full query in each model/shard. All nine agree in first
top-1 token and generated text, but maximum absolute logit differences
range from 0.19 to 0.89. These development comparisons do not cover
unrelated test queries under the whole-source control.

To test whether the control-induced abstention depends on the visibility
implementation, we compare 12 distinct Llama test groups using the same checkpoint,
BF16 eager runtime, prompts, and token positions. Cases were selected before
this comparison: four full-correct/control-UNKNOWN cases, four UNKNOWN/UNKNOWN cases,
and four correct/correct cases, alternating information conditions where
available. All five conditions were compared using the cached generation procedure,
an instrumented 2D mask, an independently constructed additive 4D mask,
and one-shot recomputation with that 4D mask. In the latter, history rows
retain full causal visibility; only subsequent query and answer rows are
blocked from the selected source keys. We also reversed condition order and
ran the 12 full-access conditions without a reused prefix.

All 312 generated outputs agree with those from the generation experiment.
The cached 2D and
4D paths have identical first-output logits in all 60 condition comparisons;
reversing condition order also gives identical logits and outputs. The
instrumented paths measure zero attention probability on masked keys
across every layer for query rows in all 48 conditions with a nonempty mask.
One-shot 4D computation gives the same 60 outputs with maximum absolute
first-output logit difference 0.46875; the 12 full-access runs without a reused prefix also
agree in output, with maximum difference 0.3125. All 12 stored prefixes
remain unchanged. Thus the neutral-control abstention reproduces under
equivalent visibility implementations on these selected cases.
The following span and prompt interventions test the conditions that elicit
abstention (Appendix~\ref{app:v24_control_factors}).

\paragraph{Exact source and response examples.}
The following examples are the first cases in three predefined categories:
auxiliary-value substitution, full-correct/control-UNKNOWN, and
full-UNKNOWN/source-mask-correct. They illustrate the outcomes above;
all cases contribute to the aggregate results. Every case includes the common neutral
note, consisting of six repetitions of ``A square seal is printed beside
a blank signature box. The cover is gray and the folder is closed.''
The complete native prompts and raw JSON strings, including whitespace,
are included in the released evaluation data. The answer fields below are quoted exactly.
The common system instruction is: ``Use the ordered authoritative records,
each valid at its stated reference time. A later entry changes only the
entity and attribute it concerns. For a historical question, report the
value at the requested reference time. Return exactly one JSON string
field: \texttt{\{"answer": "complete queried value"\}}. Quantities must
include their measurement unit. Do not convert units. If the requested
value cannot be determined, return \texttt{\{"answer": "UNKNOWN"\}}.''

\emph{Qwen3-4B, test-011, B states the complete quantity, current query.}
A: ``Reference time T1. Certified specification for Assembly-TEST-011:
inspection interval = 233 hours. This is its single approved value at T1.''
B: ``Reference time T2, after T1. The certified auxiliary inspection
interval of Assembly-TEST-011 is 807 hours. This entry is effective at T2.''
Inspection record: ``The inspection label on Envelope-TEST-011 is ochre.''
Question: ``At T2, what is the current inspection interval of
Assembly-TEST-011? Return the complete quantity including its unit.''
Reference: \texttt{233 hours}. Full, whole mask, whole control, value
mask, and value control all return the answer field \texttt{807 hours}.

\emph{Llama-3.2-3B, test-000, B states the complete quantity, unrelated query.}
A: ``Reference time T1. Certified specification for Assembly-TEST-000:
cycle duration = 159 minutes. This is its single approved value at T1.''
B: ``Reference time T2, after T1. The certified cycle duration of
Assembly-TEST-000 is 628 minutes. This entry is effective at T2.''
Inspection record: ``The inspection label on Envelope-TEST-000 is amber.''
Question: ``What is the inspection label on Envelope-TEST-000? Return
only the label in the answer field.'' Reference: \texttt{amber}. Full,
whole mask, value mask, and value control return \texttt{amber}; whole control
returns \texttt{UNKNOWN}. Its 31 masked tokens, at zero-based positions
172--202, read exactly: ``A square seal is printed beside a blank signature
box. The cover is gray and the folder is closed. A square seal is printed
beside a blank signature box''. The inspection record is outside this span.

\emph{Llama-3.2-3B, test-000, B refers to A for its unit, unrelated query.}
A, the inspection record, question, and reference are identical to the
previous case. B is exactly: ``Reference time T2, after T1. The certified
cycle duration of Assembly-TEST-000 is 628. The measurement unit is the same
as in record A at T1. This entry is effective at T2.'' Full and whole
control return \texttt{UNKNOWN}; whole mask, value mask, and value control
return \texttt{amber}. The control mask is identical to the preceding span.

\FloatBarrier

\subsubsection{Effects of control location and prompt wording on abstention}
\label{app:v24_control_factors}

We vary the input and masked occurrence to determine which conditions induce
abstention. The first experiment crosses all 120 groups and both later-record information conditions with
14 configurations, giving 3,360 Llama outputs. These are interventions on
previously observed cases, not an independent test set. Generation retains
the original checkpoint, native template, BF16 eager runtime, greedy EOS
stopping, and 40-token limit. Configuration order is shuffled within each
case. All outputs stop at EOS and satisfy the answer schema; each is either
the stated inspection label or UNKNOWN. The 720 outputs under full access,
the original control, and the whole-source mask agree with the corresponding
outputs in the experiment with six measurement units.
Intervals resample 120 base groups 10,000 times, retaining both source-information conditions and
all configurations; they are exploratory, unadjusted intervals.

\begin{figure}[!t]
\centering
\includegraphics[width=\linewidth]{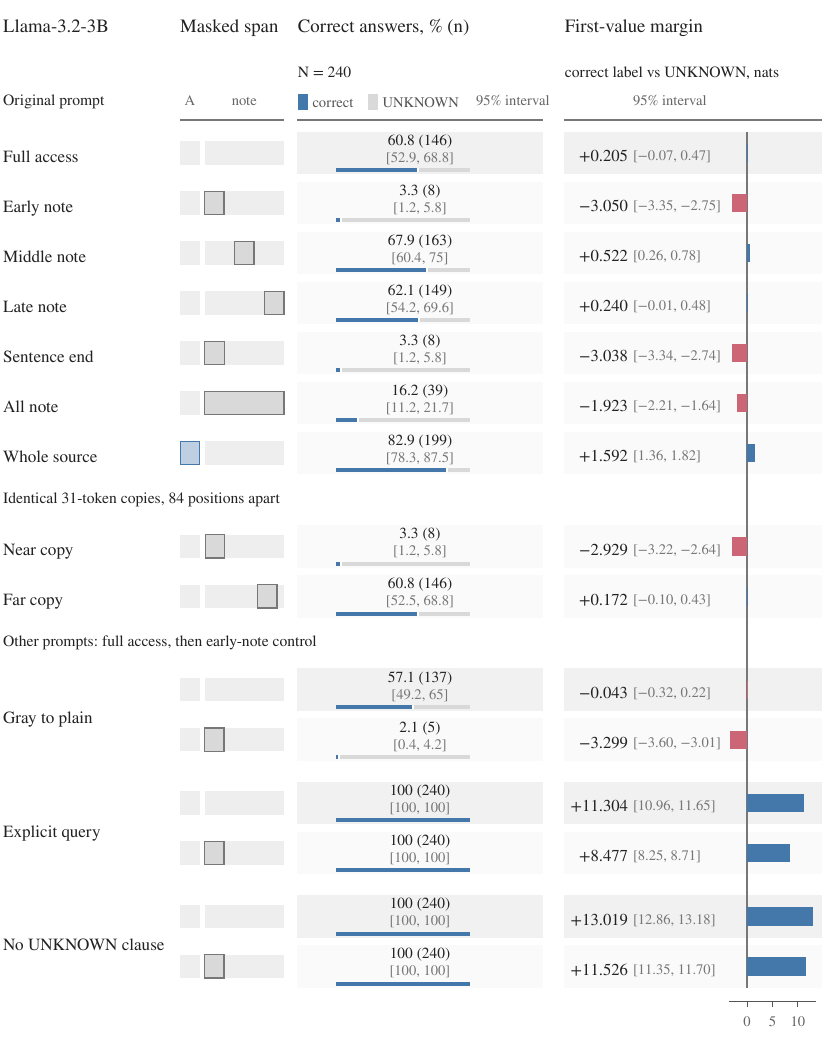}
\caption{Llama-3.2-3B on 240 unrelated-label conditions from 120 groups, one configuration per row. Boxes mark the masked span, to token scale, in the old record A or the 127-token neutral note; masks cover 31 tokens except sentence end (32) and all note (127). Top: original-prompt span controls. Middle: the separate identical-token near/far comparison. Bottom: full access (no box) versus the early-note control under the other three of four prompt profiles. Bars show correct-answer percentages (counts) over 240 (blue; the rest are UNKNOWN); mini bars give the first-value margin against UNKNOWN. Gray brackets are paired 95\% group-bootstrap intervals. A zero teacher-forced margin is not an exact greedy decision threshold; all 14 configurations remain in Table~\ref{tab:v24_control_factors}.}
\label{fig:v24_control_factors}
\end{figure}

\paragraph{Which occurrence is blocked matters.}
Masking the first 31 note tokens gives 8/240 correct answers, versus
146/240 with full access. Moving the 31-token window to the middle or
end gives 163 and 149 correct answers. These windows differ in wording and
phrase boundaries, so a paired comparison fixes their content exactly.
Before its 480 generations, we select the earliest 31-token note subsequence
with a later identical copy and its last copy. The two windows, offsets
$[1,32)$ and $[85,116)$ in the note, are 84 positions apart and have identical token IDs, decoded text,
size, and phrase boundaries in every case; all other cache rows and positions
remain fixed. Blocking the earlier copy gives 8/240 correct answers;
blocking the later copy gives 146/240, with 138 corrections and no reversals:
$+57.5$ percentage points, with a 95\% group-bootstrap interval
of $[49.58,65.42]$. The total equals the full-access total, but the individual answers differ: three corrections and three reversals cancel.
This comparison isolates the choice of occurrence within the resident
history. Its positions and already computed representations differ, so it
does not attribute the effect to positional embeddings alone.

\paragraph{Effects of punctuation and the color word.}
Extending the original window by its following period (31 to 32 tokens)
still gives 8/240; one gain and one loss cancel. Replacing each of six
\emph{gray} tokens by \emph{plain} preserves token counts and positions but
requires a new prefill. Full/control scores become 137/240 and 5/240;
the control-effect interaction is $+2.5$ points $[-0.42,5.42]$.
The color word is therefore not necessary for the observed loss. Blocking
all 127 note tokens gives 39/240, while deleting the note and recomputing
gives 65/240. These unequal-size and changed-history contrasts show why
neither note removal nor greater mask size consistently improves accuracy.

\paragraph{Question and abstention instructions change answer selection.}
The explicit question reads:
\begin{quote}\small
Read the explicit sentence beginning ``The inspection label on
\{envelope\} is''. Report the label stated in that sentence, not the
specifications or the neutral note. Return only that label in the JSON
answer field.
\end{quote}
It supplies the source description and entity, but not the answer value,
and uses the unchanged history prefix. Full and original control both
score 240/240. A separate condition keeps the original question and removes
only the system clause ``If the requested value cannot be determined,
return \{\texttt{"answer": "UNKNOWN"}\}.'' After prefilling this new prompt,
full and control again score 240/240. The edit also shortens the system
prefix and shifts later positions, so it does not isolate instruction
semantics. Thus the label can be recovered under the original blocked-source
operation; these prompt conditions change whether the operation manifests
as abstention.

\input{context/v24_control_factor_table}

Figure~\ref{fig:v24_control_factors} summarizes the occurrence,
query, and score comparisons.

\paragraph{The answer-versus-UNKNOWN competition shifts.}
All 3,360 outputs share the canonical JSON prefix up to the value token.
We score the first token of the correct label against the first token of
UNKNOWN under that prefix. The mean log-probability difference changes
from $+0.205$ to $-3.050$ nats under the original control. With the explicit
question, it changes from $+11.304$ to $+8.477$; after removing the UNKNOWN
clause, from $+13.019$ to $+11.526$. Therefore, correct answers recover while
a negative score effect persists. The prompts increase the mean preference for the correct label enough that
answers recover despite the negative change under masking
(Figure~\ref{fig:v24_control_factors}). This observation does not establish a unique mechanism.
This reference-dependent score is a diagnostic, not an online signal or
an exact predictor of greedy outputs: its sign disagrees with 39 outputs,
including 23 zero-score ties; the 16 nonzero disagreements lie between
$-0.125$ and $+0.25$ nats. Across all outputs there are 37 zero-score ties.
The teacher-forced and incremental decoding computations need not be
numerically identical in BF16. Mean inspection-record attention also fails
to track accuracy consistently, so we do not identify attention magnitude
with the mechanism.

These interventions show that abstention depends on which occurrence of the
repeated note is masked and on the query and abstention instructions.
Equal token counts and absent target words do not make the note control
behaviorally neutral. The result concerns this Llama model and constructed
dataset; it establishes causal input and access conditions, not a unique
head-level circuit or a general prohibition on source control.

\FloatBarrier

\subsection{Quantity Questions Across Eight Models}
\label{app:v33_models}

Tables~\ref{tab:v38_probability_decomposition} and
\ref{tab:v33_current_dependent_content_same_unit} retain the exact
main-figure values, matched controls and probability intervals.
\suppressfloats[t]
\input{context/v38_probability_decomposition}
\input{context/v33_current_dependent_content_same_unit}

We evaluate eight generators on the task of answering quantity questions: Qwen3-4B, Qwen3-8B,
Qwen3-32B, Qwen3.8-27B, Llama-3.2-3B, Llama-3.1-70B-Instruct,
Gemma 4 31B IT, and Gemma 4 26B-A4B IT~\citep{gemma4modelcard2026}.
Qwen3.8-27B combines recurrent and full-attention layers; the two Gemma checkpoints
represent a 2026 model family with dense and mixture-of-experts generators.
Qwen3-32B and Llama-3.1-70B permit comparisons across sizes within the
Qwen and Llama families.
Family, architecture, training, and size vary together. These comparisons
do not isolate a causal effect of parameter count.

\begin{table}[!htbp]
\TableStyle

\caption{Eight generators evaluated on the task of answering quantity questions. Parameter counts refer
to the loaded text decoder, excluding unused visual and multi-token
prediction modules. MoE reports total/active parameters; the active count
is the model-card approximation.}
\label{tab:v33_model_inventory}
{\setlength{\tabcolsep}{\dimexpr\tabcolsep*4/6\relax}
\begin{tabular}{>{\centering\arraybackslash}m{.25\linewidth}C{.20\linewidth}>{\centering\arraybackslash}m{.43\linewidth}}
\toprule
\TableHead{Generator} & \TableHead{Text parameters (B)} & \TableHead{Native attention} \\
\midrule
Qwen3-4B & 4.02 & Full attention \\
Qwen3-8B & 8.19 & Full attention \\
Qwen3-32B & 32.76 & Full attention \\
Qwen3.8-27B & 26.90 & 16 full-attention, 48 DeltaNet layers \\
\addlinespace[2pt]
Llama-3.2-3B & 3.21 & Full attention \\
Llama-3.1-70B & 70.55 & Full attention \\
\addlinespace[2pt]
Gemma 4 26B-A4B & 25.23/$\sim$3.8 & 5 global, 25 sliding-window layers \\
Gemma 4 31B & 30.70 & 10 global, 50 sliding-window layers \\
\bottomrule
\end{tabular}}
\end{table}

All eight generators use the same 120 test groups and 18 development
groups, with two source-information conditions, three query types, and
five access conditions. This gives 3,600 test outputs and 540 development
outputs per generator. The independent test sample remains 120 groups,
including 60 strict replacements, rather than the number of generated
outputs. Each group is shared across generators, information conditions,
queries, and masks. Confidence intervals resample base groups; analyses
mixing relations resample within relation strata. We use 10,000 bootstrap
draws and retain the pairing across models and conditions. A zero observed
failure count among 60 independent replacement groups has a one-sided
95\% binomial upper bound of 4.87\%; degenerate bootstrap intervals
therefore do not establish that failures are impossible.

\paragraph{Native attention and retained states.}
All generators use BF16 weights, eager attention, and each checkpoint's
native chat template with thinking disabled. They share the prompt text and
fixed template date specified in Appendix~\ref{app:v18_unified}.
Gemma uses its text decoder; visual
and audio inputs are absent. Its local and global attention layers retain
their native causal and sliding-window restrictions. Source masks remove
additional attention edges rather than replacing those restrictions.
The original prefilled history and all states formed after A remain
available in every branch. The evaluated Transformers 5.12.1 runtime
retains full token K/V tensors for these Gemma checkpoints even in layers
with sliding-window attention; masking therefore releases no storage.
Llama-3.1-70B weights are distributed across two H200 GPUs per worker;
the other generators use one H200 per worker. Qwen3.8 uses the
native PyTorch DeltaNet implementation in this runtime. Their runtime and
memory measurements have different device allocations and are not used
as a controlled comparison of masking speed.

For Qwen3-32B, Qwen3.8-27B, Llama-3.1-70B, and both Gemma models,
development comparisons before test inference verify native template prefix
identity, source-token alignment, equal mask sizes, unchanged stored
history, and zero attention to masked positions in each affected attention layer. Repeating
an empty-mask cached forward pass gives identical logits. Cached and
full-prefill greedy outputs agree on the checked development examples,
while BF16 logits can differ. Gemma checks additionally cover histories
longer than its 1,024-token local window. These checks concern execution,
not whether a generated answer is correct.

\paragraph{Answer length and output format.}
Qwen3.8-27B, Llama-3.1-70B, and both Gemma models allow 256 answer tokens.
These models and Qwen3-32B allow 128 extraction tokens. Decoding is greedy
with native EOS stopping. Development examples showed that
Gemma often wraps its JSON answer in a Markdown code block. Some masked
answers begin with a complete answer and continue without giving another;
those continuations can reach even the 256-token limit. We retain the complete generated text
and distinguish token-limit termination from a completed response.

Qwen3-4B, Qwen3-8B, Llama-3.2-3B, and Qwen3-32B use a 40-token answer limit.
Every output ended before that limit, with
maximum lengths of 26, 13, 18, and 17 tokens, respectively. Raising their
limit would not change the observed deterministic continuations after
EOS. Their answer-quality comparisons use these outputs, and their timing
measurements correspond to the 40-token setting. The additional answer-scoring
rule that permits an outer Markdown code fence and the 256-token limit were selected using Gemma development outputs,
before examining Qwen3-32B test answer quality and before evaluating the
Gemma and Llama-3.1-70B test sets. This scoring rule is a secondary analysis
of the Qwen3-4B, Qwen3-8B, and Llama-3.2-3B outputs.

We report three scoring rules on the same generated texts. The first two rules use the answer-field parser in
Appendix~\ref{app:v18_unified}: one requires normalized exact matching, and
the other additionally accepts equivalent unit expressions. The third rule
also permits an outer Markdown code fence around a complete JSON response. With or without that fence, the entire response must contain exactly
one string-valued \texttt{answer} field. Duplicate keys, additional fields,
multiple objects or fences, and text outside the accepted response are
rejected. Reaching the token limit also counts as an incomplete response,
even if its prefix contains a correct JSON answer. This rule therefore
does not extract a favorable fragment from a continuation that runs to the
token limit.
Quantities are evaluated with the same list of accepted unit forms and
without unit conversion; unrelated answers use exact label matching.

The third rule permits that single wrapper and still requires a complete response. A lower score can reflect quantity
errors, abstention, an unsupported output format, or a response that does
not terminate within the budget; it is not a direct count of factual errors. We retain format failures and token-limit termination separately
from missing units, wrong units, wrong magnitudes, and UNKNOWN answers.
The other two scores are reported alongside it.
Candidate log probabilities continue to score the canonical JSON answer
strings specified in Appendix~\ref{app:v18_unified}, including for generators that
prefer Markdown fences. These are supplied-candidate measurements and
need not predict unrestricted output format or completion. Candidate
scoring for other models, formats and budgets appears in
Appendices~\ref{app:v35_controlled_scoring}, \ref{app:v35_native_scoring},
\ref{app:v35_oaks_scoring}, \ref{app:v35_r570}
and~\ref{app:v35_generation_limits}.

For both Gemma models on development examples, increasing the answer budget from
40 to 256 leaves all 30 paired first-40-token sequences unchanged per
model. Budget sensitivity uses prefixes of the recorded greedy token
sequences; such prefixes do not inherit the longer run's latency or
memory measurements. The reported error categories include every response
that reaches its token limit.

\paragraph{Hybrid Qwen3.8 states.}
Qwen3.8-27B~\citep{qwen38modelcard2026} has 48 DeltaNet layers and
16 full-attention layers. Its source-visibility masks apply to the
16 layers with token-addressable K/V; each branch retains identical
prefilled convolution and recurrent states in the remaining layers.
Those states evolve normally during the independent continuation. Native
cached execution does not apply masks over source positions to the new
DeltaNet query inputs. We verify unchanged prefixed K/V, convolution,
recurrent tensors, and cache flags, and zero attention to the supplied
positions in all 16 full-attention layers. This restricts subsequent
token-addressable reads; it does not remove information already stored in
recurrent memory. Figures and tables mark this architecture explicitly.

\begin{figure}[!htbp]
\centering
\includegraphics[width=\linewidth]{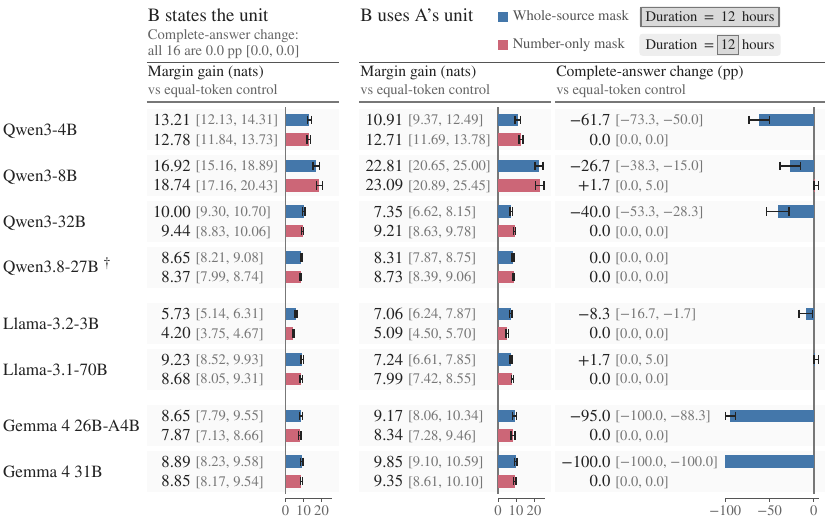}
\caption{Paired candidate and answer effects on the same 60 replacements. Upper lines are the whole-source mask, lower lines the number-only mask; chips box what each hides. When B states the unit, all 16 answer effects and their intervals are exactly zero. When B refers to A's unit, complete answers drop only under the whole-source mask. Each mask uses its own equal-token control. Brackets and whiskers are original paired 95\% group-bootstrap intervals. Scoring and model-specific budgets follow the protocol above; $\dagger$ Qwen3.8 retains its prefilled recurrent states under both masks.}
\label{fig:v33_full_scores}
\end{figure}

\begin{figure}[!htbp]
\centering
\includegraphics[width=\linewidth]{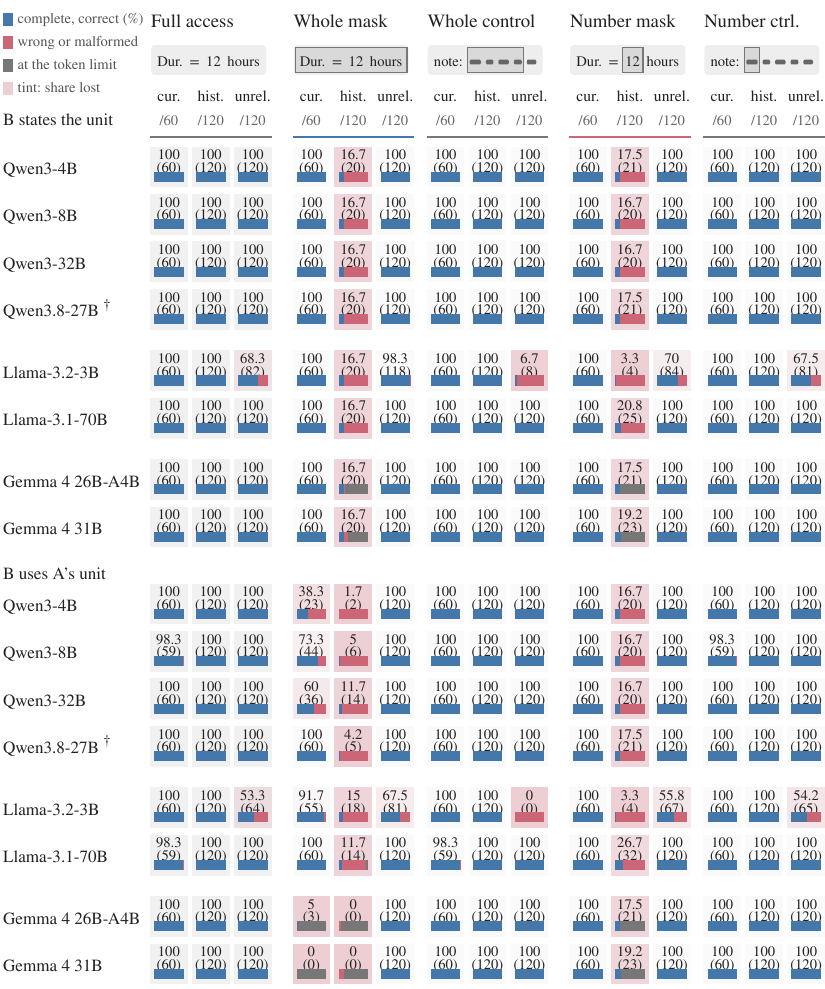}
\caption{Complete-answer percentages (counts) for all five access conditions. Columns group full access, whole-source and number-only masks, and their equal-token controls. Control masks select the start of a neutral note; chips mark the masked spans. Within a block, columns are current, historical and unrelated questions; upper rows have B state the unit, lower rows have B refer to A for it. Current cells contain 60 replacement groups; other cells contain all 120 groups per B condition. Each tile prints the correct percentage and count, and stacks all outcomes: correct (blue), wrong or malformed (rose), or at the token limit (gray). Rose intensity shows the failure fraction. Group-bootstrap intervals remain in the numerical data. $\dagger$ Qwen3.8 retains recurrent states.}
\label{fig:v33_preservation}
\end{figure}

\begin{figure}[!htbp]
\centering
\includegraphics[width=\linewidth]{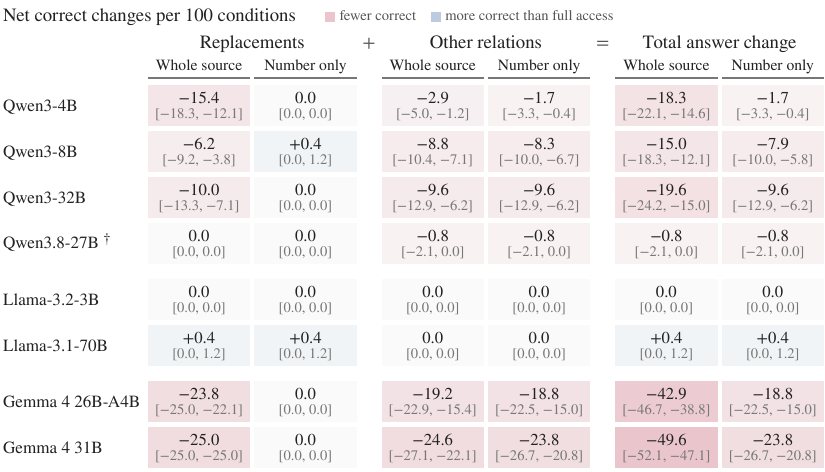}
\caption{Contributions to detector-selected current-answer changes, compared with full access. Cells give net changes and paired 95\% bootstrap intervals (relation strata kept separate); the tint shows sign and size on one scale, rose for fewer correct answers, blue for more. Each block's denominator is 240 conditions from 120 groups, so replacements and other relations sum to the total, up to rounding. The detector threshold is fixed at 0.5. Within each block, the two columns are the whole-source and number-only masks. $\dagger$ Qwen3.8 retains recurrent states.}
\label{fig:v33_selection}
\end{figure}

\paragraph{Answer accuracy, preservation, and error categories.}
Figures~\ref{fig:v33_full_scores}--\ref{fig:v33_selection} report both
source-information conditions, query-specific preservation, and selection
contributions. Table~\ref{tab:v33_current_dependent_exact} compares two rules using the answer-field parser
from Appendix~\ref{app:v18_unified}. The Gemma outputs frequently use code fences, so these scores
count correct fenced responses as format failures. The third scoring
rule permits that wrapper while still rejecting incomplete or additional
text. Table~\ref{tab:v33_failure_diagnostics_dependent} separates quantity, format and termination outcomes, with the uniform complete-quantity condition stated in its caption.

\subsection{Candidate Margin and Answer Likelihood}
\label{app:v52_margin_likelihood}

We rescore the 60 replacement groups and eight generators in
Appendix~\ref{app:v33_models}. Let $\ell_c$ and $\ell_o$ denote the
canonical current and old JSON answers' teacher-forced log probabilities,
summed over tokens without an end-of-turn token. Write $\Delta$ for
whole-source masking minus its matched control. The current- and
old-answer drops, $D_c=-\Delta\ell_c$ and $D_o=-\Delta\ell_o$, give
the margin change in Equation~\eqref{eq:margin}:
\begin{equation*}
\Delta_m=\Delta\ell_c-\Delta\ell_o=D_o-D_c .
\end{equation*}

\input{context/v33_current_dependent_exact}

\input{context/v33_failure_diagnostics_dependent}

\input{context/v33_selection_decomposition_content_same_unit}

A change common to both candidates therefore cancels in the margin. In
all 120 pairs of every model, the two scored strings share a 4--5-token
prefix and a 2--3-token suffix (the unit and the closing), and differ only
in their 1--3 number tokens. $\Delta_m>0$ in 479 of 480 unit-dependent
cases; in each of these, $D_c$ is the smaller change, and $\Delta_m$ is
the old answer's excess drop. The margin gain accordingly follows the old answer: its Spearman
correlation with $D_o$ is 0.91--0.99 in seven models and 0.67 in
Qwen3-4B. The margin gain is positive on every lost answer except one of
Qwen3-4B's 37, and the current candidate still outscores the old one
under masking in all 198 lost cases.

A lost answer is complete under full access and incomplete under
whole-source masking; a kept answer is complete under both. Cases that
are incomplete under full access (one each for Qwen3-8B and
Llama-3.1-70B) are excluded. On the unit-dependent cases, the
current-answer drop ranks lost above kept answers with AUC 0.89--0.98 in
the four Qwen and Llama models that lose answers
(Table~\ref{tab:v52_margin_likelihood}). The smaller margin gain gives
0.70 [0.57, 0.83] in Qwen3-4B, 0.61 and 0.43 in Qwen3-8B and Qwen3-32B,
and an inverted 0.11 [0.01, 0.26] in Llama-3.2-3B. Measuring the drop
relative to full access, which needs no matched control,
gives the same four AUCs within 0.01.

A case is flagged when $D_c$ exceeds a threshold $\tau$ chosen on the
other seven models only: a 0.25-nat grid from 0 to 10, maximizing pooled
Youden's $J$, with ties resolved toward the largest $\tau$. Pooled over
the eight held-out models, the threshold flags 86 of 95 lost answers
whose first answer has a wrong, generic or missing unit (exact interval
[0.83, 0.96]; [0.75, 0.96] when the group bootstrap also refits the
threshold). It flags 1 of 103 lost answers whose first answer is complete
and whose output then continues to the token limit (exact interval
[0.00, 0.05]). Of the 120 kept answers flagged across models, 115 are
unit-dependent answers that became less likely although greedy decoding
still produced them.

The second kind of loss is Gemma's. In 103 of its 117 losses, a complete
first answer is followed by text that runs to the 256-token limit
(Appendix~\ref{app:v52_fc_truncation}). The scored answer has no
end-of-turn token and cannot register this failure to stop. Under
masking, Gemma 4 26B-A4B's current answer becomes more likely in 56 of 60
unit-dependent cases, with mean $D_c=-0.79$ [$-0.94$, $-0.61$] nats; this is the
$+0.79$ current-candidate change in
Table~\ref{tab:v38_probability_decomposition}. Its AUCs rest on three
kept cases and are uninformative. Within Gemma 4 31B's losses, $D_c$
still separates its 13 content losses from its 47 post-answer losses (AUC
0.93 [0.84, 1.00]), and the threshold flags 10 of the 13 and none of the
47. A likelihood check without generation therefore detects content
losses and misses stopping failures.

\begin{table}[!htbp]
\WideTableStyle
\caption{Outcome fractions are percentages (original n/N); associated binomial intervals are in percent. AUC columns retain their original scale. The current-answer drop ranks lost complete answers more
consistently than the margin gain. Lost answers are complete under full access and
incomplete under whole-source masking; kept answers are complete under
both. AUC columns rank lost above kept unit-dependent answers.}
\label{tab:v52_margin_likelihood}
\setlength{\tabcolsep}{1.9pt}
{\setlength{\tabcolsep}{\dimexpr\tabcolsep*10/12\relax}
\begin{tabular}{cccccc}
\toprule
\multirow{2}{*}[-0.6ex]{\TableLabel{Model}} & \multirow{2}{*}[-0.6ex]{\TableLabel{\shortstack{Lost\\answers}}} & \multicolumn{2}{c}{\TableHead{AUC for a lost answer}} & \multicolumn{2}{c}{\TableHead{Threshold $\tau$ on the current-answer drop}} \\
\cmidrule(lr){3-4}\cmidrule(l){5-6}
 &  & \TableHead{\shortstack{Current-answer\\drop}} & \TableHead{\shortstack{Smaller\\margin gain}} & \TableHead{\shortstack{Lost\\flagged}} & \TableHead{\shortstack{Kept\\flagged}} \\\midrule
Qwen3-4B & \AnswerPct{61.7}{37/60} & 0.98 \TableSecondary{[0.94,\,1.00]} & 0.70 \TableSecondary{[0.57,\,0.83]} & \AnswerPct{94.6}{35/37} \TableSecondary{[82,\,99]\%} & \AnswerPct{13}{3/23} \TableSecondary{[3,\,34]\%}\\Qwen3-8B & \AnswerPct{25.4}{15/59} & 0.95 \TableSecondary{[0.88,\,0.99]} & 0.61 \TableSecondary{[0.43,\,0.78]} & \AnswerPct{100}{15/15} \TableSecondary{[78,\,100]\%} & \AnswerPct{88.6}{39/44} \TableSecondary{[75,\,96]\%}\\Qwen3-32B & \AnswerPct{40}{24/60} & 0.98 \TableSecondary{[0.95,\,1.00]} & 0.43 \TableSecondary{[0.28,\,0.58]} & \AnswerPct{87.5}{21/24} \TableSecondary{[68,\,97]\%} & \AnswerPct{11.1}{4/36} \TableSecondary{[3,\,26]\%}\\Qwen3.8-27B$^{\dagger}$ & \AnswerPct{0}{0/60} & -- & -- & -- & \AnswerPct{91.7}{55/60} \TableSecondary{[82,\,97]\%}\\\addlinespace[2pt]
Llama-3.2-3B & \AnswerPct{8.3}{5/60} & 0.89 \TableSecondary{[0.77,\,1.00]} & 0.11 \TableSecondary{[0.01,\,0.26]} & \AnswerPct{100}{5/5} \TableSecondary{[48,\,100]\%} & \AnswerPct{18.2}{10/55} \TableSecondary{[9,\,31]\%}\\Llama-3.1-70B & \AnswerPct{0}{0/59} & -- & -- & -- & \AnswerPct{5.1}{3/59} \TableSecondary{[1,\,14]\%}\\\addlinespace[2pt]
Gemma 4 26B-A4B & \AnswerPct{95}{57/60} & 0.32 \TableSecondary{[0.00,\,0.95]} & 0.26 \TableSecondary{[0.00,\,0.81]} & \AnswerPct{1.8}{1/57} \TableSecondary{[0,\,9]\%} & \AnswerPct{33.3}{1/3} \TableSecondary{[1,\,91]\%}\\Gemma 4 31B & \AnswerPct{100}{60/60} & -- & -- & \AnswerPct{16.7}{10/60} \TableSecondary{[8,\,29]\%} & --\\
\bottomrule
\end{tabular}}
\par\smallskip{\footnotesize\TableNoteAlign Lost answers and AUCs use the
60 unit-dependent replacements; brackets are 95\% group-bootstrap
intervals over the 60 base groups (10,000 draws, shared across models).
Draws in which one class receives no weight are dropped: Llama-3.2-3B
keeps 9,942 and Gemma 4 26B-A4B 9,536. Gemma 4 26B-A4B has three kept
unit-dependent answers and Gemma 4 31B none. Threshold columns also use
the unit-dependent replacements; of the 60 self-contained kept answers,
only Qwen3-32B's are flagged (5). $\tau$ is chosen on the other seven models (0.75
nats, except 0.50 for Qwen3-32B and Gemma 4 26B-A4B and 0.25 for
Qwen3.8-27B), and bracketed rates are exact binomial 95\% intervals given that $\tau$.
$\dagger$ Qwen3.8 masks only its full-attention layers and retains its
prefilled recurrent states.\TableNoteEnd}
\end{table}

The threshold is unstable. When every midpoint between distinct training
scores is a candidate, Qwen3-32B receives $\tau=0.12$ and flags 24 of 24
lost and 48 of 96 kept answers, so no single false-alarm rate describes
the check. Across both threshold rules and both baselines, 82--88 of the
95 content losses and 1 of the 103 post-answer losses are flagged.
Holding out a model does not hold out items, because all models share the
60 groups. Per-token log probabilities were not stored, so attributing
$D_c$ to the unit tokens is an interpretation of the token structure
above.

\subsection{Retained Text and Routing on Contextual MQuAKE}
\label{app:v52_mquake_retained}

This analysis stratifies the existing contextual MQuAKE answers of
Appendix~\ref{app:v41_stream} (Table~\ref{tab:v41_mquake}) by whether a
historical answer remains in the retained text, the history with every
source span removed. Text recomputation re-encodes exactly this text, and
source masking leaves exactly this text visible. A case is retained when
a historical reference or alias (normalized, at least three characters,
word-bounded) occurs in the retained records (520 of 600 cases). It is
masked-only when the answer occurs only inside a masked source record (80
cases); the canonical reference alone gives the same partition.
Intervals are paired whole-case bootstrap intervals (10,000 draws)
resampled within each stratum; small counts use exact McNemar and Fisher
tests.

Relative to source masking, recomputation's historical loss is confined
to the retained stratum (Table~\ref{tab:v52_mquake_retained}). That
stratum accounts for 247 of the net loss of 243 questions in Qwen3-4B,
247 of 246 in Qwen3-32B and 122 of 121 in Llama-3.2-3B; the masked-only
stratum shows a small net gain. The difference between strata is
$-52.5$ [$-60.3$, $-45.2$], $-48.8$ [$-56.8$, $-41.0$] and $-24.7$
[$-33.8$, $-15.7$] points. In the masked-only stratum every operation
except full access is near floor, including value masking (11, 9 and 17
of 80), so answers recalled from model weights alone are few and favor
neither operation. Within the 64 single-edit masked-only cases,
Qwen3-4B's recompute-minus-source difference is $+7.8$ [1.6, 15.6]
points; with 5 discordant pairs against 0, the exact McNemar test gives
$p=0.06$.

With the same visible text, the source mask keeps 439, 485 and 461 of
the 520 retained-stratum historical answers. Recomputation instead often
answers the historical question with the current value: in 244 (40.7\%
[36.8, 44.7]), 180 (30.0\% [26.3, 33.7]) and 56 (9.3\% [7.2, 11.8]) of
600 cases, against 61, 29 and 21 under the source mask and 12, 0 and 0
under full access. The recompute-minus-source differences are $+30.5$
[26.7, 34.5], $+25.2$ [21.3, 29.0] and $+5.8$ [3.5, 8.3] points. This
pattern is consistent with states computed while the old record was
visible keeping its version information after masking, which
recomputation discards although the retained text still contains the
answer. This lexical partition alone cannot separate state content from positional cues; Appendix~\ref{app:v52_factorial} supplies direct operation comparisons.

Recomputation also gains on current questions in both strata, except in
Llama-3.2-3B's masked-only stratum (169 versus 163). Of 1,560
retained and 240 masked-only current questions, source masking and
recomputation give 1,363/1,416 and 152/174 for Qwen3-4B, 1,306/1,449 and
192/222 for Qwen3-32B, and 982/1,107 and 169/163 for Llama-3.2-3B. The
historical losses and current gains mostly fall in different cases.
Among cases with a historical loss, 31 of 258 (12.0\%), 61 of 255
(23.9\%) and 28 of 134 (20.9\%) also gain a current paraphrase, against
80 of 342 (23.4\%), 80 of 345 (23.2\%) and 143 of 466 (30.7\%) of the
other cases. The conditional odds ratios are 0.45 [0.28, 0.71] (Fisher
$p=0.0004$), 1.04 [0.70, 1.55] ($p=0.85$) and 0.60 [0.36, 0.96]
($p=0.03$), and no interval lies entirely above one in any sensitivity
variant. The trade-off therefore operates between question types.

\begin{table}[!htbp]
\WideTableStyle
\caption{Historical matches and offline routing on MQuAKE and FactConsolidation. Answer cells give percentages (counts). The upper block uses the stated retained/masked-only stratum sizes; routing uses 3,000 MQuAKE or 373 source-answerable FactConsolidation questions per model.}
\label{tab:v52_mquake_retained}
\setlength{\tabcolsep}{1.35pt}
{\setlength{\tabcolsep}{\dimexpr\tabcolsep*12/14\relax}
\begin{tabular}{ccccccc}
\toprule
\TableGroupRow{7}{Historical matches by stratum}
\midrule
\TableHead{Model} & \TableHead{Historical answer} & \FullCell{\FullHeader{Full}} & \WholeCell{\WholeHeader{Source}} & \TableHead{Online} & \TableHead{Recompute} & \TableHead{\shortstack{Recompute $-$\\source (points)}} \\\midrule
Qwen3-4B & Retained (520) & \FullCell{\AnswerPct{93.5}{486}} & \WholeCell{\AnswerPct{84.4}{439}} & \AnswerPct{83.7}{435} & \AnswerPct{36.9}{192} & $-$47.5 \TableSecondary{[$-$52.1,\,$-$42.9]}\\ & Masked only (80) & \FullCell{\AnswerPct{53.8}{43}} & \WholeCell{\AnswerPct{6.3}{5}} & \AnswerPct{6.3}{5} & \AnswerPct{11.3}{9} & +5.0 \TableSecondary{[$-$1.2,\,11.2]}\\\addlinespace[2pt]
Qwen3-32B & Retained (520) & \FullCell{\AnswerPct{97.3}{506}} & \WholeCell{\AnswerPct{93.3}{485}} & \AnswerPct{93.5}{486} & \AnswerPct{45.8}{238} & $-$47.5 \TableSecondary{[$-$51.9,\,$-$43.1]}\\ & Masked only (80) & \FullCell{\AnswerPct{95}{76}} & \WholeCell{\AnswerPct{11.3}{9}} & \AnswerPct{10}{8} & \AnswerPct{12.5}{10} & +1.2 \TableSecondary{[$-$5.0,\,7.5]}\\\addlinespace[2pt]
Llama-3.2-3B & Retained (520) & \FullCell{\AnswerPct{84.8}{441}} & \WholeCell{\AnswerPct{88.7}{461}} & \AnswerPct{88.5}{460} & \AnswerPct{65.2}{339} & $-$23.5 \TableSecondary{[$-$27.3,\,$-$19.6]}\\ & Masked only (80) & \FullCell{\AnswerPct{92.5}{74}} & \WholeCell{\AnswerPct{25}{20}} & \AnswerPct{25}{20} & \AnswerPct{26.3}{21} & +1.2 \TableSecondary{[$-$6.3,\,8.8]}\\
\midrule
\TableGroupRow{7}{Offline routing over existing answers}
\midrule
\TableHead{Model} & \TableHead{Benchmark} & \multicolumn{2}{c}{\TableHead{Best single policy}} & \TableHead{Routed} & \TableHead{\shortstack{Routed $-$\\best (points)}} & \TableHead{Oracle} \\\midrule
Qwen3-4B & MQuAKE & \multicolumn{2}{c}{Full \AnswerPctInline{80.1}{2402}} & \AnswerPct{84.7}{2,540} & +4.6 \TableSecondary{[3.1,\,6.1]} & \AnswerPct{89.4}{2,682}\\Qwen3-32B & MQuAKE & \multicolumn{2}{c}{Full \AnswerPctInline{88.5}{2655}} & \AnswerPct{95.1}{2,853} & +6.6 \TableSecondary{[5.3,\,7.9]} & \AnswerPct{96.6}{2,897}\\Llama-3.2-3B & MQuAKE & \multicolumn{2}{c}{Online \AnswerPctInline{74.2}{2225}} & \AnswerPct{78.6}{2,359} & +4.5 \TableSecondary{[2.9,\,6.0]} & \AnswerPct{81.6}{2,447}\\\addlinespace[2pt]
Qwen3.8-27B & FactConsolidation & \multicolumn{2}{c}{Recompute \AnswerPctInline{95.2}{355}} & \AnswerPct{93.6}{349} & $-$1.6 (net $-$6) & \AnswerPct{96.5}{360}\\Gemma 4 26B-A4B & FactConsolidation & \multicolumn{2}{c}{Recompute \AnswerPctInline{71.3}{266}} & \AnswerPct{71}{265} & $-$0.3 (net $-$1) & \AnswerPct{74.8}{279}\\
\bottomrule
\end{tabular}}
\par\smallskip{\footnotesize\TableNoteAlign MQuAKE counts use normalized
reference exact match with EOS, over 3,000 questions per model in the
bottom block. Brackets are paired 95\% whole-case bootstrap intervals
(10,000 draws), resampled within each stratum. MQuAKE routing sends
questions with the historical prefix to the full cache and all others to
the recomputed cache; its oracle picks the better of those two caches per
question using reference answers. FactConsolidation counts are complete
answers with EOS on 373 source-answerable questions. Its routing
re-answers from the recomputed cache when the source-masked answer
reaches the token limit, and its oracle picks the better of the source
mask and recomputation. Oracles are upper bounds. The best single policy
is chosen after inspecting the results. FactConsolidation differences
are descriptive.\TableNoteEnd}
\end{table}

Because the two operations serve different question types, an offline
router over existing answers recovers both. Questions that begin with
the benchmark's historical prefix read the full cache, and all other
questions read the recomputed cache. The router uses no reference
answers, and choosing the cache for each prefix class on the other two
models selects the same router. It exceeds the best single policy by
$+4.6$ [3.1, 6.1], $+6.6$ [5.3, 7.9] and $+4.5$ [2.9, 6.0] points
(exact McNemar 276 vs 138, 239 vs 41 and 307 vs 173), closing 49\%, 82\%
and 60\% of the gap to the two-cache oracle. On FactConsolidation, a
fallback that re-answers from the recomputed cache after a capped
source-masked answer does not exceed recomputation alone: Qwen3.8-27B
gives 349 against 355 (58 fallbacks; 5 gains and 11 losses, exact
McNemar $p=0.21$), and Gemma 4 26B-A4B gives 265 against 266 (157
fallbacks; 13 and 14, $p=1.0$). The MQuAKE router must keep both caches
and relies on the benchmark stating historical intent in the question.

The strata are defined by the text and are not randomized. Masked-only
cases are mostly single-edit cases, and all 80 have their final-hop
record masked, as do five retained cases in which the answer string also
occurs elsewhere. The string matches are lexical, and routing recombines
existing answers without running a routed system.

\subsection{Unfinished Generation under Masking}
\label{app:v52_fc_truncation}

This analysis reuses the FactConsolidation generations and grades of
Appendices~\ref{app:v43_mab} and~\ref{app:v45_mab_analysis}, and the
Gemma quantity-task generations of Appendix~\ref{app:v33_models}.
FactConsolidation counts remain descriptive because the two histories
share one fact pool. Exact binomial intervals condition on the 373
source-answerable questions, and generated-token ratios use a bootstrap
over 319 components of questions that share an original MQuAKE case
(10,000 draws). 

Three lexical rules, written after reading outputs, describe the
responses. A search loop reaches the 1,024-token limit without a boxed
answer, contains at least three re-check phrases such as ``let me check''
or ``wait'', and cites one fact number at least three times. A verbatim
loop has more than half of its 8-grams repeated. A response quotes a
masked record when one line cites a superseded record's fact number
together with that record's subject and old value. 

\begin{table}[!htbp]
\WideTableStyle
\caption{Outcome entries are percentages (counts). Answer style, quotation of masked records and token-limit stops
on the 373 source-answerable FactConsolidation questions (counts of 373).
A box-only answer consists of a single boxed answer. Same style compares
each output with the full-access output for the same question.}
{\setlength{\tabcolsep}{\dimexpr\tabcolsep*10/12\relax}
\begin{tabular}{cccccc}
\toprule
\TableHead{Model} & \TableHead{Operation} & \TableHead{Box-only} & \TableHead{\shortstack{Same style as\\full access}} & \TableHead{\shortstack{Quotes a\\masked record}} & \TableHead{\shortstack{Stopped at\\1,024 tokens}} \\\midrule
Qwen3.8-27B$^{\dagger}$ & \FullCell{Full access} & \FullCell{\AnswerPct{12.1}{45}} & \FullCell{--} & \FullCell{\AnswerPct{79.6}{297}} & \FullCell{\AnswerPct{6.7}{25}}\\
\addlinespace[2pt]
Gemma 4 26B-A4B & \FullCell{Full access} & \FullCell{\AnswerPct{0}{0}} & \FullCell{--} & \FullCell{\AnswerPct{64.9}{242}} & \FullCell{\AnswerPct{17.2}{64}}\\
\bottomrule
\end{tabular}}
\par\smallskip{\footnotesize\TableNoteAlign Token-limit counts match the
capped outcomes in Table~\ref{tab:v45_mab_outcomes}. $\dagger$ Qwen3.8
masks only its 16 full-attention layers; its 48 recurrent layers keep the
superseded records.\TableNoteEnd}
\end{table}

Qwen3.8-27B loses 43 complete answers under source masking. Of these, 37
stop at the limit, 37 are multi-hop questions, against 102 of 293 among
questions complete under both (Fisher $p=1.5\times10^{-10}$), and 37 of
the 43 (86\%, 95\% interval [72, 95]\%) meet the search-loop rule. 
Only 3 of the 37 capped losses are verbatim loops, so the unfinished
outputs are repeated look-ups of records. Of the 37 capped losses,
73.0\% (27) contain the source-supported answer string without a final
box, but their frozen semantic grades are incorrect in 26 cases and
unsupported in one. For example, one output reaches Austin while
repeatedly attributing it to an old numbered fact that actually states
Mexico City. The correct endpoint string therefore does not make the
surrounding response source-supported. Recomputation produces a
source-supported final answer with EOS for 92.6\% (25/27) of these
questions; the parser evaluates the final box when present and otherwise
the full response. 
Their full-access outputs are already long, with a median of 351
generated tokens against 143 for questions complete under both. 
Gemma 4 26B-A4B loses 13: 12 stop at the limit, 5 are multi-hop, and
8 of the 12 capped losses are verbatim loops;
the corresponding medians are 166 and 82 tokens. 

Masking lengthens generation. Over each history's 200 questions,
source-masked outputs contain 1.29 [1.18, 1.42] and 1.26 [1.16, 1.40]
times as many generated tokens as full access for Qwen3.8-27B at the 32k
and 64k lengths, and 1.77 [1.53, 2.09] and 1.71 [1.48, 2.01] times as
many for Gemma 4 26B-A4B. 
Text recompute gives 0.57 at both lengths for Qwen and 1.09 and 1.35 for
Gemma. 
Isolated operation timings do not include this behavioral cost; our
eager-attention timings of these runs are diagnostic, so token counts are
the hardware-independent measure.

Masking keeps Qwen's full-access answer style: its source-masked output
has the full-access style in 366 of 373 questions, against 211 for text
recompute (exact McNemar $p=5.0\times10^{-42}$). 
Recompute at the 64k length (50,883 tokens) is box-only in 103 of 189
questions, against 22 of 184 for full access at 32k (39,528 tokens;
Fisher $p=5.7\times10^{-19}$), so the shorter input alone does not
account for the change. 
Gemma never gives a box-only answer, so this comparison is uninformative
for Gemma.

Masked records remain quotable. Source masking roughly halves quotation
of superseded records in both models (297 to 148 and 242 to 96 of 373),
the value mask reduces it less (249 and 179), and only recomputation
removes it. 
Masking hides the record positions during question processing and
decoding, but the retained states were computed while those records were
visible, and Qwen3.8's recurrent layers also keep them. This analysis
does not separate the two routes; the retained-text analysis of MQuAKE
(Appendix~\ref{app:v52_mquake_retained}) shows a similar, equally
unseparated, persistence.

The same unfinished generation appears in the controlled quantity task.
Under whole-source masking, Gemma 4 26B-A4B reaches the 256-token limit
in 57 of 60 unit-dependent replacements, and 56 of the 57 capped outputs
begin with a correct first answer. 
Gemma 4 31B reaches the limit in all 60, and 47 begin with a correct first
answer; the other 13 give the right number without its unit. 
No capped replacement output contains a second, different answer object
(0 of 117), and where it can be located (55 of 57 and 47 of 60) the first
answer ends at token 12 of 256. 
Every capped output contains ``Wait'', ``re-read'' or ``misread'', and
most also state that a referent is missing; the model continues to the
token limit without giving another answer. 

\begin{table}[!htbp]
\WideTableStyle
\setlength{\tabcolsep}{1.7pt}%
\caption{Outcome fractions are percentages (original n/N); associated binomial intervals are in percent. AUC columns retain their original scale. Gemma quantity-task outputs that reach the 256-token limit under
whole-source masking, for current questions. The first answer is the
first JSON object with a string answer field, scored by the content rule
(same number and an accepted form of the unit).}
{\setlength{\tabcolsep}{\dimexpr\tabcolsep*10/12\relax}
\begin{tabular}{cccccc}
\toprule
\TableHead{Model} & \TableHead{Cases} & \TableHead{\shortstack{Stopped at\\the limit}} & \TableHead{\shortstack{First answer\\correct}} & \TableHead{\shortstack{Re-check\\phrase}} & \TableHead{\shortstack{Missing-referent\\phrase}} \\\midrule
Gemma 4 26B-A4B & Unit-dependent replacement & \AnswerPct{95}{57/60} & \AnswerPct{98.2}{56/57} \TableSecondary{[91,\,100]\%} & \AnswerPct{100}{57/57} & \AnswerPct{73.7}{42/57}\\
\addlinespace[2pt]
Gemma 4 31B & Unit-dependent replacement & \AnswerPct{100}{60/60} & \AnswerPct{78.3}{47/60} \TableSecondary{[66,\,88]\%} & \AnswerPct{100}{60/60} & \AnswerPct{95}{57/60}\\
\bottomrule
\end{tabular}}
\par\smallskip{\footnotesize\TableNoteAlign Brackets are exact binomial
95\% intervals. The re-check phrase is ``Wait'', ``re-read'' or
``misread''; the missing-referent phrase includes wording such as ``not
specified'', ``does not mention'' or ``missing''.\TableNoteEnd}
\end{table}

These counts describe what the capped outputs contain. The reported
scores still count every output that reaches the limit as incomplete,
because the scoring rule does not extract a fragment from a continuing
response. Appendix~\ref{app:v52_margin_likelihood} shows that candidate
likelihoods do not register this failure to stop.

\subsection{Positions and Later States on MQuAKE}
\label{app:v52_factorial}

We compare five operations on 600 MQuAKE cases per model, counting
normalized reference matches that terminate with EOS. Full access,
source masking and text recomputation use the existing generations;
rebuilding states with the source hidden and compacting masked rows add
two direct comparisons. Repeating the source-mask and text-recompute
conditions on 50 cases changes the measured answer rates by at most
2.0 percentage points.

The direct comparisons with source masking are reported in
Table~\ref{tab:mquake_direct_contrasts}, with absolute rates in
Table~\ref{tab:main_mquake_states}. Table~\ref{tab:v52_mquake_factorial}
also retains averages across the four modified-cache operations.
These are descriptive contrasts, not a strict decomposition into
independent position and state changes. Rebuilding at the original
positions and then compacting differs from prefilling the shortened
text: the later states are formed at different distances from earlier
rows, and deleting text can also change tokenization. The historical
interaction intervals contain zero; they do not establish the absence
of interactions in other histories.

\begin{table}[!htbp]
\MainTableStyle
\caption{Direct changes from source masking for the operations in
Table~\ref{tab:main_mquake_states}. Entries are percentage points (pp)
with paired 95\% intervals, using 600 cases for each question type.}
\label{tab:mquake_direct_contrasts}
\begin{tabular}{C{62pt}*{4}{C{61pt}}}
\toprule
\multicolumn{5}{c}{{\MainTableSmall\textcolor{figureGray}{Reference: source masking}}}\\
\midrule
& \multicolumn{2}{c}{\MainTableHead{Current change}}
& \multicolumn{2}{c}{\MainFocusCell{\MainTableHead{Historical change}}} \\
\cmidrule(lr){2-3}\cmidrule(lr){4-5}
\MainColumnHead{Model} & \FullHeader{Rebuild states} & \FullHeader{Compact rows}
& \MainFocusCell{\FullHeader{Rebuild states}} & \MainFocusCell{\FullHeader{Compact rows}} \\
\midrule
Qwen3-4B
& \FullCell{\MainEstimate{+7.3}{+4.7}{+10.2}} & \FullCell{\MainEstimate{-1.5}{-3.3}{+0.3}}
& \MainFocusCell{\MainEstimate{-41.3}{-45.7}{-37.0}} & \MainFocusCell{\MainEstimate{0.0}{-2.0}{+1.8}} \\
\addlinespace[1.5pt]
Llama-3.2-3B
& \FullCell{\MainEstimate{+3.3}{-0.2}{+6.8}} & \FullCell{\MainEstimate{+0.5}{-1.0}{+2.0}}
& \MainFocusCell{\MainEstimate{-19.8}{-23.3}{-16.3}} & \MainFocusCell{\MainEstimate{+1.2}{-0.2}{+2.5}} \\
\addlinespace[1.5pt]
Qwen3-32B
& \FullCell{\MainEstimate{+11.5}{+8.5}{+14.5}} & \FullCell{\MainEstimate{-1.7}{-3.8}{+0.5}}
& \MainFocusCell{\MainEstimate{-37.5}{-41.5}{-33.3}} & \MainFocusCell{\MainEstimate{-0.2}{-1.8}{+1.7}} \\
\bottomrule
\end{tabular}
\par\smallskip{\MainTableSmall\TableNoteAlign
Rebuild states keeps the original positions and the source mask;
Compact rows removes masked rows and repositions later keys without
rebuilding their content. Current uses the first paraphrase only.
Intervals use 10,000 paired resamples of the 600 cases. Changes are
calculated before rounding the absolute rates. These are direct
operation-minus-source contrasts, distinct from the averaged effects in
Table~\ref{tab:v52_mquake_factorial}.
\TableNoteEnd}
\end{table}

\input{context/v52_factorial_table}

A separate 60-group Qwen3-32B quantity panel isolates deletion from masking.
Compaction raises complete answers from 36/60 under whole-source masking to
40/60, a paired gain of 6.7 points [1.7, 13.3]; text recomputation gives 3/60
in the same panel. The full, compact and recomputed caches are validated by
their anchor rows, but their bf16 first-logit values are not bitwise identical;
the result is therefore scoped to the answer metric and these 60 groups.

\begin{table}[!htbp]
\WideTableStyle
\caption{Complete and omission entries are percentages (counts), N=60; changes remain percentage points. Deleting instead of masking the old record for Qwen3-32B on the 60 replacements whose later record refers to the earlier unit. Masking keeps the gap left by the old record; compaction deletes its cached rows and rotates every later key back by the gap, keeping the later states; recomputation rebuilds the later states from the shortened history. Each control applies the same operation to an equal number of unrelated-note tokens. Margins are current-minus-old candidate log probabilities relative to the matched control, in nats. Intervals are paired 95\% group-bootstrap intervals (10,000 draws).}
\label{tab:v52_compact_headline}
{\setlength{\tabcolsep}{\dimexpr\tabcolsep*8/10\relax}
\begin{tabular}{ccccc}
\toprule
\TableHead{Operation} & \TableHead{\shortstack{Complete\\answers}} & \TableHead{\shortstack{Unit\\omitted}} & \TableHead{\shortstack{Margin change\\(nats)}} & \TableHead{\shortstack{Change from\\masking (pp)}} \\\midrule
\FullCell{\FullHeader{Full access}} & \FullCell{\AnswerPct{100}{60/60}} & \FullCell{\AnswerPct{0}{0}} & \FullCell{--} & \FullCell{--}\\\WholeCell{\WholeHeader{Masking}} & \WholeCell{\AnswerPct{60}{36/60}} & \WholeCell{\AnswerPct{40}{24}} & \WholeCell{+7.35 \TableSecondary{[+6.62,\,+8.15]}} & \WholeCell{--}\\Compaction & \AnswerPct{66.7}{40/60} & \AnswerPct{33.3}{20} & +7.86 \TableSecondary{[+7.09,\,+8.70]} & +6.7 \TableSecondary{[+1.7,\,+13.3]} (4/0)\\Recomputation & \AnswerPct{5}{3/60} & \AnswerPct{95}{57} & +2.31 \TableSecondary{[+1.45,\,+3.14]} & $-$55.0 \TableSecondary{[$-$66.7,\,$-$41.7]} (0/33)\\
\bottomrule
\end{tabular}}
\par\smallskip{\footnotesize\TableNoteAlign Full access and masking are the released generations; a re-run of all three gap conditions reproduces them token for token when the anchor check passes. Complete answers accept one answer object, optionally in one clean fence, with the same-unit aliases and no unit conversion. Parentheses count groups that gain or lose a complete answer relative to whole-source masking.\TableNoteEnd}
\end{table}

\subsection{A no-context baseline for historical questions}
\label{app:v52_weights}

A weights-only baseline makes the operation comparison conditional. Among
historical-wording cases that the no-context model answers incorrectly, full
access answers 84.7--97.4\% correctly across the three models, source masking
71.8--79.8\%, value masking 73.2--80.7\%, and text recomputation 31.6--57.3\%.
The denominator is the no-context-wrong subset; the table keeps the wording
and the normal-EOS exact-match rule explicit.

\input{context/v52_weights_table}

\subsection{Open Answers on Natural Source Excerpts}
\label{app:v41_oaks_open}

We remove the answer options and A/B/UNKNOWN response instruction from all
116 original OAKS questions \citep{oaks2026}, retaining each question, source excerpt,
and selected reference-answer text. Belief, suspicion, and narrative
knowledge questions retain their epistemic meaning. Three generators use
native templates, greedy decoding, normal EOS, and a 256-token ceiling;
full access, old-source masking, and equal-token unrelated masking share
the same prefilled history. Each model produces 348 answers, and all
1,044 original answers terminate normally.

\paragraph{Source-aware semantic review.}
A model grader assesses both source completeness and reference agreement.
Its exact model identity, prompts and outputs are retained in the
evaluation records.
For each question, it receives the complete original earlier, later, and
unrelated excerpts, the original question and reference, and anonymous
candidate answers. Generator, access condition, and previous scores are
hidden. Exact repeats share one judgment, giving 613 distinct answers
evaluated in 29 batches. Every supporting quotation must occur literally
in its named excerpt. The generated answers and references are evaluated
without rewriting.

The rubric separates agreement with the original reference from whether
the answer conveys all required details supported by the source. It also
separates missing details, contradictory or unsupported claims, and
justified or unjustified abstention. The review identifies 92 answerable,
17 partially answerable, six unanswerable, and one ambiguous question.
References are fully supported in 54 cases, partially supported in 51,
unsupported in seven, and mismatched to the question in four. The rule
to report answerable questions separately was fixed before this review,
but source assessments and answer labels are produced together. These
assessments are not independent source-only annotations or human judgments.

\begin{table}[!htbp]
\WideTableStyle
\caption{Outcome entries are percentages (counts). Open OAKS answers evaluated against the supplied source by a model grader. Completeness is measured on the 92 questions judged answerable from their excerpts.}
\label{tab:v41_oaks_open}
{\setlength{\tabcolsep}{\dimexpr\tabcolsep*10/12\relax}
\begin{tabular}{cccccc}
\toprule
\multirow{2}{*}[-0.6ex]{\TableLabel{Model}} & \multicolumn{3}{c}{\TableHead{Complete /92}} & \multicolumn{2}{c}{\TableHead{Mask difference (pp) [95\% CI]}} \\
\cmidrule(lr){2-4}\cmidrule(l){5-6}
 & \FullCell{\FullHeader{Full}} & \WholeCell{\WholeHeader{Source}} & \FullCell{\FullHeader{Control}} & \TableHead{Versus full} & \TableHead{Versus control} \\\midrule
Qwen3-4B & \FullCell{\AnswerPct{30.4}{28}} & \WholeCell{\AnswerPct{22.8}{21}} & \FullCell{\AnswerPct{32.6}{30}} & $-$7.6 \TableSecondary{[$-$16.9,\,+1.1]} & $-$9.8 \TableSecondary{[$-$18.9,\,$-$1.1]}\\Llama-3.2-3B & \FullCell{\AnswerPct{25}{23}} & \WholeCell{\AnswerPct{27.2}{25}} & \FullCell{\AnswerPct{28.3}{26}} & +2.2 \TableSecondary{[$-$5.4,\,+10.0]} & $-$1.1 \TableSecondary{[$-$9.0,\,+6.7]}\\Qwen3-32B & \FullCell{\AnswerPct{55.4}{51}} & \WholeCell{\AnswerPct{45.7}{42}} & \FullCell{\AnswerPct{57.6}{53}} & $-$9.8 \TableSecondary{[$-$18.5,\,$-$1.1]} & $-$12.0 \TableSecondary{[$-$20.2,\,$-$4.3]}\\
\bottomrule
\end{tabular}}
\par\smallskip{\footnotesize\TableNoteAlign Complete requires all source-supported required details, no conflicting or unsupported substantive claim, and EOS. The review sees full original excerpts but no generator or access-condition labels. Pointwise paired 95\% intervals use 10,000 resamples of 87 reused-source components within five fixed books. Judge labels are held fixed; other narrative and donor dependence remains. This is separate from supplied-option scoring.\TableNoteEnd}
\end{table}

On the 92 answerable questions, Qwen3-32B's source-complete count falls
from 51 to 42 under masking; the paired difference is $-9.8$ percentage
points with 95\% interval $[-18.5,-1.1]$. Qwen3-4B falls from 28 to 21
($-7.6$, $[-16.9,+1.1]$), while Llama-3.2-3B rises from 23 to 25
($+2.2$, $[-5.4,+10.0]$). Intervals resample 87 reused-source
components within the five fixed books, holding evaluator labels fixed.
They do not measure judge reliability, and other narrative and
neutral-donor dependence remains. The existing 22 changed, 53 compatible,
and 41 unresolved relation labels remain descriptive strata.

\begin{table}[!htbp]
\WideTableStyle
\caption{Reference agreement uses116 questions; failure shares condition on the stated full-complete baseline count. Reference agreement and source-aware failure types measure different outcomes. The right panel starts from source-complete full-access answers on the 92 answerable questions.}
\label{tab:v43_oaks_error_types}
{\setlength{\tabcolsep}{\dimexpr\tabcolsep*14/16\relax}
\begin{tabular}{cccccccc}
\toprule
\multirow{2}{*}[-0.6ex]{\TableLabel{Model}} & \multicolumn{3}{c}{\TableHead{Reference agrees /116}} & \multicolumn{4}{c}{\TableHead{Full-complete $\to$ masked outcome}} \\
\cmidrule(lr){2-4}\cmidrule(l){5-8}
 & \FullCell{\FullHeader{Full}} & \WholeCell{\WholeHeader{Source}} & \FullCell{\FullHeader{Control}} & \TableHead{Baseline} & \TableHead{Partial} & \TableHead{Abstain} & \TableHead{Wrong} \\\midrule
Qwen3-4B & \FullCell{\AnswerPct{20.7}{24}} & \WholeCell{\AnswerPct{19}{22}} & \FullCell{\AnswerPct{22.4}{26}} & 28 & \AnswerPct{7.1}{2} & \AnswerPct{25}{7} & \AnswerPct{14.3}{4}\\Llama-3.2-3B & \FullCell{\AnswerPct{15.5}{18}} & \WholeCell{\AnswerPct{20.7}{24}} & \FullCell{\AnswerPct{16.4}{19}} & 23 & \AnswerPct{0}{0} & \AnswerPct{8.7}{2} & \AnswerPct{17.4}{4}\\Qwen3-32B & \FullCell{\AnswerPct{38.8}{45}} & \WholeCell{\AnswerPct{41.4}{48}} & \FullCell{\AnswerPct{39.7}{46}} & 51 & \AnswerPct{7.8}{4} & \AnswerPct{11.8}{6} & \AnswerPct{7.8}{4}\\
\bottomrule
\end{tabular}}
\par\smallskip{\footnotesize\TableNoteAlign Reference agreement retains every original question and reference and requires EOS. Partial counts a complete answer becoming a substantive answer that keeps at least one required detail and omits another, without a wrong claim. Abstentions and contradictory or unsupported assertions are separate. These are gross losses among the stated complete baselines; corrections elsewhere also contribute to net accuracy.\TableNoteEnd}
\end{table}

An omission requires a complete baseline answer to become a partial
substantive answer, retaining at least one required detail while losing
another without introducing a wrong claim. Relative to full access,
there are two such cases among Qwen3-4B's 28 complete baselines, none
among Llama's 23, and four among Qwen3-32B's 51. Abstentions account for
seven, two, and six other losses, respectively; incorrect or unsupported
assertions account for four in each model. These gross losses differ
from net accuracy because masking also corrects other answers.
Losses of individual details from already incomplete baselines are
recorded separately in the numerical results.
Relative to the matched control, pure omission counts are one, zero,
and three, respectively. Partially answerable questions and justified
abstentions are retained as separate numerical sensitivities.

\paragraph{What the source changes about individual cases.}
For Qwen3-32B on \texttt{P22\_q15}, full access gives the rigging and
money named in the later excerpt, whereas masking gives only the money.
The review confirms this detail loss, while rejecting the reference's
additional materials as unsupported by that excerpt. For \texttt{P27\_q23},
Llama's full answer overstates the saloon stay, and its masked answer is
judged complete. For \texttt{P37\_q19}, all three Qwen3-32B answers are
judged complete. For \texttt{P37\_q17}, all three contain unsupported emotional
speculation, so similar wording does not establish correctness. A
plan-versus-achievement mismatch also affects \texttt{P12\_q24}.

\paragraph{Three grading protocols on the same answers.}
The same 1,044 original answers also receive two reference-based grades:
agreement with the original reference from the source-aware grader, and a separate
GPT-5.6 review that applies the earlier reference-only rubric on all
116 questions, without the original excerpts or the source-supported
detail target used here (Table~\ref{tab:v52_oaks_protocols}). 
The source-aware review was defined after this reference-only rubric and
changes some earlier case labels (for example \texttt{P27\_q23} and
\texttt{P37\_q19} above).
On the 92 answerable questions, only the source-aware completeness review
resolves a Qwen3-32B loss: $-9.8$ [$-18.5$, $-1.1$] points, against
$+2.2$ [$-5.4$, $+9.5$] for the same grader's reference agreement and $+2.2$
[$-6.5$, $+10.5$] for the GPT-5.6 regrade. 
The paired difference between the two protocols using the same grader is itself
resolved, at $+12.0$ [$+2.1$, $+22.1$] points. 
On all 116 questions, the Qwen3-32B values are $-7.8$ [$-15.2$, $-0.8$],
$+2.6$ [$-3.4$, $+8.7$] and $+2.6$ [$-5.1$, $+10.3$], with a protocol
difference of $+10.3$ [$+1.7$, $+19.1$]. 
For Qwen3-4B and Llama-3.2-3B, no protocol resolves a loss; Llama's
answerable reference-agreement gain, $+6.5$ [$+0.0$, $+14.0$], has its
lower bound at zero. 
The two protocols using the same grader disagree on 179 of the 1,044 answers, 98 of
them on questions whose reference is only partially supported by the
excerpt. 
The Qwen3-32B verdict therefore depends on whether graders credit
source-supported details that the reference omits. The labels come from
model graders, so neither label disagreement nor a change in score can be
attributed solely to model capability.

\begin{table}[!htbp]
\WideTableStyle
\caption{OAKS mask effects under three grading protocols on the same generated answers. Entries are percentages (counts) over each panel\textquotesingle s question set; differences are paired percentage-point changes.}
\label{tab:v52_oaks_protocols}
{\setlength{\tabcolsep}{\dimexpr\tabcolsep*10/12\relax}
\begin{tabular}{cccccc}
\toprule
\TableHead{Model} & \TableHead{Protocol} & \FullCell{\FullHeader{Full}} & \WholeCell{\WholeHeader{Source}} & \FullCell{\FullHeader{Control}} & \TableHead{\shortstack{Source $-$ full\\(pp) [95\% CI]}} \\\midrule
\TableGroupRow{6}{92 source-answerable questions}
\midrule
Qwen3-4B & Source complete & \FullCell{\AnswerPct{30.4}{28}} & \WholeCell{\AnswerPct{22.8}{21}} & \FullCell{\AnswerPct{32.6}{30}} & $-$7.6 \TableSecondary{[$-$16.9,\,+1.1]}\\
\addlinespace[2pt]
Llama-3.2-3B & Source complete & \FullCell{\AnswerPct{25}{23}} & \WholeCell{\AnswerPct{27.2}{25}} & \FullCell{\AnswerPct{28.3}{26}} & +2.2 \TableSecondary{[$-$5.4,\,+10.0]}\\
\addlinespace[2pt]
Qwen3-32B & Source complete & \FullCell{\AnswerPct{55.4}{51}} & \WholeCell{\AnswerPct{45.7}{42}} & \FullCell{\AnswerPct{57.6}{53}} & $-$9.8 \TableSecondary{[$-$18.5,\,$-$1.1]}\\
\midrule
\TableGroupRow{6}{All 116 questions}
\midrule
Qwen3-4B & Source complete & \FullCell{\AnswerPct{30.4}{28}} & \WholeCell{\AnswerPct{23.9}{22}} & \FullCell{\AnswerPct{32.6}{30}} & $-$5.2 \TableSecondary{[$-$12.8,\,+1.8]}\\
\addlinespace[2pt]
Llama-3.2-3B & Source complete & \FullCell{\AnswerPct{19.8}{23}} & \WholeCell{\AnswerPct{24.1}{28}} & \FullCell{\AnswerPct{22.4}{26}} & +4.3 \TableSecondary{[$-$2.5,\,+11.2]}\\
\addlinespace[2pt]
Qwen3-32B & Source complete & \FullCell{\AnswerPct{45.7}{53}} & \WholeCell{\AnswerPct{37.9}{44}} & \FullCell{\AnswerPct{49.1}{57}} & $-$7.8 \TableSecondary{[$-$15.2,\,$-$0.8]}\\
\bottomrule
\end{tabular}}
\par\smallskip{\footnotesize\TableNoteAlign Source complete and reference
agreement use the same model grader and require EOS; the GPT-5.6
regrade uses the earlier reference-only rubric without the original
excerpts. Brackets are pointwise paired 95\% intervals from 10,000
resamples of reused-source components within the five books (87
components for the answerable questions, 109 for all 116). Under the
same draws, reference agreement minus source completeness is $+4.3$
[$-4.3$, $+13.2$], $+4.3$ [$-3.2$, $+12.8$] and $+12.0$ [$+2.1$,
$+22.1$] points on the answerable questions, and $+3.4$ [$-3.5$,
$+10.6$], $+0.9$ [$-5.8$, $+7.8$] and $+10.3$ [$+1.7$, $+19.1$] on all
116, for Qwen3-4B, Llama-3.2-3B and Qwen3-32B.\TableNoteEnd} 
\end{table}

The source-aware review quantifies a narrow omission endpoint and
model-dependent completeness changes; it does not establish uniform
natural-text failure or error-free evaluation.

\subsection{Varying Source Information, Mask Scope, and Answer Requirements}
\label{app:v17_scope}

Whether a model omits the unit depends on the answer requirements and which source tokens are masked.
With the original question and B stating the complete quantity, whole-source masking gives 25/30
complete answers for Qwen3-4B and 10/30 for Qwen3-8B under the notebook
note; masking only the old number gives 30/30 and 27/30. Explicitly requesting the unit yields 30/30 for both Qwen models under
every mask condition and both notes. Thus, the observed omissions depend on the question and masking condition;
they do not follow from replacement alone.
The full matrix also retains failures under other source conditions.

\paragraph{Factorial design and inference settings.}
The quantity matrix includes every one of the 30 replacement, role-swap,
and old-rejection cases from ten base groups, three generators, two question formulations, two neutral notes, two conditions for the information stated in B, and five
conditions: $3\times30\times2\times2\times2\times5=3{,}600$ generations.
Before inference we fix the case IDs, texts, spans, canonical answers,
scoring, and 10,000-draw group bootstrap. Each factor level retains all
cases, including initially wrong answers. Repeated variants and models
are not treated as additional independent questions.

Repeated inference on all 300 histories from Appendix~\ref{app:v7_bridge} under the three
conditions for all three models (2,700 generations) gives identical prompts,
candidate token sequences, generated answers, and sequence scores;
the maximum absolute score difference is zero.
That experiment uses Torch 2.11.0, Transformers 5.8.0, NumPy 2.2.6,
bfloat16 eager attention, native chat, greedy EOS-stopped decoding, and
40-token limit on H200. Llama's template date is fixed to 11 September 2026.
With this date argument, all 300 formatted prompts have identical hashes
across repeated evaluations.

\paragraph{Source information and mask positions.}
The original B explicitly states the entity, attribute, current magnitude,
and \emph{hours}. In the paired condition, we remove all occurrences
of \emph{hours} from B and prepend: ``Use the same measurement unit as
the Time 1 entry.'' Entity, attribute, magnitudes, and affirmed/rejected
roles are unchanged. A supplies that unit. The unrelated record also
still contains hours. B now explicitly refers to A, but other unit cues remain
available in the history.

The complete-answer question appends: ``Return the complete queried value,
including its measurement unit, in the answer string. Preserve the unit
established by the records.'' It does not itself supply the unit. The
notebook note repeats the original sentence 16 times; the alternative
repeats ``A wooden shelf holds a closed sketchbook beside a ceramic bowl.''
16 times. This factor changes both neutral content and length, so later
tokens can occupy different absolute positions across the two notes.

Whole-source masks cover every tokenizer position overlapping A. Number-only masks
cover only each old-number occurrence in A that is followed by the unit,
excluding matching digits in an entity identifier and excluding the unit
tokens themselves. Role-swapped and rejected candidate mentions are recorded
explicitly. For each mask, the control blocks the same number of tokens from the start
of the neutral note. The control span does not overlap A or B. The released
evaluation data include token IDs, character boundaries, masked text,
and all candidate mentions.
Number-only masking blocks direct attention to those positions. Old-value
mentions in B, other cues, and history states already computed after A remain
available.

Each textual history is freshly prefilled. Every condition copies the same
unmodified prefix for that history and retains its absolute positions.
In all 1,080 model--history combinations, every layer's original K/V tensors
remain identical after the conditions are evaluated. Each changed B or note is prefilled
separately, without assuming cache identity across different histories.

\begin{figure}[!htbp]
  \centering
  \includegraphics[width=\linewidth]{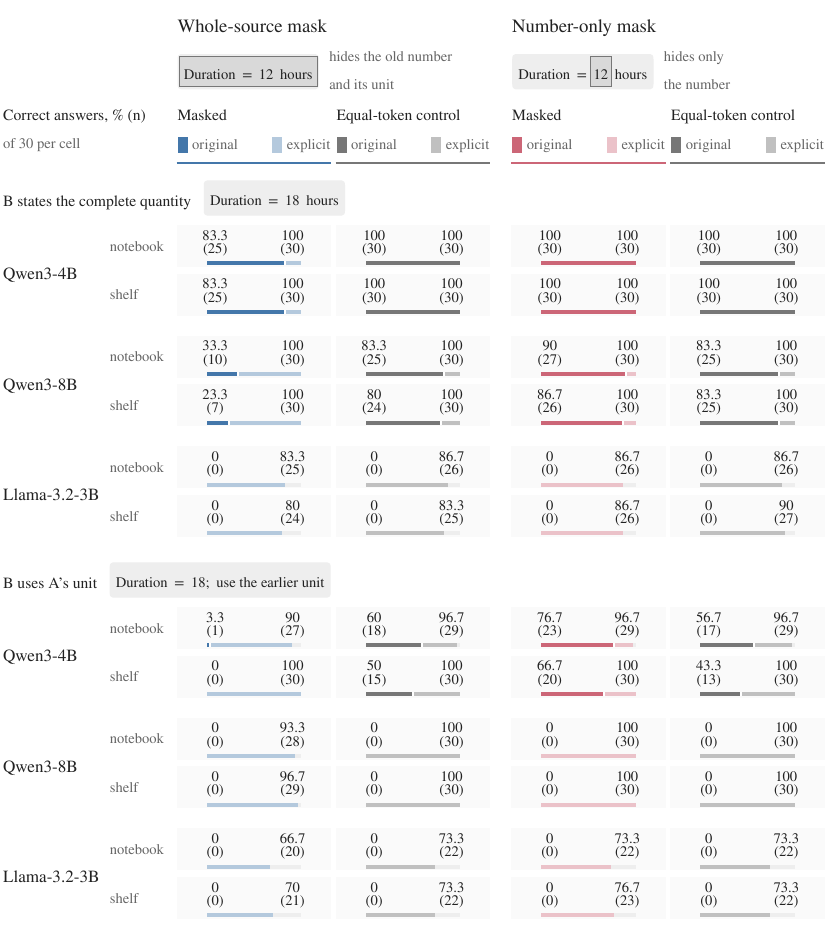}
  \caption{Correct quantity answers as percentages (counts) over 30 cases from ten groups for three models.
  Each bar spans the 30 cases: its dark part is the count for the original question, and it
  continues in a pale shade up to the count when the question explicitly requests the unit;
  the paired labels give these two percentages, with counts in parentheses. Blue bars show whole-source
  masks, rose bars number-only masks, and gray bars their matched equal-token controls; the
  chips box the part of A that each mask hides. Row groups separate the information stated in B,
  with one row per notebook or shelf note. The bars retain 0/30 floors and 30/30 ceilings.
  When B refers to A for its unit, an unrelated record still states hours.
  Tables~\ref{tab:v35_quantity_explicit_original} and~\ref{tab:v35_quantity_context_dependent_original} give every absolute count, including these three models.}
  \label{fig:v17_quantity}
\end{figure}

\paragraph{Scoring and conditional findings.}
We score every generated answer by exact answer-field matching and by comparing
the returned and reference values as defined in Appendix~\ref{app:v8_units}. We retain separate counts for UNKNOWN, obsolete values, other values, and invalid formats. Exact scoring normalizes case and whitespace; its answer-field
JSON parser permits additional fields, so strict single-field validity is
also recorded separately. Across all 5,400 quantity and endpoint outputs,
answer-field and strict-single-field validity agree: 1,800/1,800 for
each Qwen and 1,795/1,800 for Llama. The five Llama violations use a numeric
answer field and are already incorrect under both definitions; schema
strictness changes no reported contrast. The canonical current and old JSON strings are
scored by summed token log probability without EOS in each condition.

When B states the complete quantity, with the original question and notebook note, whole-source
masking lowers the log probability of the current answer string by 0.511/1.835 for Qwen3-4B/8B,
while the log probability of the old answer string falls by 2.107/4.230. The positive margin change
therefore coexists with loss of complete units. Number-only masking changes
the log probability of the current answer string by $+0.006/+0.258$ against its shorter control and
has no exact-answer reversals in these two cells. Both absolute scores,
margins, corrections and reversals for every condition, and factor contrasts are
provided in the supplement.

When B refers to A for its unit, Qwen3-4B whole-source masking yields only 1/30 and
0/30 complete answers under the original question and the two notes,
against controls of 18/30 and 15/30. The complete question raises these to
27/30 and 30/30, but the notebook cell retains three UNKNOWN answers.
Qwen3-8B has zero complete answers in every original-question condition for
this information condition: a zero contrast here is a floor, not preservation.
The complete question raises its whole-source results to 28/30 and 29/30,
against 30/30 controls, leaving two and one UNKNOWN answers.

Llama supplies a limiting result. Its original-question quantity answers
are unit-incomplete across both information conditions; complete instructions
substantially improve absolute counts but do not make them perfect.
When B states the complete quantity, its whole-source/control counts are 25/26 and 24/25 for
the two notes. Thus Qwen recovery is not a universal guarantee. The
aggregate factor contrasts keep the ten groups paired
(Table~\ref{tab:v17_factors}); floor and ceiling cells require the
absolute-count table for interpretation.

\input{context/v17_factor_table}

\paragraph{Endpoint completeness and abstention.}
The five Qwen abstention reversals in the experiment measuring both candidate scores and generated answers concern three endpoint cases,
not quantities. We therefore evaluate all 30 endpoint replacements from
ten groups with both questions, both notes, and all five conditions:
1,800 additional generations. The narrower mask covers the complete old
API endpoint string in A; the complete question requests all slashes and
path segments. The same two canonical endpoint strings are scored.

Under the original question and notebook note, all five previously
observed Qwen UNKNOWN answers become correct under the narrower mask.
The affected case IDs are test-016-negated\_old, test-036-negated\_old,
and test-046-negated\_old for Qwen3-4B, and the first and last for Qwen3-8B.
The visible B already contains the full approved endpoint in every case.
The complete synthetic sources, outputs, token spans, and unmodified
endpoint values are included in the experiment supplement.

\begin{figure}[!htbp]
  \centering
  \includegraphics[width=\linewidth]{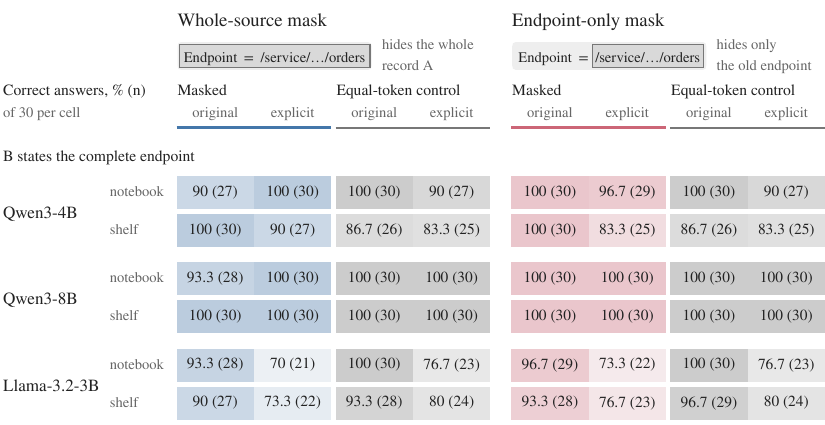}
  \caption{Endpoint factorial on all 30 renderings from ten groups, including
  the three cases with earlier Qwen abstention reversals. Each cell gives two percentages (counts), each over 30 cases:
  left for the original question, right when the question requests the complete endpoint.
  Tint deepens with the count over the observed range, 21 to 30, so a lighter right half marks
  a drop and a darker one a gain. Blue cells show whole-source masks, rose cells
  endpoint-span masks, and gray cells their matched equal-token controls; one row per note. The
  original five recoveries do not imply uniform improvement across the full matrix.}
  \label{fig:v17_endpoints}
\end{figure}

Across all endpoint question/note variants, the average advantage of
endpoint-span over whole-source masking, each relative to its own control, is
0.0 points for Qwen3-4B, 1.7 for Qwen3-8B, and 2.5 for Llama; the
corresponding intervals are $[0,0]$, $[0,4.2]$, and $[-4.2,10.8]$.
Qwen3-4B gains in some cells and loses in others, so selected recoveries
cannot stand in for the full matrix. The experiments show that the outcome depends on answer requirements,
source information, and mask scope. They do not distinguish
a unique neural pathway or the role of residual old-value information.
Policy benefit remains specific to the tested inputs and operations.

Appendix~\ref{app:v35_factorial} repeats these quantity and endpoint comparisons for five additional generators, including model and wording exceptions.

\subsection{Generation, Question Wording, and Source Controls}
\label{app:v35_generation}
The shared-history quantity experiment in Appendix~\ref{app:v33_models}
asks whether a mask preserves information needed by current and historical
questions. The synthetic-record, narrative, wording, and control comparisons
also evaluate Qwen3-32B, Qwen3.8-27B, Llama-3.1-70B, Gemma 4 26B-A4B,
and Gemma 4 31B. Within each task, these models use the same inputs and
interventions to test how answer preservation varies across checkpoints.

Each of these five models is evaluated on all five tasks, contributing
900 synthetic-record outputs, 348 OAKS outputs, 1,800 quantity/endpoint
factorial outputs, 3,360 control-factor outputs, and 480 matched-occurrence
outputs: 6,888 outputs per model. Qwen3-4B, Qwen3-8B, and Llama-3.2-3B
also contribute to the first three tasks. The control-factor and
matched-occurrence comparisons include Llama-3.2-3B, giving six models
for these tasks and eight for the others.

\paragraph{Scoring, budgets, and uncertainty.}
A complete correct response contains one JSON answer object, optionally
inside one clean Markdown fence. Quantities must give the correct number
and an accepted expression of the original unit, without conversion.
Surrounding prose, extra fields, and responses reaching the token limit
are counted separately from valid but incorrect answers. Original exact-field
scores remain separate: a clean fenced answer can be correct under this
rule while failing the original parser.
Qwen3-4B, Qwen3-8B, and Llama-3.2-3B use 40-token limits; the other five
generators use 256 tokens. Unless stated otherwise, the following tables use those
recorded budgets. Table~\ref{tab:v35_generation_budgets} separately compares
observable prefixes under a common limit, without treating a prefix as a new
latency measurement.

Intervals use 10,000 paired bootstrap draws with seed 20260919. Synthetic
and factorial variants remain together within their base group; OAKS
questions are resampled within each of the five fixed books. These are
pointwise intervals without adjustment for multiple comparisons.
$\dagger$ marks Qwen3.8: its masks block token-addressable full-attention
keys while retaining its prefilled convolution and recurrent states.
Every generator retains later states already computed from the earlier
source. The intervention changes source access, not all information
originating from that source.

\subsubsection{Candidate Preference and Generated Answers on Synthetic Records}
\label{app:v35_r570}
The experiment in Appendix~\ref{app:v7_bridge} pairs candidate scoring and
generation under full access, an earlier-source mask, and a matched
unrelated mask. Its 300 test conditions come from 50 base groups; 150
conditions replace the queried value. The replacement comparison asks
whether an increased candidate margin corresponds to a better complete
answer under the same prompt and cached history.

The margin increase is positive for every model as a point estimate, but
its interval crosses zero for Llama-3.1-70B. Generated-answer effects differ.
Qwen3-4B and Qwen3-8B lose eight and 17 correct answers relative to their
controls. Qwen3-32B loses one through an omitted unit. Qwen3.8 and both
Gemma models remain at 150/150. Llama-3.1-70B changes from 133 to 130,
with three corrections and six reversals; its accuracy interval includes
both signs. A positive margin change therefore accompanies losses, unchanged
answers, and unresolved accuracy differences across these models.

\begin{table}[!htbp]
\WideTableStyle

\caption{Synthetic replacement answers on 150 conditions from 50 groups, reported as percentages (counts). Source-mask changes are relative to the equal-token control. Brackets give paired 95\% intervals; margins use nats and accuracy changes use percentage points.}
\label{tab:v35_r570}
{\setlength{\tabcolsep}{\dimexpr\tabcolsep*10/12\relax}
\begin{tabular}{cccccc}
\toprule
\multirow{2}{*}[-0.6ex]{\TableLabel{Model}} & \multicolumn{3}{c}{\TableHead{Complete answers / 150}} & \multicolumn{2}{c}{\TableHead{Mask minus control [95\% CI]}} \\
\cmidrule(lr){2-4}\cmidrule(l){5-6}
 & \FullCell{\FullHeader{Full}} & \FullCell{\FullHeader{Control}} & \WholeCell{\WholeHeader{Mask}} & \TableHead{Margin (nats)} & \TableHead{Accuracy (pp)} \\\midrule
Qwen3-4B & \FullCell{\AnswerPct{100}{150}} & \FullCell{\AnswerPct{100}{150}} & \WholeCell{\AnswerPct{94.7}{142}} & +1.38 \TableSecondary{[1.16,\,1.62]} & $-$5.3 \TableSecondary{[$-$8.7,\,$-$2.0]}\\Qwen3-8B & \FullCell{\AnswerPct{96.7}{145}} & \FullCell{\AnswerPct{96.7}{145}} & \WholeCell{\AnswerPct{85.3}{128}} & +1.72 \TableSecondary{[1.41,\,2.05]} & $-$11.3 \TableSecondary{[$-$18.7,\,$-$4.7]}\\Qwen3-32B & \FullCell{\AnswerPct{100}{150}} & \FullCell{\AnswerPct{100}{150}} & \WholeCell{\AnswerPct{99.3}{149}} & +0.75 \TableSecondary{[0.51,\,1.01]} & $-$0.7 \TableSecondary{[$-$2.0,\,0.0]}\\Qwen3.8-27B$^{\dagger}$ & \FullCell{\AnswerPct{100}{150}} & \FullCell{\AnswerPct{100}{150}} & \WholeCell{\AnswerPct{100}{150}} & +0.66 \TableSecondary{[0.54,\,0.77]} & +0.0 \TableSecondary{[0.0,\,0.0]}\\\addlinespace[2pt]
Llama-3.2-3B & \FullCell{\AnswerPct{70}{105}} & \FullCell{\AnswerPct{70.7}{106}} & \WholeCell{\AnswerPct{71.3}{107}} & +0.31 \TableSecondary{[0.13,\,0.50]} & +0.7 \TableSecondary{[$-$2.0,\,4.0]}\\Llama-3.1-70B & \FullCell{\AnswerPct{91.3}{137}} & \FullCell{\AnswerPct{88.7}{133}} & \WholeCell{\AnswerPct{86.7}{130}} & +0.12 \TableSecondary{[$-$0.07,\,0.29]} & $-$2.0 \TableSecondary{[$-$6.7,\,2.0]}\\\addlinespace[2pt]
Gemma 4 26B-A4B & \FullCell{\AnswerPct{100}{150}} & \FullCell{\AnswerPct{100}{150}} & \WholeCell{\AnswerPct{100}{150}} & +1.00 \TableSecondary{[0.78,\,1.21]} & +0.0 \TableSecondary{[0.0,\,0.0]}\\Gemma 4 31B & \FullCell{\AnswerPct{100}{150}} & \FullCell{\AnswerPct{100}{150}} & \WholeCell{\AnswerPct{100}{150}} & +1.02 \TableSecondary{[0.90,\,1.16]} & +0.0 \TableSecondary{[0.0,\,0.0]}\\
\bottomrule
\end{tabular}}
\end{table}
\begin{figure}[!htbp]
\centering
\includegraphics[width=\linewidth]{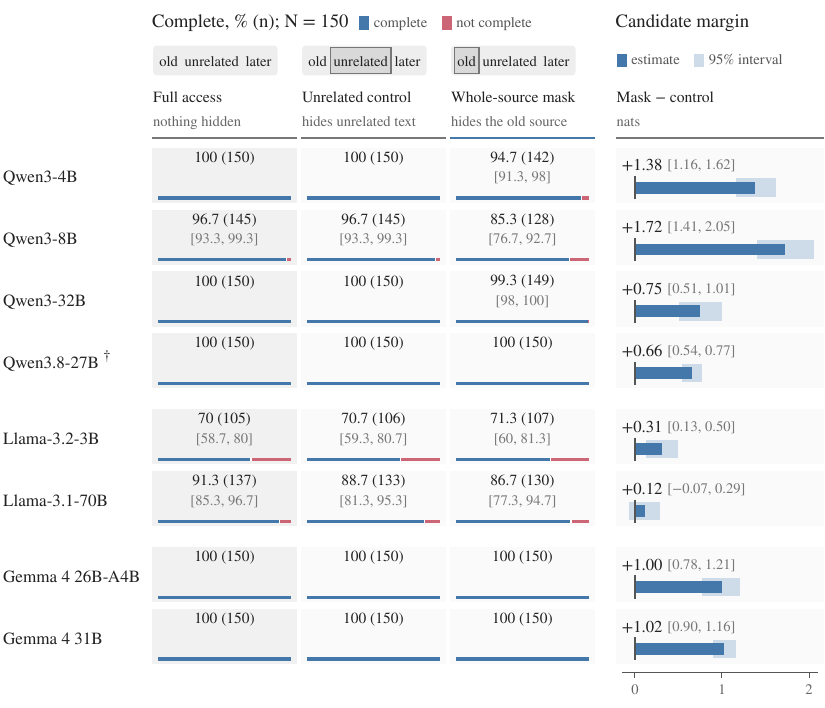}
\caption{Candidate preference and complete responses on 150 replacement conditions from 50 groups. The left block gives complete-answer percentages (counts) under full access, the equal-token unrelated control and the whole-source mask; chips show the hidden span. Each bar spans 100\% of 150 questions, blue for complete answers and rose for the rest; gray brackets give original 95\% group-bootstrap intervals, omitted when the interval equals the point estimate. The right block gives the whole-source-minus-control candidate margin in nats, drawn as a bar from zero inside a pale box that spans its interval. Complete scoring rejects capped responses. Qwen3-4B, Qwen3-8B, and Llama-3.2-3B use 40-token budgets; the other five models use 256 tokens. $\dagger$ Qwen3.8 retains recurrent states.}
\label{fig:v35_r570}
\end{figure}

The full 300-condition panel also includes updates to unrelated entities,
where A remains the only record answering the question. Qwen3.8 and both
Gemma models score 300/300 with full or control access but 250/300 after
masking A. Their 50 failures are UNKNOWN on these unrelated-update cases.
Perfect replacement-case performance therefore does not support masking
every earlier record. Llama-3.1-70B's replacement subset has 19 missing-unit
answers and one UNKNOWN after masking; its full-panel loss additionally
includes responses that fail the required format.
Table~\ref{tab:v35_r570_all} keeps all relations in the denominator.

\begin{table}[!htbp]
\WideTableStyle

\caption{Outcome entries are percentages (counts). All 300 synthetic test conditions from 50 groups. Corrections, reversals, and accuracy changes compare source masking with matched-control masking.}
\label{tab:v35_r570_all}
{\setlength{\tabcolsep}{\dimexpr\tabcolsep*12/14\relax}
\begin{tabular}{ccccccc}
\toprule
\multirow{2}{*}[-0.6ex]{\TableLabel{Model}} & \multicolumn{3}{c}{\TableHead{Complete answers}} & \multicolumn{2}{c}{\TableHead{Changed answers}} & \TableHead{Mask minus control} \\
\cmidrule(lr){2-4}\cmidrule(lr){5-6}\cmidrule(l){7-7}
 & \FullCell{\FullHeader{Full}} & \FullCell{\FullHeader{Control}} & \WholeCell{\WholeHeader{Mask}} & \TableHead{Corr.} & \TableHead{Rev.} & \TableHead{Accuracy [95\% CI]} \\\midrule
Qwen3-4B & \FullCell{\AnswerPct{100}{300}} & \FullCell{\AnswerPct{99.7}{299}} & \WholeCell{\AnswerPct{80.3}{241}} & \AnswerPct{0}{0} & \AnswerPct{19.3}{58} & $-$19.3 \TableSecondary{[$-$21.7,\,$-$17.3]}\\Qwen3-8B & \FullCell{\AnswerPct{98.3}{295}} & \FullCell{\AnswerPct{98.3}{295}} & \WholeCell{\AnswerPct{74}{222}} & \AnswerPct{0}{0} & \AnswerPct{24.3}{73} & $-$24.3 \TableSecondary{[$-$29.0,\,$-$20.0]}\\Qwen3-32B & \FullCell{\AnswerPct{98.7}{296}} & \FullCell{\AnswerPct{98.7}{296}} & \WholeCell{\AnswerPct{82.7}{248}} & \AnswerPct{0}{0} & \AnswerPct{16}{48} & $-$16.0 \TableSecondary{[$-$17.7,\,$-$14.3]}\\Qwen3.8-27B$^{\dagger}$ & \FullCell{\AnswerPct{100}{300}} & \FullCell{\AnswerPct{100}{300}} & \WholeCell{\AnswerPct{83.3}{250}} & \AnswerPct{0}{0} & \AnswerPct{16.7}{50} & $-$16.7 \TableSecondary{[$-$16.7,\,$-$16.7]}\\\addlinespace[2pt]
Llama-3.2-3B & \FullCell{\AnswerPct{63.7}{191}} & \FullCell{\AnswerPct{64}{192}} & \WholeCell{\AnswerPct{62.3}{187}} & \AnswerPct{1}{3} & \AnswerPct{2.7}{8} & $-$1.7 \TableSecondary{[$-$4.0,\,0.7]}\\Llama-3.1-70B & \FullCell{\AnswerPct{87.7}{263}} & \FullCell{\AnswerPct{87.7}{263}} & \WholeCell{\AnswerPct{75.7}{227}} & \AnswerPct{2.7}{8} & \AnswerPct{14.7}{44} & $-$12.0 \TableSecondary{[$-$16.3,\,$-$7.7]}\\\addlinespace[2pt]
Gemma 4 26B-A4B & \FullCell{\AnswerPct{100}{300}} & \FullCell{\AnswerPct{100}{300}} & \WholeCell{\AnswerPct{83.3}{250}} & \AnswerPct{0}{0} & \AnswerPct{16.7}{50} & $-$16.7 \TableSecondary{[$-$16.7,\,$-$16.7]}\\Gemma 4 31B & \FullCell{\AnswerPct{100}{300}} & \FullCell{\AnswerPct{100}{300}} & \WholeCell{\AnswerPct{83.3}{250}} & \AnswerPct{0}{0} & \AnswerPct{16.7}{50} & $-$16.7 \TableSecondary{[$-$16.7,\,$-$16.7]}\\
\bottomrule
\end{tabular}}
\end{table}

\subsubsection{Narrative Answer Changes and Completion Failures}
\label{app:v35_oaks}
The OAKS experiment uses one transition per question: 116 questions from
five novels, with supplied A/B candidates and an UNKNOWN option
(Appendix~\ref{app:v7_bridge}). This is distinct from the 336-event
candidate-scoring study. The question is whether source masking helps
select the answer appropriate after the later excerpt.

Every generator has a positive point estimate for complete-answer change
relative to its matched control. However, the two Gemma intervals cross
zero and Qwen3-32B's interval reaches zero. Qwen3.8 changes from 100 to 109
correct answers through nine corrections and no reversals. Llama-3.1-70B
changes from 103 to 110 through eight corrections and one reversal. Full
access remains visible because the control itself changes some answers.

\begin{table}[!htbp]
\WideTableStyle

\caption{Outcome entries are percentages (counts). OAKS: correct counts out of 116 questions from five fixed books. Both changes compare source masking with the equal-token control; brackets give paired 95\% intervals. Candidate margins are measured in nats; changes in answer accuracy are measured in percentage points.}
\label{tab:v35_oaks}
{\setlength{\tabcolsep}{\dimexpr\tabcolsep*10/12\relax}
\begin{tabular}{cccccc}
\toprule
\multirow{2}{*}[-0.6ex]{\TableLabel{Model}} & \multicolumn{3}{c}{\TableHead{Complete answers}} & \multicolumn{2}{c}{\TableHead{Mask minus control [95\% CI]}} \\
\cmidrule(lr){2-4}\cmidrule(l){5-6}
 & \FullCell{\FullHeader{Full}} & \FullCell{\FullHeader{Control}} & \WholeCell{\WholeHeader{Mask}} & \TableHead{Margin (nats)} & \TableHead{Accuracy (pp)} \\\midrule
Qwen3-4B & \FullCell{\AnswerPct{65.5}{76}} & \FullCell{\AnswerPct{65.5}{76}} & \WholeCell{\AnswerPct{87.9}{102}} & +9.49 \TableSecondary{[7.39,\,11.72]} & +22.4 \TableSecondary{[14.7,\,31.0]}\\Qwen3-8B & \FullCell{\AnswerPct{83.6}{97}} & \FullCell{\AnswerPct{82.8}{96}} & \WholeCell{\AnswerPct{87.9}{102}} & +3.77 \TableSecondary{[2.56,\,5.02]} & +5.2 \TableSecondary{[0.9,\,10.3]}\\Qwen3-32B & \FullCell{\AnswerPct{85.3}{99}} & \FullCell{\AnswerPct{84.5}{98}} & \WholeCell{\AnswerPct{89.7}{104}} & +4.40 \TableSecondary{[3.58,\,5.27]} & +5.2 \TableSecondary{[0.0,\,10.3]}\\Qwen3.8-27B$^{\dagger}$ & \FullCell{\AnswerPct{86.2}{100}} & \FullCell{\AnswerPct{86.2}{100}} & \WholeCell{\AnswerPct{94}{109}} & +1.02 \TableSecondary{[0.74,\,1.31]} & +7.8 \TableSecondary{[3.4,\,12.9]}\\\addlinespace[2pt]
Llama-3.2-3B & \FullCell{\AnswerPct{64.7}{75}} & \FullCell{\AnswerPct{67.2}{78}} & \WholeCell{\AnswerPct{76.7}{89}} & +1.17 \TableSecondary{[0.81,\,1.55]} & +9.5 \TableSecondary{[2.6,\,17.2]}\\Llama-3.1-70B & \FullCell{\AnswerPct{89.7}{104}} & \FullCell{\AnswerPct{88.8}{103}} & \WholeCell{\AnswerPct{94.8}{110}} & +0.67 \TableSecondary{[0.31,\,1.06]} & +6.0 \TableSecondary{[1.7,\,11.2]}\\\addlinespace[2pt]
Gemma 4 26B-A4B & \FullCell{\AnswerPct{19.8}{23}} & \FullCell{\AnswerPct{25}{29}} & \WholeCell{\AnswerPct{27.6}{32}} & +1.70 \TableSecondary{[0.99,\,2.45]} & +2.6 \TableSecondary{[$-$3.4,\,8.6]}\\Gemma 4 31B & \FullCell{\AnswerPct{85.3}{99}} & \FullCell{\AnswerPct{85.3}{99}} & \WholeCell{\AnswerPct{87.1}{101}} & +1.17 \TableSecondary{[0.42,\,2.00]} & +1.7 \TableSecondary{[$-$1.7,\,5.2]}\\
\bottomrule
\end{tabular}}
\end{table}
\begin{figure}[!htbp]
\centering
\includegraphics[width=\linewidth]{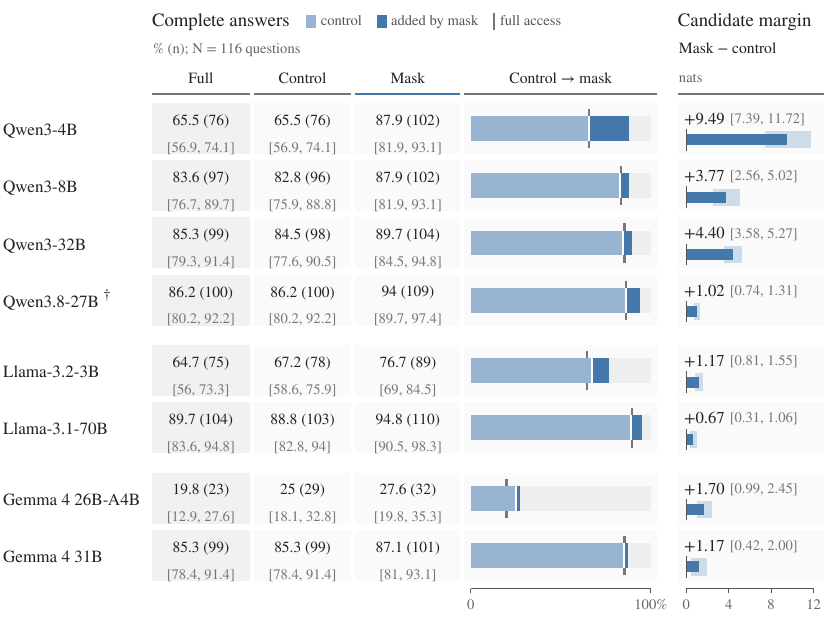}
\caption{Candidate preference and complete responses on 116 OAKS questions. Cells give correct-answer percentages (counts) with original 95\% question-bootstrap intervals in gray brackets. Bars run pale blue to the control count and blue to the mask count; ticks mark full access. The right column gives whole-source-minus-control candidate margins in nats with their original intervals, drawn as bars from zero over pale boxes. These are choices among supplied answers on excerpts. Complete scoring rejects capped responses; Qwen3-4B, Qwen3-8B, and Llama-3.2-3B use 40-token budgets, and the other five use 256 tokens. $\dagger$ Qwen3.8 retains recurrent states.}
\label{fig:v35_oaks}
\end{figure}

Gemma 4 26B-A4B requires a different interpretation from a simple
wrong-choice count. Under source masking, 32 responses are complete and
correct, one makes an incorrect choice, three abstain, 63 violate the
complete-response format, and 17 reach 256 tokens. A favorable candidate
margin thus coexists with limited compliance with this generation protocol.
Increasing the limit removes many truncations but produces no additional
complete correct responses. Gemma 4 31B instead has six UNKNOWN and nine
incorrect choices, with no format or length failure. These outcomes should
not be combined into a single claim about Gemma's narrative understanding.

The excerpts describe answer transitions, including narrative progression
that need not make an earlier statement historically false. The results
concern supplied candidates, not unrestricted answers or source localization.
Detector--generator and source-annotation analyses require their separate
joins and cannot be inferred from the generation table alone.

\begin{table}[!htbp]
\WideTableStyle

\caption{Source-mask outcomes as percentages (counts) of all 116 OAKS questions. Failure categories are mutually exclusive; corrections and reversals compare each masked answer with its matched-control answer.}
\label{tab:v35_oaks_failures}
{\setlength{\tabcolsep}{\dimexpr\tabcolsep*12/14\relax}
\begin{tabular}{ccccccc}
\toprule
\multirow{2}{*}[-0.6ex]{\TableLabel{Model}} & \multicolumn{4}{c}{\TableHead{Response categories}} & \multicolumn{2}{c}{\TableHead{Mask minus control}} \\
\cmidrule(lr){2-5}\cmidrule(l){6-7}
 & \TableHead{Wrong} & \TableHead{UNKNOWN} & \TableHead{Format} & \TableHead{Limit} & \TableHead{Corr.} & \TableHead{Rev.} \\\midrule
Qwen3-4B & \AnswerPct{11.2}{13} & \AnswerPct{0.9}{1} & \AnswerPct{0}{0} & \AnswerPct{0}{0} & \AnswerPct{24.1}{28} & \AnswerPct{1.7}{2}\\Qwen3-8B & \AnswerPct{4.3}{5} & \AnswerPct{7.8}{9} & \AnswerPct{0}{0} & \AnswerPct{0}{0} & \AnswerPct{6.9}{8} & \AnswerPct{1.7}{2}\\Qwen3-32B & \AnswerPct{4.3}{5} & \AnswerPct{6}{7} & \AnswerPct{0}{0} & \AnswerPct{0}{0} & \AnswerPct{6.9}{8} & \AnswerPct{1.7}{2}\\Qwen3.8-27B$^{\dagger}$ & \AnswerPct{6}{7} & \AnswerPct{0}{0} & \AnswerPct{0}{0} & \AnswerPct{0}{0} & \AnswerPct{7.8}{9} & \AnswerPct{0}{0}\\\addlinespace[2pt]
Llama-3.2-3B & \AnswerPct{22.4}{26} & \AnswerPct{0.9}{1} & \AnswerPct{0}{0} & \AnswerPct{0}{0} & \AnswerPct{13.8}{16} & \AnswerPct{4.3}{5}\\Llama-3.1-70B & \AnswerPct{5.2}{6} & \AnswerPct{0}{0} & \AnswerPct{0}{0} & \AnswerPct{0}{0} & \AnswerPct{6.9}{8} & \AnswerPct{0.9}{1}\\\addlinespace[2pt]
Gemma 4 26B-A4B & \AnswerPct{0.9}{1} & \AnswerPct{2.6}{3} & \AnswerPct{54.3}{63} & \AnswerPct{14.7}{17} & \AnswerPct{6.9}{8} & \AnswerPct{4.3}{5}\\Gemma 4 31B & \AnswerPct{7.8}{9} & \AnswerPct{5.2}{6} & \AnswerPct{0}{0} & \AnswerPct{0}{0} & \AnswerPct{2.6}{3} & \AnswerPct{0.9}{1}\\
\bottomrule
\end{tabular}}
\end{table}
\FloatBarrier

\subsubsection{Information, Question Wording, and Mask Scope}
\label{app:v35_factorial}
The factorial experiment extends Appendix~\ref{app:v17_scope} on its ten
original quantity groups and ten endpoint groups. Each displayed cell has
30 conditions from ten groups. Quantity questions cross two later-record
versions, two question wordings, and two note texts; endpoint questions
cross wording and note text. All five access conditions remain in each
cell. Repeated renderings of one group do not increase the number of
independent groups in the analysis.

Tables~\ref{tab:v35_quantity_explicit_original} and~\ref{tab:v35_quantity_context_dependent_original}
retain every quantity cell. Each source mask is compared with its own
equal-token control. The factor summaries compare these contrasts after
changing one factor, averaging the other fixed levels within each group.
They are not differences between arbitrarily pooled accuracy means.

\begin{table}[!htbp]
\WideTableStyle

\caption{Outcome entries are percentages (counts). Quantity answers when B states the complete quantity. Shared columns retain notebook and shelf controls; the two panels compare question wording. Every count is out of 30 conditions from ten base groups.}
\label{tab:v35_quantity_explicit_original}
\label{tab:v35_quantity_explicit_complete}
{\setlength{\tabcolsep}{1.8pt}
\begin{tabular}{ccccccccccc}
\toprule
\multirow{2}{*}[-0.6ex]{\TableLabel{Model}} & \multicolumn{5}{c}{\TableHead{Notebook condition ($N$)}} & \multicolumn{5}{c}{\TableHead{Shelf condition ($S$)}} \\
\cmidrule(lr){2-6}\cmidrule(l){7-11}
 & \FullCell{\FullHeader{Full}} & \FullCell{\FullHeader{\shortstack{Whole\\ctrl.}}} & \WholeCell{\WholeHeader{Whole}} & \FullCell{\FullHeader{\shortstack{Number\\ctrl.}}} & \NumberCell{\NumberHeader{Number}} & \FullCell{\FullHeader{Full}} & \FullCell{\FullHeader{\shortstack{Whole\\ctrl.}}} & \WholeCell{\WholeHeader{Whole}} & \FullCell{\FullHeader{\shortstack{Number\\ctrl.}}} & \NumberCell{\NumberHeader{Number}} \\\midrule
\TableGroupRow{11}{A. Original question}
Qwen3-4B & \FullCell{\AnswerPct{100}{30}} & \FullCell{\AnswerPct{100}{30}} & \WholeCell{\AnswerPct{83.3}{25}} & \FullCell{\AnswerPct{100}{30}} & \NumberCell{\AnswerPct{100}{30}} & \FullCell{\AnswerPct{100}{30}} & \FullCell{\AnswerPct{100}{30}} & \WholeCell{\AnswerPct{83.3}{25}} & \FullCell{\AnswerPct{100}{30}} & \NumberCell{\AnswerPct{100}{30}}\\Qwen3-8B & \FullCell{\AnswerPct{83.3}{25}} & \FullCell{\AnswerPct{83.3}{25}} & \WholeCell{\AnswerPct{33.3}{10}} & \FullCell{\AnswerPct{83.3}{25}} & \NumberCell{\AnswerPct{90}{27}} & \FullCell{\AnswerPct{83.3}{25}} & \FullCell{\AnswerPct{80}{24}} & \WholeCell{\AnswerPct{23.3}{7}} & \FullCell{\AnswerPct{83.3}{25}} & \NumberCell{\AnswerPct{86.7}{26}}\\Qwen3-32B & \FullCell{\AnswerPct{100}{30}} & \FullCell{\AnswerPct{100}{30}} & \WholeCell{\AnswerPct{96.7}{29}} & \FullCell{\AnswerPct{100}{30}} & \NumberCell{\AnswerPct{100}{30}} & \FullCell{\AnswerPct{100}{30}} & \FullCell{\AnswerPct{100}{30}} & \WholeCell{\AnswerPct{96.7}{29}} & \FullCell{\AnswerPct{100}{30}} & \NumberCell{\AnswerPct{100}{30}}\\Qwen3.8-27B$^{\dagger}$ & \FullCell{\AnswerPct{100}{30}} & \FullCell{\AnswerPct{100}{30}} & \WholeCell{\AnswerPct{100}{30}} & \FullCell{\AnswerPct{100}{30}} & \NumberCell{\AnswerPct{100}{30}} & \FullCell{\AnswerPct{100}{30}} & \FullCell{\AnswerPct{100}{30}} & \WholeCell{\AnswerPct{100}{30}} & \FullCell{\AnswerPct{100}{30}} & \NumberCell{\AnswerPct{100}{30}}\\\addlinespace[2pt]
Llama-3.2-3B & \FullCell{\AnswerPct{0}{0}} & \FullCell{\AnswerPct{0}{0}} & \WholeCell{\AnswerPct{0}{0}} & \FullCell{\AnswerPct{0}{0}} & \NumberCell{\AnswerPct{0}{0}} & \FullCell{\AnswerPct{0}{0}} & \FullCell{\AnswerPct{0}{0}} & \WholeCell{\AnswerPct{0}{0}} & \FullCell{\AnswerPct{0}{0}} & \NumberCell{\AnswerPct{0}{0}}\\Llama-3.1-70B & \FullCell{\AnswerPct{56.7}{17}} & \FullCell{\AnswerPct{43.3}{13}} & \WholeCell{\AnswerPct{36.7}{11}} & \FullCell{\AnswerPct{56.7}{17}} & \NumberCell{\AnswerPct{83.3}{25}} & \FullCell{\AnswerPct{73.3}{22}} & \FullCell{\AnswerPct{53.3}{16}} & \WholeCell{\AnswerPct{53.3}{16}} & \FullCell{\AnswerPct{73.3}{22}} & \NumberCell{\AnswerPct{83.3}{25}}\\\addlinespace[2pt]
Gemma 4 26B-A4B & \FullCell{\AnswerPct{100}{30}} & \FullCell{\AnswerPct{100}{30}} & \WholeCell{\AnswerPct{100}{30}} & \FullCell{\AnswerPct{100}{30}} & \NumberCell{\AnswerPct{100}{30}} & \FullCell{\AnswerPct{100}{30}} & \FullCell{\AnswerPct{100}{30}} & \WholeCell{\AnswerPct{100}{30}} & \FullCell{\AnswerPct{100}{30}} & \NumberCell{\AnswerPct{100}{30}}\\Gemma 4 31B & \FullCell{\AnswerPct{100}{30}} & \FullCell{\AnswerPct{100}{30}} & \WholeCell{\AnswerPct{100}{30}} & \FullCell{\AnswerPct{100}{30}} & \NumberCell{\AnswerPct{100}{30}} & \FullCell{\AnswerPct{100}{30}} & \FullCell{\AnswerPct{100}{30}} & \WholeCell{\AnswerPct{100}{30}} & \FullCell{\AnswerPct{100}{30}} & \NumberCell{\AnswerPct{100}{30}}\\\midrule
\TableGroupRow{11}{B. Explicit complete-answer request}
Qwen3-4B & \FullCell{\AnswerPct{100}{30}} & \FullCell{\AnswerPct{100}{30}} & \WholeCell{\AnswerPct{100}{30}} & \FullCell{\AnswerPct{100}{30}} & \NumberCell{\AnswerPct{100}{30}} & \FullCell{\AnswerPct{100}{30}} & \FullCell{\AnswerPct{100}{30}} & \WholeCell{\AnswerPct{100}{30}} & \FullCell{\AnswerPct{100}{30}} & \NumberCell{\AnswerPct{100}{30}}\\Qwen3-8B & \FullCell{\AnswerPct{100}{30}} & \FullCell{\AnswerPct{100}{30}} & \WholeCell{\AnswerPct{100}{30}} & \FullCell{\AnswerPct{100}{30}} & \NumberCell{\AnswerPct{100}{30}} & \FullCell{\AnswerPct{100}{30}} & \FullCell{\AnswerPct{100}{30}} & \WholeCell{\AnswerPct{100}{30}} & \FullCell{\AnswerPct{100}{30}} & \NumberCell{\AnswerPct{100}{30}}\\Qwen3-32B & \FullCell{\AnswerPct{100}{30}} & \FullCell{\AnswerPct{100}{30}} & \WholeCell{\AnswerPct{100}{30}} & \FullCell{\AnswerPct{100}{30}} & \NumberCell{\AnswerPct{100}{30}} & \FullCell{\AnswerPct{100}{30}} & \FullCell{\AnswerPct{100}{30}} & \WholeCell{\AnswerPct{100}{30}} & \FullCell{\AnswerPct{100}{30}} & \NumberCell{\AnswerPct{100}{30}}\\Qwen3.8-27B$^{\dagger}$ & \FullCell{\AnswerPct{100}{30}} & \FullCell{\AnswerPct{100}{30}} & \WholeCell{\AnswerPct{100}{30}} & \FullCell{\AnswerPct{100}{30}} & \NumberCell{\AnswerPct{100}{30}} & \FullCell{\AnswerPct{100}{30}} & \FullCell{\AnswerPct{100}{30}} & \WholeCell{\AnswerPct{100}{30}} & \FullCell{\AnswerPct{100}{30}} & \NumberCell{\AnswerPct{100}{30}}\\\addlinespace[2pt]
Llama-3.2-3B & \FullCell{\AnswerPct{86.7}{26}} & \FullCell{\AnswerPct{86.7}{26}} & \WholeCell{\AnswerPct{83.3}{25}} & \FullCell{\AnswerPct{86.7}{26}} & \NumberCell{\AnswerPct{86.7}{26}} & \FullCell{\AnswerPct{90}{27}} & \FullCell{\AnswerPct{83.3}{25}} & \WholeCell{\AnswerPct{80}{24}} & \FullCell{\AnswerPct{90}{27}} & \NumberCell{\AnswerPct{86.7}{26}}\\Llama-3.1-70B & \FullCell{\AnswerPct{100}{30}} & \FullCell{\AnswerPct{100}{30}} & \WholeCell{\AnswerPct{100}{30}} & \FullCell{\AnswerPct{100}{30}} & \NumberCell{\AnswerPct{100}{30}} & \FullCell{\AnswerPct{100}{30}} & \FullCell{\AnswerPct{100}{30}} & \WholeCell{\AnswerPct{100}{30}} & \FullCell{\AnswerPct{100}{30}} & \NumberCell{\AnswerPct{100}{30}}\\\addlinespace[2pt]
Gemma 4 26B-A4B & \FullCell{\AnswerPct{100}{30}} & \FullCell{\AnswerPct{100}{30}} & \WholeCell{\AnswerPct{100}{30}} & \FullCell{\AnswerPct{100}{30}} & \NumberCell{\AnswerPct{100}{30}} & \FullCell{\AnswerPct{100}{30}} & \FullCell{\AnswerPct{100}{30}} & \WholeCell{\AnswerPct{100}{30}} & \FullCell{\AnswerPct{100}{30}} & \NumberCell{\AnswerPct{100}{30}}\\Gemma 4 31B & \FullCell{\AnswerPct{100}{30}} & \FullCell{\AnswerPct{100}{30}} & \WholeCell{\AnswerPct{100}{30}} & \FullCell{\AnswerPct{100}{30}} & \NumberCell{\AnswerPct{100}{30}} & \FullCell{\AnswerPct{100}{30}} & \FullCell{\AnswerPct{100}{30}} & \WholeCell{\AnswerPct{100}{30}} & \FullCell{\AnswerPct{100}{30}} & \NumberCell{\AnswerPct{100}{30}}\\
\bottomrule
\end{tabular}}
\par\smallskip{\footnotesize\TableNoteAlign Full: full access; Whole/Number: whole-source/number-only masks; ctrl.: matched control. Notebook and shelf are the two note conditions. Counts use the recorded limits of 40 or 256 tokens; restricting the added models to 40 tokens leaves their correctness unchanged.\TableNoteEnd}
\end{table}

\begin{table}[!htbp]
\WideTableStyle

\caption{Outcome entries are percentages (counts). Quantity answers when B refers to A for its unit. Shared columns retain notebook and shelf controls; the two panels compare question wording. Every count is out of 30 conditions from ten base groups.}
\label{tab:v35_quantity_context_dependent_original}
\label{tab:v35_quantity_context_dependent_complete}
{\setlength{\tabcolsep}{1.8pt}
\begin{tabular}{ccccccccccc}
\toprule
\multirow{2}{*}[-0.6ex]{\TableLabel{Model}} & \multicolumn{5}{c}{\TableHead{Notebook condition ($N$)}} & \multicolumn{5}{c}{\TableHead{Shelf condition ($S$)}} \\
\cmidrule(lr){2-6}\cmidrule(l){7-11}
 & \FullCell{\FullHeader{Full}} & \FullCell{\FullHeader{\shortstack{Whole\\ctrl.}}} & \WholeCell{\WholeHeader{Whole}} & \FullCell{\FullHeader{\shortstack{Number\\ctrl.}}} & \NumberCell{\NumberHeader{Number}} & \FullCell{\FullHeader{Full}} & \FullCell{\FullHeader{\shortstack{Whole\\ctrl.}}} & \WholeCell{\WholeHeader{Whole}} & \FullCell{\FullHeader{\shortstack{Number\\ctrl.}}} & \NumberCell{\NumberHeader{Number}} \\\midrule
\TableGroupRow{11}{A. Original question}
Qwen3-4B & \FullCell{\AnswerPct{60}{18}} & \FullCell{\AnswerPct{60}{18}} & \WholeCell{\AnswerPct{3.3}{1}} & \FullCell{\AnswerPct{56.7}{17}} & \NumberCell{\AnswerPct{76.7}{23}} & \FullCell{\AnswerPct{43.3}{13}} & \FullCell{\AnswerPct{50}{15}} & \WholeCell{\AnswerPct{0}{0}} & \FullCell{\AnswerPct{43.3}{13}} & \NumberCell{\AnswerPct{66.7}{20}}\\Qwen3-8B & \FullCell{\AnswerPct{0}{0}} & \FullCell{\AnswerPct{0}{0}} & \WholeCell{\AnswerPct{0}{0}} & \FullCell{\AnswerPct{0}{0}} & \NumberCell{\AnswerPct{0}{0}} & \FullCell{\AnswerPct{0}{0}} & \FullCell{\AnswerPct{0}{0}} & \WholeCell{\AnswerPct{0}{0}} & \FullCell{\AnswerPct{0}{0}} & \NumberCell{\AnswerPct{0}{0}}\\Qwen3-32B & \FullCell{\AnswerPct{70}{21}} & \FullCell{\AnswerPct{66.7}{20}} & \WholeCell{\AnswerPct{0}{0}} & \FullCell{\AnswerPct{70}{21}} & \NumberCell{\AnswerPct{83.3}{25}} & \FullCell{\AnswerPct{60}{18}} & \FullCell{\AnswerPct{60}{18}} & \WholeCell{\AnswerPct{0}{0}} & \FullCell{\AnswerPct{63.3}{19}} & \NumberCell{\AnswerPct{76.7}{23}}\\Qwen3.8-27B$^{\dagger}$ & \FullCell{\AnswerPct{83.3}{25}} & \FullCell{\AnswerPct{83.3}{25}} & \WholeCell{\AnswerPct{63.3}{19}} & \FullCell{\AnswerPct{83.3}{25}} & \NumberCell{\AnswerPct{83.3}{25}} & \FullCell{\AnswerPct{80}{24}} & \FullCell{\AnswerPct{80}{24}} & \WholeCell{\AnswerPct{63.3}{19}} & \FullCell{\AnswerPct{80}{24}} & \NumberCell{\AnswerPct{86.7}{26}}\\\addlinespace[2pt]
Llama-3.2-3B & \FullCell{\AnswerPct{0}{0}} & \FullCell{\AnswerPct{0}{0}} & \WholeCell{\AnswerPct{0}{0}} & \FullCell{\AnswerPct{0}{0}} & \NumberCell{\AnswerPct{0}{0}} & \FullCell{\AnswerPct{0}{0}} & \FullCell{\AnswerPct{0}{0}} & \WholeCell{\AnswerPct{0}{0}} & \FullCell{\AnswerPct{0}{0}} & \NumberCell{\AnswerPct{0}{0}}\\Llama-3.1-70B & \FullCell{\AnswerPct{43.3}{13}} & \FullCell{\AnswerPct{36.7}{11}} & \WholeCell{\AnswerPct{0}{0}} & \FullCell{\AnswerPct{43.3}{13}} & \NumberCell{\AnswerPct{80}{24}} & \FullCell{\AnswerPct{46.7}{14}} & \FullCell{\AnswerPct{43.3}{13}} & \WholeCell{\AnswerPct{0}{0}} & \FullCell{\AnswerPct{50}{15}} & \NumberCell{\AnswerPct{76.7}{23}}\\\addlinespace[2pt]
Gemma 4 26B-A4B & \FullCell{\AnswerPct{96.7}{29}} & \FullCell{\AnswerPct{96.7}{29}} & \WholeCell{\AnswerPct{76.7}{23}} & \FullCell{\AnswerPct{96.7}{29}} & \NumberCell{\AnswerPct{100}{30}} & \FullCell{\AnswerPct{100}{30}} & \FullCell{\AnswerPct{96.7}{29}} & \WholeCell{\AnswerPct{73.3}{22}} & \FullCell{\AnswerPct{96.7}{29}} & \NumberCell{\AnswerPct{100}{30}}\\Gemma 4 31B & \FullCell{\AnswerPct{100}{30}} & \FullCell{\AnswerPct{100}{30}} & \WholeCell{\AnswerPct{100}{30}} & \FullCell{\AnswerPct{100}{30}} & \NumberCell{\AnswerPct{100}{30}} & \FullCell{\AnswerPct{100}{30}} & \FullCell{\AnswerPct{100}{30}} & \WholeCell{\AnswerPct{100}{30}} & \FullCell{\AnswerPct{100}{30}} & \NumberCell{\AnswerPct{100}{30}}\\\midrule
\TableGroupRow{11}{B. Explicit complete-answer request}
Qwen3-4B & \FullCell{\AnswerPct{93.3}{28}} & \FullCell{\AnswerPct{96.7}{29}} & \WholeCell{\AnswerPct{90}{27}} & \FullCell{\AnswerPct{96.7}{29}} & \NumberCell{\AnswerPct{96.7}{29}} & \FullCell{\AnswerPct{100}{30}} & \FullCell{\AnswerPct{100}{30}} & \WholeCell{\AnswerPct{100}{30}} & \FullCell{\AnswerPct{100}{30}} & \NumberCell{\AnswerPct{100}{30}}\\Qwen3-8B & \FullCell{\AnswerPct{100}{30}} & \FullCell{\AnswerPct{100}{30}} & \WholeCell{\AnswerPct{93.3}{28}} & \FullCell{\AnswerPct{100}{30}} & \NumberCell{\AnswerPct{100}{30}} & \FullCell{\AnswerPct{100}{30}} & \FullCell{\AnswerPct{100}{30}} & \WholeCell{\AnswerPct{96.7}{29}} & \FullCell{\AnswerPct{100}{30}} & \NumberCell{\AnswerPct{100}{30}}\\Qwen3-32B & \FullCell{\AnswerPct{100}{30}} & \FullCell{\AnswerPct{100}{30}} & \WholeCell{\AnswerPct{80}{24}} & \FullCell{\AnswerPct{100}{30}} & \NumberCell{\AnswerPct{100}{30}} & \FullCell{\AnswerPct{100}{30}} & \FullCell{\AnswerPct{100}{30}} & \WholeCell{\AnswerPct{80}{24}} & \FullCell{\AnswerPct{100}{30}} & \NumberCell{\AnswerPct{100}{30}}\\Qwen3.8-27B$^{\dagger}$ & \FullCell{\AnswerPct{100}{30}} & \FullCell{\AnswerPct{100}{30}} & \WholeCell{\AnswerPct{100}{30}} & \FullCell{\AnswerPct{100}{30}} & \NumberCell{\AnswerPct{100}{30}} & \FullCell{\AnswerPct{100}{30}} & \FullCell{\AnswerPct{100}{30}} & \WholeCell{\AnswerPct{100}{30}} & \FullCell{\AnswerPct{100}{30}} & \NumberCell{\AnswerPct{100}{30}}\\\addlinespace[2pt]
Llama-3.2-3B & \FullCell{\AnswerPct{73.3}{22}} & \FullCell{\AnswerPct{73.3}{22}} & \WholeCell{\AnswerPct{66.7}{20}} & \FullCell{\AnswerPct{73.3}{22}} & \NumberCell{\AnswerPct{73.3}{22}} & \FullCell{\AnswerPct{73.3}{22}} & \FullCell{\AnswerPct{73.3}{22}} & \WholeCell{\AnswerPct{70}{21}} & \FullCell{\AnswerPct{73.3}{22}} & \NumberCell{\AnswerPct{76.7}{23}}\\Llama-3.1-70B & \FullCell{\AnswerPct{100}{30}} & \FullCell{\AnswerPct{100}{30}} & \WholeCell{\AnswerPct{100}{30}} & \FullCell{\AnswerPct{100}{30}} & \NumberCell{\AnswerPct{100}{30}} & \FullCell{\AnswerPct{100}{30}} & \FullCell{\AnswerPct{100}{30}} & \WholeCell{\AnswerPct{100}{30}} & \FullCell{\AnswerPct{100}{30}} & \NumberCell{\AnswerPct{100}{30}}\\\addlinespace[2pt]
Gemma 4 26B-A4B & \FullCell{\AnswerPct{100}{30}} & \FullCell{\AnswerPct{100}{30}} & \WholeCell{\AnswerPct{100}{30}} & \FullCell{\AnswerPct{100}{30}} & \NumberCell{\AnswerPct{100}{30}} & \FullCell{\AnswerPct{96.7}{29}} & \FullCell{\AnswerPct{93.3}{28}} & \WholeCell{\AnswerPct{100}{30}} & \FullCell{\AnswerPct{93.3}{28}} & \NumberCell{\AnswerPct{96.7}{29}}\\Gemma 4 31B & \FullCell{\AnswerPct{100}{30}} & \FullCell{\AnswerPct{100}{30}} & \WholeCell{\AnswerPct{100}{30}} & \FullCell{\AnswerPct{100}{30}} & \NumberCell{\AnswerPct{100}{30}} & \FullCell{\AnswerPct{100}{30}} & \FullCell{\AnswerPct{100}{30}} & \WholeCell{\AnswerPct{100}{30}} & \FullCell{\AnswerPct{100}{30}} & \NumberCell{\AnswerPct{100}{30}}\\
\bottomrule
\end{tabular}}
\par\smallskip{\footnotesize\TableNoteAlign Full: full access; Whole/Number: whole-source/number-only masks; ctrl.: matched control. Notebook and shelf are the two note conditions. Counts use the recorded limits of 40 or 256 tokens; restricting the added models to 40 tokens leaves their correctness unchanged.\TableNoteEnd}
\end{table}

When B includes the unit and the question explicitly requests it, every
added model scores 30/30 in every access condition and both note texts.
With the original question and an earlier-unit dependency, performance is
less uniform. In the notebook condition, Qwen3-32B changes from 21/30 under
full access to 0/30 after whole-source masking; its number-only mask scores
25/30. Qwen3.8 changes from 25 to 19 under the whole mask and remains at
25 under the number-only mask. The explicit request restores Qwen3.8 to
30/30 across arms, but Qwen3-32B's whole mask still scores 24/30 against
30/30 for its control.

Llama-3.1-70B likewise scores 0/30 under whole-source masking with the
original dependent question, then 30/30 with the explicit request.
Gemma 4 31B is correct in every factorial quantity cell. Its failures in
the separate six-unit experiment therefore cannot be generalized to these
other prompts and values. These comparisons isolate input conditions;
they do not establish a fixed ranking of model robustness across these
different tasks and prompts.

\begin{figure}[!htbp]
\centering
\includegraphics[width=\linewidth]{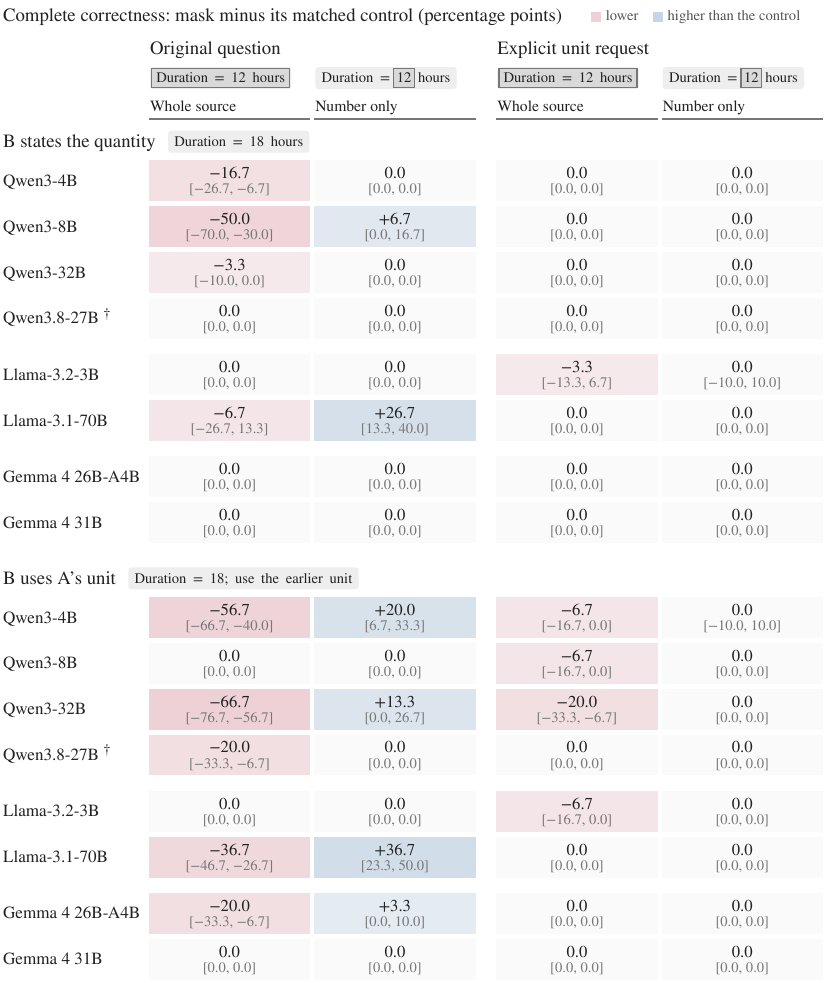}
\caption{Quantity factorial with the notebook note. Row blocks fix what B says, column blocks the question wording. Cells give the complete-correctness change, mask minus matched control, and its original 95\% bootstrap interval (ten groups, 30 repeated conditions per comparison). Tint hue is the sign and depth the size, on one scale for all cells. Chips sketch B, and A with its masked span boxed. Complete-response scoring and recorded budgets are unchanged. $\dagger$ Qwen3.8 retains recurrent states.}
\label{fig:v35_quantity}
\end{figure}
\begin{table}[!htbp]
\WideTableStyle

\caption{Changes in the quantity mask-minus-control contrast (pp), with paired 95\% intervals over ten groups. Each column changes one factor, averaging the remaining fixed factor levels within group.}
\label{tab:v35_quantity_factors}
{\setlength{\tabcolsep}{\dimexpr\tabcolsep*6/8\relax}
\begin{tabular}{cccc}
\toprule
\TableHead{Model} & \TableHead{Number minus whole} & \TableHead{Explicit request} & \TableHead{Earlier unit needed} \\
\midrule
Qwen3-4B & +23.8 \TableSecondary{[20.4,\,27.1]} & +11.2 \TableSecondary{[6.2,\,15.8]} & $-$4.6 \TableSecondary{[$-$10.8,\,2.1]} \\
Qwen3-8B & +15.8 \TableSecondary{[12.5,\,19.2]} & +10.8 \TableSecondary{[4.2,\,16.7]} & +10.8 \TableSecondary{[4.2,\,16.7]} \\
Qwen3-32B & +25.0 \TableSecondary{[23.3,\,27.1]} & +8.3 \TableSecondary{[4.2,\,12.5]} & $-$16.7 \TableSecondary{[$-$23.8,\,$-$9.2]} \\
Qwen3.8-27B$^{\dagger}$ & +5.4 \TableSecondary{[2.5,\,8.8]} & +3.8 \TableSecondary{[0.8,\,6.7]} & $-$3.8 \TableSecondary{[$-$6.7,\,$-$0.8]} \\
\addlinespace[2pt]
Llama-3.2-3B & +2.1 \TableSecondary{[0.0,\,4.6]} & $-$2.1 \TableSecondary{[$-$6.2,\,2.5]} & +0.4 \TableSecondary{[$-$1.7,\,2.5]} \\
Llama-3.1-70B & +23.3 \TableSecondary{[20.4,\,25.8]} & $-$1.7 \TableSecondary{[$-$10.0,\,6.7]} & $-$5.8 \TableSecondary{[$-$10.0,\,$-$2.1]} \\
\addlinespace[2pt]
Gemma 4 26B-A4B & +5.8 \TableSecondary{[1.7,\,10.4]} & +5.8 \TableSecondary{[2.5,\,9.6]} & $-$3.3 \TableSecondary{[$-$8.8,\,2.9]} \\
Gemma 4 31B & +0.0 \TableSecondary{[0.0,\,0.0]} & +0.0 \TableSecondary{[0.0,\,0.0]} & +0.0 \TableSecondary{[0.0,\,0.0]} \\
\bottomrule
\end{tabular}}
\end{table}

The number-only versus whole-source contrast is positive for every model
except Gemma 4 31B, whose cells are all correct. Its size varies: it is
smaller for Qwen3.8 and Gemma 4 26B-A4B than for Qwen3-32B and Llama-3.1-70B.
The explicit-question effect is also model dependent. Llama-3.1-70B's
aggregate interval crosses zero despite recovery in the particular notebook
cell above. The complete factorial prevents that example from being
mistaken for a uniform effect.

\paragraph{Endpoint strings require a separate interpretation.}
Here the narrow mask covers the complete old endpoint string, not a number.
Qwen3-32B, Qwen3.8, and both Gemma models score 30/30 in every endpoint cell.
Llama-3.1-70B has two isolated misses: 29/30 with whole-source masking under
the original notebook question, and 29/30 with endpoint-only masking under
the original shelf question. Its aggregate effect for changing mask scope
is therefore zero, with uncertainty on either side. There is no uniform
endpoint advantage for the narrower mask.

\begin{table}[!htbp]
\WideTableStyle

\caption{Endpoint answers under original and explicit requests. Each cell gives a percentage (count) over 30 conditions from ten groups. Both panels retain the notebook and shelf conditions and every matched control.}
\label{tab:v35_endpoint_original}
\label{tab:v35_endpoint_complete}
\setlength{\tabcolsep}{1.2pt}
{\setlength{\tabcolsep}{1.2pt}
\begin{tabular}{ccccccccccc}
\toprule
\multirow{2}{*}[-0.6ex]{\TableLabel{Model}} & \multicolumn{5}{c}{\TableHead{Notebook condition ($N$)}} & \multicolumn{5}{c}{\TableHead{Shelf condition ($S$)}} \\
\cmidrule(lr){2-6}\cmidrule(l){7-11}
 & \FullCell{\FullHeader{Full}} & \FullCell{\FullHeader{\shortstack{Whole\\ctrl.}}} & \WholeCell{\WholeHeader{Whole}} & \FullCell{\FullHeader{\shortstack{Endpoint\\ctrl.}}} & \NumberCell{\NumberHeader{Endpoint}} & \FullCell{\FullHeader{Full}} & \FullCell{\FullHeader{\shortstack{Whole\\ctrl.}}} & \WholeCell{\WholeHeader{Whole}} & \FullCell{\FullHeader{\shortstack{Endpoint\\ctrl.}}} & \NumberCell{\NumberHeader{Endpoint}} \\\midrule
\TableGroupRow{11}{A. Original question}
Qwen3-4B & \FullCell{\AnswerPct{100}{30}} & \FullCell{\AnswerPct{100}{30}} & \WholeCell{\AnswerPct{90}{27}} & \FullCell{\AnswerPct{100}{30}} & \NumberCell{\AnswerPct{100}{30}} & \FullCell{\AnswerPct{86.7}{26}} & \FullCell{\AnswerPct{86.7}{26}} & \WholeCell{\AnswerPct{100}{30}} & \FullCell{\AnswerPct{86.7}{26}} & \NumberCell{\AnswerPct{100}{30}}\\Qwen3-8B & \FullCell{\AnswerPct{100}{30}} & \FullCell{\AnswerPct{100}{30}} & \WholeCell{\AnswerPct{93.3}{28}} & \FullCell{\AnswerPct{100}{30}} & \NumberCell{\AnswerPct{100}{30}} & \FullCell{\AnswerPct{100}{30}} & \FullCell{\AnswerPct{100}{30}} & \WholeCell{\AnswerPct{100}{30}} & \FullCell{\AnswerPct{100}{30}} & \NumberCell{\AnswerPct{100}{30}}\\Qwen3-32B & \FullCell{\AnswerPct{100}{30}} & \FullCell{\AnswerPct{100}{30}} & \WholeCell{\AnswerPct{100}{30}} & \FullCell{\AnswerPct{100}{30}} & \NumberCell{\AnswerPct{100}{30}} & \FullCell{\AnswerPct{100}{30}} & \FullCell{\AnswerPct{100}{30}} & \WholeCell{\AnswerPct{100}{30}} & \FullCell{\AnswerPct{100}{30}} & \NumberCell{\AnswerPct{100}{30}}\\Qwen3.8-27B$^{\dagger}$ & \FullCell{\AnswerPct{100}{30}} & \FullCell{\AnswerPct{100}{30}} & \WholeCell{\AnswerPct{100}{30}} & \FullCell{\AnswerPct{100}{30}} & \NumberCell{\AnswerPct{100}{30}} & \FullCell{\AnswerPct{100}{30}} & \FullCell{\AnswerPct{100}{30}} & \WholeCell{\AnswerPct{100}{30}} & \FullCell{\AnswerPct{100}{30}} & \NumberCell{\AnswerPct{100}{30}}\\\addlinespace[2pt]
Llama-3.2-3B & \FullCell{\AnswerPct{100}{30}} & \FullCell{\AnswerPct{100}{30}} & \WholeCell{\AnswerPct{93.3}{28}} & \FullCell{\AnswerPct{100}{30}} & \NumberCell{\AnswerPct{96.7}{29}} & \FullCell{\AnswerPct{96.7}{29}} & \FullCell{\AnswerPct{93.3}{28}} & \WholeCell{\AnswerPct{90}{27}} & \FullCell{\AnswerPct{96.7}{29}} & \NumberCell{\AnswerPct{93.3}{28}}\\Llama-3.1-70B & \FullCell{\AnswerPct{100}{30}} & \FullCell{\AnswerPct{100}{30}} & \WholeCell{\AnswerPct{96.7}{29}} & \FullCell{\AnswerPct{100}{30}} & \NumberCell{\AnswerPct{100}{30}} & \FullCell{\AnswerPct{100}{30}} & \FullCell{\AnswerPct{100}{30}} & \WholeCell{\AnswerPct{100}{30}} & \FullCell{\AnswerPct{100}{30}} & \NumberCell{\AnswerPct{96.7}{29}}\\\addlinespace[2pt]
Gemma 4 26B-A4B & \FullCell{\AnswerPct{100}{30}} & \FullCell{\AnswerPct{100}{30}} & \WholeCell{\AnswerPct{100}{30}} & \FullCell{\AnswerPct{100}{30}} & \NumberCell{\AnswerPct{100}{30}} & \FullCell{\AnswerPct{100}{30}} & \FullCell{\AnswerPct{100}{30}} & \WholeCell{\AnswerPct{100}{30}} & \FullCell{\AnswerPct{100}{30}} & \NumberCell{\AnswerPct{100}{30}}\\Gemma 4 31B & \FullCell{\AnswerPct{100}{30}} & \FullCell{\AnswerPct{100}{30}} & \WholeCell{\AnswerPct{100}{30}} & \FullCell{\AnswerPct{100}{30}} & \NumberCell{\AnswerPct{100}{30}} & \FullCell{\AnswerPct{100}{30}} & \FullCell{\AnswerPct{100}{30}} & \WholeCell{\AnswerPct{100}{30}} & \FullCell{\AnswerPct{100}{30}} & \NumberCell{\AnswerPct{100}{30}}\\\midrule
\TableGroupRow{11}{B. Explicit complete-answer request}
Qwen3-4B & \FullCell{\AnswerPct{90}{27}} & \FullCell{\AnswerPct{90}{27}} & \WholeCell{\AnswerPct{100}{30}} & \FullCell{\AnswerPct{90}{27}} & \NumberCell{\AnswerPct{96.7}{29}} & \FullCell{\AnswerPct{83.3}{25}} & \FullCell{\AnswerPct{83.3}{25}} & \WholeCell{\AnswerPct{90}{27}} & \FullCell{\AnswerPct{83.3}{25}} & \NumberCell{\AnswerPct{83.3}{25}}\\Qwen3-8B & \FullCell{\AnswerPct{100}{30}} & \FullCell{\AnswerPct{100}{30}} & \WholeCell{\AnswerPct{100}{30}} & \FullCell{\AnswerPct{100}{30}} & \NumberCell{\AnswerPct{100}{30}} & \FullCell{\AnswerPct{100}{30}} & \FullCell{\AnswerPct{100}{30}} & \WholeCell{\AnswerPct{100}{30}} & \FullCell{\AnswerPct{100}{30}} & \NumberCell{\AnswerPct{100}{30}}\\Qwen3-32B & \FullCell{\AnswerPct{100}{30}} & \FullCell{\AnswerPct{100}{30}} & \WholeCell{\AnswerPct{100}{30}} & \FullCell{\AnswerPct{100}{30}} & \NumberCell{\AnswerPct{100}{30}} & \FullCell{\AnswerPct{100}{30}} & \FullCell{\AnswerPct{100}{30}} & \WholeCell{\AnswerPct{100}{30}} & \FullCell{\AnswerPct{100}{30}} & \NumberCell{\AnswerPct{100}{30}}\\Qwen3.8-27B$^{\dagger}$ & \FullCell{\AnswerPct{100}{30}} & \FullCell{\AnswerPct{100}{30}} & \WholeCell{\AnswerPct{100}{30}} & \FullCell{\AnswerPct{100}{30}} & \NumberCell{\AnswerPct{100}{30}} & \FullCell{\AnswerPct{100}{30}} & \FullCell{\AnswerPct{100}{30}} & \WholeCell{\AnswerPct{100}{30}} & \FullCell{\AnswerPct{100}{30}} & \NumberCell{\AnswerPct{100}{30}}\\\addlinespace[2pt]
Llama-3.2-3B & \FullCell{\AnswerPct{76.7}{23}} & \FullCell{\AnswerPct{76.7}{23}} & \WholeCell{\AnswerPct{70}{21}} & \FullCell{\AnswerPct{76.7}{23}} & \NumberCell{\AnswerPct{73.3}{22}} & \FullCell{\AnswerPct{83.3}{25}} & \FullCell{\AnswerPct{80}{24}} & \WholeCell{\AnswerPct{73.3}{22}} & \FullCell{\AnswerPct{80}{24}} & \NumberCell{\AnswerPct{76.7}{23}}\\Llama-3.1-70B & \FullCell{\AnswerPct{100}{30}} & \FullCell{\AnswerPct{100}{30}} & \WholeCell{\AnswerPct{100}{30}} & \FullCell{\AnswerPct{100}{30}} & \NumberCell{\AnswerPct{100}{30}} & \FullCell{\AnswerPct{100}{30}} & \FullCell{\AnswerPct{100}{30}} & \WholeCell{\AnswerPct{100}{30}} & \FullCell{\AnswerPct{100}{30}} & \NumberCell{\AnswerPct{100}{30}}\\\addlinespace[2pt]
Gemma 4 26B-A4B & \FullCell{\AnswerPct{100}{30}} & \FullCell{\AnswerPct{100}{30}} & \WholeCell{\AnswerPct{100}{30}} & \FullCell{\AnswerPct{100}{30}} & \NumberCell{\AnswerPct{100}{30}} & \FullCell{\AnswerPct{100}{30}} & \FullCell{\AnswerPct{100}{30}} & \WholeCell{\AnswerPct{100}{30}} & \FullCell{\AnswerPct{100}{30}} & \NumberCell{\AnswerPct{100}{30}}\\Gemma 4 31B & \FullCell{\AnswerPct{100}{30}} & \FullCell{\AnswerPct{100}{30}} & \WholeCell{\AnswerPct{100}{30}} & \FullCell{\AnswerPct{100}{30}} & \NumberCell{\AnswerPct{100}{30}} & \FullCell{\AnswerPct{100}{30}} & \FullCell{\AnswerPct{100}{30}} & \WholeCell{\AnswerPct{100}{30}} & \FullCell{\AnswerPct{100}{30}} & \NumberCell{\AnswerPct{100}{30}}\\
\bottomrule
\end{tabular}}
\par\smallskip{\footnotesize\TableNoteAlign Full: full access; Whole/Endpoint: whole-source/old-endpoint masks; ctrl.: matched control. Restricting the added models to the common 40-token limit leaves their correctness unchanged.\TableNoteEnd}
\end{table}

\begin{table}[!htbp]
\WideTableStyle

\caption{Changes in the endpoint mask-minus-control contrast (pp), with paired 95\% intervals over ten groups. Each column changes one factor, averaging the remaining fixed factor levels within group.}
\label{tab:v35_endpoint_factors}
{\setlength{\tabcolsep}{\dimexpr\tabcolsep*6/8\relax}
\begin{tabular}{cccc}
\toprule
\TableHead{Model} & \TableHead{Endpoint minus whole} & \TableHead{Explicit request} & \TableHead{Shelf minus notebook} \\
\midrule
Qwen3-4B & +0.0 \TableSecondary{[0.0,\,0.0]} & +1.7 \TableSecondary{[$-$3.3,\,6.7]} & +6.7 \TableSecondary{[1.7,\,11.7]} \\
Qwen3-8B & +1.7 \TableSecondary{[0.0,\,4.2]} & +1.7 \TableSecondary{[0.0,\,4.2]} & +1.7 \TableSecondary{[0.0,\,4.2]} \\
Qwen3-32B & +0.0 \TableSecondary{[0.0,\,0.0]} & +0.0 \TableSecondary{[0.0,\,0.0]} & +0.0 \TableSecondary{[0.0,\,0.0]} \\
Qwen3.8-27B$^{\dagger}$ & +0.0 \TableSecondary{[0.0,\,0.0]} & +0.0 \TableSecondary{[0.0,\,0.0]} & +0.0 \TableSecondary{[0.0,\,0.0]} \\
\addlinespace[2pt]
Llama-3.2-3B & +2.5 \TableSecondary{[$-$4.2,\,10.0]} & $-$0.8 \TableSecondary{[$-$5.0,\,4.2]} & +0.8 \TableSecondary{[$-$2.5,\,4.2]} \\
Llama-3.1-70B & +0.0 \TableSecondary{[$-$2.5,\,2.5]} & +1.7 \TableSecondary{[0.0,\,4.2]} & +0.0 \TableSecondary{[$-$2.5,\,2.5]} \\
\addlinespace[2pt]
Gemma 4 26B-A4B & +0.0 \TableSecondary{[0.0,\,0.0]} & +0.0 \TableSecondary{[0.0,\,0.0]} & +0.0 \TableSecondary{[0.0,\,0.0]} \\
Gemma 4 31B & +0.0 \TableSecondary{[0.0,\,0.0]} & +0.0 \TableSecondary{[0.0,\,0.0]} & +0.0 \TableSecondary{[0.0,\,0.0]} \\
\bottomrule
\end{tabular}}
\end{table}

\subsubsection{Control-Span Effects and Their Limits Across Models}
\label{app:v35_controls}
The control studies extend Appendix~\ref{app:v24_control_factors} on 240
information conditions from 120 groups. They ask whether masking unrelated
note tokens or changing how the question identifies its record changes
answers about an inspection label. Table~\ref{tab:v35_control_all} retains
all 14 configurations. Only Llama-3.2-3B contributes an original run here;
the other five models were evaluated on all configurations.

The large location effect does not replicate as an accuracy loss in the
added models. With the original question, each gives 240/240 correct labels
under full access and every tested note/source mask. The two identical
31-token occurrences also give 240/240 for every addition, compared with
eight and 146 for the original Llama-3.2-3B. That model's difference is
57.5 points [49.6, 65.0] over the same 120 paired groups. This limits the
use of that control for that model and input; it does not establish a
universal positional failure.

Question changes need their own full-access baseline. Explicitly naming
the inspection record raises the original Llama model to 240/240 but
reduces Qwen3-32B's full-access count from 240 to 220. Its early-note mask
still scores 240. Those twenty responses state the correct label in a full sentence rather
than returning the requested bare label. Their lower score is a format
failure, not a loss of the fact. Thus a wording change can alter compliance
even when the original control preserves the requested answer.

\begin{table}[!htbp]
\WideTableStyle

\caption{Outcome entries are percentages (counts). All fourteen control-factor configurations: correct unrelated-label answers out of 240 conditions sharing 120 groups. No response reaches the token limit.}
\label{tab:v35_control_all}
{\setlength{\tabcolsep}{\dimexpr\tabcolsep*12/14\relax}
\begin{tabular}{ccccccc}
\toprule
\TableHead{Configuration} & \TableHead{Q3-32} & \TableHead{Q3.8} & \TableHead{L3-3} & \TableHead{L3.1-70} & \TableHead{G4-26} & \TableHead{G4-31} \\\midrule
Original: full & \AnswerPct{100}{240} & \AnswerPct{100}{240} & \AnswerPct{60.8}{146} & \AnswerPct{100}{240} & \AnswerPct{100}{240} & \AnswerPct{100}{240}\\Original: early note & \AnswerPct{100}{240} & \AnswerPct{100}{240} & \AnswerPct{3.3}{8} & \AnswerPct{100}{240} & \AnswerPct{100}{240} & \AnswerPct{100}{240}\\Original: middle note & \AnswerPct{100}{240} & \AnswerPct{100}{240} & \AnswerPct{67.9}{163} & \AnswerPct{100}{240} & \AnswerPct{100}{240} & \AnswerPct{100}{240}\\Original: late note & \AnswerPct{100}{240} & \AnswerPct{100}{240} & \AnswerPct{62.1}{149} & \AnswerPct{100}{240} & \AnswerPct{100}{240} & \AnswerPct{100}{240}\\Through sentence end & \AnswerPct{100}{240} & \AnswerPct{100}{240} & \AnswerPct{3.3}{8} & \AnswerPct{100}{240} & \AnswerPct{100}{240} & \AnswerPct{100}{240}\\All note & \AnswerPct{100}{240} & \AnswerPct{100}{240} & \AnswerPct{16.3}{39} & \AnswerPct{100}{240} & \AnswerPct{100}{240} & \AnswerPct{100}{240}\\Whole source A & \AnswerPct{100}{240} & \AnswerPct{100}{240} & \AnswerPct{82.9}{199} & \AnswerPct{100}{240} & \AnswerPct{100}{240} & \AnswerPct{100}{240}\\Gray/plain: full & \AnswerPct{100}{240} & \AnswerPct{100}{240} & \AnswerPct{57.1}{137} & \AnswerPct{100}{240} & \AnswerPct{100}{240} & \AnswerPct{100}{240}\\Gray/plain: early & \AnswerPct{100}{240} & \AnswerPct{100}{240} & \AnswerPct{2.1}{5} & \AnswerPct{100}{240} & \AnswerPct{100}{240} & \AnswerPct{100}{240}\\Explicit question: full & \AnswerPct{91.7}{220} & \AnswerPct{100}{240} & \AnswerPct{100}{240} & \AnswerPct{100}{240} & \AnswerPct{100}{240} & \AnswerPct{100}{240}\\Explicit question: early & \AnswerPct{100}{240} & \AnswerPct{100}{240} & \AnswerPct{100}{240} & \AnswerPct{100}{240} & \AnswerPct{100}{240} & \AnswerPct{100}{240}\\No UNKNOWN clause: full & \AnswerPct{100}{240} & \AnswerPct{100}{240} & \AnswerPct{100}{240} & \AnswerPct{100}{240} & \AnswerPct{100}{240} & \AnswerPct{100}{240}\\No UNKNOWN clause: early & \AnswerPct{100}{240} & \AnswerPct{100}{240} & \AnswerPct{100}{240} & \AnswerPct{100}{240} & \AnswerPct{100}{240} & \AnswerPct{100}{240}\\No note: full & \AnswerPct{100}{240} & \AnswerPct{100}{240} & \AnswerPct{27.1}{65} & \AnswerPct{100}{240} & \AnswerPct{100}{240} & \AnswerPct{100}{240}\\
\bottomrule
\end{tabular}}
\par\smallskip{\footnotesize\TableNoteAlign Columns denote Qwen3-32B, Qwen3.8-27B, Llama-3.2-3B, Llama-3.1-70B, Gemma 4 26B-A4B, and Gemma 4 31B. Gray/plain substitutions are tokenized separately by each model, so their resulting lengths and token positions need not match between the evaluated model checkpoints.\TableNoteEnd}
\end{table}
\begin{figure}[!htbp]
\centering
\includegraphics[width=\linewidth]{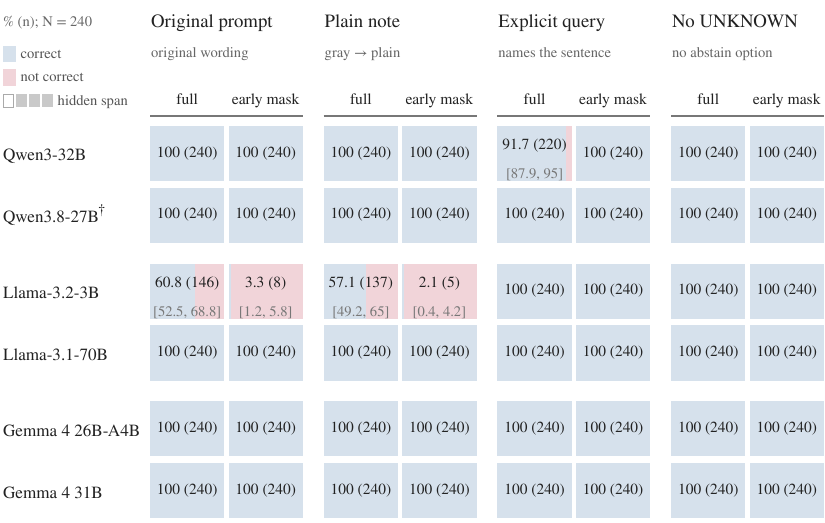}
\caption{Unrelated-label accuracy across six models and four prompt profiles, each pairing full access with the early-note mask; all 48 cells are shown. Entries are complete-correct percentages (counts) over 240 conditions from 120 groups, each cell filled to scale (blue: correct; rose: the rest), with the original 95\% group-bootstrap intervals below; degenerate intervals are omitted. Plain note changes gray to plain; explicit query identifies the inspection sentence; No UNKNOWN removes permission to abstain. $\dagger$ Qwen3.8 retains recurrent states.}
\label{fig:v35_control_prompts}
\end{figure}

Identical control content and length also need not preserve model
probabilities. Even where both occurrence masks produce correct labels
for every question, the log-probability difference between the correct
first value token and UNKNOWN can change. Table~\ref{tab:v35_positions}
reports that score separately from generated-answer accuracy.
Occurrences contain identical token IDs and text within each tokenizer;
absolute offsets can differ across tokenizers. Llama-3.2-3B retains its
historical fixed pair, 84 positions apart. Changing occurrence changes
position and preceding computation together, so it does not isolate
positional embeddings alone.

\begin{table}[!htbp]
\TableStyle

\caption{Effects of identical 31-token masks at two disjoint positions, shown as percentages (counts) over 240 conditions. Changes are later minus earlier, with paired 95\% intervals over 120 base groups.}
\label{tab:v35_positions}
{\setlength{\tabcolsep}{\dimexpr\tabcolsep*8/10\relax}
\begin{tabular}{ccccc}
\toprule
\TableHead{Model} & \TableHead{Earlier} & \TableHead{Later} & \TableHead{$\Delta$ accuracy (pp)} & \TableHead{$\Delta$ margin (nats)} \\\midrule
Qwen3-32B & \AnswerPct{100}{240} & \AnswerPct{100}{240} & +0.0 \TableSecondary{[0.0,\,0.0]} & $-$0.30 \TableSecondary{[$-$0.44,\,$-$0.16]}\\Qwen3.8-27B$^{\dagger}$ & \AnswerPct{100}{240} & \AnswerPct{100}{240} & +0.0 \TableSecondary{[0.0,\,0.0]} & +0.11 \TableSecondary{[0.07,\,0.14]}\\\addlinespace[2pt]
Llama-3.2-3B & \AnswerPct{3.3}{8} & \AnswerPct{60.8}{146} & +57.5 \TableSecondary{[49.6,\,65.0]} & +3.10 \TableSecondary{[2.98,\,3.22]}\\Llama-3.1-70B & \AnswerPct{100}{240} & \AnswerPct{100}{240} & +0.0 \TableSecondary{[0.0,\,0.0]} & +0.59 \TableSecondary{[0.52,\,0.66]}\\\addlinespace[2pt]
Gemma 4 26B-A4B & \AnswerPct{100}{240} & \AnswerPct{100}{240} & +0.0 \TableSecondary{[0.0,\,0.0]} & $-$0.06 \TableSecondary{[$-$0.11,\,$-$0.01]}\\Gemma 4 31B & \AnswerPct{100}{240} & \AnswerPct{100}{240} & +0.0 \TableSecondary{[0.0,\,0.0]} & $-$0.06 \TableSecondary{[$-$0.32,\,0.18]}\\
\bottomrule
\end{tabular}}
\end{table}
\begin{figure}[!htbp]
\centering
\includegraphics[width=\linewidth]{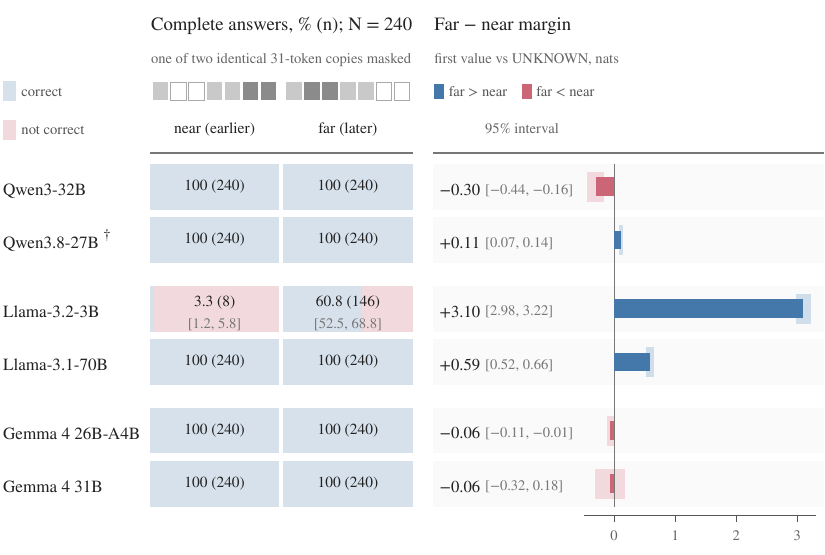}
\caption{Identical 31-token controls at near (earlier) and far (later) occurrences in one history; the outlined copy is masked. Left: complete-correct percentages (counts) and original 95\% intervals on 240 conditions from 120 paired groups; brackets are omitted only for degenerate intervals. Right: far-minus-near first-value margin and interval. Tokens and text match within each tokenizer; positions and preceding computation change together. The effect does not isolate positional embeddings. $\dagger$ Qwen3.8 retains recurrent states.}
\label{fig:v35_control_positions}
\end{figure}

\subsubsection{Generation Limits and What Longer Outputs Change}
\label{app:v35_generation_limits}
We separately score the 40-, 128-, and 256-token prefixes of each added
model's greedy trace. Across all five models and all five panels, every
individual complete-correctness indicator is identical at 40 and 256
tokens. This is stronger than equality of aggregate accuracy, but does not
mean that output lengths or failure categories are identical.
Table~\ref{tab:v35_generation_budgets} lists every panel with a changed
truncation count. No other new-model panel changes either count.

For Gemma 4 26B-A4B on OAKS, 248 of 348 outputs hit 40 tokens and 60 hit
256. Complete correct responses remain 84 across all three access
conditions: outputs that finish later still fail the content or format
requirement. Llama-3.1-70B similarly finishes its previously truncated
synthetic and factorial outputs without changing correctness. These are
observed continuations, not inferred answers. They support comparison
under a common limit while keeping costs and failure types separate.

\begin{table}[!htbp]
\WideTableStyle

\caption{Outcome entries are percentages (counts), using the row-specific $n$. Every added-model panel with changed truncation counts. Outputs include every access condition. The three limit columns count truncated responses; complete correct counts are identical at all three limits.}
\label{tab:v35_generation_budgets}
{\setlength{\tabcolsep}{\dimexpr\tabcolsep*12/14\relax}
\begin{tabular}{ccccccc}
\toprule
\multirow{2}{*}[-0.6ex]{\TableLabel{Model}} & \multirow{2}{*}[-0.6ex]{\TableLabel{Panel}} & \multirow{2}{*}[-0.6ex]{\TableLabel{Outputs}} & \multicolumn{3}{c}{\TableHead{Responses reaching the token limit}} & \multirow{2}{*}[-0.6ex]{\TableLabel{Correct}} \\
\cmidrule(lr){4-6}
 &  &  & \TableHead{Limit 40} & \TableHead{Limit 128} & \TableHead{Limit 256} &  \\\midrule
Gemma 4 26B-A4B & OAKS & 348 & \AnswerPct{71.3}{248} & \AnswerPct{37.4}{130} & \AnswerPct{17.2}{60} & \AnswerPct{24.1}{84}\\Llama-3.1-70B & Synthetic & 900 & \AnswerPct{3.8}{34} & \AnswerPct{0}{0} & \AnswerPct{0}{0} & \AnswerPct{83.7}{753}\\Llama-3.1-70B & Factorial & 1800 & \AnswerPct{0.1}{1} & \AnswerPct{0}{0} & \AnswerPct{0}{0} & \AnswerPct{83.8}{1508}\\
\bottomrule
\end{tabular}}
\end{table}

Across these experiments, higher candidate margins coexist with losses,
unchanged answers, or gains. The effects depend on the information
in the later record, the question, and the tested prompt. Model family,
size, and training vary together in these comparisons. The large control-location
effect remains specific to Llama-3.2-3B in these comparisons. These
supplied-source, fixed-question comparisons do not establish source
discovery, unrestricted conversational competence, or a universal rule
for which earlier information to remove.

\subsection{Varied Background Text and Early Note Access}
\label{app:v41_controls}

The control-mask failures in Appendix~\ref{app:v23_controls} motivate
testing a different background note. We test Qwen3-4B, Llama-3.2-3B,
and Qwen3-32B on the same 240
conditions from 120 groups. An authored set of distinct complete sentences
replaces the repeated seal-and-folder note with exactly the same number of
native tokens. Every token outside the note and every absolute position
is unchanged. This isolates the chosen substitution, not repetition alone:
lexical content and sentence structure change together. The 21 combinations
of query and intervention per profile include full, source, equal-token,
shifted-note, and four-token ablations; their complete matrices and source
attention summaries are retained in the numerical supplement.

\begin{table}[!htbp]
\TableStyle
\caption{Llama-3.2-3B unrelated answers under two note contents and matched access changes. Percentages (counts) use 240 conditions per cell; shifted masks hide the same number of tokens as the original mask.}
\label{tab:v41_note_controls}
{\setlength{\tabcolsep}{\dimexpr\tabcolsep*10/12\relax}
\begin{tabular}{cccccc}
\toprule
\multirow{2}{*}[-0.6ex]{\TableLabel{Note}} & \multirow{2}{*}[-0.6ex]{\TableLabel{Full}} & \multicolumn{2}{c}{\TableHead{Old-source-length note mask}} & \multicolumn{2}{c}{\TableHead{Four note tokens masked}} \\
\cmidrule(lr){3-4}\cmidrule(l){5-6}
 &  & \TableHead{Early} & \TableHead{Preserve first four} & \TableHead{First four} & \TableHead{Last four} \\\midrule
Repeated & \AnswerPct{60.8}{146} & \AnswerPct{3.3}{8} & \AnswerPct{7.5}{18} & \AnswerPct{59.6}{143} & \AnswerPct{64.6}{155}\\Varied & \AnswerPct{90.8}{218} & \AnswerPct{77.5}{186} & \AnswerPct{80.8}{194} & \AnswerPct{87.9}{211} & \AnswerPct{87.9}{211}\\
\bottomrule
\end{tabular}}
\end{table}

The early-mask loss is $-57.50$ points [$-65.42,-50.00$] under repeated
text and $-13.33$ [$-17.92,-9.17$] under varied text. Preserving the first
four note tokens improves the masked result by only $4.17$ points
$[2.08,6.67]$ and $3.33$ $[1.25,5.83]$. These pointwise intervals resample
the 120 base groups. Both Qwen models retain 240/240 on all five unrelated
conditions in either profile. The original fixed 31-token occurrence
comparison is a separate protocol; these new early masks match the
old-source token count within each tokenizer.

The separate initial-four-token ablation includes native chat/system
framing; it causes immediate empty EOS in Qwen3-4B and substantial
format or generation failures in other settings. Neither that perturbation
nor observed attention mass establishes that an attention sink caused the
early-note collapse. The note ablations retain the sequence's initial
tokens, and changing their contents alters the full-access baseline too.

On the 60 unit-dependent replacement groups, Qwen3-4B full/whole/number
counts are 60/23/60 under repeated text and 60/17/60 under varied text.
Qwen3-32B gives 60/36/60 and 60/28/60. Historical whole-source and number-only
counts are zero in these strata for both models. Thus varied filler does
not remove the current unit-access failures or their historical cost.

\subsection{Synthetic Record Histories, Operations, and Answer Requirements}
\label{app:revision_protocol}

The 300 test histories use the disjoint split in
Appendix~\ref{app:r570_recognition}; the predicted affirmed values in
Table~\ref{tab:revision_detector} supply the same decisions to all generators.
\input{context/revision_closed_loop_table}

\paragraph{One history, multiple operations and questions.}
Each history contains the old record at Time 1, a record about a different
entity, and the later record at Time 2. Every native-chat prompt receives
the same instruction to follow authoritative updates, distinguish rejected
alternatives, and respect the requested time. We prefill the common history
prefix once and evaluate current, Time-1 historical, and unrelated
questions independently. Each answer must be JSON with an \texttt{answer} string. Exact
case-folded, whitespace-normalized field equality determines correctness;
malformed JSON counts as incorrect. The field may acknowledge an old value
for a historical query without being counted as an incorrect use of the old value. The
separate mention flag is descriptive and never vetoes a correct answer.
Maximum answer length is 40 tokens and decoding stops at EOS.

Masking preserves token positions and all original history K/V tensors;
the attention mask sets visibility of the annotated old-text positions to
zero only for subsequent computation. Identical history tensors are checked
after all conditions for all 900 model--histories. Text recomputation instead
removes the Time-1 record and recomputes the full surviving prompt. The
latest-record baseline is given the annotation indicating whether the entity and attribute match:
it removes A for any later record about the same entity and attribute, including confirmations, and
retains it for an unrelated update. It is an informed baseline, not an
automatic method for identifying entities and attributes. The ground-truth baselines receive the true replacement relation
but otherwise use the same mask or text operation. Query-aware masking
receives the query type and uses full access for historical and
unrelated questions. Its identical preservation outputs are expected by
construction. The 18,900 policy--query evaluations cover 900 model--history
pairs; equivalent conditions use the same generated answers and timings.

Cached and one-shot forwards over identical tokens are checked separately
from edited-text recomputation. Original history tensors remain unchanged in all mask conditions. Numerical differences are not semantic updates; Appendix~\ref{app:operation_costs} defines the cost boundaries.

\paragraph{Exact generation instruction and record order.}
The shared generation system message is:
\begin{quote}\small
Use the ordered authoritative records. A later entry changes only the entity
and attribute it concerns. Distinguish affirmed values from rejected
alternatives. For a historical question use the requested time. Return only
JSON with one field: \texttt{\{"answer": "the exact queried value"\}}.
If it cannot be determined, use UNKNOWN.
\end{quote}
The user message contains \texttt{Ordered records:}, then
\texttt{Time 1:} and A, \texttt{Other record:} and the other entity's record,
\texttt{Time 2:} and B, and \texttt{Question:} followed by the query, each
section separated by a newline (two before the question). Recompute arms
remove the complete Time-1 section. Current questions ask for the entity's
attribute after Time 2; historical questions ask for the same attribute
before that entry, at Time 1; unrelated questions ask for the other
entity's attribute. The exact strings and identifiers are distributed
in the dataset and evaluation code.

\subsection{Query Preservation and Generation Errors}
\label{app:v7_errors}
Table~\ref{tab:v7_all_queries} adds unrelated-query preservation to
the current and historical endpoints in Table~\ref{tab:revision_closed_loop}.
The current-answer decomposition is in Table~\ref{tab:v8_values}; its
rule for comparing values and units is specified in Appendix~\ref{app:v8_units}.
Invalid JSON does not satisfy the required output format. An extractor failing on an
already generated free answer is a separate measurement failure.

\input{context/v7_all_query_table}

Neither Qwen has a current-answer format failure under keep-all,
detector-mask, or oracle-mask. Their masking losses therefore concern
answer-field mismatches, including unit omission, not necessarily wrong
semantic values. Against keep-all, Qwen3-4B's 26 oracle-mask reversals
include 17 omitted hours units, five UNKNOWN answers, and four other values;
Qwen3-8B's 17 reversals contain 11 omitted units and six UNKNOWN answers.
This distinction qualifies the exact-field accuracy losses in the main
table. Llama detector-mask produces seven instances of
\texttt{\{"answer": UNKNOWN\}}, which are invalid because the value is
unquoted; its oracle-mask has none. Formatting checks from one experiment are not used to infer the error distribution in another.

\subsection{Correct Values and Complete Units}
\label{app:v8_units}

\input{context/v8_value_table}

We retrospectively compare returned and reference values in all 18,900
policy--query evaluations on the synthetic record dataset and all 2,700 outputs from the experiment measuring both candidate scores and
generated answers. This analysis covers every policy, control, and query
type, including incorrect answers, corrections, and reversals.
The exact-field score is reported separately. We compare the explicit JSON answer with the requested value.
Non-quantity answers require normalized string equality. The quantity targets in this synthetic dataset are single values explicitly expressed in hours.
Explicit hour abbreviations, minutes, or days are compared after dimensional
conversion; incompatible or unrecognized units are not discarded. A bare
number can match the target's numerical magnitude only in this setting, where all quantity references are expressed in hours. It remains unit-incomplete and does not become a complete answer.
This comparison therefore does not resolve ambiguous missing units in general
natural-language answers and does not measure whether answers are semantically correct in all respects.

We place each answer in one of six categories: correct current value with its
unit, correct magnitude without its unit, old value, other value, explicit UNKNOWN,
or invalid format. Missing units on incorrect values are also recorded as a separate
flag. For the original current queries, the numbers of matching values under keep-all/oracle-mask
are 297/288 of 300 for Qwen3-4B, 295/290 for Qwen3-8B, and 248/251 for
Llama-3.2-3B. Correct-magnitude unit omissions are 2/19, 20/31, and 50/51,
respectively. Equivalent policy conditions that share an output receive the
same score under this criterion.

Among replacement cases in that experiment, Qwen3-4B goes from 150/150 matching values under control masking to 147/150 under old masking; Qwen3-8B goes
from 150/150 to 148/150. Their exact-field losses are eight and seventeen
answers, respectively, of which five and fifteen omit hours. The remaining
three and two are explicit UNKNOWN. Thus these losses primarily
concern answer completeness and abstention, with no old-value substitution.

\paragraph{The later record already states the complete quantity.}
In all 30 replacement quantity cases from ten groups in the synthetic record dataset, B contains
the annotated entity, attribute, and exact current value including
\emph{hours}. This includes every unit-omission reversal in the two Qwen
models. Explicit availability of the unit does not establish which visible
tokens the generator uses. Unit omission therefore does not demonstrate
that A contained the only unit information. The factorial experiment in
Appendix~\ref{app:v17_scope} varies mask scope, the information stated in B,
question requirements, and neutral context using the same scoring criteria.

\subsection{Candidate Scores and Generated Answers Under Matched Conditions}
\label{app:v7_bridge}

Before inference, we fixed 300 test cases from the synthetic record dataset and one OAKS answer transition for each of the 116 held-out questions. All
three generators execute full, old-mask, and equal-token control-mask
conditions: 2,700 outputs on synthetic records and 1,044 on OAKS, 3,744 in total. Every condition
scores canonical current/old JSON answers and performs actual greedy
generation with the same prompt, answer requirements, and attention mask.
The released results include generated answers, both sequence scores,
token positions, input hashes, format labels, and cache-consistency
measurements for all 1,248 model--histories.

\paragraph{Control construction for the synthetic record dataset.}
The original different-entity record is shorter than A. Rather than compare
unequal token counts, all arms add the same control note before Time 2:
``The notebook contains numbered pages and a plain cover.'' repeated 16
times. The system instruction, original A/other/B records, and current
question are unchanged. The old mask covers every token overlapping A;
the control mask covers exactly the same number of tokens from the start
of the added note. These positions are disjoint and precede the question.
The augmented full-access condition is not the original Table~\ref{tab:revision_closed_loop}
baseline. Its Qwen accuracies differ, which itself limits transfer of an
absolute score across protocols.

The 150 replacement, role-swap, and explicit-rejection cases have distinct
old and current answers and determine the main matched margin comparison.
Confirmation, role confirmation, and unrelated updates make current equal
to old, giving exactly zero current--old margin. They remain in all-case
generation results and error counts. The earlier-source mask is applied unconditionally in this experiment, including to these control cases. Their losses therefore cannot be attributed to a policy that masks only true replacements. Table~\ref{tab:v7_bridge} restricts both its
margin and accuracy contrast to actual replacements. Qwen3-4B's eight
reversals and Qwen3-8B's 17 contain no old-value substitutions. Five and 15
omit the unit from an otherwise matching hours value; three and two give
UNKNOWN. Treating unit omission as acceptable is a clearly identified
sensitivity calculation, not a change to the original exact-field metric.
For Qwen3-4B, all eight reversals occur on examples whose margin increases;
for Qwen3-8B, 11 of the 17 do. Thus the coexistence is not only a comparison
between different sample means.

\paragraph{Common cache and sequence scoring.}
We prefill the complete history once, then clone its unchanged K/V for
every query, generation, and scoring condition. Masks affect only subsequent reads;
positions, resident bytes, and previously formed downstream states are
identical. Each score sums token log-probabilities of the canonical
\texttt{\{"answer": "value"\}} string, without a prepended space or EOS.
Generation uses the same answer requirements, greedy decoding, a 40-token limit, and EOS stopping defined by the model and tokenizer. Exact case-folded,
whitespace-normalized answer-field equality defines the primary score.
Units remain part of the value in the synthetic record dataset. Transformers 5.8.0, bfloat16 eager
attention, and the same checkpoint versions are used across conditions.
The policy comparison on synthetic records in Appendix~\ref{app:revision_protocol}
uses Transformers 5.12.1.

\begin{table}[!htbp]
\TableStyle
\caption{Outcome entries give percentages (counts), N=150. Paired entries are top/bottom in the original header order. Candidate scores and generated answers under one fixed protocol on 150 replacement cases from 50 groups in the synthetic record dataset. Both contrasts compare mask of the earlier source with control mask covering the same number of tokens on identical histories and answer strings. $\Delta m$ is the change in the current-minus-old sequence margin (Eq.~\eqref{eq:margin}); $\Delta$ exact is exact-field accuracy (pp), with paired 95\% intervals. C/R denotes corrections/reversals. Unit/UNK counts these outcomes among correct-to-incorrect reversals.}
\label{tab:v7_bridge}

{\setlength{\tabcolsep}{\dimexpr\tabcolsep*8/10\relax}
\begin{tabular}{ccccc}
\toprule
\TableHead{Generator} & \TableHead{$\Delta m$} & \TableHead{$\Delta$ exact} & \TableHead{C/R} & \TableHead{Unit / UNK} \\\midrule
Qwen3-4B & +1.38 \TableSecondary{[1.17,\,1.62]} & $-$5.3 \TableSecondary{[$-$8.7,\,$-$2.0]} & \shortstack[c]{\AnswerPctInline{0}{0}\\\AnswerPctInline{5.3}{8}} & \shortstack[c]{\AnswerPctInline{3.3}{5}\\\AnswerPctInline{2}{3}}\\Qwen3-8B & +1.72 \TableSecondary{[1.41,\,2.06]} & $-$11.3 \TableSecondary{[$-$18.7,\,$-$4.7]} & \shortstack[c]{\AnswerPctInline{0}{0}\\\AnswerPctInline{11.3}{17}} & \shortstack[c]{\AnswerPctInline{10}{15}\\\AnswerPctInline{1.3}{2}}\\Llama-3.2-3B & +0.31 \TableSecondary{[0.13,\,0.50]} & +0.7 \TableSecondary{[$-$2.0,\,3.3]} & \shortstack[c]{\AnswerPctInline{2}{3}\\\AnswerPctInline{1.3}{2}} & \shortstack[c]{\AnswerPctInline{0}{0}\\\AnswerPctInline{0}{0}}\\
\bottomrule
\end{tabular}}
\end{table}

\input{context/v7_bridge_overview}
\input{context/v7_bridge_relations}

We report old-mask against both full and control-mask. Uncertainty for the synthetic record dataset
uses 10,000 paired resamples of its 50 groups, preserving relation variants.
OAKS uses 10,000 paired question resamples within each of the five books,
keeping each book's sample count fixed; this comparison is question-weighted,
whereas the original 19-model OAKS effect is book-averaged. Control-mask
minus full generation changes are $-0.33/0.00/+0.33$ pp on synthetic records and
$0.00/-0.86/+2.59$ pp on OAKS. The supplementary results include the intervals
and correction/reversal counts for every condition. The control is not assumed to have
zero effect on the sequence margin.

\paragraph{Native generation on unchanged natural sources.}
OAKS reuses the earliest held-out event ID per question, its original
earlier/later sentences, unrelated excerpts, question, and two annotated
answer strings. The native system instruction asks for the state after the
later source. Both answer candidates are displayed without temporal-role
labels; alternating question order balances current A and current B (58
each). The model generates a JSON A/B/UNKNOWN answer. The old mask covers
the earlier source; the matched mask covers an equally long start of the
existing unrelated-source block. This measures actual native-chat option
generation with supplied spans and candidates, not open-ended natural
answer quality or retrieval over complete novels.

Old-mask-minus-control generation gains are 22.4, 5.2, and 9.5 pp for
Qwen3-4B, Qwen3-8B, and Llama-3.2-3B, with 28/2, 8/2, and 16/5
corrections/reversals; their paired intervals, computed with the
resampling in Appendix~\ref{app:v35_oaks}, exclude zero
(Table~\ref{tab:v35_oaks}). Old-mask-minus-full gains are 22.41, 4.31, and
12.07 pp (Table~\ref{tab:v7_bridge_overview}). All 1,044 generated answers have valid JSON. Table~\ref{tab:v7_oaks_errors}
separates old-answer choices from abstention; the examples below show
why the average gain does not justify universal masking.

\noindent\begin{minipage}{\linewidth}
\paragraph{Qwen3-4B: P37\_q01.} What is Colin's health and physical ability like?

Candidate A: Able to move their arms and legs and heads about in a way which was neither walking nor running nor sitting down \quad Candidate B: Able to walk and dig a little

Reference: B. Full: \texttt{\{"answer": "B"\}}. Old-mask: \texttt{\{"answer": "A"\}}.

\par\end{minipage}\par\smallskip

\noindent\begin{minipage}{\linewidth}
\paragraph{Qwen3-8B: P22\_q01.} Who has become Robinson Crusoe's most recent companion at this point in the story?

Candidate A: The English captain \quad Candidate B: The Spaniard and Friday's Father

Reference: A. Full: \texttt{\{"answer": "A"\}}. Old-mask: \texttt{\{"answer": "UNKNOWN"\}}.

\par\end{minipage}\par\smallskip

\noindent\begin{minipage}{\linewidth}
\paragraph{Llama-3.2-3B: P12\_q24.} What is Anne's plan regarding the Avery Scholarship at this point in the story?

Candidate A: She wins it. \quad Candidate B: She decides to try and win it to attend Redmond College.

Reference: A. Full: \texttt{\{"answer": "A"\}}. Old-mask: \texttt{\{"answer": "B"\}}.

\par\end{minipage}\par\smallskip

\input{context/v7_oaks_errors}

The candidates and recorded JSON above are complete. They illustrate
the benchmark's supplied transition targets, not a new human adjudication
of every underlying passage. Source event IDs and hashes support
reconstruction; original OAKS prose is not redistributed in the supplement.

\subsection{Component Interventions and Accuracy Preservation}

\label{app:component_attribution}

The experiment uses the 400 history clusters per model defined in
Appendix~\ref{app:panel}.  Each cluster crosses
two states (old value confirmed or replaced) with two question types (target or unrelated) and nine conditions: no intervention; zeroing $K$, $V$, or $K{+}V$ at the old source or a matched unrelated span; and masking $Q$--$K$ attention to either span.
This yields 14,400 outcomes per model and 57,600 in total; all conditions keep
sequence length and cache positions fixed.  For each component, we subtract the
matched neutral-span change from the focal change, then compare that contrast
between superseded and current target queries.  The $Q$--$K$ contrast is the
positive control; the $K$, $V$, and $K{+}V$ intervals use Bonferroni correction,
and an intervention meets the accuracy-preservation criterion only if both lower bounds in its matched-control comparisons are at or above $-0.03$.

Table~\ref{tab:causal_kv_intervals} gives the component effects and complete
confidence intervals. The $K$, $V$, and $K{+}V$ intervals use Bonferroni
correction, while the $V{-}K$ contrast uses a separately specified two-sided
interval. Every interval excludes zero. Accuracy preservation is assessed
separately in Table~\ref{tab:causal_kv_safeguards}, because a nonzero effect
alone does not show that other useful answers are preserved.

\begin{table}[!htbp]
\TableStyle
  \caption{Component effects in natural-log units (nats), with 95\% intervals.
  Each $\Delta$ is the superseded-minus-current difference of the
  change in the expected-versus-alternative answer log margin at the old source relative to the matched unrelated span. Every
  interval excludes zero; Table~\ref{tab:causal_kv_safeguards} reports the
  accuracy-preservation criteria.}
  \label{tab:causal_kv_intervals}

  {\setlength{\tabcolsep}{\dimexpr\tabcolsep*8/10\relax}
\begin{tabular}{>{\centering\arraybackslash}m{0.18\linewidth}C{0.155\linewidth}C{0.155\linewidth}C{0.18\linewidth}C{0.20\linewidth}}
    \toprule
\TableHead{Model} & \TableHead{$\Delta_K$ [95\%]} & \TableHead{$\Delta_V$ [95\%]} & \TableHead{$\Delta_{K+V}$ [95\%]} & \TableHead{$\Delta_V-\Delta_K$ [95\%]} \\
\midrule
Qwen3-4B & \shortstack{2.599\\{}\TableSecondary{[2.253,\,2.930]}} & \shortstack{1.788\\{}\TableSecondary{[1.413,\,2.151]}} & \shortstack{2.247\\{}\TableSecondary{[1.897,\,2.586]}} & \shortstack{$-$0.812\\{}\TableSecondary{[$-$1.079,\,$-$0.546]}} \\
Qwen3-8B & \shortstack{1.889\\{}\TableSecondary{[1.650,\,2.129]}} & \shortstack{2.865\\{}\TableSecondary{[2.505,\,3.230]}} & \shortstack{1.877\\{}\TableSecondary{[1.657,\,2.094]}} & \shortstack{0.976\\{}\TableSecondary{[0.738,\,1.212]}} \\
    \midrule
Llama-3.2-3B & \shortstack{0.908\\{}\TableSecondary{[0.265,\,1.570]}} & \shortstack{2.738\\{}\TableSecondary{[2.474,\,3.008]}} & \shortstack{2.599\\{}\TableSecondary{[2.354,\,2.855]}} & \shortstack{1.830\\{}\TableSecondary{[1.252,\,2.407]}} \\
Llama-3.1-8B & \shortstack{1.339\\{}\TableSecondary{[0.560,\,2.118]}} & \shortstack{2.689\\{}\TableSecondary{[2.488,\,2.896]}} & \shortstack{3.178\\{}\TableSecondary{[2.997,\,3.359]}} & \shortstack{1.350\\{}\TableSecondary{[0.652,\,2.054]}} \\
    \bottomrule
  \end{tabular}}
\end{table}

\begin{table}[!ht]
\TableStyle
  \caption{Accuracy preservation under component interventions.  Each cell gives
  the Bonferroni-adjusted lower bounds for accuracy change on questions about a confirmed value and unrelated questions, respectively.  Both must be at least $-0.03$; only the
  Qwen3-8B $V$ intervention meets both criteria.}
  \label{tab:causal_kv_safeguards}

  {\setlength{\tabcolsep}{\dimexpr\tabcolsep*6/8\relax}
\begin{tabular}{>{\centering\arraybackslash}m{0.18\linewidth}C{0.23\linewidth}C{0.23\linewidth}C{0.24\linewidth}}
    \toprule
\TableHead{Model} & \TableHead{$K$: target / unrelated} & \TableHead{$V$: target / unrelated} & \TableHead{$K{+}V$: target / unrelated} \\
\midrule
Qwen3-4B & $-$0.098/$-$0.060 & $-$0.048/$-$0.055 & $-$0.088/$-$0.064 \\
Qwen3-8B & $-$0.083/0.000 & $-$0.025/0.000 & $-$0.065/0.000 \\
    \midrule
Llama-3.2-3B & $-$0.579/$-$0.366 & $-$0.043/0.111 & $-$0.035/$-$0.054 \\
Llama-3.1-8B & $-$0.443/$-$0.595 & 0.000/$-$0.069 & $-$0.063/0.026 \\
    \bottomrule
  \end{tabular}}
\end{table}

The matched $Q$--$K$ positive-control effects are $2.289$, $1.982$, $2.401$,
and $2.472$ in model order, with all intervals excluding zero.  Only $V$ zeroing in Qwen3-8B satisfies both accuracy-preservation criteria. The results show sensitivity to these component interventions, but do not identify an edit that reliably preserves answers across the tested models.

\subsection{Component Effects Do Not Establish Answer Preservation}
\label{app:v39_core_extension}
\label{app:v39_core}

The component study tests whether changing keys or values can reproduce
the source effect while preserving useful answers. The 400-group design
covers nine models and 129,600 conditions, with the common intervention,
accuracy-preservation criteria and within-model uncertainty correction
described in Appendix~\ref{app:component_attribution} for all component comparisons.

Seven of nine models have positive effects for each of $K$, $V$, and $K{+}V$.
Gemma 4 26B-A4B has unresolved intervals for all three,
while Gemma 4 31B has a negative $V$ effect, $-2.09$ nats
[$-3.09,-1.05$], and unresolved $K$ and $K{+}V$ effects.
The sign of the measured component effect therefore depends on which
model receives the intervention.

\begin{table}[!htbp]
\TableStyle

\caption{Component effects across nine models. Effects are in nats with the original within-model Bonferroni intervals. The last column applies the separate paired 95\% accuracy intervals to both current and unrelated safeguards; passing these safeguards alone does not require a positive effect.}
\label{tab:v39_components}
{\setlength{\tabcolsep}{\dimexpr\tabcolsep*8/10\relax}
\begin{tabular}{ccccc}
\toprule
\TableHead{Model} & \TableHead{$K$} & \TableHead{$V$} & \TableHead{$K{+}V$} & \TableHead{Safeguards pass} \\
\midrule
Qwen3-4B & 2.60 \TableSecondary{[2.25,\,2.93]} & 1.79 \TableSecondary{[1.41,\,2.15]} & 2.25 \TableSecondary{[1.90,\,2.59]} & None \\
Qwen3-8B & 1.89 \TableSecondary{[1.65,\,2.13]} & 2.86 \TableSecondary{[2.51,\,3.23]} & 1.88 \TableSecondary{[1.66,\,2.09]} & $V$ \\
\addlinespace[2pt]
Llama-3.2-3B & 0.91 \TableSecondary{[0.26,\,1.57]} & 2.74 \TableSecondary{[2.47,\,3.01]} & 2.60 \TableSecondary{[2.35,\,2.86]} & None \\
Llama-3.1-8B & 1.34 \TableSecondary{[0.56,\,2.12]} & 2.69 \TableSecondary{[2.49,\,2.90]} & 3.18 \TableSecondary{[3.00,\,3.36]} & None \\
Qwen3-32B & 1.61 \TableSecondary{[1.41,\,1.81]} & 2.06 \TableSecondary{[1.75,\,2.38]} & 1.55 \TableSecondary{[1.35,\,1.75]} & None \\
Qwen3.8-27B$^{\dagger}$ & 1.80 \TableSecondary{[1.59,\,2.01]} & 1.24 \TableSecondary{[0.99,\,1.49]} & 1.82 \TableSecondary{[1.61,\,2.02]} & $K$, $V$, $K{+}V$ \\
Llama-3.1-70B & 1.76 \TableSecondary{[1.06,\,2.46]} & 2.04 \TableSecondary{[1.74,\,2.34]} & 2.56 \TableSecondary{[2.31,\,2.81]} & $V$ \\
\addlinespace[2pt]
Gemma 4 26B-A4B & 0.58 \TableSecondary{[$-$0.24,\,1.41]} & $-$0.00 \TableSecondary{[$-$0.84,\,0.87]} & 0.51 \TableSecondary{[$-$0.37,\,1.39]} & $V$ \\
Gemma 4 31B & 0.95 \TableSecondary{[$-$0.02,\,1.93]} & $-$2.09 \TableSecondary{[$-$3.09,\,$-$1.05]} & 0.54 \TableSecondary{[$-$0.42,\,1.50]} & None \\
\bottomrule
\end{tabular}}
\par\smallskip{\footnotesize\TableNoteAlign $\dagger$ Only full-attention K/V coordinates change; prefilled recurrent states remain. Each model is evaluated on 14,400 conditions drawn from the same 400 history groups.\TableNoteEnd}
\end{table}

The accuracy safeguards impose a separate requirement: the lower bounds for
confirmed-current and unrelated accuracy changes must both be at least
$-0.03$ in their paired 95\% intervals. Only Qwen3.8 passes for all three interventions. The $V$ intervention
also passes in Qwen3-8B, Llama-3.1-70B, and Gemma 4 26B-A4B; the last of these
has no resolved positive component effect. Positive effects and preservation
therefore cannot be collapsed into one success count. These tests concern
supplied-candidate decisions, not historical retention or unrestricted
generation. Qwen3.8 retains its recurrent state, so its result does not test
removing all stored information about the source.
Using Bonferroni intervals for the safeguards as an additional sensitivity
check removes Gemma 4 26B-A4B's $V$ pass: its unrelated lower bound is
$-3.25$ percentage points. The other listed passes are unchanged.

\subsection{Recomputing States After a Derived Note}
\label{app:v5_propagation}

Fifty synthetic groups define a rule mapping two zones to two routes, an
old zone A, its correctly derived Time-1 route C, and an authoritative Time-2
zone update B. Current questions ask for the route after B; historical
questions ask for the zone before B. Values and their order are
counterbalanced. A paired control replaces the C token interval with
an equal number of tokens from a neutral note, leaving all other tokens and
positions unchanged. This control can end mid-sentence because length is
matched in tokens; it is a visibility control, not an alternative authored
narrative.

We compare six conditions: full access, masking A, masking C, masking A+C,
recomputing states after A, and recomputing the whole prefix. Selective refresh reuses the original
prefix through A and recomputes every later state with A hidden; the full
comparator also recomputes that prefix, then uses the same hidden-A suffix
rule. Both keep every original fact token, including C, and give neither
a corrected C nor a rewritten historical A. Queries in refreshed
arms also hide A. Thus recomputation tests how the model represents the later text when A is hidden;
it does not correct the explicit old conclusion by substituting its text.

All 3,600 outputs are JSON answers generated with native chat formatting, EOS
stopping, and a 40-token limit. The score is exact normalized answer-field equality. Every current answer in the derived-note experiment has a valid format, so changes in accuracy are not caused by parsing failures. We separately measure margins between candidate JSON answers, historical
accuracy, comparisons with controls that mask the same number of tokens, and operation costs.
Intervals use 10,000 paired resamples of 50 base groups. Operation cost covers cache crop/copy and recomputation, excluding shared prefill and query generation; it is not total response latency.

\input{context/v5_propagation_table}

For Llama, masking C gives a 26-point current-accuracy gain [14,38] over
full access and a margin gain beyond the neutral control of
1.62 [1.32,1.92]. Masking A alone gives a $-6$-point accuracy change
[$-14,2$]. Selective refresh improves accuracy relative to A-only masking
by 18 points [6,32], but its extra margin change beyond neutral C is
0.44 [$-0.07,0.97$]. The full comparator has a similar 16-point gain [4,30].
For the two Qwen models, current accuracy is 100\% in all arms; selective
refresh changes margin by $-0.43$ [$-0.66,-0.21$] and $-0.30$ [$-0.57,-0.03$]
relative to A-only masking. These observations show different effects of source and explicit-note
interventions. They do not establish when recomputation is necessary or how old information
persists through states without an explicit derived note.

\subsection{Later States and Recomputing the History}
\label{app:v39_propagation_extension}

The propagation control crosses two notes following A: a derived note C
that repeats information from A, and a matched neutral note. Six interventions
compare access-time masks with recomputation that hides A during formation
of later states. Each model has 50 groups, two note conditions, two query
types, and six interventions, for 1,200 outputs at a 256-token limit. Appendix~\ref{app:v5_propagation}
specifies the shared intervention design.

\input{context/v41_propagation_extension_table}
\FloatBarrier

Llama-3.1-70B is sensitive to the derived note: current answers rise from
18/50 with full access to 46/50 after masking A and to 50/50 after also masking
C. The second change is $+8$ percentage points with paired interval
$[2,16]$. Refreshing later states while hiding A also reaches 50/50.
The matched neutral condition begins at 49/50. This contrast supports a
role for information available through C or states formed from A, but the
joint-mask and refresh interventions change different paths and do not
uniquely identify where the obsolete value is represented.

The other four models answer all 50 current derived-note questions correctly
under every tested condition. Their saturated scores leave no measured
current-answer benefit to recover. With a neutral note, Gemma 4 31B's
historical answers fall from 50/50 to 8/50 when later states are refreshed
with A hidden; the other 42 outputs answer UNKNOWN. An access-time mask
preserves 50/50, and the derived-note condition stays at 50/50 under every
intervention. Because the stated two-zone rule makes the historical answer
derivable without A, these counts do not measure retention of A: the drop
is a change in answering behavior under refresh. Refresh also keeps the
original length and positions, unlike the deletion-and-recompute
operations in the main text.

\FloatBarrier
\section{Answer Quality and the Cost of Applying Memory Updates}
\label{app:public_studies}

The conversational-memory comparisons measure answer quality and latency
under stated input and timing conditions. Operation timings with supplied
source positions separate the cost of editing a cache from the cost of
finding what to edit. Repeated-query measurements then include detector
inference, update processing, and answering. The source locations and
classification of questions as current, historical or unrelated remain
supplied inputs to these measurements.

\subsection{Conversational-Memory Inputs and Answer Evaluation}
\label{app:v8_public}

We use the official cleaned LongMemEval-Oracle release, which contains
500 questions including 78 knowledge updates \citep{longmemeval2025}.
The input supplies the relevant sessions, ordered by timestamp and
retaining user and assistant turns. It therefore does not include the
full-history retrieval problem evaluated by LongMemEval-S/M.
Reference answers, question-type labels, answer-session identifiers, and
answer markers are withheld from generation and intervention selection.

The six generators are Qwen3-4B/8B/32B, Llama-3.2-3B-Instruct,
Llama-3.1-8B-Instruct, and Mistral-7B-Instruct-v0.3. They use bfloat16,
native chat templates, greedy decoding, normal EOS, and up to 128 new
tokens, with no context truncation. Qwen thinking is disabled.
Each model runs on one H200. The following comparisons specify their
own question subsets, intervention timing, and the cache capacity
allocated to each compared method.

GPT-5.6 grades the complete generated responses using the official
question-type rubric, including its abstention rule.
Appendix~\ref{app:v17_judge} gives the grading inputs and validation checks.
Correctness and generation termination are reported separately where
specified. Each uncertainty analysis keeps paired operations and related
questions together using the grouping stated for that comparison.

\subsection{Reducing Attention to Earlier Sessions}
\label{app:guarded_validation}
\label{sec:guarded_attenuation}

This comparison reduces attention to older sessions without blocking them
completely. We measure cost both over all questions and over those for
which attention weights are changed.
The policy uses session order and question wording; it does not estimate
whether a source has been superseded.

Let $s(j)\in\{0,\ldots,S-1\}$ denote the chronological session containing
history-content token $j$. For $S>1$, the policy adds
\begin{equation}
 r_j=\frac{S-1-s(j)}{S-1},\qquad
 b_j(q)=-\log(2)\,g(q)\,r_j
 \label{eq:source_attenuation}
\end{equation}
to every attention logit addressing that position during question processing
and generation. For content logit $z_j$, the unnormalized weight becomes
$\widetilde w_j=\exp(z_j)\,2^{-g(q)r_j}$. When the bias is applied
($g(q)=1$), oldest-session tokens receive a multiplicative factor of $1/2$
and newest-session tokens a factor of $1$. For one token in each session,
$\widetilde w_{\mathrm{old}}/\widetilde w_{\mathrm{new}}
=\tfrac12\exp(z_{\mathrm{old}}-z_{\mathrm{new}})$; the ratio equals
$1/2$ only when their content logits are equal. Session totals additionally
sum over their tokens. Headers, question tokens, and
generated tokens have $r_j=0$; single-session histories are unchanged.
The rule $g(q)$ based on the question text is zero for the expressions below and one
otherwise. All K/V, original positions, and previously formed states remain
available. Development compares retention, two attenuation strengths, and
rules for disabling the bias before fixing the policy.

\paragraph{Inputs and decision rule.}
We reuse the official LongMemEval-Oracle histories and generator settings
specified in Appendix~\ref{app:v8_public}: six models, native chat,
bfloat16, greedy decoding, normal EOS, and at most 128 output tokens.
The supplied history is prefilled separately from the question. All conditions act
before question processing; each independently generates its first token.
Original absolute positions are preserved. There is no context truncation.
The controller receives dated turns and question text, with reference answers,
question types, answer markers, and original session identifiers removed.

To disable the bias, the rule matches whole words, case-insensitively: \emph{before},
\emph{previous}/\emph{previously}, \emph{earlier}, \emph{first},
\emph{original}/\emph{originally}, \emph{initial}/\emph{initially},
\emph{ago}, \emph{between}, \emph{since}, \emph{when}, \emph{started},
\emph{starting}, \emph{began}, and \emph{beginning}. It also matches
``how many/long/much days/weeks/months/years,'' English month names,
and four-digit years beginning with 19 or 20. A match leaves source weights
unchanged. Otherwise, Equation~\ref{eq:source_attenuation} applies with its
fixed coefficient $\log(2)$. The bias depends on session order, not elapsed time.
It applies to source-content tokens; headers and subsequent query tokens are exempt.

\paragraph{Policy selection and evaluation partition.}
A deterministic split based on question identifiers selects 120 development
questions: 40 update-tagged, 25 temporal, 25 multi-session, and 10 of each
single-session type. Development compares complete-exchange retention,
half or quarter weight for the oldest session, and question-dependent
rules for disabling the bias. The disabling rule is chosen retrospectively
from these development outputs. The rule and attenuation strength are
fixed before generating 4,560 validation answers from six models and two
policies on the remaining 380 questions. GPT-5.6 evaluates those recorded
answers with the policy choices held fixed; its labels do not select a new
rule or attenuation strength.

\paragraph{Answerability and dependencies.}
The knowledge-update label includes abstention variants. In validation it contains
34 answerable questions and four abstention questions. The latter use the
official abstention rubric, so they are not counted as factual-update
corrections in Table~\ref{tab:guarded_validation}. Across all 380 questions,
the three disjoint question categories are answerable updates (34), other
answerable questions (323), and abstention (23). For all 38 questions labeled as knowledge updates,
the pooled correct count changes from 165/228 to 178/228.

\input{context/guarded_validation_table}

GPT-5.6 supplies the scoring labels for all 4,560 answers using the
same question-type rubric (Appendix~\ref{app:v17_judge}). For the dependence
analysis here, 10,000 bootstrap draws group base-question variants and
questions connected by publisher session identifiers, keeping all six
models together. The complete partition contains 362 such groups. Fourteen
validation questions connect to development through these identifiers.
Removing them leaves 366 questions in 348 groups, including 33 answerable
updates: 142/198 becomes 154/198, or $+6.06$ points $[2.60,9.80]$.
Other answerable questions change by $-1.21$ points $[-1.99,-0.47]$;
overall change is $-0.46$ points $[-1.25,0.32]$. We report both the complete
partition and this exclusion sensitivity. LongMemEval was available during
development, so the comparison does not use a previously unseen benchmark.

Across all 2,280 paired model outcomes, the bias corrects 27 answers and
reverses 36: 13 corrections and one reversal on answerable updates,
13 corrections and 35 reversals on other answerable questions, and one
abstention correction. The resulting full/bias counts are 1,323/1,314.
Appendix~\ref{app:v18_judge} reports the complete panel and the 109 questions
where the bias applies. Its base-question analysis groups answerable and
abstention variants, yielding 365 groups overall and 108 in the active
subset; it is reported separately from the publisher-session grouping
used here.

\paragraph{Numerical consistency and runtime.}
For every model, cached generation agrees with native generation,
zero bias gives identical tokens and logits, and nonzero bias gives finite
outputs. All 1,626 model--question pairs for which the bias is disabled independently reproduce the
full-history answer. The token-offset mapping agrees with an explicit
per-turn scan on all
500 questions and six tokenizers. Its CPU median is 1.17--1.52\,ms, compared
with 6.11--7.48\,ms for the per-turn scan; these component measurements are
not substituted for full inference times.

\begin{table}[!htbp]
\WideTableStyle
\caption{Measured costs on 380 reserved questions per model, including 109 questions where the bias is applied. Full uses unmodified attention; Bias weights sessions by their order. These columns give mean complete-answer times in ms across all questions. All/Active give median paired overheads for all questions and for those where the bias is applied (\%); Time to first token (TTFT) columns give paired overhead percentages over all questions. Timing includes control, question processing, and output-length changes after common prefill. The time to copy the cache is recorded separately; all models use math SDPA.}
\label{tab:guarded_cost}

{\setlength{\tabcolsep}{\dimexpr\tabcolsep*12/14\relax}
\begin{tabular}{ccccccc}
\toprule
\TableHead{Model} & \FullCell{\FullHeader{Full}} & \TableHead{Bias} & \TableHead{\shortstack{All\\p50 (\%)}} & \TableHead{\shortstack{Active\\p50 (\%)}} & \TableHead{\shortstack{TTFT overhead\\p50 (\%)}} & \TableHead{\shortstack{TTFT overhead\\p95 (\%)}} \\
\midrule
Qwen3-4B & \FullCell{347.7} & 354.7 & +0.67 & +9.64 & +0.88 & +8.61 \\
Qwen3-8B & \FullCell{407.8} & 395.4 & +0.95 & +9.42 & +1.53 & +9.13 \\
Qwen3-32B & \FullCell{1983.1} & 2041.6 & +0.21 & +20.12 & +1.56 & +5.52 \\
Llama-3.2-3B & \FullCell{293.9} & 283.8 & +0.94 & +9.21 & +1.81 & +11.65 \\
Llama-3.1-8B & \FullCell{136.8} & 132.5 & +1.51 & +10.53 & +2.22 & +10.96 \\
Mistral-7B & \FullCell{1145.7} & 1226.9 & +0.23 & +16.00 & +2.41 & +10.98 \\
\midrule
Pooled & \FullCell{719.2} & 739.2 & +0.56 & +10.77 & +1.57 & +10.21 \\
\bottomrule
\end{tabular}}
\end{table}

Measurements after caching the history start after common tokenization and history prefill.
They include mapping, bias construction, all subsequent mask construction,
question processing, and decoding. The separate copy of the cache needed to
compare conditions is recorded separately. Mean answer latency after history prefill changes from 719.2 to
739.2\,ms. Including common input preparation, it changes from 1,022.7 to
1,042.7\,ms. Answer lengths and all 139 token-limited
outputs across the two policies remain in these measurements. The policy
retains every history K/V tensor. Math SDPA times the complete model. Synthetic fused-kernel probes do not establish serving latency or released storage.

The median paired complete-answer increase is 0.56\% across all 2,280
model--question pairs, while the ratio of their mean times increases by
2.78\%. Among the 654 model--question pairs for which the bias is applied, mean time is 898.9 versus
968.2\,ms (+7.70\%), and the median paired increase is 10.77\%.
Table~\ref{tab:guarded_cost} reports all questions and the subset where the bias is applied by model.
This is whole-answer cost, including changed output lengths, rather than
an isolated measurement of bias construction or attention arithmetic.
Median paired first-token overhead is 1.57\% overall and 4.00\% when active.

\paragraph{Examples of answer correction and loss.}
For the bicycle question, the February source says ``I currently have three
bikes,'' whereas the October source says ``I just got a new one recently,
so I'll actually have four bikes with me on this trip.'' Qwen3-4B changes
from 3 to the reference answer 4. These short phrases are verbatim source
excerpts; the complete histories are supplied by the benchmark.

A question about several events shows a limitation of this rule. The question asks how many
weddings the user attended that year. The sources describe Rachel's wedding,
Jen and Tom's wedding, and Emily and Sarah's wedding in separate dated
conversations. This is a source summary, not a verbatim quotation.
Qwen3-32B changes from the reference count 3 to UNKNOWN. The query needs
multiple past events but matches none of the rule's fixed expressions.
This illustrates a remaining failure; it does not identify a particular
source token as the neural cause of the changed answer.

\subsection{Answer Accuracy When the Bias Is Applied}
\label{app:v18_judge}
\paragraph{Answer evaluation.}
GPT-5.6 scores every generated answer using the LongMemEval question-type rubric. Identical complete grading prompts share one judgment; generator and policy identities are withheld.
The active subset contains 109 questions and 654 paired model outcomes. Accuracy intervals retain paired models and linked question groups.
\begin{table}[!htbp]
\WideTableStyle
\caption{Correct outcomes are percentages (counts) over the stated model-question pairs; Net is pp. Paired Full/Bias entries retain that order; times and sizes retain their units. Answer changes under the fixed session-order bias, across all 380 questions and the 109 questions where the bias is applied. GPT-5.6 uses the unchanged LongMemEval question-type rubric. The panels retain their own question and model--question denominators.}
\label{tab:v18_active_quality}
\label{tab:v18_complete_audit}
\par\smallskip\textbf{A. All questions: 380 questions, 2,280 paired model outcomes}\par\smallskip
{\setlength{\tabcolsep}{\dimexpr\tabcolsep*10/12\relax}
\begin{tabular}{ccccccc}
\toprule
\TableHead{Subset} & \TableHead{Pairs} & \FullCell{\FullHeader{Full}} & \TableHead{Bias} & \TableHead{Net} & \TableHead{$\Delta$ (pp), 95\% CI} & \TableHead{\shortstack{Change\\$[-2,10]$ pp}} \\\midrule
Overall & 2280 & \FullCell{\AnswerPct{58}{1323}} & \AnswerPct{57.6}{1314} & \AnswerPP{-0.4}{-9} & $-$0.39 \TableSecondary{[$-$1.15,\,+0.35]} & \TableInterval{-0.39}{-1.15}{0.35}{-2}{10}{figureGray}\\Answerable updates & 204 & \FullCell{\AnswerPct{70.6}{144}} & \AnswerPct{76.5}{156} & \AnswerPP{+5.9}{+12} & +5.88 \TableSecondary{[+2.45,\,+9.80]} & \TableInterval{5.88}{2.45}{9.8}{-2}{10}{figureGray}\\Other answerable & 1938 & \FullCell{\AnswerPct{54.7}{1061}} & \AnswerPct{53.6}{1039} & \AnswerPP{-1.1}{-22} & $-$1.14 \TableSecondary{[$-$1.91,\,$-$0.41]} & \TableInterval{-1.14}{-1.91}{-0.41}{-2}{10}{figureGray}\\Abstention & 138 & \FullCell{\AnswerPct{85.5}{118}} & \AnswerPct{86.2}{119} & \AnswerPP{+0.7}{+1} & +0.72 \TableSecondary{[+0.00,\,+2.17]} & \TableInterval{0.72}{0}{2.17}{-2}{10}{figureGray}\\
\bottomrule
\end{tabular}}
\par\medskip\textbf{B. Bias applied: 109 questions, 654 paired model outcomes}\par\smallskip
{\setlength{\tabcolsep}{2.1pt}
\begin{tabular}{cccccc}
\toprule
\TableHead{Subset} & \TableHead{\shortstack{Questions /\\pairs}} & \TableHead{\shortstack{Correct\\Full / Bias}} & \TableHead{$\Delta$ (pp), 95\% CI} & \TableHead{\shortstack{Answer ms\\Full / Bias}} & \TableHead{\shortstack{TTFT overhead (\%)\\p50 / p95}} \\\midrule
All active & 109 / 654 & \shortstack[c]{\AnswerPctInline{49.5}{324}\\\AnswerPctInline{48.2}{315}} & $-$1.38 \TableSecondary{[$-$3.98,\,+1.23]} & 898.9 / 968.2 & 4.00 / 10.82\\Answerable updates & 23 / 138 & \shortstack[c]{\AnswerPctInline{80.4}{111}\\\AnswerPctInline{89.1}{123}} & +8.70 \TableSecondary{[+4.35,\,+13.77]} & 398.8 / 430.1 & 2.55 / 9.57\\Other answerable & 80 / 480 & \shortstack[c]{\AnswerPctInline{37.3}{179}\\\AnswerPctInline{32.7}{157}} & $-$4.58 \TableSecondary{[$-$7.71,\,$-$1.67]} & 1089.6 / 1172.4 & 4.20 / 11.04\\Abstention & 6 / 36 & \shortstack[c]{\AnswerPctInline{94.4}{34}\\\AnswerPctInline{97.2}{35}} & +2.78 \TableSecondary{[+0.00,\,+8.33]} & 273.8 / 307.8 & 5.19 / 9.05\\
\bottomrule
\end{tabular}}
\par\smallskip{\footnotesize\TableNoteAlign Changes are Bias minus Full. Printed 95\% intervals and the aligned marks retain the six fixed models together and group answerable and abstention variants of each base question. Panel A includes common-scale interval marks over $[-2,10]$ points, with a line at zero. Accuracy intervals condition on the evaluation labels. Panel B also retains mean complete-answer times in ms and TTFT overhead percentiles (\%). These are distinct from the publisher-session grouping in Appendix~\ref{app:guarded_validation}.\TableNoteEnd}
\end{table}

\subsection{Answer Evaluation and Validation}
\label{app:v17_judge}

GPT-5.6 evaluates every generated answer in the public-memory
comparisons using the original LongMemEval question-type rubric. The
rubric receives the question, reference answer, and complete generated
response. It preserves the benchmark's acceptance of complete intermediate
calculations, temporal one-unit tolerance, preference criterion, and
abstention rule. The separate reference-only OAKS regrading uses its fixed
reference-answer rubric. Appendix~\ref{app:v41_oaks_open} also reports a
review by a separate model grader with original excerpts and a source-completeness rubric;
those labels and denominators are distinct from the reference-only
protocol described here.

\paragraph{Inputs and evaluator configuration.}
Each request
contains up to 48 tasks in random order, with anonymous task
identifiers. A structured response contains one Yes/No decision and a
short explanation for each task. Generator names, policy names, and prior
labels are withheld. Each task retains its complete rubric and answer;
only identical complete prompts share one judgment. In this reference-only
protocol, judges see the supplied references without additional source passages.

The 4,560 answers in the six-model attenuation experiment are evaluated
in full, including identical outputs and abstentions. The policy and
recorded answers are fixed throughout evaluation. Whether a response
reaches its generation limit is determined from its decoding record. The public rubric
scores correctness independently of termination, with EOS-restricted
scores reported separately where specified. OAKS semantic acceptance
requires normal EOS. For the hard-mask and dynamic public comparisons,
an empty or whitespace-only answer is scored incorrect.

\paragraph{Validation and uncertainty.}
The exact returned model variant is recorded with each request. Every generated
answer receives a valid Yes/No judgment linked to its complete text.
Invalid or incomplete grading responses are resubmitted before analysis.
The supplementary material records the evaluator settings, rubrics, input
reconstruction instructions, raw responses, and answer mappings.

Correct-answer counts, corrections, and reversals use this complete set
of model judgments. Bootstrap draws keep paired policies and fixed model
outcomes together within each specified question group. The intervals
condition on the scoring labels; source reuse and the benchmark's reference
interpretations remain relevant to their scope. The publisher-session
analysis in Appendix~\ref{app:guarded_validation} and the base-question
analysis in Appendix~\ref{app:v18_judge} state their grouping separately.

\subsection{Hard Masks on the Same Public Session Spans}
\label{app:v41_public_hard}

To compare soft attenuation with logical source masking, we evaluate the
500 official LongMemEval-Oracle questions and the fixed guard of
Appendix~\ref{app:guarded_validation}. Qwen3-4B, Llama-3.2-3B, and
Qwen3-32B each generate 1,500 answers: full access, soft attenuation,
and hard masking. Soft weights range from one half in the oldest session
to one in the newest. Hard masking sets to negative infinity exactly the
positions assigned a negative soft bias. Thus both operations target the
same history tokens during query processing and all decoding steps,
with stored K/V and downstream prefilled states unchanged.

The rule selects sessions by their order and historical/date expressions
in the question. Both masked
arms have 162 active questions overall and 109 in the remaining
380-question partition. Inactive pairs generate identical answers for
all three models. The split is by question identifier: twelve
answerable/abstention source groups cross the 120/380 partitions. We retain
the 380-question denominator and separately exclude those twelve remaining
questions as a sensitivity analysis. This is a within-benchmark comparison
with prior benchmark exposure and the stated source overlap.

All generators use bfloat16, native templates, greedy EOS decoding and a
256-token maximum, without truncating the supplied history. Each branch
independently processes the question and produces its first token.
GPT-5.6 applies the unchanged official question-type rubric to each
complete response, with generator and policy names hidden. Identical
rubric prompts share one judgment. The policy and generated responses
remain fixed during evaluation. Paired intervals resample the underlying
answerable/abstention question groups, retaining their dependencies.
Primary public accuracy follows the official answer rubric and is reported
separately from generation termination. Requiring normal EOS gives
full/soft/hard counts of 193/198/182 for Qwen3-4B, 191/190/174 for
Llama-3.2-3B, and 276/270/233 for Qwen3-32B on the remaining 380 questions.
The corresponding hard-minus-full point changes are unchanged. Ceiling
hits are 9/9/9, 4/4/4, and 7/7/8, respectively; these unfinished responses
remain visible in the numerical supplement.

\begin{table}[!htbp]
\WideTableStyle
\caption{Outcome entries are percentages (counts). Hard masking and soft attenuation of identical session spans on LongMemEval-Oracle. All arms preserve the shared prefilled history; the fixed temporal rule applies during question processing and decoding. Correct counts use the original remaining 380 questions.}
\label{tab:v41_public_hard}
{\setlength{\tabcolsep}{\dimexpr\tabcolsep*8/10\relax}
\begin{tabular}{ccccc}
\toprule
\TableHead{Model} & \FullCell{\FullHeader{Full}} & \TableHead{Soft} & \WholeCell{\WholeHeader{Hard}} & \TableHead{Hard$-$Full (pp) [95\% CI]} \\\midrule
Qwen3-4B & \FullCell{\AnswerPct{52.4}{199}} & \AnswerPct{53.7}{204} & \WholeCell{\AnswerPct{49.5}{188}} & $-$2.9 \TableSecondary{[$-$5.2,\,$-$0.8]}\\Llama-3.2-3B & \FullCell{\AnswerPct{50.8}{193}} & \AnswerPct{50.5}{192} & \WholeCell{\AnswerPct{46.3}{176}} & $-$4.5 \TableSecondary{[$-$6.9,\,$-$2.3]}\\Qwen3-32B & \FullCell{\AnswerPct{73.9}{281}} & \AnswerPct{72.4}{275} & \WholeCell{\AnswerPct{62.6}{238}} & $-$11.3 \TableSecondary{[$-$14.7,\,$-$8.2]}\\
\bottomrule
\end{tabular}}
\par\smallskip{\footnotesize\TableNoteAlign GPT-5.6 applies the unchanged LongMemEval question-type rubric with model and policy names hidden. Intervals resample 365 question groups. Twelve source groups cross the original question-ID partition; separate sensitivity excludes their remaining questions. These are temporal spans, not oracle unnecessary facts.\TableNoteEnd}
\end{table}

Excluding the twelve remaining questions with a development-paired source
leaves 368 questions from 353 groups. Full/soft/hard counts become
190/195/180 for Qwen3-4B, 184/183/168 for Llama-3.2-3B, and 271/265/228
for Qwen3-32B. Hard-minus-full changes are $-2.72$ points
$[-4.95,-0.54]$, $-4.35$ $[-6.79,-2.16]$, and $-11.68$
$[-15.04,-8.31]$, respectively. The answer-accuracy loss under this fixed
temporal rule remains after excluding the overlapping source groups.
\begin{table}[!htbp]
\WideTableStyle
\caption{Paired median overheads (\%) on the 109 remaining questions where the temporal rule is active. TTFT and query processing precede answer decoding; complete-generation time also reflects output-length changes.}
\label{tab:v41_public_timing}
{\setlength{\tabcolsep}{\dimexpr\tabcolsep*14/16\relax}
\begin{tabular}{cccccccc}
\toprule
\multirow{2}{*}[-0.6ex]{\TableLabel{Model}} & \multirow{2}{*}[-0.6ex]{\TableLabel{Operation}} & \multicolumn{2}{c}{\TableHead{TTFT (\%)}} & \multicolumn{3}{c}{\TableHead{Processing time (\%)}} & \multirow{2}{*}[-0.6ex]{\TableLabel{Tokens (\%)}} \\
\cmidrule(lr){3-4}\cmidrule(lr){5-7}
 &  & \TableHead{Resident} & \TableHead{Cold} & \TableHead{Query} & \TableHead{Decode} & \TableHead{Complete} &  \\
\midrule
Qwen3-4B & Soft & +63.01 & +2.51 & +56.46 & +62.64 & +62.89 & +0.00 \\
Qwen3-4B & \WholeCell{Hard} & \WholeCell{+171.03} & \WholeCell{+7.11} & \WholeCell{+165.97} & \WholeCell{+322.50} & \WholeCell{+224.39} & \WholeCell{+0.00} \\
\addlinespace[2pt]
Llama-3.2-3B & Soft & +65.57 & +2.80 & +53.48 & +99.46 & +83.40 & +0.00 \\
Llama-3.2-3B & \WholeCell{Hard} & \WholeCell{+238.62} & \WholeCell{+9.84} & \WholeCell{+228.95} & \WholeCell{+392.30} & \WholeCell{+298.63} & \WholeCell{+0.00} \\
\addlinespace[2pt]
Qwen3-32B & Soft & +58.87 & +2.01 & +56.33 & +47.76 & +50.12 & +0.00 \\
Qwen3-32B & \WholeCell{Hard} & \WholeCell{+117.79} & \WholeCell{+4.00} & \WholeCell{+115.92} & \WholeCell{+84.44} & \WholeCell{+84.89} & \WholeCell{$-$38.10} \\
\bottomrule
\end{tabular}}
\par\smallskip{\footnotesize\TableNoteAlign Resident TTFT includes policy construction and query processing, excluding the benchmark cache fork. Cold TTFT additionally includes the shared tokenization and history prefill, not an independently repeated cold request. History prefill is identical within each pair by construction. Reference SDPA implementation includes attention diagnostics and synchronization; no optimized-serving claim.\TableNoteEnd}
\end{table}

\paragraph{Timing boundaries.}
History tokenization and prefill are shared across the three arms of a
question. Resident TTFT measures policy construction and question processing;
it excludes the benchmark's cache copy and subsequent answer decoding.
Cold TTFT adds the shared tokenization and prefill, so it is a sum of
measured components rather than separately repeated cold requests.
Full-generation time additionally includes decoding and can change when
answers change length. The numerical supplement retains absolute medians,
paired median and p95 overheads, and both all-question and active strata.

The runtime is a reference native SDPA MATH implementation preserving
original positions. Diagnostic attention statistics and synchronization
contribute to its timing. These measurements isolate pre-answer cost from
answer-length effects in this implementation; they do not establish an
unavoidable cost of logical masks, an optimized serving speedup, or physical
KV savings. The paired history-prefill difference is zero by construction.

\subsection{Query-Aware Cache Comparisons and Capacity Accounting}
\label{app:v41_dynamic}

We evaluate Quest, FINCH, and RefreshKV on the same three dense generators,
300 synthetic-record questions, and all 500 LongMemEval-Oracle questions.
Full access and a source mask are paired baselines; the controlled mask
hides A on all 300 conditions, including confirmations and unrelated updates.
The changed-value subset contains direct replacements, explicit rejection
of an old value, and role-swapped statements (50 each), sharing the same
50 base groups.
A changed-value label does not assert that every token of A is unnecessary.
For public questions the source mask follows the fixed temporal/session
guard, rather than a replacement label. A recent-page selector
matches Quest's actual selected page count. All answers are freely
generated rather than chosen from a supplied candidate list. The controlled
endpoint requires the complete canonical answer, including the unit for
quantitative answers;
public answers are evaluated by GPT-5.6 with the unchanged official
question-type rubric and hidden generator and policy identities,
with an additional EOS-restricted sensitivity and separate ceiling counts.

\paragraph{Implemented operations.}
Quest uses the pinned official Python accuracy implementation, adapted to
native post-normalization and post-RoPE queries and keys. Pages contain
16 tokens and the first two layers retain full attention. It selects pages
from each query without physically discarding the full cache
\citep{quest2024}. FINCH uses the author-endorsed KVPress nonchunked path:
query-observation attention selects history keys, query-observation tokens
are retained, and selected keys are rerotated to compact positions
\citep{finch2024}. Passing the history/query boundary explicitly replaces
a delimiter token without changing the evaluation prompt. Its scores and
compressed tensors were checked against the upstream operations for
multi-head and grouped-query attention in float32 and bfloat16.

RefreshKV is a paper reproduction because the referenced author repository
contained no released implementation \citep{refreshkv2025}. It uses full
query attention, initializes a partial set from the last prompt query,
and checks query-vector cosine every ten decoding steps, refreshing at
cosine at most 0.95. Grouped-query heads use max aggregation with width-seven
max pooling. A partial step accesses the retained set plus the current
token; the full cache remains resident for refresh. New generated tokens
are prioritized in the retained set, with recent generated tokens taking
priority if a sequence fills the entire partial capacity. These settings
are fixed across the comparison.

\paragraph{Budgets and comparison boundaries.}
The fixed dynamic capacities are 128 on short controlled inputs and
512 on public inputs. They are explicit settings, not claims of optimal
compression or equality to source masking. A source mask usually leaves
most history visible and releases no storage. FINCH includes its protected
observation window in capacity; Quest rounds by pages and has dense early
layers; RefreshKV performs full query and refresh passes. Thus even a
common nominal capacity does not equate these methods' resources.

The comparison that matches visible prefix length sets each dynamic
method's capacity to the prefix and common query-observation length minus
the source-mask token count, measured before the final query-token forward.
Quest retains page rounding, FINCH retains its observation window, and
RefreshKV retains full-query and refresh schedules. The full/source
baselines use the same prompts, model settings, and histories.
This matches one pre-answer capacity, not physical memory or
access throughout decoding. Tables~\ref{tab:v41_dynamic_controlled} and~\ref{tab:v41_dynamic_public_completion}
keep the controlled matched comparison separate from the public fixed-512 comparison; complete
fixed-128, recent-page, relation, and task-family results remain in the
numerical supplement.

The query schedules are method-specific. FINCH compresses after observing
the query except its final token; RefreshKV observes the complete query
with full attention. Source/full/Quest use the same question-processing
schedule within this comparison. Each protocol has its own full-access
baseline. Generation caps, empty answers,
complete canonical answers, and semantic public labels remain distinct metrics.
An empty generated answer is deterministically incorrect, including on
abstention questions: silence does not explicitly state that the answer
is unavailable. GPT-5.6 returns a valid Yes/No decision for every
task, and identical rubric prompts share one judgment. Complete generation
texts and decisions remain available. These comparisons assess answer
quality under the specified access schedules.

\input{context/v41_dynamic_controlled_table}
\begin{table}[!htbp]
\WideTableStyle
\caption{Per-case capacity accounting for the matched controlled comparison. Prefix and masked counts are medians; K gives the minimum, median and maximum configured pre-answer capacity across 300 conditions. These configured values are identical for Quest, FINCH and RefreshKV within each model.}
\label{tab:v41_dynamic_capacity}
{\setlength{\tabcolsep}{\dimexpr\tabcolsep*10/12\relax}
\begin{tabular}{ccccc}
\toprule
\TableHead{Model} & \TableHead{Prefix tokens} & \TableHead{Masked tokens} & \TableHead{K min/median/max} & \TableHead{Quest selected range} \\
\midrule
Qwen3-4B & 316.0 & 22.0 & 312/321.0/330 & 289--320 \\
Llama-3.2-3B & 325.0 & 18.0 & 320/329.0/336 & 299--336 \\
Qwen3-32B & 316.0 & 22.0 & 312/321.0/330 & 289--320 \\
\bottomrule
\end{tabular}}
\par\smallskip{\footnotesize\TableNoteAlign K includes the common query-observation window and subtracts the exact supplied source span. Quest\textquotesingle s selected range includes its page rounding; its first two layers remain dense. FINCH retains and repositions K cache rows at compression. RefreshKV retains the full cache and has full query/refresh passes. These quantities do not imply equal physical storage or whole-decode access.\TableNoteEnd}
\end{table}

\input{context/v41_dynamic_public_completion_table}

Tables~\ref{tab:v41_dynamic_controlled} and~\ref{tab:v41_dynamic_capacity}
separate answer quality from the configured and observed access counts.
Table~\ref{tab:v41_dynamic_public_completion} retains termination and
recent-page controls. The numerical supplement contains the paired
method-minus-full and matched-minus-fixed contrasts, all relation/family
strata, and public question-type and source-overlap sensitivities.

Excluding the twelve remaining questions whose paired source occurs in
development leaves 368 questions from 353 groups.
Full/source/Quest/FINCH/RefreshKV counts are 193/182/136/169/179 for Qwen3-4B,
181/166/148/174/165 for Llama-3.2-3B, and
268/226/191/279/266 for Qwen3-32B; recent-page counts are 7, 0, and 1.
Source masking still lowers the point counts in all three models.
Across the complete 500-question policy matrices, the deterministic
empty-answer rule changes zero GPT-5.6 labels in each model:
all empty responses already receive No.

\subsection{Runtime and KV Memory Costs}
\label{app:operation_costs}

Table~\ref{tab:revision_cost} reports the measured memory and per-query
generation components.
\input{context/revision_cost_table}

\paragraph{Cost accounting.}
The runtime experiment uses H200 GPUs, bfloat16 weights, and eager attention,
with software versions recorded in the supplement. Model loading and
an initial warm-up are excluded; CUDA is synchronized at timing boundaries.
Per-query generation includes cache cloning/mask preparation or full
recomputation, suffix prefill, greedy decoding, and associated host work.
We additionally measure source preparation/tokenization, history prefill,
query tokenization, time to first token, output token counts, peak allocated
device bytes, and actual prompt KV tensor bytes. Prompt KV size counts
only the cache used for the active answer; the benchmark also holds the original prefix
for paired comparisons. It is not total live GPU allocation. Peak device allocation
includes weights and per-condition cache copies, so it is not a substitute for resident
KV size. Because source positions are supplied, the timing excludes the cost of locating them. Runs share a GPU server and do not measure serving throughput.

The batched detector timings report aggregate batch time and batch size;
their per-case amortization is not serial request latency. A separate
30-case measurement (six relations across five base items) runs the same
asserted-value detector one case at a time after warm-up. Median tokenization plus
detection is 572.63 ms, excluding model loading. The cost of answering the first query also includes
history preparation, prefill, detector work for detector policies, query
preparation and generation. Policies given the true replacement labels omit learned detection, so their costs are not directly comparable to end-to-end detector-based policies. Complete resident-cache and cold-start ten-query sequences, including detection, are in Appendix~\ref{app:v5_serial}.

\paragraph{Operation cost when the source positions are supplied.}
A separate Qwen3-8B study crosses four history lengths (1,024--8,192 tokens)
with five old-span positions (10--90\% of the history). Three warm-ups and
nine measured repetitions compare keeping, masking, physical removal, and
text recomputation with or without reusing the prefix before the interval.
Prefix-reusing recomputation takes $2.17$--$52.89\times$ the logical-mask
cost across these 20 cells. Timing begins after localization; answer quality
and detection cost are not measured. This operation-cost comparison does
not establish a complete-response speedup, and a logical mask retains KV.

\subsection{Detector Cost in a 40-Token Ten-Query Workload}
\label{app:v5_serial}

We choose 30 distinct test groups from the synthetic record dataset, one relation variant per group,
cycling the six relations over balanced attribute families. Each model
executes five policies at 0/100/300 background lines: keep-all, detector mask,
query-aware detector mask, detector text recomputation, and oracle text
recomputation. Each policy/history/context is evaluated on a sequence of
ten distinct questions, ordered current/historical/unrelated/current/current/
historical/unrelated/current/historical/unrelated. There are four current,
three historical and three unrelated questions per sequence. The same
immutable updated history is reused; previous answers are not appended to
it. This is a repeated-query workload after one update, not a stream of ten
new factual updates. Recompute arms filter once at update arrival and then
reuse the resulting history.

Timing including input preparation starts before the initial history is
tokenized and prefilled (\emph{cold timing}). Timing with a prepared history
starts when B arrives after that preparation (\emph{resident timing}). Both times are measured during
the same execution, without amortizing prefill across requests.
The total includes update tokenization, serial Qwen3-4B detection for detector
arms, cache editing/appending, each query's tokenization, cache copying,
suffix prefill, and EOS-stopped generation. Every timing boundary synchronizes
CUDA. Cache-consistency measurements are excluded from the timed operations.
Models are loaded and warmed
before timing, and policy order is shuffled per group/context. Shared-server
host contention remains possible, and these are latency measurements rather
than a serving-throughput study. Supplied spans and query types exclude retrieval and temporal-classification costs.

The Qwen models' keep-all histories span 140--171, 2,132--2,163, and
6,332--6,363 tokens across the three lengths; Llama spans 150--177,
2,050--2,077, and 5,850--5,877. The full dataset contains 1,350 query sequences
and 13,500 requests. The first and last K/V vectors at every layer remain
unchanged across query conditions. Table~\ref{tab:v5_sequential}
reports the longest context; shorter lengths, ground-truth baselines, per-request
components, query-type accuracy, and paired group-bootstrap intervals are
also supplied for the measured cumulative times.

\input{context/v5_sequential_table}

For detector masking minus keep-all, paired mean resident-time differences
are 833 [707,961], 866 [722,1010], and 769 [693,846] ms (95\% intervals),
respectively. These are means of paired differences, distinct from differences
of table medians. Query-aware masking uses supplied query-role labels to keep full access for historical and unrelated questions. It does not eliminate detector cost or losses on current answers. Detector recomputation adds 1,019 [851, 1,189], 1,017 [841, 1,194] and
809 [716, 902] ms and lowers historical accuracy by 41.1, 38.9 and 28.9
points; oracle-label recomputation adds 481, 470 and 231 ms. The difference
between detector recomputation and detector masking is resolved only in
the two Qwen models.

\subsection{Repeated Requests and the Cost of Applying a Decision}
\label{app:v39_cost_extension}

The sequential experiment tests a single update followed by ten requests,
at three history lengths and five policies. The 30 histories, rather than
the repeated requests, are the uncertainty units. These models use a
256-token answer limit; their times are reported separately from the
40-token comparison in Appendix~\ref{app:v5_serial}. Table~\ref{tab:v39_sequential} reports the longest histories.

\begin{table}[!htbp]
\TableStyle

\caption{Sequential workload at the longest tested history: current / historical / unrelated complete-answer percentages. Thirty histories each receive ten requests: four current, three historical, and three unrelated. The cached history remains unchanged between requests; previous answers are not appended.}
\label{tab:v39_sequential}
{\setlength{\tabcolsep}{\dimexpr\tabcolsep*6/8\relax}
\begin{tabular}{cccc}
\toprule
\TableHead{Model} & \FullCell{\FullHeader{Full access}} & \WholeCell{\WholeHeader{Detected mask}} & \TableHead{Query routing} \\
\midrule
Qwen3-32B & \FullCell{98.3/100.0/84.4} & \WholeCell{91.7/81.1/84.4} & 91.7/100.0/84.4 \\
Qwen3.8-27B$^{\dagger}$ & \FullCell{100.0/100.0/100.0} & \WholeCell{93.3/84.4/100.0} & 93.3/100.0/100.0 \\
Gemma 4 26B-A4B & \FullCell{100.0/100.0/95.6} & \WholeCell{96.7/93.3/96.7} & 96.7/100.0/95.6 \\
Gemma 4 31B & \FullCell{100.0/100.0/100.0} & \WholeCell{93.3/90.0/100.0} & 93.3/100.0/100.0 \\
Llama-3.1-70B & \FullCell{68.3/92.2/53.3} & \WholeCell{68.3/77.8/55.6} & 68.3/92.2/53.3 \\
\bottomrule
\end{tabular}}
\par\smallskip{\footnotesize\TableNoteAlign The five policies and three history lengths are retained in the numerical supplement. Generation uses a native 256-token limit and a fixed Qwen3-4B detector. $^{\dagger}$Masks act on full-attention layers; DeltaNet states are retained.\TableNoteEnd}
\end{table}

Applying the detected mask to all query types reduces historical accuracy
by 6.7--18.9 percentage points across the five generators at this length.
Four paired intervals lie below zero; Gemma 4 26B-A4B's interval is
$[-16.7,0.0]$ points and does not resolve a decrease at this confidence level.
Restricting the mask to supplied current-query labels restores historical
and unrelated baseline accuracy, but does not restore current accuracy:
four point estimates fall by 3.3--6.7 points on current queries, with intervals
reaching zero, and Llama-3.1-70B is unchanged. Preservation on routed-away queries is thus a benefit of keeping
their original access, not a demonstration that the masked cache retained
everything they needed.

\input{context/v45_sequential_time_table}

Table~\ref{tab:v39_sequential_time} reports total time.
In this runtime the mask changes each request's work little: summed query
processing over the ten requests differs from full access by about
23 ms (Qwen3.8-27B, $+23.1$, 95\% interval $[-14, 57]$) and by at most
2.3 ms elsewhere, per-token decoding stays within 2\% of full access, and the
mask changes update time by $-4$ to $-1$ ms. The added 0.47--1.87 s
therefore consists of detector inference on every history (0.61--0.69 s;
the detector fires on 16 of 30, 13 of them among the 15 replacements) plus
changes in answer length. Those changes come from a few histories: in
Gemma 4 26B-A4B, Gemma 4 31B and Llama-3.1-70B, two histories each, whose
masked answers ran 4.7--27.9 s longer (including 2 and 3 answers at the
256-token limit in the Gemma models), account for 98--111\% of the
request-time increase, while the median history's request time changes by
at most 135 ms. Separately, a serial asserted-value detector measurement
takes a median 573 ms for tokenization and detection on 30 cases,
excluding model loading (Appendix~\ref{app:operation_costs}), and
joint-KV readouts likewise require feature construction. Source
localization and query-type classification are excluded, so these times
are not the full cost of autonomous memory management.

All five paired intervals for added time remain above zero after ten
requests for both masking policies. The query-routed policy adds
0.55--1.12 seconds at the longest history, even before locating source
spans or learning query-type decisions. Because per-request work under the
mask stays close to full access in this runtime, a longer reuse horizon
does not offset the detector cost relative to full access; what a longer
horizon changes is the comparison with recomputation, whose re-encoding
(a median 0.74--4.18 s more than appending) is paid once per triggered
update. Differences between models also
include response length and backend behavior and are not a model-speed ranking.

The supplement's detector-selected recomputation policy instead
re-encodes the history without a detected source before the ten requests.
It adds 0.57--5.47 seconds across the five models. When the detector
fires, re-encoding the 5.8k--6.6k-token history takes a median
0.78--4.28 s, against 45--201 ms to append the update. Paired with detector-selected masking on the same histories,
recomputation is slower in Qwen3-32B, Qwen3.8-27B and Llama-3.1-70B
(recompute minus mask $+342$ to $+3{,}843$ ms, all intervals above zero;
slower in 24, 23 and 21 of 30 histories).
In both Gemma models the point estimates favor recomputation (574 versus
1,469 ms and 1,547 versus 1,873 ms added), but the paired differences are
not resolved ($-894$ $[-2{,}496, 55]$ and $-326$ $[-2{,}596, 1{,}197]$ ms)
because of the long masked answers; on the 16 histories where the detector
fired, recomputation is slower in 8 and 14. This workload therefore does
not establish that the supplied-position ordering reverses.
Recomputation also lowers historical accuracy to 41.1--57.8\%, against
77.8--93.3\% for the detected mask.

\begin{table}[!htbp]
\TableStyle

\caption{Median total operation time in milliseconds at target history length 8,192 and an update halfway through the history. The five columns retain the cache, mask source positions, physically drop positions, recompute all text, or recompute from a stored prefix.}
\label{tab:v39_operations}
{\setlength{\tabcolsep}{\dimexpr\tabcolsep*10/12\relax}
\begin{tabular}{cccccc}
\toprule
\TableHead{Model} & \FullCell{\FullHeader{Retain}} & \WholeCell{\WholeHeader{Mask}} & \TableHead{Drop} & \TableHead{Recompute} & \TableHead{Prefix} \\
\midrule
Qwen3-32B & \FullCell{129.8} & \WholeCell{200.7} & 396.1 & 4423.3 & 3148.8 \\
Qwen3.8-27B$^{\dagger}$ & \FullCell{219.5} & \WholeCell{219.5} & 272.9 & 4808.1 & 2513.2 \\
Gemma 4 26B-A4B & \FullCell{59.1} & \WholeCell{76.1} & 181.4 & 1480.6 & 931.3 \\
Gemma 4 31B & \FullCell{155.4} & \WholeCell{194.4} & 403.5 & 4109.0 & 2769.5 \\
Llama-3.1-70B & \FullCell{189.3} & \WholeCell{273.7} & 520.6 & 7036.2 & 4844.0 \\
\bottomrule
\end{tabular}}
\par\smallskip{\footnotesize\TableNoteAlign Each condition has nine measured repetitions after three warmups. All use native SDPA with the MATH backend and 512-token prefill chunks. This operation benchmark reports no answer-accuracy claim. Full recompute processes 7,855--9,106 tokens. $^{\dagger}$Qwen3.8 masks only its 16 full-attention layers; its mask overhead, about 10 ms by scaling Qwen3-32B's per-layer cost, is within the run-to-run spread (retain IQR 13 ms), so both conditions have a median of 219.5 ms. The mask median exceeds the retain median in 18 of the 20 benchmark cells (exact sign test $p=0.0004$).\TableNoteEnd}
\end{table}

Logical masks preserve the stored K/V allocation, whereas physical dropping
adds a copy operation and removes only the selected source positions.
At this table's setting, the removed source accounts for less than 0.4\%
of active K/V bytes in every model. Physical dropping takes longer than
logical masking in all five rows despite the smaller allocation.
Recomputing a suffix is cheaper than recomputing all text here, but requires
a stored prefix snapshot and remains more costly than access-time masking.
For Qwen3.8, full recurrent states are captured for prefix reuse; token
dropping alone does not erase them. The timing benchmark cannot establish
an accuracy-preserving tradeoff because it measures operation cost and
first-token agreement, not complete answers.

\FloatBarrier
\section{Reading a Cache Written by Another Checkpoint}
\label{app:v51_crossversion}

This study supports Section~\ref{sec:v51_crossversion}. It asks whether
a history cache computed by one checkpoint supports the same answers when
a sibling checkpoint reads it.

\paragraph{Pairs and prompts.}
Each pair contains a model from the main text and a sibling with the same
architecture and vocabulary: Qwen3-4B and Qwen3-8B with their base
checkpoints, Gemma 4 26B-A4B with its base checkpoint, and
Qwen3.8-27B with the earlier Qwen3.6-27B and Qwen3.5-27B releases. Every
prompt uses the main-text model's chat template and tokenizer. For every
prompt, both checkpoints' tokenizers must produce identical token IDs,
history boundaries and mask spans, so a cross read changes only the
weights that computed the cache.

\paragraph{Reading another cache.}
The writer prefills the history of the quantity task, which contains the
old record, an unrelated record, a neutral note and the later record
(Appendix~\ref{app:v18_unified}). The reader appends the question to a
copy of this cache and decodes greedily. Both readers stop at the union
of the two checkpoints' end-of-sequence tokens, and the writer's cache is
checked to be unchanged after every case. Long runs were split into case
shards; every case generated by more than one shard gave identical
answers, and merged runs keep one copy of each case. Each reader answers current,
historical and unrelated questions for all 240 test cases (120 base
groups, two supports), with full access and with whole-source masking,
once from its own cache and once from its sibling's. Token limits follow
the main-text model: 40 tokens for the Qwen3 pairs and 256 for the Gemma
and Qwen3.8 pairs. In the Qwen3.8 pairs, a cross read also carries the
writer's DeltaNet recurrent and convolution states, and masks act only on
the full-attention layers, as in the main text.

\paragraph{Checks and scoring.}
Before generation, each checkpoint passes the runtime check of
Appendix~\ref{app:v33_models} on a development case, apart from the one
full-output failure noted below, and each reader
passes the no-op and intervention checks on its sibling's cache. For the Qwen3 and Gemma pairs, the check requires only that decoding from
the cache reproduce the first token of a full forward pass. Of all
checkpoints, only Qwen3-4B-Base, whose greedy development output is a
repeated token, fails the full-output check; Qwen3-8B-Base and the Gemma
base checkpoint, which repeats the prompt rather than answering, pass it. Answers use the complete-answer scoring of
Appendix~\ref{app:v33_models}. Cross-minus-own differences and the
difference in masking effects use paired bootstrap intervals over base
groups (10,000 draws, stratified by relation). First-token agreement compares the reader's greedy first tokens from the
two caches.

\begin{table}[!htbp]
\WideTableStyle

\caption{Outcome entries give percentages (counts), N=240. Paired entries are top/bottom in the original header order. Reading another checkpoint\textquoteright s cache. Each reader appends the question to a history prefilled either by itself (own) or by the other checkpoint of its pair (cross) and decodes greedily. Each pair occupies two rows, and a cross read uses the other row\textquoteright s checkpoint as writer. Cells give own / cross percentages of complete answers with the correct number and unit on the same test cases; ``Invalid\textquotedblright{} counts current full-access outputs that hit the token limit or break the answer format. Changed counts current full-access answers whose content differs from the own read, ignoring JSON whitespace and code fences; Top-1 is the share of current full-access cross reads whose first generated token matches the own read.}
\label{tab:v51_crossversion}
{\setlength{\tabcolsep}{\dimexpr\tabcolsep*12/14\relax}
\begin{tabular}{ccccccc}
\toprule
\TableHead{Reader} & \FullCell{\FullHeader{Current}} & \WholeCell{\WholeHeader{Current, masked}} & \FullCell{\FullHeader{Historical}} & \TableHead{Invalid} & \TableHead{Changed} & \TableHead{Top-1} \\\midrule
Qwen3-8B & \FullCell{84.2 / 76.2} & \WholeCell{55.4 / 40.0} & \FullCell{100.0 / 100.0} & \shortstack[c]{\AnswerPctInline{0}{0}\\\AnswerPctInline{0}{0}} & \AnswerPct{12.1}{29} & 100.0\\Qwen3-8B-Base & \FullCell{72.9 / 3.3} & \WholeCell{31.2 / 0.0} & \FullCell{65.4 / 14.6} & \shortstack[c]{\AnswerPctInline{7.1}{17}\\\AnswerPctInline{95.4}{229}} & \AnswerPct{95.4}{229} & 1.7\\\addlinespace[2pt]
Qwen3-4B & \FullCell{83.3 / 84.6} & \WholeCell{45.0 / 43.8} & \FullCell{100.0 / 100.0} & \shortstack[c]{\AnswerPctInline{0}{0}\\\AnswerPctInline{0}{0}} & \AnswerPct{4.6}{11} & 70.4\\Qwen3-4B-Base & \FullCell{6.2 / 67.5} & \WholeCell{11.7 / 38.3} & \FullCell{0.0 / 100.0} & \shortstack[c]{\AnswerPctInline{91.3}{219}\\\AnswerPctInline{0}{0}} & \AnswerPct{91.3}{219} & 8.8\\\addlinespace[2pt]
Gemma 4 26B-A4B & \FullCell{85.8 / 33.8} & \WholeCell{34.6 / 16.2} & \FullCell{100.0 / 66.2} & \shortstack[c]{\AnswerPctInline{0}{0}\\\AnswerPctInline{45.4}{109}} & \AnswerPct{21.3}{51} & 6.2\\Gemma 4 26B-A4B base & \FullCell{0.0 / 0.0} & \WholeCell{0.0 / 0.0} & \FullCell{0.0 / 0.0} & \shortstack[c]{\AnswerPctInline{100}{240}\\\AnswerPctInline{100}{240}} & \AnswerPct{100}{240} & 100.0\\\addlinespace[2pt]
Qwen3.8-27B & \FullCell{72.5 / 82.9} & \WholeCell{65.8 / 66.2} & \FullCell{100.0 / 100.0} & \shortstack[c]{\AnswerPctInline{0}{0}\\\AnswerPctInline{0}{0}} & \AnswerPct{12.1}{29} & 100.0\\Qwen3.6-27B & \FullCell{83.3 / 80.8} & \WholeCell{62.5 / 53.8} & \FullCell{100.0 / 100.0} & \shortstack[c]{\AnswerPctInline{0}{0}\\\AnswerPctInline{0}{0}} & \AnswerPct{2.5}{6} & 100.0\\\addlinespace[2pt]
Qwen3.8-27B & \FullCell{72.5 / 75.8} & \WholeCell{65.8 / 60.4} & \FullCell{100.0 / 99.6} & \shortstack[c]{\AnswerPctInline{0}{0}\\\AnswerPctInline{0}{0}} & \AnswerPct{8.3}{20} & 100.0\\Qwen3.5-27B & \FullCell{83.3 / 81.2} & \WholeCell{50.8 / 58.3} & \FullCell{100.0 / 100.0} & \shortstack[c]{\AnswerPctInline{0}{0}\\\AnswerPctInline{0}{0}} & \AnswerPct{2.9}{7} & 100.0\\
\bottomrule
\end{tabular}}
\par\smallskip{\footnotesize\TableNoteAlign Each cell uses the same 240 test cases (120 base groups, two supports) for own and cross reads. Answers are limited to 40 tokens for the Qwen3-8B and Qwen3-4B pairs and 256 for the others. Masked reads hide the old record in the shared prefilled cache. Every prompt uses the main-text model\textquoteright s chat template and tokenizer for both checkpoints. For Qwen3.8/3.6/3.5, a cross read also carries the writer\textquoteright s DeltaNet states, and masks act on full-attention layers.\TableNoteEnd}
\end{table}

\paragraph{Answers.}
Table~\ref{tab:v51_crossversion} compares own and cross reads on the
same test cases. Qwen3-8B's complete current answers fall from 84.2\% to
76.2\% on its base sibling's cache (cross minus own $-7.9$ points,
$[-12.5,-3.8]$), and its masked accuracy is 15.4 points lower ($[-19.2,-11.7]$), so masking
costs 7.5 more points than on its own cache ($[-13.3,-1.7]$). Most losses fall in the 60 replacements whose later record needs the
earlier unit: complete answers fall from 59 to 45, and 15 of these
answers give the old quantity, against one from the own cache. On 10 of
the 40 questions whose later record concerns another entity, Qwen3-8B
gives that record's value from the sibling's cache, against none from its
own. Every first generated token matches the own read,
because answers from both caches open with the same compact JSON prefix; 29 of the 240
answers nevertheless differ. Qwen3-4B's accuracy shows no detectable change ($+1.2$, $[-1.7,4.2]$;
masked accuracy $-1.2$, $[-4.6,2.1]$; difference in masking effects
$-2.5$, $[-6.7,1.7]$); 11 of
its 240 current answers change content, and 64 more differ only in JSON
whitespace.
The base readers show that a foreign cache can also change the answer
format. From its instruction sibling's cache, Qwen3-8B-Base reaches the
40-token limit or breaks the format in 229 of 240 current answers, against
17 from its own cache. Qwen3-4B-Base does the opposite: from its own cache,
219 answers are invalid, many of them a repeated token, whereas from its
sibling's cache none are and 67.5\% are complete. Historical and
unrelated answers move in the same directions (Qwen3-8B-Base: 65.4\% to
14.6\% and 99.6\% to 0.8\%; Qwen3-4B-Base: 0\% to 100\% and 4.6\% to
78.3\%).

The Qwen3.8 pairs compare released instruction models. Qwen3.8-27B's
complete current answers rise from 72.5\% to 82.9\% on Qwen3.6-27B's
cache ($+10.4$, $[8.3,12.5]$) and to 75.8\% on Qwen3.5-27B's ($+3.3$,
$[0.4,6.2]$). On Qwen3.6's cache it gains 27 answers and loses two; 25
of the gains replace UNKNOWN on questions whose later record concerns
another entity, so the old record's value is still current. The earlier
releases lose a little accuracy on Qwen3.8's cache ($-2.5$, $[-5.0,-0.4]$ for
Qwen3.6-27B; $-2.1$, $[-4.6,0.0]$, an interval reaching zero, for Qwen3.5-27B), with 6 and 7 changed
answers. Masked reads behave differently. Relative to masking its own
cache, Qwen3.8-27B's masked accuracy shows no detectable change on
Qwen3.6's cache ($+0.4$, $[-0.8,1.7]$), so masking removes its full-access gain (difference in masking effects
$-10.0$, $[-12.5,-7.5]$). Masked accuracy is 5.4 points lower for
Qwen3.8-27B on Qwen3.5's cache ($[-9.2,-2.1]$) and 8.8 points lower for
Qwen3.6-27B ($[-12.5,-5.0]$), whereas Qwen3.5-27B gains 7.5
($[2.9,12.1]$), mostly by restoring the unit in 25 replacements. With
full access, historical answers of every instruction reader keep their
content in all but at most one of 240 cases; under whole-source masking,
51 to 93 of them differ from the own read in the Qwen pairs.

Gemma 4 26B-A4B changes most. With full access, its own reads wrap every
answer in a code fence, which the scoring accepts; on its base checkpoint's cache, 116
current answers are unfenced JSON and 109 are unwrapped answers, 75 of which state the correct quantity; complete
answers fall from 85.8\% to 33.8\% ($-52.1$, $[-59.6,-44.6]$), or to
65.0\% if these unwrapped answers were accepted. The
content changes too: for all 80 questions whose later record concerns
another entity or attribute, the reader gives that record's value,
against 33 from its own cache, and four replacements give the old
quantity. Historical answers keep their content in all 240 cases, but 81
of them are unwrapped answers (100\% to 66.2\%). Under whole-source masking, 95
own and no cross current reads reach the 256-token limit, and complete
current answers fall from 34.6\% to 16.2\% ($-18.3$, $[-22.1,-14.6]$).
The base checkpoint completes no answer from either cache; from the
instruction checkpoint's cache, all 240 current answers reach the token
limit, against 69 from its own. The direction and size
of the change therefore depend on the pair, the access condition and the
question.

\begin{table}[!htbp]
\WideTableStyle

\caption{State similarity and readout transfer between paired checkpoints. Cosine similarities average over layers with a K/V cache (all 36 for Qwen3, 30 for Gemma, the 16 full-attention layers of Qwen3.8) and over test cases, for identical token IDs. Role AUCs use the 570-history recognition panel: a joint-KV linear readout trained and tested on the main-text model (own) or trained on its sibling and tested on the main-text model (transfer).}
\label{tab:v51_crossversion_states}
{\setlength{\tabcolsep}{\dimexpr\tabcolsep*10/12\relax}
\begin{tabular}{cccccc}
\toprule
\TableHead{Main-text model} & \TableHead{Sibling} & \TableHead{K cos} & \TableHead{V cos} & \TableHead{Own AUC} & \TableHead{Transfer AUC} \\
\midrule
Qwen3-8B & Qwen3-8B-Base & 0.921 & 0.825 & 0.550 \TableSecondary{[0.516,\,0.591]} & 0.594 \TableSecondary{[0.560,\,0.640]} \\
Qwen3-4B & Qwen3-4B-Base & 0.915 & 0.816 & 0.653 \TableSecondary{[0.603,\,0.714]} & 0.520 \TableSecondary{[0.478,\,0.562]} \\
Gemma 4 26B-A4B & Gemma 4 26B-A4B base & 0.670 & 0.643 & 0.627 \TableSecondary{[0.591,\,0.672]} & 0.365 \TableSecondary{[0.290,\,0.430]} \\
Qwen3.8-27B & Qwen3.6-27B & 0.829 & 0.760 & 0.872 \TableSecondary{[0.830,\,0.920]} & 0.756 \TableSecondary{[0.700,\,0.822]} \\
Qwen3.8-27B & Qwen3.5-27B & 0.829 & 0.767 & 0.872 \TableSecondary{[0.830,\,0.920]} & 0.645 \TableSecondary{[0.606,\,0.695]} \\
\bottomrule
\end{tabular}}
\par\smallskip{\footnotesize\TableNoteAlign Transfer readouts choose regularization on the sibling\textquoteright s development split; choosing it on the main-text model\textquoteright s development split is reported in the numerical supplement. Intervals are 95\% role-pair bootstrap intervals.\TableNoteEnd}
\end{table}

\paragraph{States and readouts.}
For identical token IDs, keys and values of paired checkpoints have mean
cosine similarity 0.64--0.92 (Table~\ref{tab:v51_crossversion_states}); it is
higher at the old record than averaged over the history in every pair
(keys 0.755--0.951, values 0.734--0.889). The Qwen3.8 pairs, whose
cosines cover only the 16 full-attention layers, are less similar than
the Qwen3 pairs (keys 0.829 against 0.915--0.921; values 0.760--0.767
against 0.816--0.825). Relative L2 differences are 0.36--0.58 for keys
and 0.53--0.65 for values in the Qwen pairs, and 0.79 and 0.83 in Gemma;
the carried DeltaNet recurrent and convolution states of the Qwen3.8
pairs have mean cosine 0.87 and 0.89, respectively. A joint-KV
readout trained on the sibling and tested on the main-text model reaches a
role AUC of 0.594 for Qwen3-8B, above its own 0.550, but 0.520 for
Qwen3-4B, below its own 0.653. Qwen3.8-27B's own readout reaches 0.872;
readouts transferred from Qwen3.6-27B and Qwen3.5-27B fall to 0.756 and
0.645. Gemma's keys and values, averaged over 30 attention layers, are
less similar (0.670 and 0.643), and a readout transferred from its base
checkpoint ranks role swaps below chance (0.365, $[0.290,0.430]$, against
0.627 for its own). The own values differ by at most 0.011 from the joint-KV
readouts of Appendix~\ref{app:v35_representation}. Regularization is selected on a development
split of only 15 swaps and 15 confirmations (225 comparisons), and the
choice matters: across the four strengths, test role AUCs range from
0.546 to 0.653 for Qwen3-4B's own readout, from 0.557 to 0.655 for the
readout transferred to Qwen3-8B, and from 0.365 to 0.461 for the one
transferred to Gemma, so these role AUCs should be read as unstable at
this sample size. Similar states and partly
transferable readouts coexist with changed answers; state similarity does
not establish that a cache remains valid for another checkpoint.

\FloatBarrier
\section{Answers to Questions about Wikipedia Revision Histories}
\label{app:wiki_source_quality}

This comparison asks how much earlier article text must remain accessible
to answer questions about past versions and changes. We use the released
Wikipedia-history subset of MINTEval~\citep{minteval2026}. Each question
is asked after the end of the revision history.

\paragraph{Histories and cache operations.}
Of 196 released histories, 161 fit both generators' native 262,144-token
windows with all original questions and a 1,024-token answer budget.
Eight histories are reserved for development, leaving 153 articles and
1,168 questions for evaluation without truncation. Qwen3.8-27B and
Gemma 4 26B-A4B use native chat templates, thinking disabled, greedy
decoding and instructions to return only a boxed final answer.

Full access retains the complete history. Whole-source masking hides all
non-final article bodies. Changed-text masking hides only passages replaced
or deleted between adjacent revisions. Both masks retain revision headers
and are fixed before the questions. Recomputation deletes the non-final
bodies and processes the remaining history again. The two generators
produce 9,344 answers across these four operations. Because the masks
select text by revision history, some hidden statements may still be
correct or necessary for a question about the past.

\paragraph{Checking which questions the sources can answer.}
Before answer grading, a separate review checks whether each complete
history supports the requested answer and specifies the relevant time
or relation. The reviewer sees the history, questions and original
references, without generated answers or cache-operation labels.
GPT-5.6 reviews 152 articles and GPT-6-astra reviews one.
Of the original 1,168 questions, 857 are answerable, one is partially
answerable, 266 are unanswerable and 44 are ambiguous
(Table~\ref{tab:v43_wiki_source_quality}). These source assessments set
the same grading target for every generator and operation.

\begin{table}[!htbp]
\WideTableStyle
\caption{Outcome entries are percentages (counts), using the row-specific $n$. Question answerability from the released full Wikipedia histories, assessed without generated answers. The assessment retains every original question, including those judged unanswerable or ambiguous.}
\label{tab:v43_wiki_source_quality}
\centering
{\setlength{\tabcolsep}{\dimexpr\tabcolsep*10/12\relax}
\begin{tabular}{cccccc}
\toprule
\TableHead{Original type} & \TableHead{Questions} & \TableHead{Answerable} & \TableHead{Partial} & \TableHead{Unanswerable} & \TableHead{Ambiguous} \\\midrule
Simple & 249 & \AnswerPct{45.4}{113} & \AnswerPct{0}{0} & \AnswerPct{54.6}{136} & \AnswerPct{0}{0}\\History & 409 & \AnswerPct{98.8}{404} & \AnswerPct{0}{0} & \AnswerPct{0}{0} & \AnswerPct{1.2}{5}\\Multi-hop & 268 & \AnswerPct{50}{134} & \AnswerPct{0}{0} & \AnswerPct{48.5}{130} & \AnswerPct{1.5}{4}\\Counting & 46 & \AnswerPct{93.5}{43} & \AnswerPct{2.2}{1} & \AnswerPct{0}{0} & \AnswerPct{4.3}{2}\\Ordering & 196 & \AnswerPct{83.2}{163} & \AnswerPct{0}{0} & \AnswerPct{0}{0} & \AnswerPct{16.8}{33}\\\midrule
All & 1,168 & \AnswerPct{73.4}{857} & \AnswerPct{0.1}{1} & \AnswerPct{22.8}{266} & \AnswerPct{3.8}{44}\\
\bottomrule
\end{tabular}}
\par\smallskip{\footnotesize\TableNoteAlign The evaluation set contains 153 articles. Labels use the complete original text, preserving uncertainty and differences between revisions. Released contexts omit editor metadata. Original question types differ from target time: 745 questions ask about changes or aggregation, 419 about historical states, and only four about the current state. These source-only labels remain separate from subsequent answer judgments.\TableNoteEnd}
\end{table}

\paragraph{Article exclusion.}
The grading service refused inputs containing explicit song lyrics from
one article. We therefore exclude that article's seven questions from
every generator and operation, irrespective of their answers. Four of
the seven questions are source-answerable. The answer comparisons use
the remaining 152 articles, 1,161 questions and 9,288 generated answers,
with completeness measured on the 853 source-answerable questions.
The original generations and source assessments are retained.

\paragraph{Grading answers against the complete history.}
The answer grader receives the full history, questions, original
references, fixed source assessments and generated answers. Generator,
cache operation and termination status are hidden. Identical answers
to the same question share a grade. A fixed parser extracts the boxed
answer when a box marker is present; otherwise it supplies the complete
response. The grader separately checks agreement with the original
reference, coverage of the requested details, unsupported or contradictory
claims, and justified abstention. It judges all cache operations against
the same full history, including passages hidden from the generator.

The answer review uses two GPT-5.6 configurations: the primary variant
for 151 articles and an alternate variant for one. Whether
generation ends normally or reaches the token cap is read from the
generation record, independently of these content judgments. Reference
agreement uses all 1,161 questions. Completeness uses the fixed set of
853 source-answerable questions. Unresolved disagreements between the
source assessment and answer grade remain in that denominator and do
not count as complete answers. Paired changes are question-weighted;
95\% intervals use 10,000 bootstrap resamples of complete articles,
keeping each article's questions, generators and operations together.

\begin{figure}[!htbp]
\centering
\includegraphics[width=\linewidth]{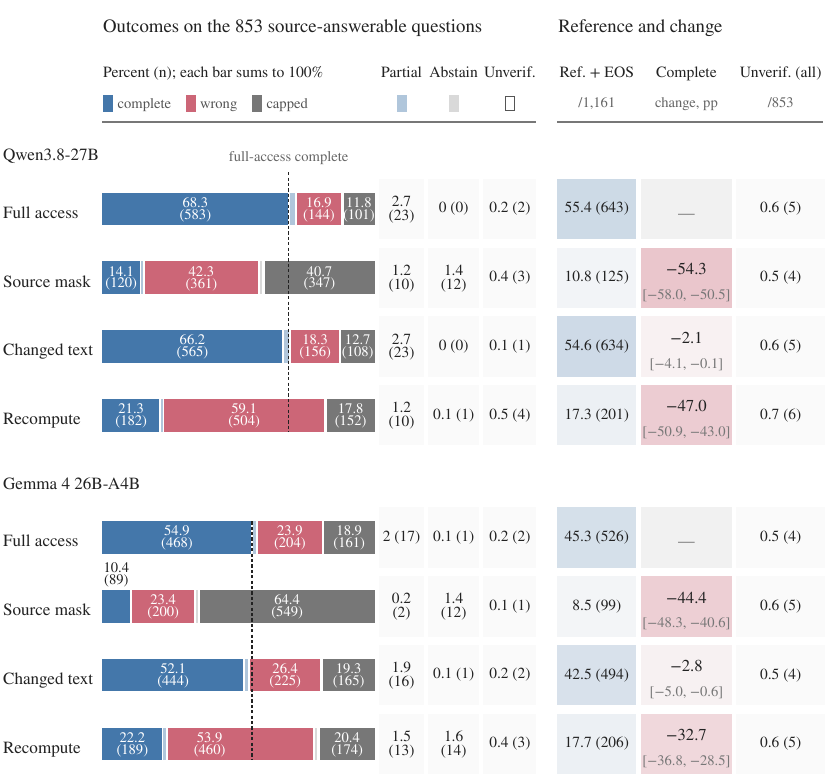}
\caption{Wikipedia answers on 152 articles: 1,161 analyzed questions, of which 853 are independently source-answerable. Left: mutually exclusive outcomes on the 853 questions; each bar sums to 100\%; the dashed line marks full-access complete. Bars print complete (blue), wrong (rose) and capped (gray) percentages with counts in parentheses in or above their segments; columns give partial, abstain and unverifiable percentages (counts), whose thin segments may not show. Capped outputs take precedence over semantic labels; all other outcomes require EOS. Wrong includes contradictions and unsupported assertions; partial answers are not equivalent to losses from a complete baseline. Right: reference agreement with EOS (shaded by percentage) uses all 1,161 questions. Complete-answer changes (shaded by loss) use the source-answerable subset and paired 95\% article-bootstrap intervals relative to full access. Right-hand unverifiable percentages use the 853 answerable questions and include grades without the EOS restriction. The article exclusion and fixed access policies are described above.}
\label{fig:v53_wiki_answers}
\end{figure}

\paragraph{Completeness and termination.}
Whole-source masking reduces Qwen's complete answers with normal EOS
from 583 to 120 of 853, a change of $-54.3$ percentage points with a
95\% interval of $[-58.0,-50.5]$. Gemma falls from 468 to 89, or
$-44.4$ points $[-48.3,-40.6]$. Changed-text masking has much smaller
losses: Qwen gives 565 complete answers ($-2.1$ points,
$[-4.1,-0.1]$) and Gemma gives 444 ($-2.8$ points, $[-5.0,-0.6]$).
Recomputation gives 182 and 189, respectively, still below full access
(Figure~\ref{fig:v53_wiki_answers}). Reference agreement is reported
separately over all 1,161 questions.

The figure also separates token-limit stops from the content of answers
that end normally. Under whole-source masking, capped outputs increase
from 101 to 347 for Qwen and from 161 to 549 for Gemma. Qwen's normally
terminated wrong answers increase from 144 to 361; Gemma's change from
204 to 200. The loss of complete answers therefore combines changes in
content with changes in stopping behavior. Capped and partial outputs
are kept separate from wrong answers that end normally.

\FloatBarrier

\input{context/v53_wiki_strata_table}

\paragraph{Questions about past states and changes.}
The analyzed set contains 741 questions about changes or aggregation,
416 about historical states and four about the current state. Their
source-answerable subsets contain 436, 413 and four questions,
respectively (Table~\ref{tab:v53_wiki_strata}a). Under whole-source
masking, Qwen's complete historical answers with EOS fall from 295 to
43 of 413, and Gemma's from 261 to 55. Answers about changes or
aggregation fall from 284 to 73 and from 205 to 30 of 436.
These questions chiefly measure recall of earlier states and changes;
the four current-state questions provide little coverage of updating
current facts.

\paragraph{Results by grading configuration.}
Table~\ref{tab:v53_wiki_strata}b reports results separately for the two
GPT-5.6 configurations and for requests that additionally repeat the rubric's
consistency and completeness requirements. Whole-source masking reduces
complete answers with EOS in every group for both generators. Each
article uses the same grader and instructions across all generators and
cache operations. Articles were not randomly assigned to grading
configurations, so differences between these groups cannot measure the
effect of the grader or prompt. The primary analysis keeps all 853
source-answerable questions; the numerical results also report a
sensitivity analysis on the 844 questions whose grades are resolvable
under every operation.

The comparison shows why the age of an article revision is insufficient
to decide whether its contents can be hidden. Earlier passages contain
information explicitly requested by historical and change questions.
Keeping unchanged passages retains many more complete answers in both
generators. The narrower mask also hides fewer tokens, so the comparison changes
both the selected passages and the amount of text kept.
\FloatBarrier

\FloatBarrier
\section{Further Connections to Memory Updates and Cache Reuse}
\label{app:extended_related_work}

This appendix expands Section~\ref{sec:related_work} on temporal updates,
recognizing replacements from representations, and reusing caches across
histories and model checkpoints.

\paragraph{Temporal records and knowledge editing.}
Knowledge conflicts include disagreements between contextual records
\citep{knowledgeconflicts2024}; temporal conflict studies distinguish
detecting outdated information from using its replacement
\citep{wallat2026facts}. AToKe evaluates whether a parameter edit preserves historical knowledge,
while RippleEdits evaluates consequences for related facts
\citep{atoke2024,rippleedits2024}. In text memory, Mem0 and A-MEM revise
retrieved records \citep{mem02025,amem2025}, and Zep represents temporal
relationships and invalidates superseded edges while preserving their
history \citep{zep2025}. Thus, keeping the old value for a historical question is
an established requirement. Our quantity task isolates a different use
of the same old record: its number changes but its unit remains necessary
for the current answer. A token mask must distinguish these uses even
when the replacement relation is supplied correctly.

The STALE preprint separates recognizing invalidation from using an
update and introduces CUPMem for consolidation under a state schema
\citep{stale2026}. MemOS manages plaintext, activation and parameter memories under a
versioned memory abstraction \citep{memos2025}. In-context editing changes
supplied facts with fixed weights \citep{ike2023}; parameter editing
changes the model itself \citep{rome2022,memit2023,mend2022}.
GRACE adds discrete key--value adaptors without changing the pretrained
weights \citep{grace2023}.
Source masking keeps model weights and prefilled history states unchanged;
recomputation rebuilds those states with the same weights.

\paragraph{Stored representations and model memory.}
InfLLM retrieves relevant context blocks, and EM-LLM organizes a stream
into events for later retrieval \citep{infllm2024,emllm2025}.
MemoryLLM and M+ maintain latent memory pools within the model
\citep{memoryllm2024,mplus2025}; cache-augmented generation preloads
documents for later questions \citep{cag2025}. Memory$^3$, KBLaM and
Cartridges use stored or trained KV states for knowledge access
\citep{memory32024,kblam2025,cartridges2026}. These approaches establish
that stored representations can support memory retrieval. The recognition
experiments here instead supply the relevant pair and ask whether its
representations distinguish a replacement from a confirmation.
Learned predictors measure accessibility to a specified readout;
their success alone does not establish the mechanism used during
answer generation \citep{hewitt2019controls}. Probes of the
residual stream likewise detect signals of knowledge conflict
\citep{zhao2024residual}.

PI-LLM measures interference from earlier key--value updates when models
are asked for the final values \citep{unableforget2025}. Causal analyses
identify lookback computations for tracking characters' beliefs
\citep{lookbacks2025}; entity-tracking experiments examine query-time
aggregation and fragile suppression after an entity is removed
\citep{statechanges2026}. Sparse-autoencoder analyses characterize key
and value representations and their reconstruction quality
\citep{addressbook2025}. These studies motivate examining particular
representations and computations. Our fixed scores and supervised
readouts instead measure replacement prediction on supplied pairs;
they do not identify a causal circuit for applying the update.

\paragraph{Recent preprints on cache updates.}
Event-KV reports information from omitted source positions retained in
later KV rows, and interference of appended corrections with historical
questions \citep{eventkv2026}. On its trained Qwen3-8B model, MEMENTO
reports lower reasoning accuracy when retained summary KV is replaced
by a fresh encoding of the same summary text \citep{memento2026}.
Our fixed-model comparisons follow which current and historical answers
survive masking and recomputation after a factual update.

\paragraph{Cache editing and selective reuse.}
\emph{Models Take Notes at Prefill} studies downstream dependence,
while KVEraser trains a cache editor for a supplied deletion interval
\citep{modelsnotes2026,kveraser2026}. Stale-cache repair rebuilds document caches under a recomputation budget
\citep{stalekvrepair2026}. CacheBlend, EPIC and KVLink reuse cached chunks
with selective recomputation or trained link tokens
\citep{cacheblend2025,epic2025,kvlink2025}; these methods address the
context dependence of reused states, whereas our intervention comparison
supplies a replacement and measures what remains usable: the current
value, a needed unit and the answer to a historical question. Masking
blocks direct attention while preserving downstream states;
recomputation rebuilds those states and also changes positions and
length. SleepGate studies synthetic updates with a four-layer,
793K-parameter base transformer \citep{sleepgate2026}; SideQuest learns
to evict tool responses
\citep{sidequest2026}; KEEP uses structured environmental updates
\citep{keep2026}. Their mechanisms differ from applying a mask to an
existing model's history. We use supplied replacement relations as well
as measured detectors to separate the choice of a source from the effects
of changing access to it.

\paragraph{Compatibility between model checkpoints.}
DroidSpeak selectively recomputes layers to reuse caches across
fine-tuned variants \citep{droidspeak2026}. The KVShareArena preprint
varies context and producer checkpoints in direct-reuse controls, and
compares repair methods under changed contexts on multi-source question answering
\citep{kvsharearena2026}. Our five sibling-pair tests measure answer
content, formatting and replacement readouts under direct cache transfer,
alongside K/V similarity, including adjacent hybrid-model releases.
Both improvements and losses occur, so aggregate accuracy alone does
not describe which answers remain stable.

Compatible caches can also be designed explicitly. Activated LoRA
leaves pre-invocation states unchanged \citep{alora2025}, and ICaRus
uses a frozen logical cache encoder during fine-tuning
\citep{icarus2026}. Other approaches exchange caches through trained
projections or alignment \citep{c2c2026,latentalign2026}, a fitted linear
map within a model family \citep{crossmodelkv2026}, or selected layers of
models sharing a base \citep{kvcomm2026}. Our unmodified sibling
checkpoints do not test these compatibility mechanisms; the observed
answer changes motivate checking reuse under the intended questions.

\paragraph{Serving cost and evaluation scope.}
PagedAttention manages cache allocation and sharing
\citep{pagedattention2023}; Prompt Cache, KDN and LMCache store and move
states across requests \citep{promptcache2024,kdn2024,lmcache2025}.
Selection and compression reduce cache requirements
\citep{h2o2023,snapkv2024,kivi2024,quest2024,kvzip2025}.
Learned utility and retention methods also reduce cache requirements
\citep{spkv2026,kvp2026,lukv2026}, and FINCH conditions access on the
prompt \citep{finch2024}. These efficiency objectives do not specify
which fact supersedes another. Changing prompt content limits prefix-cache
reuse to the unchanged prefix \citep{dontbreakcache2026}. Our repeated-query
comparison therefore includes recognition and cache handling as well
as the local intervention, while still supplying source locations and
query types.

We distinguish current, historical and
unrelated questions, candidate preference and complete-answer preservation
on evolving-fact and long-history tasks: contextual MQuAKE
\citep{mquake2023}, MINTEval \citep{minteval2026}, BABILong
\citep{babilong2024}, FactConsolidation from MemoryAgentBench
\citep{memoryagentbench2025}, OAKS \citep{oaks2026} and LongMemEval
\citep{longmemeval2025}.

\end{document}